\documentclass{opendatalab}
\newcommand\blfootnote[1]{%
  \begingroup
  \renewcommand\thefootnote{}\footnote{#1}%
  \addtocounter{footnote}{-1}%
  \endgroup
}
\usepackage[utf8]{inputenc}
\usepackage{amsmath,amssymb}
\usepackage{amsthm}
\usepackage{nicefrac}
\usepackage{booktabs}
\usepackage{array}
\usepackage{multirow}

\newcommand{\cmark}{\ensuremath{\checkmark}}
\newcommand{\pmark}{\ensuremath{\triangle}}
\newcommand{\xmark}{\ensuremath{\times}}

\DeclareMathOperator*{\argmin}{arg\,min}
\DeclareMathOperator*{\argmax}{arg\,max}
\usepackage{xcolor}

\usepackage[most]{tcolorbox}
\usepackage{longtable}
\usepackage{listings}
\usepackage[numbers]{natbib}
\usepackage{enumitem}
\usepackage{graphicx}
\usepackage{algorithm}
\usepackage{algpseudocode}
\usepackage{xspace}
\usepackage{hyperref}
\usepackage{algorithm}%
\usepackage{algorithmicx}%
\usepackage{algpseudocode}%
\usepackage{listings}%
\usepackage{wrapfig}
\usepackage{booktabs,multirow,makecell,colortbl,graphicx,xcolor,amsmath,amssymb}
\definecolor{bestbg}{RGB}{217,234,211}
\definecolor{worstbg}{RGB}{244,204,204}
\definecolor{basebg}{RGB}{239,239,239}

\usepackage{tikz}
\usepackage[edges]{forest}
\usetikzlibrary{mindmap, trees, shadows, arrows.meta}
\useforestlibrary{edges}
\usepackage{mathtools}

\definecolor{hidden-red}{RGB}{205, 44, 36}
\definecolor{hidden-blue}{RGB}{194,232,247}
\definecolor{hidden-orange}{RGB}{243,202,120}
\definecolor{hidden-green}{RGB}{34,139,34}
\definecolor{hidden-pink}{RGB}{255,245,247}
\definecolor{hidden-black}{RGB}{20,68,106}

\definecolor{codegreen}{rgb}{0,0.6,0}
\definecolor{codegray}{rgb}{0.5,0.5,0.5}
\definecolor{codepurple}{rgb}{0.58,0,0.82}
\definecolor{backcolour}{rgb}{0.95,0.95,0.92}
\definecolor{promptcolor}{HTML}{D1D0F2}
\definecolor{promptcolorheader}{HTML}{bdbcec}
\definecolor{bestbg}{RGB}{217,234,211}
\definecolor{basebg}{RGB}{239,239,239}
\definecolor{warnbg}{RGB}{252,229,205}
\definecolor{oursbg}{RGB}{229,243,255}
\newcommand{\bestnum}[1]{\cellcolor{bestbg}\textbf{#1}}

\newcommand{\github}{\raisebox{-1.5pt}{\includegraphics[height=1.05em]{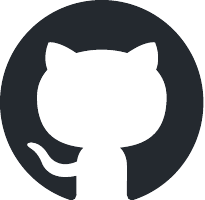}}\xspace}
\newcommand{\web}{\raisebox{-1.5pt}{\includegraphics[height=1.05em]{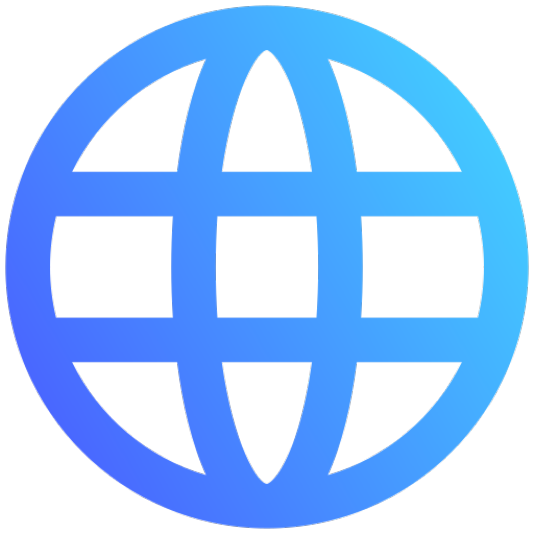}}\xspace}
\newcommand{\huggingface}{\raisebox{-1.5pt}{\includegraphics[height=1.05em]{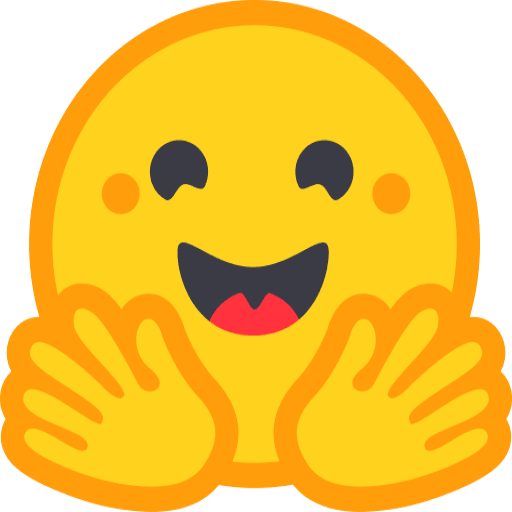}}\xspace}

\definecolor{promptcolor}{HTML}{E3F0FA}
\definecolor{promptcolorheader}{HTML}{B5D6ED}
\definecolor{prompttitletext}{HTML}{1B3A5C}
\definecolor{optHiddenBlue}{HTML}{4a90e2}
\definecolor{optHiddenBlack}{HTML}{333333}

\definecolor{famT1}{RGB}{80,120,200}   
\definecolor{famT2}{RGB}{120,90,180}   
\definecolor{famT3}{RGB}{230,120,120}  
\definecolor{famT4}{RGB}{90,200,170}   
\definecolor{famT5}{RGB}{220,160,60}   

\definecolor{famO1}{RGB}{96,160,96}    
\definecolor{famO2}{RGB}{214,126,68}   
\definecolor{famO3}{RGB}{86,148,196}   
\definecolor{famO4}{RGB}{166,118,190}  
\definecolor{famO5}{RGB}{188,150,64}   
\definecolor{famO6}{RGB}{192,98,122}   

\usepackage{booktabs}
\usepackage{tabularx}
\usepackage{makecell}
\usepackage{array}
\newcolumntype{L}[1]{>{\raggedright\arraybackslash}p{#1}}
\newcolumntype{Y}{>{\raggedright\arraybackslash}X}

\usepackage{xcolor}
\usepackage[most]{tcolorbox}

\definecolor{main}{RGB}{70, 120, 180}
\definecolor{sub}{RGB}{245, 248, 252}

\definecolor{famS0}{HTML}{6A8292} 
\definecolor{famS1}{HTML}{1E88E5} 
\definecolor{famS2}{HTML}{26A69A} 
\definecolor{famS3}{HTML}{E67E22} 
\definecolor{famS4}{HTML}{8E44AD} 
\definecolor{famS5}{HTML}{1565C0} 
\newtcolorbox{takeaway}{
  enhanced,
  breakable,
  boxrule = 0pt,
  colback = sub,
  borderline west = {2pt}{0pt}{main},
  borderline east  = {2pt}{0pt}{main},
  sharp corners,
  boxsep = 3pt,
  left = 8pt,
  right = 8pt,
  top = 5pt,
  bottom = 5pt,
  before skip = 6pt,
  after skip = 6pt,
}

\newtcolorbox{promptbox}[1][]{
  enhanced, breakable,
  top=0.3em,bottom=0.3em,left=0.5em,right=0.5em,
  toptitle=0.3em,bottomtitle=0.2em,boxsep=0pt,
  colframe=promptcolorheader, colback=promptcolor!50, boxrule=0.5pt,
  width=\columnwidth, 
  coltitle=prompttitletext,
  title={\footnotesize #1} 
}
\lstdefinestyle{promptstyle}{
    backgroundcolor=\color{backcolour},   
    commentstyle=\color{codegreen},
    keywordstyle=\color{magenta},
    numberstyle=\tiny\color{codegray},
    stringstyle=\color{codepurple},
    basicstyle=\ttfamily\footnotesize,
    breakatwhitespace=false,         
    breaklines=true,                 
    captionpos=b,
    keepspaces=true,
    numbers=left,
    numbersep=5pt,
    showspaces=false,          
    showstringspaces=false,
    showtabs=false,
    tabsize=2
}
\title{OmniOpt: Taxonomy, Geometry, and Benchmarking of Modern Optimizers}

\author[1,3]{Siyuan Li$^{*}$}
\author[1,2]{Jiabao Pan$^{*}$}
\author[1,4]{Yumou Liu$^{*}$}
\author[7]{Zhuoli Ouyang$^{*}$}
\author[3]{Xin Jin$^{*}$}
\author[5]{Xinglong Xu}
\author[5]{Jingxuan Wei}
\author[2]{Shengye Pang$^{\dagger}$}
\author[6]{Jintao Chen}
\author[4]{Xuanhe Zhou}
\author[1]{Conghui He}
\author[1]{Cheng Tan$^{\dagger}$}

\affiliation[1]{Shanghai Artificial Intelligence Laboratory}
\affiliation[2]{Shanghai University}
\affiliation[3]{Westlake University}
\affiliation[4]{Shanghai Jiao Tong University}
\affiliation[5]{UCAS}
\affiliation[6]{Zhejiang University}
\affiliation[7]{Southern University of Science and Technology}

\abstract{
Optimizer selection for large-scale model training has become a system-level
design decision constrained jointly by compute, memory, tuning budget, and task
diversity, yet the landscape of over one hundred methods remains fragmented.
We therefore present \textbf{OmniOpt},
a unified survey and benchmark cookbook of optimizers for the research community.
OmniOpt rests on four coupled components. First, we treat every optimizer update
as a structured transformation through a five-stage meta-pipeline, and show that
most methods engage only one or two of these stages. Second, we use
norm-constrained linear minimization oracles (LMOs) to unify different
optimizers. Third, these two views ground a dual-dimension taxonomy, one
dimension assigning each method to a mechanism family and the other recording the
measurable training objectives it aims to improve. Fourth, and at the core of
this paper, we instantiate the full taxonomy in a unified cross-domain benchmark
spanning representative optimizers, model scales, and training regimes from
language model pretraining to image classification, systematically analyzing each
method family across multiple effect objectives and laying out their trade-offs.
OmniOpt thus supplies the research community with an operational
coordinate system for selecting optimizers under explicit mechanism and
objective assumptions, and charts a direction for the future development of the
optimizer community.
}

\metadatainfo{
    \small\makebox[\linewidth][c]{
        \github~\href{https://github.com/OpenRaiser/OmniOpt}{\textbf{Code}} \quad
        \web~\href{https://openraiser.github.io/OmniOpt/}{\textbf{Website}} \quad
        \huggingface~\href{https://huggingface.co/datasets/OpenRaiser/collections}{\textbf{Hugging Face}}
    }
}

\begin{document}


\maketitle

\begin{figure}[H]
  \centering
  \vspace{-0.5em}
  \includegraphics[width=1.0\linewidth]{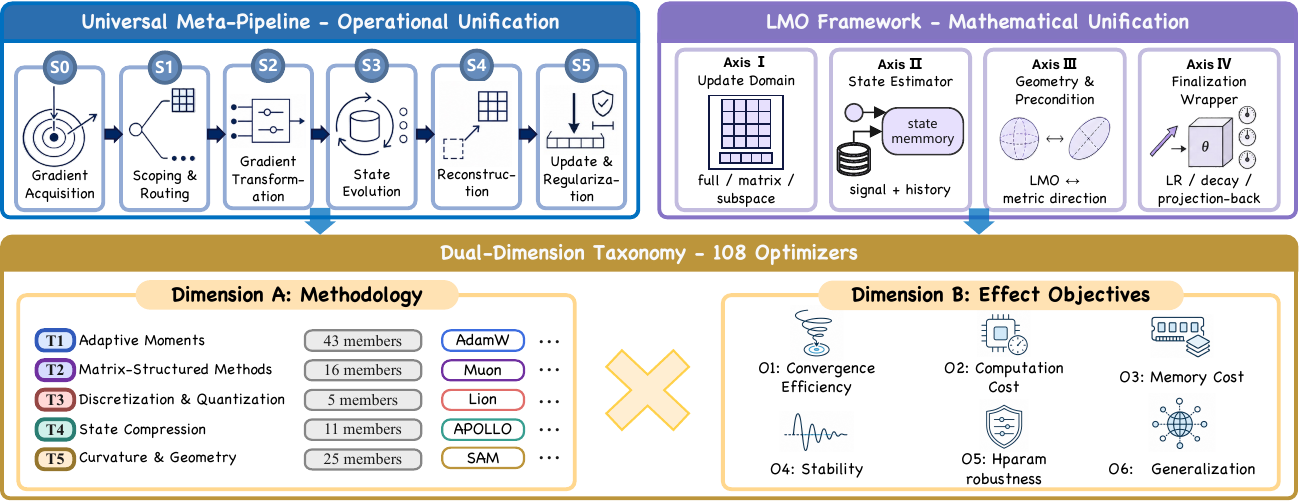}
  \caption{Overview of the proposed survey and benchmark framework for a wild range of optimizers. The paper first introduces a universal meta-pipeline, then develops an LMO-driven four-axis decomposition, builds a process-aligned methodological taxonomy and an effect-objective taxonomy, and finally connects both taxonomies to a large-scale benchmark study.}
  \vspace{-3.5em}
  \label{fig:overview_abstract}
\end{figure}
\blfootnote{$^{*}$Equal contribution.\quad $^{\dagger}$Corresponding authors.}

\clearpage
\tableofcontents
\clearpage

\section{Introduction}
\label{sec:intro}

\subsection{Background and Motivation}
\label{sec:intro-background}

Optimizers have become a system-level design choice in model training.  Whether a modern training run succeeds depends on whether a stochastic update can reduce the empirical loss, and equally on a fixed budget of accelerators, optimizer-state memory, communication bandwidth, batch size, data scale, tuning effort, and downstream evaluation cost.  
Adam and AdamW established the dominant element-wise adaptive-moment template for deep learning~\cite{kingma2015adam,loshchilov2019decoupled}. Recent LLM-oriented optimizers, however, no longer fit cleanly into this single template.  Sign-based methods change the direction map~\cite{chen2023symbolic}, matrix-level methods exploit structured curvature or orthogonalized directions~\cite{gupta2018shampoo}, low-rank methods reduce state in projected subspaces~\cite{zhao2024galore}, and sharpness-aware methods regularize the geometry of the final update~\cite{foret2021sharpness}.  
Optimizer selection has therefore become an empirical, mathematical, and engineering problem at the same time.

This rapidly expanding literature is difficult to navigate for three reasons. First, optimizer papers are usually organized around a local mechanism, such as a new moment estimator, preconditioner, projection, quantizer, or post-update correction. Such naming is useful for implementation, but it does not reveal whether two optimizers intervene at the same stage of the update process, whether their components are orthogonal, or whether they are compositionally compatible.  Second, empirical claims are highly protocol-sensitive.  Recent LLM optimizer studies show that conclusions can change with model scale, batch size, training duration, data-to-model ratio, learning-rate schedule, warmup, weight decay, and hyperparameter tuning budget~\cite{zhao2025deconstructing,semenov2025benchmarking,wen2025fantastic}.
A method that improves pretraining loss may increase memory, wall-clock time, or tuning burden.  Conversely, a memory-saving method may change both the update direction and the state representation. 
Third, the mathematical accounts of modern optimizers remain fragmented.  Element-wise adaptive methods, matrix preconditioners, sign optimizers, low-rank projections, and geometry-aware regularizers are often analyzed with different notation and different assumptions.

\begin{figure}[ht]
  \centering
  \includegraphics[width=1.0\linewidth]{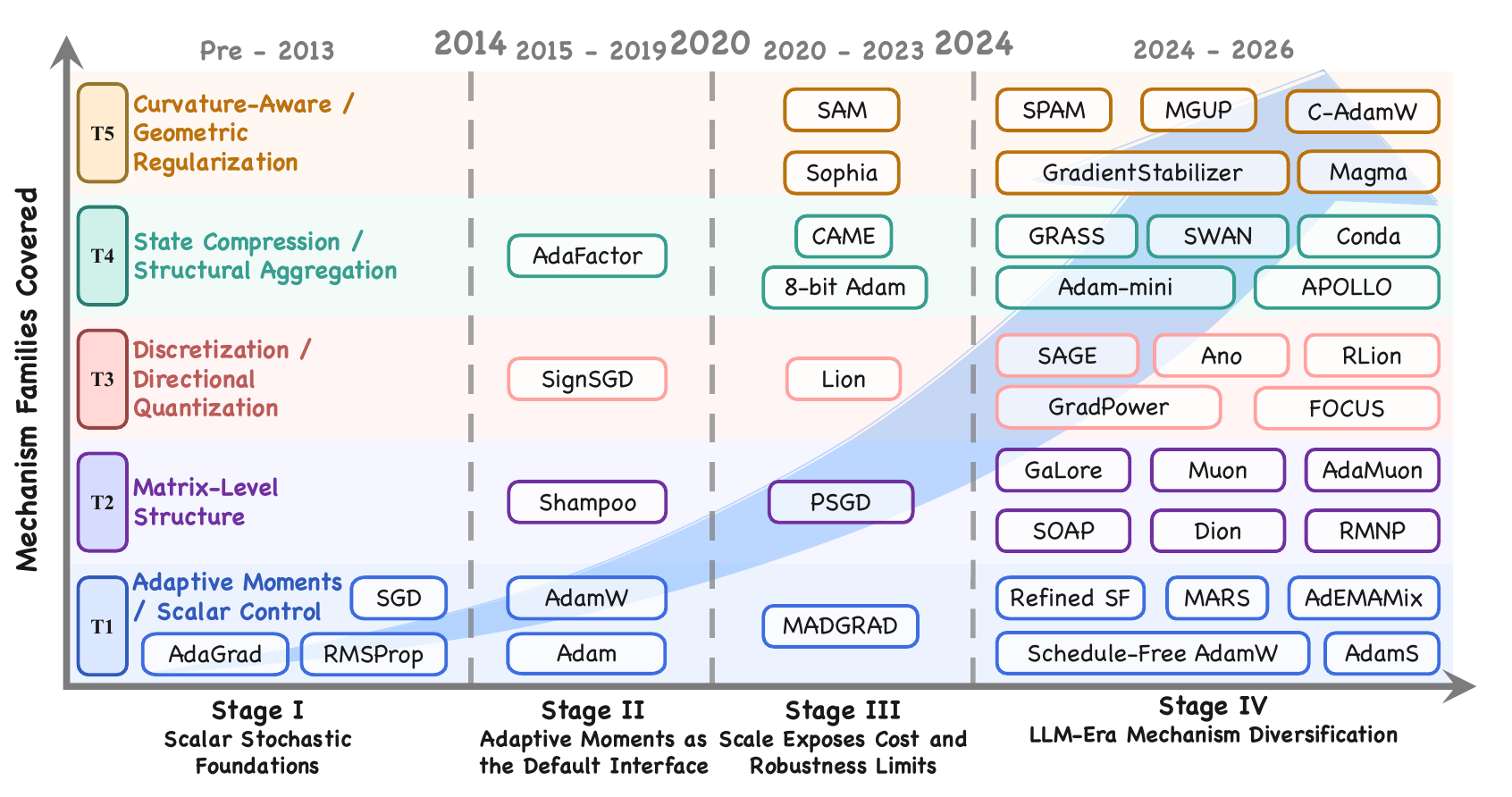}
  \caption{Evolution of optimizer design for deep learning and LLM training. The expanding coverage of T1--T5 mechanism lanes and the diagonal trend arrow indicate increasing method count and mechanism diversity, not a performance ranking or exact method counts.}
  \label{fig:timeline}
\end{figure}

This paper addresses these gaps by treating an optimizer as a structured transformation from a stochastic training signal to a parameter update. The central premise is that most modern optimizers are sparse modifications of a shared meta-pipeline: they route parameters, transform the gradient signal, evolve internal state, reconstruct an update, and finally apply scaling or regularization. This operational view induces a process-aligned methodological taxonomy in which each optimizer family is grouped by the stage where its primary non-identity operation occurs.  On top of this view, we use linear minimization oracles (LMOs) as a geometric primitive for explaining norm-induced update directions, including sign directions, orthogonal matrix directions, and preconditioned directions~\cite{bernstein2024oldoptimizer,pethick2025normlmo}.  The resulting four-axis decomposition extends idealized LMO direction selection to practical optimizer updates by adding curvature estimation, gradient-estimation quality, and state-compression choices. 
A second, effect-oriented taxonomy then records which training objectives each method is designed or expected to improve, such as convergence efficiency, per-step cost, memory consumption, stability, hyperparameter robustness, and generalization. Together, these views form a mechanism-aware framework for answering three linked questions: where an optimizer acts, why its update has a particular mathematical form, and what practical training objective it is meant to improve.

\subsection{Contributions and Paper Organization}
\label{sec:intro-contributions}

This paper is a survey and benchmark study of optimizers for LLM training that aims to provide a mechanism-aware map for comparing, combining, and evaluating optimizers under explicit assumptions.  Which optimizer is best depends on the specific training constraints.  The four components below are intentionally coupled: the meta-pipeline defines non-overlapping operation sites, the methodological taxonomy is derived from those sites, the LMO-driven analysis explains the mathematical direction selected by each family, and the objective taxonomy turns optimizer claims into measurable evaluation axes.  The main contributions are as follows.

\begin{enumerate}[leftmargin=*]
  \item \textbf{A universal meta-pipeline for optimizers.}  We
  formulate a five-step operational pipeline that aligns modern optimizers
  under a common update process, namely parameter scoping and routing, gradient
  transformation, state evolution, update reconstruction, and update
  finalization.  This view exposes an identity-mapping principle, that most
  optimizers perform nontrivial work in only one or two stages while leaving
  the remaining stages as defaults.
  \item \textbf{An LMO-driven four-axis decomposition.}  We use
  norm-constrained LMOs to unify direction selection and extend this view to a
  four-axis description of practical optimizer updates, whose axes specify the
  update domain, the state estimator, the geometry-and-precondition operator,
  and the finalization wrapper.  Under this view, the update directions selected
  by different optimizers are unified by a single geometric language.
  \item \textbf{A dual-dimension taxonomy for more than one hundred
  optimizers.}  We organize optimizer methods along two complementary
  dimensions.  The methodological dimension assigns each optimizer to a
  non-overlapping primary mechanism family, and the objective-oriented dimension
  records the measurable training objectives each method is intended to improve.
  \item \textbf{A mechanism-aligned benchmark study.}  This is the core of the
  paper.  The empirical part instantiates the full taxonomy in a unified
  cross-domain benchmark, spanning language model pretraining from 60M to 1B
  parameters, across four architectures, and at context lengths from 256 to 32k
  tokens, together with image classification on CIFAR100~\cite{krizhevsky2009learning}.  The evaluation is carried
  out at the family and axis level, systematically comparing representative
  optimizers across six effect objectives (convergence, cost, memory, stability,
  hyperparameter robustness, and generalization) and laying out their trade-offs
  one by one.  We further perform a mechanistic ablation of the matrix-structured
  optimizer Muon~\cite{jordan2024muon} to examine how its sub-operations compose across scale and
  architecture.  The benchmark yields several findings that recur throughout the
  paper.  No single optimizer dominates this multi-objective frontier.  Aggressive
  state compression excels under short context but degrades sharply as input
  complexity grows.  Structured-matrix methods transfer most stably across
  architectures and tasks but at substantial per-step cost.  Optimizer rankings
  exhibit systematic crossings with scale, context length, and domain.
\end{enumerate}

Figure~\ref{fig:overview_abstract} summarizes the planned organization of
the paper. The key design is that the meta-pipeline, LMO-driven
decomposition, methodological taxonomy, objective taxonomy, and benchmark
study are mutually aligned and reinforce one another as an integrated whole.


\subsection{Positioning Against Existing Surveys and Benchmarks}
\label{sec:intro-positioning}

\begin{table*}[!t]
  \centering
  \caption{Relation to optimizer surveys and unifying-theory work.}
  \label{tab:related_survey_theory}
  \scriptsize
  \setlength{\tabcolsep}{0.75mm}
  \renewcommand{\arraystretch}{1.22}
  \begin{tabular}{@{}>{\raggedright\arraybackslash}m{0.225\textwidth}>{\raggedright\arraybackslash}m{0.100\textwidth}>{\raggedright\arraybackslash}m{0.170\textwidth}>{\centering\arraybackslash}m{0.075\textwidth}|>{\centering\arraybackslash}m{0.080\textwidth}>{\centering\arraybackslash}m{0.085\textwidth}|>{\centering\arraybackslash}m{0.085\textwidth}>{\centering\arraybackslash}m{0.075\textwidth}@{}}
    \toprule
    \multirow[c]{2}{0.225\textwidth}[-4.0ex]{\centering\arraybackslash\textbf{Work}} &
    \multirow[c]{2}{0.100\textwidth}[-4.0ex]{\centering\arraybackslash\textbf{Type}} &
    \multicolumn{2}{c}{\textbf{Scope}} &
    \multicolumn{2}{c}{\textbf{Mathematical lens}} &
    \multicolumn{2}{c}{\textbf{Evidence and taxonomy}} \\
    \cmidrule(lr){3-4}\cmidrule(lr){5-6}\cmidrule(l){7-8}
    & & \makecell[c]{\textbf{Primary}\\\textbf{object}} &
    \makecell[c]{\textbf{LLM}\\\textbf{optimizer}\\\textbf{coverage}} &
    \makecell[c]{\textbf{Optimizer}\\\textbf{update}\\\textbf{formulas}} &
    \makecell[c]{\textbf{Geometric}\\\textbf{unification}\\\textbf{lens}} &
    \makecell[c]{\textbf{Benchmark}\\\textbf{included}} &
    \makecell[c]{\textbf{Update}\\\textbf{stage}\\\textbf{taxonomy}} \\
    \midrule
    \multicolumn{8}{@{}l}{\textbf{Survey and benchmark papers}} \\
    \midrule
    Evolution Methods~\citep{zhang2026evolution}
    & Survey+bench. & Broad optimizers & \cmark & \cmark & \pmark & \cmark & \xmark \\
    DL Optimizers~\citep{altinel2025development}
    & Survey & DL optimizer history & \pmark & \cmark & \xmark & \xmark & \xmark \\
    LLM Valley~\citep{ranganath2026navigating}
    & Survey & LLM optimizers & \cmark & \pmark & \xmark & \pmark & \xmark \\
    Mem.-Eff. PT~\citep{glentis2025scalable}
    & Survey+bench. & Memory-efficient LLM & \cmark & \pmark & \xmark & \cmark & \xmark \\
    \midrule
    \multicolumn{8}{@{}l}{\textbf{Unifying mathematical theory}} \\
    \midrule
    Old Optimizer, New Norm~\citep{bernstein2024oldoptimizer}
    & Theory & Norm geometry & \pmark & \cmark & \cmark & \xmark & \xmark \\
    Norm-Constrained LMOs~\citep{pethick2025normlmo}
    & Theory+Algo. & LMO training & \xmark & \cmark & \cmark & \pmark & \xmark \\
    Lions/Muons~\citep{sfyraki2025lions}
    & Theory & Lion/Muon geometry & \pmark & \cmark & \cmark & \xmark & \xmark \\
    \midrule
    \rowcolor{oursbg}
    \textbf{Ours}
    & Survey+bench. & 100+ LLM optimizers & \cmark & \cmark & \cmark & \cmark & \cmark \\
    \bottomrule
  \end{tabular}
  \vspace{2pt}
  \begin{minipage}{0.98\textwidth}
    \footnotesize \emph{Note.} \cmark indicates that the criterion is central
    to the work; \pmark indicates partial or secondary coverage; \xmark indicates
    that the criterion is not central to the work. Bench., Algo., DL, and PT
    abbreviate benchmark, algorithm, deep learning, and pretraining.
  \end{minipage}
\end{table*}
\FloatBarrier

\subsubsection{Prior Literature Strands and Survey-Theory Baselines}

The closest prior work falls into three interlocking lines: optimizer surveys,
unifying mathematical theory, and large-language-model optimizer benchmarks.
General optimizer surveys describe the historical progression from stochastic
gradient descent and momentum to adaptive, second-order, memory-efficient, and
matrix-aware methods.  Theoretical work supplies a complementary explanation of
why some updates can be viewed as norm-induced steepest descent, stochastic
Frank--Wolfe steps, or directions produced by a linear minimization oracle.
Empirical benchmarks then show how these mechanisms behave under realistic
pretraining or fine-tuning protocols.  Taken separately, however, these lines do
not identify where a method intervenes in the optimizer update pipeline, which
mechanisms are compatible with one another, or how a taxonomy should determine
the axes of a benchmark.

Table~\ref{tab:related_survey_theory} places representative surveys and
unifying-theory papers along three groups of criteria.  The scope columns ask
what object each work primarily studies and whether large-language-model
optimizers are central to its scope.  The mathematical-lens columns distinguish
formula-level treatment of optimizer updates from geometric interpretations
through norms, linear minimization oracles, or Frank--Wolfe geometry.  The final
columns record whether a work includes benchmark evidence and whether it
organizes optimizers by the stage at which they intervene in an update.  In
Tables~\ref{tab:related_survey_theory} and~\ref{tab:related_benchmarks},
\cmark, \pmark, and \xmark indicate primary coverage, partial or secondary
coverage, and a non-central topic, respectively.

The broad surveys by Zhang et al.~\citep{zhang2026evolution} and Altinel and
Ozcan~\citep{altinel2025development} are useful references for update rules,
optimizer families, and the historical development of deep-learning
optimization, but their organization is mainly driven by gradient-information
order, chronology, or application setting.  Ranganath et
al.~\citep{ranganath2026navigating} move closer to the setting of this paper by
covering adaptive, sign-based, memory-efficient, low-rank, and matrix-based
optimizers for large language models.  Glentis et
al.~\citep{glentis2025scalable} focus more narrowly on parameter- and
memory-efficient pretraining, which is especially relevant to optimizer-state
memory and low-rank training.  The unifying-theory papers play a different
role.  Bernstein et al.~\citep{bernstein2024oldoptimizer} connect several
optimizers to architecture-dependent norms, Pethick et
al.~\citep{pethick2025normlmo} develop training algorithms based on
norm-constrained linear minimization oracles, and Sfyraki et
al.~\citep{sfyraki2025lions} relate Lion and Muon to stochastic Frank--Wolfe.
These works supply the geometric language used later in this paper, but they do
not provide a survey-scale taxonomy that maps optimizer mechanisms to update
stages and then connects those stages to benchmark objectives.

\subsubsection{Benchmark Evidence and Taxonomy Mapping}

The second group consists of large-language-model optimizer benchmarks and
mechanistic empirical studies.  Their common message is that optimizer
comparison is not a stable global ranking problem.  Adam-like methods, sign
methods, matrix-aware methods, curvature-aware methods, and memory-efficient
variants can exchange relative positions when the tuning budget, model size,
batch size, training duration, data regime, or evaluation endpoint changes.
This makes it necessary to report effect-specific trade-offs, including
convergence, wall-clock cost, optimizer-state memory, robustness to
hyperparameters, and downstream generalization.

Table~\ref{tab:related_benchmarks} summarizes this empirical literature by
training setting, method coverage, and evaluation focus.  The method-coverage
columns use the Dimension-A families introduced in
Section~\ref{sec:taxonomy}: T1 element-wise adaptive moment and scalar control,
T2 matrix-level structural methods, T3 discretization and directional
quantization, T4 state compression and structural aggregation, and T5
curvature-aware and geometric regularization.  Coverage is assigned by mapping the
optimizers actually benchmarked in each work to their primary mechanism labels
according to what each optimizer actually does.  Thus Lion and Signum
count as T3 rather than T4 even when discussed as memory-saving methods, whereas
AdaFactor and Adam-mini count as T4 because they change optimizer-state
representation.  Zeroth-order fine-tuning is treated as a boundary case: the
signal-acquisition mechanism is outside the main T1--T5 taxonomy, but ZO-Adam
and ZO-SGD-Sign can still be recorded as T1-like and T3-like update analogues.
The evaluation columns distinguish hyperparameter tuning, protocol variation,
cost accounting, downstream evaluation, and mechanistic analysis.  Here
\emph{protocol variation} includes model size, batch size, training length,
data-to-model ratio, architecture, training stage, and task setting, while
\emph{cost accounting} includes compute, memory, runtime, and query cost.  The
row for this paper describes its benchmark design, and the actual experimental
results appear in the main text.
\begin{table*}[!t]
  \centering
  \caption{Relation to large-language-model optimizer benchmarks and mechanistic empirical
  studies.}
  \label{tab:related_benchmarks}
  \scriptsize
  \setlength{\tabcolsep}{0.4mm}
  \renewcommand{\arraystretch}{1.14}
  \begin{tabular}{@{}>{\raggedright\arraybackslash}m{0.230\textwidth}>{\raggedright\arraybackslash}m{0.100\textwidth}|*{5}{>{\centering\arraybackslash}m{0.070\textwidth}}|*{5}{>{\centering\arraybackslash}m{0.047\textwidth}}@{}}
    \toprule
    \multirow[c]{2}{0.230\textwidth}[-1.5ex]{\centering\arraybackslash\textbf{Work}} &
    \multicolumn{1}{>{\centering\arraybackslash}m{0.100\textwidth}}{\multirow[c]{2}{0.100\textwidth}[-1.5ex]{\centering\arraybackslash\shortstack{\textbf{Training}\\\textbf{setting}}}} &
    \multicolumn{5}{c}{\textbf{Method coverage}} &
    \multicolumn{5}{c}{\textbf{Evaluation focus}} \\
    \cmidrule(lr){3-7}\cmidrule(l){8-12}
    & \multicolumn{1}{>{\raggedright\arraybackslash}m{0.100\textwidth}}{} & \shortstack{\textbf{Adaptive}\\\textbf{moment}} & \shortstack{\textbf{Matrix}\\\textbf{methods}} & \shortstack{\textbf{Direction}\\\textbf{discre-}\\\textbf{tization}} & \shortstack{\textbf{State}\\\textbf{compres-}\\\textbf{sion}} & \multicolumn{1}{>{\centering\arraybackslash}m{0.070\textwidth}}{\shortstack{\textbf{Geometry}\\\textbf{regular-}\\\textbf{ization}}} & \textbf{Tune} & \shortstack{\textbf{Proto-}\\\textbf{col}} & \textbf{Cost} & \shortstack{\textbf{Down-}\\\textbf{stream}} & \shortstack{\textbf{Mech-}\\\textbf{anism}} \\
    \midrule
    \mbox{Deconstructing Optimizers~\citep{zhao2025deconstructing}}
    & \mbox{PT} & \cmark & \xmark & \cmark & \cmark & \cmark
    & \cmark & \cmark & \pmark & \xmark & \cmark \\
    \mbox{Benchmarking LLM Opt.~\citep{semenov2025benchmarking}}
    & \mbox{PT} & \cmark & \cmark & \cmark & \xmark & \cmark
    & \cmark & \cmark & \pmark & \xmark & \cmark \\
    \mbox{Fantastic PT Opt.~\citep{wen2025fantastic}}
    & \mbox{PT} & \cmark & \cmark & \cmark & \cmark & \cmark
    & \cmark & \cmark & \cmark & \xmark & \pmark \\
    \mbox{Budgeted LLM PT~\citep{schlotthauer2025budget}}
    & \mbox{Budget PT} & \cmark & \xmark & \cmark & \xmark & \cmark
    & \pmark & \pmark & \cmark & \cmark & \pmark \\
    \mbox{ZO FT Benchmark~\citep{zhang2024zerothorder}}
    & \mbox{ZO FT} & \pmark & \xmark & \pmark & \xmark & \xmark
    & \pmark & \pmark & \cmark & \cmark & \pmark \\
    \mbox{Muon Fine-tuning Study~\citep{qu2026muonfinetune}}
    & \mbox{FT} & \cmark & \cmark & \xmark & \xmark & \xmark
    & \pmark & \pmark & \xmark & \cmark & \cmark \\
    \mbox{Muon Spectral Study~\citep{shumaylov2026muonnot}}
    & \mbox{Mech.} & \pmark & \cmark & \xmark & \xmark & \xmark
    & \xmark & \xmark & \xmark & \xmark & \cmark \\
    \rowcolor{oursbg}
    \textbf{Ours}
    & \mbox{PT} & \cmark & \cmark & \cmark & \cmark & \cmark
    & \cmark & \cmark & \cmark & \cmark & \cmark \\
    \bottomrule
  \end{tabular}
  \vspace{2pt}
  \begin{minipage}{0.98\textwidth}
    \footnotesize \emph{Note.} \cmark indicates explicit coverage of the
    family; \pmark indicates analogue, boundary, or baseline-only coverage;
    \xmark indicates no coverage under these labels.  Opt., PT, ZO, FT, and
    Mech. abbreviate optimizers, pretraining, zeroth-order, fine-tuning, and
    mechanistic, respectively; Budget PT denotes budgeted pretraining.
  \end{minipage}
\end{table*}
\FloatBarrier

\subsubsection{Position of This Paper and Roadmap}

Zhao et al.~\citep{zhao2025deconstructing} and Semenov et
al.~\citep{semenov2025benchmarking} are the closest pretraining benchmarks to
our setting.  Both compare Adam-like baselines with sign, curvature-aware, or
memory-efficient variants, and both show that conclusions depend on
hyperparameter sweeps and protocol choices.  Wen et
al.~\citep{wen2025fantastic} sharpen this point by revisiting claimed optimizer
speedups under stricter hyperparameter search, end-of-training evaluation,
hyperparameter transfer, and scaling conditions.  Schlotthauer et
al.~\citep{schlotthauer2025budget} focus on resource-limited pretraining, where
the optimizer affects not only loss but also graphics-processing-unit hours and
downstream quality.  Zhang et al.~\citep{zhang2024zerothorder} study
zeroth-order fine-tuning, which marks a boundary case for our taxonomy, since
its central innovation is how the training signal is acquired, whereas our main
taxonomy begins after a signal enters the optimizer update.  The Muon
studies~\citep{qu2026muonfinetune,shumaylov2026muonnot} expose a different
boundary by showing that matrix-structured directions interact with
pretraining-versus-fine-tuning stage, low-rank adaptation subspaces, spectral
exponents, and local descent geometry.  Together, these studies motivate
benchmark reporting by method family and by objective rather than by a single
aggregate rank.

This paper connects these three lines through a mechanism-aware organization.
The universal meta-pipeline specifies where an optimizer changes a training
update.  The LMO-driven four-axis decomposition specifies how the update
direction, curvature estimate, gradient signal, and state representation are
formed.  The dual-dimension taxonomy specifies how optimizers should be grouped
and which training effects should be measured.  The benchmark component is
therefore not an isolated leaderboard but an empirical test of whether the
taxonomy separates mechanisms that lead to distinct convergence, cost, memory,
stability, tuning, and generalization behavior.

The remainder of the paper follows the same progression from foundations to
evaluation.  Section~\ref{sec:prelim} introduces the LLM training problem,
classical adaptive optimization, and notation.
Section~\ref{sec:unified-theoretical-framework} presents the unified
theoretical framework of this work.  Subsection~\ref{sec:meta-pipeline} introduces the
universal meta-pipeline, its five operational stages, and representative
instantiations organized around the identity-mapping principle, while
Subsection~\ref{sec:lmo-four-axis} develops the LMO-driven four-axis
decomposition as a mathematical explanation of practical optimizer updates.
Section~\ref{sec:taxonomy} introduces the dual-dimension taxonomy and explains
how the method dimension is aligned with pipeline operations while the
objective dimension captures effect targets.
Section~\ref{sec:optimizer-families} consolidates the major optimizer families,
with Subsections~\ref{sec:t1}--\ref{sec:t5} surveying T1--T5 through the
meta-pipeline, LMO-driven coordinates, and objective-oriented assessments.
Section~\ref{sec:benchmark} reports the benchmark study.  Sections~\ref{sec:discussion} and~\ref{sec:conclusion}
discuss limitations, open problems, and conclusions.
\section{Preliminaries}
\label{sec:prelim}

This section lays the groundwork for the rest of the paper and fixes the
optimization framework, model-structural conventions, and mathematical notation
used in the analysis that follows.  Section~\ref{sec:prelim-formulation} first
formalizes large-language-model training as a stochastic optimization problem
and clarifies its characteristic parameter topology.  Building on this,
Section~\ref{sec:prelim-background} reviews the core mechanisms of classical
optimization methods, ranging from SGD and momentum to adaptive moment
estimation.  It also introduces the preconditioning perspective behind the
Newton method, the natural gradient, and Fisher or Hessian curvature, so that the
later sections have the background they need to develop the unified theoretical
framework.

\subsection{Problem Formulation}
\label{sec:prelim-formulation}

\paragraph{Training objective and stochastic gradients.}
Let $\mathcal{D}=\{(\mathbf{x}_i,\mathbf{y}_i)\}_{i=1}^N$ denote a training
corpus and let $\theta\in\mathbb{R}^d$ collect all trainable parameters of a
language model.  Pretraining and supervised fine-tuning both reduce to empirical
risk minimization,
\begin{equation}
  \min_{\theta\in\mathbb{R}^d}\;\mathcal{L}(\theta)
  = \frac{1}{N}\sum_{i=1}^{N}\ell(\theta;\,\mathbf{x}_i,\mathbf{y}_i),
  \label{eq:objective}
\end{equation}
where $\ell$ is usually a token-level cross-entropy or an instruction-tuning
loss derived from it. At step $t$, the optimizer receives a mini-batch
gradient,
\begin{equation}
  g_t = \frac{1}{|\mathcal{B}_t|}\sum_{i\in\mathcal{B}_t}
        \nabla_\theta\ell(\theta_t;\,\mathbf{x}_i,\mathbf{y}_i),
  \label{eq:mbgrad}
\end{equation}
where $\mathcal{B}_t\subset\{1,\ldots,N\}$.  When a stochastic-oracle model is
needed, we write $\mathbb{E}[g_t\mid\theta_t]=\nabla\mathcal{L}(\theta_t)$ and
$\mathbb{E}[\|g_t-\nabla\mathcal{L}(\theta_t)\|^2]\le\sigma^2$.  In practice,
$g_t$ reflects data sampling, sequence packing, distributed reduction, mixed
precision, gradient accumulation, clipping, and loss scaling.  These system
details are not separate from optimizer behavior.  In fact, they are what
determine the noise level, numerical range, and memory traffic seen by the
update rule.

\paragraph{Transformer parameter topology.}
Transformer-style LLMs~\cite{vaswani2017attention} contain heterogeneous
parameter tensors.  For optimizer design, the most important distinction is not
the layer name but the tensor geometry.  Two-dimensional matrices
$W\in\mathbb{R}^{m\times n}$, such as attention projections
$W_Q,W_K,W_V,W_O$, feed-forward projections, and embedding tables, admit
row-column operations including singular-value decompositions, Kronecker
factorization, matrix orthogonalization, and low-rank projection.  Vector-like
parameters, such as biases and normalization gains or shifts, do not have the
same matrix geometry and are usually routed to element-wise rules or excluded
from weight decay.  Some methods introduce finer scopes, e.g., attention-head
blocks, per-layer groups, or module-specific rules.  This parameter topology is
the basis for the scoping-and-routing stage of the universal meta-pipeline in
Section~\ref{sec:meta-pipeline}.

\paragraph{Memory budget under mixed-precision training.}
Under BF16 forward and backward computation with FP32 master weights, the
training memory can be estimated as
\begin{equation}
  M_{\mathrm{train}}
  \approx 4d + 2d + 2d + S_{\mathrm{opt}}d + M_{\mathrm{act}}.
  \label{eq:memory-budget}
\end{equation}
The first three terms correspond to FP32 master parameters, BF16 model copies,
and BF16 gradients, $S_{\mathrm{opt}}$ is the optimizer-state footprint in bytes
per parameter, and $M_{\mathrm{act}}$ denotes activation and temporary-buffer
memory.  Adam stores two FP32 state tensors $m_t$ and $v_t$, so each parameter
carries an extra 8 bytes.  A 7B-parameter model therefore needs about 56\,GB for
Adam states alone, before activations, temporary buffers, and distributed-training
overheads.  State-compressed methods in Section~\ref{sec:t4} are therefore not
merely engineering variants.  Under a fixed hardware budget, reducing
$S_{\mathrm{opt}}$ changes the feasible model size, batch size, and tuning regime.

\subsection{Classical Optimization Background}
\label{sec:prelim-background}

The modern LLM optimizer literature can be read as a sequence of modifications
to a small set of classical design choices.  These choices are whether to
smooth gradients, whether to maintain a curvature proxy, how to scale the
direction, and how to regularize the final write.  The following review fixes the baseline equations
used by later sections.

\paragraph{Newton method and the perspective of preconditioning.}
A natural starting point for our subsequent preconditioning analysis is the Newton method. Near the current iterate, a second-order Taylor expansion yields:
\begin{equation}
  \mathcal{L}(\theta_t+\delta)
  \approx
  \mathcal{L}(\theta_t)
  + g_t^\top\delta
  + \frac{1}{2}\delta^\top H_t\delta,
  \quad
  H_t=\nabla^2\mathcal{L}(\theta_t).
  \label{eq:newton-quadratic}
\end{equation}
Minimizing this local quadratic model gives the damped Newton update
\begin{equation}
  \theta_{t+1}
  = \theta_t - \eta_t H_t^{-1}g_t,
  \label{eq:newton-update}
\end{equation}
where the scalar $\eta_t$ absorbs damping, line search, or trust-region control.  This equation gives the classical preconditioning viewpoint, since the optimizer does not only choose a gradient signal but also chooses the geometry in which that signal is scaled.  SGD can be read as the first-order simplification that replaces $H_t^{-1}$ by the identity.  Modern large-model optimizers replace the exact inverse Hessian with tractable approximations or geometry-induced direction operators, where Adam uses a diagonal second-moment or Fisher-like proxy, Shampoo and SOAP use structured matrix factors or eigenbases, and Muon uses a polar or Gram-induced matrix direction.  This connection motivates the preconditioned and matrix-structured views used throughout the rest of the survey.

\paragraph{SGD and momentum.}
Stochastic gradient descent (SGD) writes the raw mini-batch gradient into the
parameters:
\begin{equation}
  \theta_{t+1} = \theta_t - \eta_t\,g_t.
  \label{eq:sgd}
\end{equation}
Momentum replaces the instantaneous gradient with an exponential moving
average (EMA):
\begin{equation}
  m_t = \beta_1\,m_{t-1} + (1-\beta_1)\,g_t,\quad
  \theta_{t+1} = \theta_t - \eta_t\,m_t.
  \label{eq:sgdm}
\end{equation}
SGD and momentum use one global scale, or a schedule of global scales, for all
coordinates.  They therefore do not adapt to coordinate-wise gradient
magnitudes or curvature proxies.
In the notation of Eq.~\eqref{eq:newton-update}, SGD corresponds to the
identity preconditioner, $H_t=I$, so $H_t^{-1}g_t=g_t$.  Momentum keeps the
same identity geometry but replaces the instantaneous signal $g_t$ with the
smoothed signal $m_t$.

\paragraph{AdaGrad and RMSProp.}
AdaGrad~\cite{duchi2011adaptive} introduced a diagonal, history-dependent
preconditioner by accumulating squared gradients:
\begin{equation}
  G_t = \sum_{\tau=1}^{t}g_\tau^2,\quad
  \theta_{t+1}
  = \theta_t - \eta_t\,\frac{g_t}{\sqrt{G_t}+\epsilon}.
  \label{eq:adagrad}
\end{equation}
Here and below, powers and divisions are element-wise unless stated otherwise.
The monotone accumulator is useful for sparse features but can make effective
learning rates decay too aggressively in long training runs.  RMSProp
~\cite{tieleman2012rmsprop} replaces the cumulative sum with an EMA:
\begin{equation}
  v_t = \beta_2\,v_{t-1}+(1-\beta_2)\,g_t^2,\quad
  \theta_{t+1}
  = \theta_t - \eta_t\,\frac{g_t}{\sqrt{v_t}+\epsilon}.
  \label{eq:rmsprop}
\end{equation}
Equivalently, AdaGrad uses the effective diagonal metric
$H_t^{\mathrm{AdaGrad}}=\operatorname{diag}(\sqrt{G_t}+\epsilon)$, whereas
RMSProp uses $H_t^{\mathrm{RMSProp}}=\operatorname{diag}(\sqrt{v_t}+\epsilon)$.
Both should be read as cheap diagonal preconditioners, that is, approximate
proxies for Fisher or Hessian curvature.

\paragraph{Adam.}
Adam~\cite{kingma2015adam} combines momentum and RMSProp-style second moments,
then corrects the initialization bias of both EMAs:
\begin{align}
  m_t &= \beta_1\,m_{t-1}+(1-\beta_1)\,g_t,\quad v_t = \beta_2\,v_{t-1}+(1-\beta_2)\,g_t^2,
  \label{eq:adam-moments}\\
  \hat{m}_t &= m_t/(1-\beta_1^t),\quad \hat{v}_t = v_t/(1-\beta_2^t),
  \label{eq:adam-bias}\\
  \theta_{t+1} &= \theta_t - \eta_t\,\hat{m}_t/(\sqrt{\hat{v}_t}+\epsilon).
  \label{eq:adam-update}
\end{align}
The diagonal preconditioner of Adam remains the reference point for most LLM
optimizers.  Later methods may alter the moment time scale, replace the
gradient estimator, change the direction map, compress the state, or add a
wrapper around the final update.  Nevertheless, they are often compared against
this Adam-like template.
In the same preconditioning notation, Adam uses the effective signal
$\hat{m}_t$ and the diagonal metric
$H_t^{\mathrm{Adam}}=\operatorname{diag}(\sqrt{\hat{v}_t}+\epsilon)$, so the
preconditioned direction is $H_t^{-1}\hat{m}_t
=\hat{m}_t/(\sqrt{\hat{v}_t}+\epsilon)$.

\paragraph{AdamW: decoupled weight decay.}
In Adam, adding $\ell_2$ regularization to the loss causes the regularization
gradient to be divided by the adaptive denominator. AdamW
~\cite{loshchilov2019decoupled} decouples weight decay from this adaptive
scaling:
\begin{equation}
  \theta_{t+1} = \theta_t
    - \eta_t\,\frac{\hat{m}_t}{\sqrt{\hat{v}_t}+\epsilon}
    - \eta_t\,\lambda\,\theta_t .
  \label{eq:adamw}
\end{equation}
This change is small at the formula level but important as a baseline convention: throughout the survey, an ``Adam-like'' optimizer means a method whose main state evolution is inherited from AdamW unless stated otherwise.
AdamW retains the Adam diagonal preconditioner $H_t^{\mathrm{Adam}}$ for data-gradient components, moving weight decay outside the preconditioned update direction and incorporating it directly in the final parameter assignment step.

\paragraph{Natural gradient and the Fisher information matrix.}
The natural gradient~\cite{amari1998natural} motivates many preconditioned
updates.  For a parameterized distribution $p_\theta(z)$, the Fisher
information matrix is:
\begin{equation}
  F(\theta)=\mathbb{E}_{z\sim p_\theta}\!\left[
    \nabla_\theta\log p_\theta(z)\,
    \nabla_\theta\log p_\theta(z)^\top
  \right].
  \label{eq:fisher-matrix}
\end{equation}
The corresponding update
$\theta_{t+1}=\theta_t-\eta_t F(\theta_t)^{-1}\nabla\mathcal{L}(\theta_t)$
uses the geometry of the predictive distribution, whereas ordinary gradient
descent uses the Euclidean geometry of raw parameters.  Exact storage and inversion of $F$ are infeasible
for modern LLMs, so practical optimizers use diagonal, block-diagonal,
Kronecker, low-rank, or stochastic approximations.
Thus, natural gradient follows the same Newton-style template, but it
replaces $H_t$ with the Fisher metric $F(\theta_t)$.  This makes it a useful
bridge to later matrix and curvature-aware methods, which differ mainly in how
they approximate, factor, or simplify this metric.

\paragraph{Gradient covariance and curvature.}
For negative log-likelihood objectives, the outer product of score gradients
defines the Fisher matrix.  Under standard regularity conditions, the gradient
covariance, the Fisher matrix, and the Hessian are approximately equal, as
given by
\begin{equation}
  \mathbb{E}\!\left[g_t g_t^\top\mid\theta_t\right]
  \approx F(\theta_t),
  \quad
  F(\theta_t) \approx H(\theta_t)=\nabla^2\mathcal{L}(\theta_t).
  \label{eq:fisher-identity}
\end{equation}
This distinction matters for a survey.  The Adam $v_t=\mathrm{EMA}(g_t^2)$ serves as a diagonal gradient-covariance estimate.  The Shampoo factors $L_t=\mathrm{EMA}(G_tG_t^\top)$ and $R_t=\mathrm{EMA}(G_t^\top G_t)$ form a Kronecker-style curvature proxy~\cite{gupta2018shampoo}, with $H_t^{\mathrm{Shampoo}}\approx L_t^{1/4}\otimes R_t^{1/4}$. Sophia estimates a clipped diagonal Hessian using stochastic second-order information~\cite{liu2023sophia}.  Section~\ref{sec:lmo-four-axis} turns these choices into Axis~II (state and curvature estimator) and Axis~III
(geometry and precondition operator) of the LMO-driven four-axis decomposition.

\paragraph{Notation.}
Table~\ref{tab:notation} collects the mathematical symbols used throughout
the paper and closes the preliminary setup before the meta-pipeline is
introduced.

\begin{table}[!t]
  \centering
  \caption{Summary of the mathematical notations used throughout this survey, including optimization variables, matrix operators, second-order quantities, and framework-specific symbols. The table serves as a reference for the formulations presented in the following sections.}
  \label{tab:notation}
  \footnotesize
  \setlength{\tabcolsep}{2pt}
  \begin{tabularx}{\textwidth}{@{}>{\raggedright\arraybackslash}p{0.17\textwidth}>{\raggedright\arraybackslash}X|>{\raggedright\arraybackslash}p{0.17\textwidth}>{\raggedright\arraybackslash}X@{}}
    \toprule
    \textbf{Symbol} & \textbf{Meaning} &
    \textbf{Symbol} & \textbf{Meaning} \\
    \midrule

    $d$ & Total number of model parameters &
    $W_t\in\mathbb{R}^{m\times n}$ & Weight matrix \\

    $\theta_t\in\mathbb{R}^{d}$ & Parameter vector &
    $G_t\in\mathbb{R}^{m\times n}$ & Gradient matrix \\

    $\mathcal{L}(\theta)$ & Training loss &
    $M_t\in\mathbb{R}^{m\times n}$ & Matrix momentum \\

    $g_t$ & Mini-batch stochastic gradient &
    $W=U\Sigma V^\top$ & Singular value decomposition \\

    $m_t$ & First-order momentum &
    $\langle A,B\rangle=\mathrm{tr}(A^\top B)$ & Matrix inner product \\

    $v_t$ & Second-order moment estimate &
    $\odot$ & Hadamard product \\

    $\eta_t$ & Learning rate &
    $\|\cdot\|_p$ & Vector $\ell_p$ norm \\

    $\beta_1,\beta_2$ & EMA decay coefficients &
    $\|\cdot\|_F$ & Frobenius norm \\

    $\lambda$ & Weight-decay coefficient &
    $\|\cdot\|_{\mathrm{op}}$ & Spectral norm \\

    $\epsilon$ & Numerical stability constant &
    $\|\cdot\|_\ast$ or $\|\cdot\|_*$ & Nuclear or dual norm \\

    $F(\theta)$ & Fisher information matrix &
    $Q_L,Q_R$ & Axis-I coordinate bases \\

    $H(\theta)=\nabla^2\mathcal{L}(\theta)$ & Hessian matrix &
    $\Phi_t$ & Axis-III direction operator \\

    $\hat{H}_t$ & Axis-II curvature proxy &
    $\alpha\in[0,1]$ & Preconditioner exponent \\

    $\mathrm{EMA}_\beta(x_t)$ & Exponential moving average &
    $\mathrm{lmo}_{\mathcal{C}}(s)$ & Linear minimization oracle \\

    $\tilde{G}_t$ & Variance-reduced gradient estimate &
    $\rho^{(i)}$ & Meta-pipeline routing label \\

    $[\cdot]^\sharp$ & Dual map &
    $T_1$--$T_5$ & Dimension-A method families \\

    && $O_1$--$O_6$ & Dimension-B effect objectives \\

    \bottomrule
  \end{tabularx}
\end{table}

\section{Unified Theoretical Framework for Optimizers}
\label{sec:unified-theoretical-framework}

\subsection{Universal Meta-Pipeline}
\label{sec:meta-pipeline}

\subsubsection{Operational Steps}
\label{sec:meta-pipeline-steps}

Modern LLM optimizers are usually named after their most visible local
mechanism: an adaptive moment estimator, a matrix preconditioner, a low-rank
projection, a sign map, a compressed state, or a sharpness-aware correction.
Such descriptions are useful for implementation, but they make cross-method
comparison difficult because they do not say \emph{where} the mechanism enters
the update.  We therefore introduce a \emph{Universal Meta-Pipeline}: a
process-level abstraction that aligns the optimizers considered in this survey
to a common single-step update template.  Its purpose is to expose the
operation site of each method, to
separate primary mechanisms from defaults, and to prepare the bridge to the
LMO-driven geometric view in Section~\ref{sec:lmo-four-axis}.

\begin{figure}[t]
  \centering
  \includegraphics[width=1.0\linewidth]{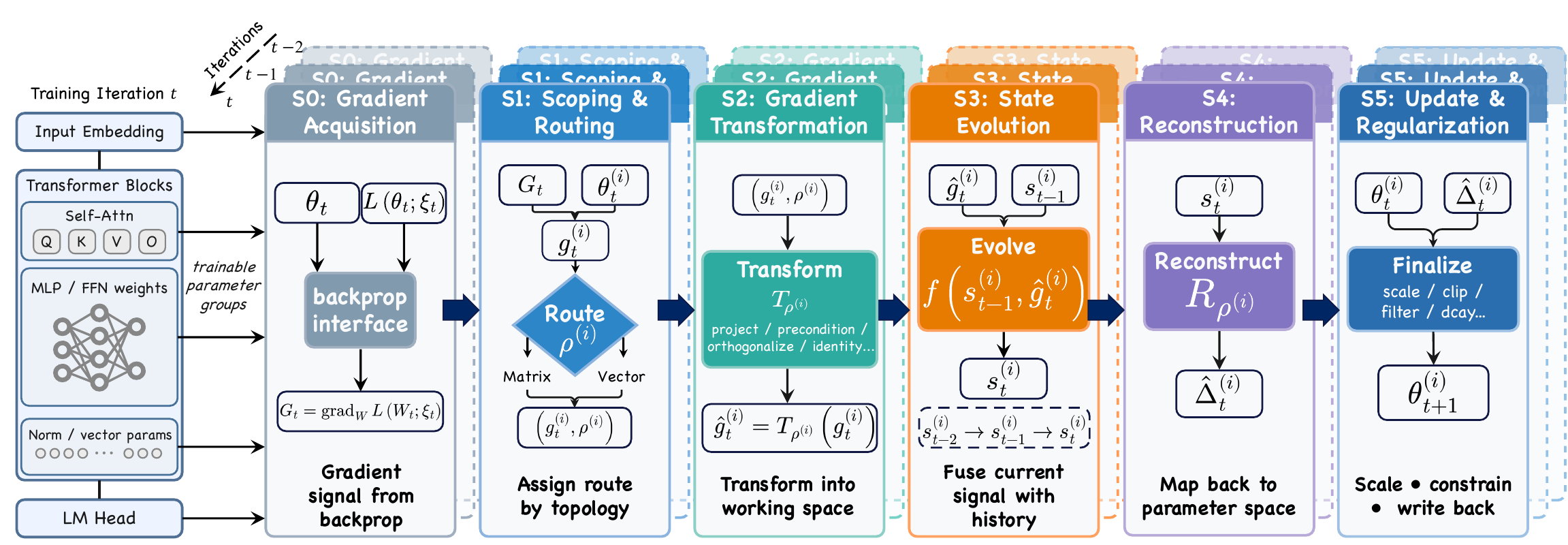}
  \caption{Universal meta-pipeline for one optimizer step.  The training system
  provides a gradient or higher-order signal (S0), which the optimizer
  processes through five internal stages for each parameter group.}
  \label{fig:meta_pipeline}
\end{figure}

\paragraph{Formal setup.}
Let $W_t\in\mathbb{R}^{m\times n}$ be a two-dimensional parameter tensor,
$G_t\in\mathbb{R}^{m\times n}$ its training signal, and
$\mathcal{S}_{t-1}$ the optimizer state before the update.  Five internal
operators $\mathbf{S1}$--$\mathbf{S5}$, each of which may degenerate to the
identity, transform $G_t$ into a signed parameter increment $\Delta_t$:
\begin{equation}
  \Delta_t =
  \mathbf{S5}\!\left(
    \mathbf{S4}\!\left(
      \mathbf{S3}\!\left(
        \mathbf{S2}\!\left(\mathbf{S1}(G_t)\right);\,\mathcal{S}_{t-1}
      \right);\,\mathcal{S}_{t-1}
    \right);\,\mathcal{S}_{t-1},\,W_t
  \right),\quad
  W_{t+1}=W_t+\Delta_t .
  \label{eq:meta_pipeline}
\end{equation}
The internal state $\mathcal{S}_t$ is updated within $\mathbf{S3}$ and is made
available to all downstream stages.  One-dimensional parameters (biases,
normalization coefficients, and other vector-like tensors) follow the same
pipeline, with matrix operations at $\mathbf{S2}$ and $\mathbf{S4}$ contracting
to element-wise equivalents or identity maps.  Figure~\ref{fig:meta_pipeline}
illustrates the pipeline for one training step.

\paragraph{S0: Training Signal Acquisition.}
Before the internal optimizer stages begin, the training system supplies a
signal.  We write this interface as S0 because it determines what information
the update pipeline is allowed to use.  The standard case is
\emph{first-order backpropagation} (FO),
$G_t=\nabla_W\mathcal{L}(W_t;\xi_t)$.  Some optimizers replace $G_t$ with a
\emph{variance-reduced} estimator, for example a STORM-style correction
$\tilde{G}_t=G_t-G_{t-1}^{\xi_t}+M_{t-1}$.  Others add
\emph{curvature-augmented} signals through Hessian-vector products or
Hutchinson diagonal estimates,
$h_t\approx u\odot (Hu)$ with $u\sim\mathcal{N}(0,I)$, as in
Sophia~\cite{liu2023sophia}.  Zeroth-order methods are treated as a boundary
case in this paper: their main contribution is often signal acquisition from
function values rather than the downstream update transformation, so they are
not part of the main T1--T5 update-mechanism taxonomy.

\paragraph{S1: Parameter Scoping and Routing.}
The first internal stage partitions the full parameter set
$\{\theta^{(i)}_{t-1}\}$ and assigns a routing label $\rho^{(i)}$ to each
group based on tensor topology and module type:
\begin{equation}
  \rho^{(i)} = \mathcal{R}\!\bigl(
    \mathrm{shape}(\theta^{(i)}),\;
    \mathrm{module\text{-}type}(\theta^{(i)})
  \bigr).
  \label{eq:routing}
\end{equation}
Equivalently, S1 outputs a partition
$\mathcal{P}=\{P_k\}_{k=1}^{K}$ together with the processing rule attached to
each group.  All downstream stages condition on $\rho^{(i)}$.  The common
partition separates two-dimensional matrices---which admit matrix-geometric
operations such as SVD decomposition, Kronecker factorization, and low-rank
projection---from vector-like parameters handled by element-wise rules.  Finer
partitions include attention-head blocks, per-layer groups, and
topology-dependent hybrid routes.  When all parameters share a single route,
S1 is an identity, as in standard AdamW.  When routing is nontrivial, the same
optimizer may apply different downstream operators to matrix, vector, or
module-specific parameter groups.

\paragraph{S2: Gradient Transformation.}
This stage applies a structured operator $\mathcal{T}$ to the routed gradient,
potentially altering its dimension or manifold:
\begin{equation}
  \hat{G}_t = \mathcal{T}\!\left(G_t;\,\mathcal{S}_{t-1}\right)
  \in \mathbb{R}^{r\times s}.
  \label{eq:s2}
\end{equation}
When $r=m$ and $s=n$, the transformation is dimension-preserving.  When
$r\ll m$ or $s\ll n$, it is a compression.  Representative transformations
include:
(i)~the identity $\hat{G}_t = G_t$ (T1, T4, most T5 methods);
(ii)~Newton--Schulz spectral orthogonalization
$\hat{G}_t = \mathrm{NS}_k(M_t)$ such that $\hat{G}_t^\top\hat{G}_t\approx I$
(Muon, T2.1);
(iii)~Kronecker-factored preconditioning
$\hat{G}_t = L_t^{-1/4}G_tR_t^{-1/4}$ where
$L_t\approx\mathbb{E}[G_tG_t^\top]$ (Shampoo, T2.2);
(iv)~low-rank orthogonal projection
$\hat{G}_t = P_t^\top G_t\in\mathbb{R}^{r\times n}$ with
$P_t\in\mathbb{R}^{m\times r}$, $r\ll m$ (GaLore, T2.3);
(v)~sign or quantized-direction discretization
$\hat{G}_t = \mathrm{sign}(\beta_1 m_{t-1}+(1-\beta_1)G_t)$
(Lion, T3).
Stages S2 and S4 form a mathematical dual: whenever S2 introduces a
dimension-reducing or basis-rotating map, S4 must define the corresponding
return map to the parameter space.  This duality is exact for orthogonal basis
rotations and approximate for projection or factorized-state methods.

\paragraph{S3: State Evolution.}
This stage updates the internal memory of the optimizer:
\begin{equation}
  \mathcal{S}_t = f\!\left(\mathcal{S}_{t-1},\,\hat{G}_t\right).
  \label{eq:s3}
\end{equation}
For Adam-family methods, $\mathcal{S}_t$ contains first- and second-order
moment EMAs:
\begin{equation}
  m_t = \beta_1 m_{t-1}+(1-\beta_1)\hat{G}_t, \quad
  v_t = \beta_2 v_{t-1}+(1-\beta_2)\hat{G}_t^{\odot 2}.
  \label{eq:adam-state}
\end{equation}
Alternative state forms include Kronecker factors
$L_t=\mathrm{EMA}_\beta(G_tG_t^\top)$,
$R_t=\mathrm{EMA}_\beta(G_t^\top G_t)$ (Shampoo, T2.2); row-column factored
states $r_t\in\mathbb{R}^m$, $c_t\in\mathbb{R}^n$ with
$v_t\approx r_t c_t^\top$ (AdaFactor, T4.1); INT8-quantized states with
error-feedback compensation (8-bit Adam, T4.2); and multi-timescale EMA
streams (AdEMAMix, T1.2).  Stateless methods either bypass S3 or pass the
current transformed signal directly downstream.
As established in Section~\ref{sec:prelim-background}, the second-order moment
$v_t=\mathrm{EMA}(g_t^2)$ is an online estimator of the diagonal Fisher
information matrix.  This connects S3 to Axis~III of the LMO-driven four-axis
decomposition (Section~\ref{sec:four-axis-decomposition}).

\paragraph{S4: Update Reconstruction.}
If S2 introduced a change of representation, this stage applies the
corresponding inverse operator $\mathcal{R}$ to restore the full-space update:
\begin{equation}
  \hat{\Delta}_t = \mathcal{R}\!\left(\tilde{\Delta}_t;\,\mathcal{S}_t\right)
  \in \mathbb{R}^{m\times n},
  \label{eq:s4}
\end{equation}
where $\tilde{\Delta}_t$ is the update direction derived from the evolved
state.
For low-rank methods, $\hat{\Delta}_t = P_t\tilde{\Delta}_t$;
for Kronecker-factor methods,
$\hat{\Delta}_t = Q_L\tilde{\Delta}_t Q_R^\top$ (exact, as $Q_L,Q_R$ are
orthogonal); for row-column-factored preconditioners, the reconstruction is a
rank-1 outer-product approximation.
When S2 is the identity, S4 is also the identity.
The exactness of the inverse map governs the fidelity of the final update
direction and is analyzed per family in
Subsections~\ref{sec:t2}--\ref{sec:t4}.

\paragraph{S5: Update Finalization.}
The reconstructed direction is converted into the actual parameter writeback
by applying scaling, regularization, and geometric constraints:
\begin{equation}
  W_{t+1} = W_t
    - \eta_t \cdot \phi_t^{(l)} \cdot
      \mathcal{C}\!\left(\mathcal{F}(\hat{\Delta}_t)\right)
    - \eta_t\lambda W_t,
  \label{eq:s5}
\end{equation}
where $\eta_t$ is the global learning rate, $\phi_t^{(l)}$ is an optional
layer-wise trust-ratio factor, $\mathcal{F}$ is a post-update filter operator,
$\mathcal{C}$ is a clipping operator, and $\lambda$ is the weight-decay
coefficient.
Representative sub-operations include:
decoupled weight decay (AdamW~\cite{loshchilov2019decoupled});
global gradient clipping;
element-wise trust-region clipping
$\hat{\Delta}_t\leftarrow\mathrm{clip}\!\left(\hat{\Delta}_t/
\max(\gamma h_t,\varepsilon),1\right)$
(Sophia~\cite{liu2023sophia});
layer-wise trust-ratio scaling
$\phi_t^{(l)} = \Phi(\|W^{(l)}\|/\|\hat{\Delta}^{(l)}\|)$ (LAMB);
direction-consistency masking (Cautious Optimizers);
and adversarial-perturbation final corrections
(SAM~\cite{foret2021sharpness}, T5.1).
These sub-operations are optional and often composable, but their order must be
specified when two mechanisms modify the same quantity.  Their shared role is
to convert the update direction $\hat{\Delta}_t$ into the physically written
parameter change.

Algorithm~\ref{alg:meta_optimizer} gives the full abstract procedure.  The
operator names \textsc{Route}, \textsc{Transform}, \textsc{Evolve},
\textsc{Reconstruct}, and \textsc{Finalize} are abstract placeholders.  A
concrete optimizer family replaces one or two of them with a structured
operator and leaves the remainder as identity maps or standard defaults.

\begin{algorithm}[!t]
\caption{Universal Meta-Pipeline for Modern Optimizer Updates.}
\label{alg:meta_optimizer}
\begin{algorithmic}[1]
  \Require Parameters $\Theta_0$, loss $\mathcal{L}$, 
           learning-rate schedule $\{\eta_t\}_{t=1}^T$
  \Require Initial optimizer state $\mathcal{S}_0$, 
           regularization coefficient $\lambda$
  \Ensure  Final parameters $\Theta_T$
  \For{$t = 1, \ldots, T$}
    \State \textcolor{famS0}{$G_t \gets \textsc{AcquireSignal}(\Theta_{t-1}, \mathcal{L})$
      \Comment{S0: FO / VR / curvature-augmented}}
    \For{each $\theta^{(i)}_{t-1} \in \Theta_{t-1}$}
      \State $g_t^{(i)} \gets \textsc{Select}(G_t,\,\theta^{(i)}_{t-1})$
      \State \textcolor{famS1}{$\rho^{(i)} \gets \textsc{Route}(\theta^{(i)}_{t-1})$ 
        \Comment{S1: parameter scoping and routing}}
      \State \textcolor{famS2}{$\hat{g}_t^{(i)} \gets \textsc{Transform}(g_t^{(i)},\,\rho^{(i)})$ 
        \Comment{S2: gradient transformation}}
      \State \textcolor{famS3}{$s_t^{(i)} \gets \textsc{Evolve}(s^{(i)}_{t-1},\,\hat{g}_t^{(i)},\,\rho^{(i)})$ 
        \Comment{S3: state evolution}}
      \State \textcolor{famS4}{$\hat{\Delta}_t^{(i)} \gets \textsc{Reconstruct}(s_t^{(i)},\,\rho^{(i)})$ 
        \Comment{S4: update reconstruction}}
      \State \textcolor{famS5}{$\theta_t^{(i)} \gets \textsc{Finalize}(\theta^{(i)}_{t-1},\,\hat{\Delta}_t^{(i)},\,\eta_t,\,\lambda)$ 
        \Comment{S5: update finalization}}
        
    \EndFor
    \State $\mathcal{S}_t \gets \{s_t^{(i)}\}_i$
  \EndFor
\end{algorithmic}
\end{algorithm}

The six stages should be read as an ordered checklist for locating
non-identity operations.  In the default path, S0 supplies a first-order
mini-batch gradient, S1 assigns all parameters to a single element-wise route,
S2 and S4 are identity maps, S3 is either stateless or a standard momentum
buffer, and S5 writes the update with a global learning rate.  Non-default
optimizers deviate from this path by changing the acquired signal in S0,
routing tensors or modules differently in S1, transforming the gradient
representation in S2, storing richer temporal or curvature information in S3,
reconstructing a full-space direction in S4, or applying decay, clipping,
trust-ratio scaling, masks, or sharpness corrections in S5.  Because practical
optimizers often combine several such deviations, the meta-pipeline is not a
one-to-one stage-to-family lookup.  It is instead an operational vocabulary for
describing where a method intervenes before the later taxonomy discusses
families at a coarser mechanism level.

\subsubsection{Identity Mapping and Representative Instantiations}
\label{sec:meta-pipeline-identity}

A central consequence of the pipeline formulation is the
\emph{identity-mapping principle}: most optimizers make a nontrivial design
choice at only one or two pipeline stages and leave the remaining stages as
identity maps or standard defaults.  This principle provides a compact
characterization of representative optimizers and supports the methodological
taxonomy in Section~\ref{sec:taxonomy}, while allowing individual families to
span multiple stages when their mechanisms are coupled.

\paragraph{AdamW~\textup{(T1).}}
All parameters are assigned to a single element-wise route, so S1 is
effectively an identity.  Gradient Transformation~(S2) is also the identity:
$G_t$ enters State Evolution~(S3) unchanged.  S3 maintains first- and
second-order moment EMAs as in Eq.~\eqref{eq:adam-state}.  This is the
defining stage of the T1 family.  Update Reconstruction~(S4) is the identity
because moment states reside in the full parameter space.  Update
Finalization~(S5) applies bias-corrected adaptive scaling and decoupled weight
decay.  In the pipeline view, \textbf{AdamW is primarily an S3/S5 method}.

\paragraph{Muon~\textup{(T2.1).}}
Parameter Scoping and Routing~(S1) separates two-dimensional weight matrices
from vector parameters.  Gradient Transformation~(S2) applies Newton--Schulz
orthogonalization to the gradient momentum, producing a spectrally normalized
matrix $\hat{G}_t$ with $\hat{G}_t^\top\hat{G}_t\approx I$.  State
Evolution~(S3) maintains standard momentum.  Update Reconstruction~(S4) is
trivially satisfied because the orthogonalization is dimension-preserving.
Update Finalization~(S5) follows an SGD-style writeback.
\textbf{Muon is primarily an S1/S2 method}.

\paragraph{GaLore~\textup{(T2.3).}}
Routing~(S1) identifies tensors eligible for low-rank treatment.  Gradient
Transformation~(S2) projects gradients into an $r$-dimensional subspace via
$\hat{G}_t = P_t^\top G_t$.  State Evolution~(S3) runs a full Adam update in
the reduced space.  Update Reconstruction~(S4) maps the subspace update back
via $\hat{\Delta}_t = P_t\tilde{\Delta}_t$.  The primary contribution is the
relocation of both computation and optimizer state into a
lower-dimensional space through the S2/S4 dual.
\textbf{GaLore spans S1--S4, with primary novelty at S2 and S4}.

\paragraph{Lion~\textup{(T3).}}
Gradient Transformation~(S2) discretizes the momentum-interpolated gradient
into the signed direction
\begin{equation}
  d_t = \mathrm{sign}\!\left(\beta_1 m_{t-1}+(1-\beta_1)G_t\right),
\end{equation}
whose entries lie in $\{-1,+1\}$ and whose writeback has an $\ell_\infty$
norm set by the learning rate.  State Evolution~(S3) maintains only a
first-order momentum (no second-order state).  Update Reconstruction~(S4) is
the identity since the discrete direction already has the parameter shape.
\textbf{Lion is primarily an S2/S3 method}.

\paragraph{SAM~\textup{(T5.1).}}
A first backward pass computes the mini-batch gradient $G_t$.  An adversarial
perturbation $\epsilon^* = \rho G_t/\|G_t\|$ is then injected, and a second
backward pass at $W_t+\epsilon^*$ supplies the sharpness-aware update signal.
Stages S1--S4 execute default element-wise operations.  The sharpness
correction is realized entirely within the Training Signal Acquisition~(S0)
and Update Finalization~(S5).
\textbf{SAM is primarily an S0/S5 method}.

Table~\ref{tab:meta_pipeline_examples} summarizes these instantiations.
Active-stage labels identify the pipeline positions that implement the
defining mechanism of each method, and the family label in the method cell
links each example to Subsections~\ref{sec:t1}--\ref{sec:t5}.

\begin{table*}[t]
  \centering
  \caption{Representative optimizer families viewed through the universal
  meta-pipeline. Only active stages and defining mechanisms are shown; inactive
  stages follow identity maps or standard defaults.}
  \label{tab:meta_pipeline_examples}
  \footnotesize
  \setlength{\tabcolsep}{6pt}
  \renewcommand{\arraystretch}{1.10}
  \begin{tabular}{@{}lcl@{}}
    \toprule
    Method (family) &
    \multicolumn{1}{c}{Active stages} &
    \multicolumn{1}{c}{Core mechanism} \\
    \midrule
    AdamW (T1.1)
      & S3, S5
      & Moment EMAs (S3) with decoupled weight decay (S5) \\
    Muon (T2.1)
      & S1, S2
      & Matrix routing (S1) with Newton--Schulz spectral orthogonalization (S2) \\
    GaLore (T2.3)
      & S1--S4
      & Low-rank projection (S1/S2), subspace Adam state (S3), and inverse projection (S4) \\
    Lion (T3)
      & S2, S3
      & Momentum interpolation (S3) followed by sign discretization (S2) \\
    SAM (T5.1)
      & S0, S5
      & Perturbation-induced gradient (S0) with neighborhood-regularized writeback (S5) \\
    \bottomrule
  \end{tabular}
\end{table*}

The same view also exposes composition constraints.  Mechanisms placed in
different stages are often naturally stackable.  For example, a variance-reduced signal
(S0) can feed AdamW, Lion, or Shampoo.  Low-rank projection (S2/S4) can be
combined with state quantization (S3/S4), and a final trust-ratio wrapper (S5)
can be placed after a standard adaptive update.  By contrast, mechanisms that
occupy the same slot require an explicit ordering.  For example, two S2
operations such as low-rank projection and spectral orthogonalization must
specify whether the optimizer projects before orthogonalizing or
orthogonalizes before projecting, because the resulting descent direction and
state memory differ.  This is why the meta-pipeline is useful not only for
classification, but also for reasoning about optimizer composition.

The meta-pipeline answers an \emph{operational} question: where does an
optimizer intervene in the update process?  The LMO-driven analysis in
Section~\ref{sec:lmo-four-axis} answers a complementary \emph{geometric}
question: why does the resulting direction take a particular mathematical
form?  The two views are aligned.  S2 corresponds to applying an analysis
basis and a direction function (Axes~I and~III of the four-axis decomposition).
S3 corresponds to maintaining a curvature estimate and its storage form
(Axes~II and~III).  S5 encompasses the weight-decay, clipping, and
trust-ratio mechanisms orthogonal to direction geometry.  This alignment
allows each optimizer to be characterized by a concise four-axis coordinate
tuple in addition to its active pipeline stages.  The full instantiation table
is given in Section~\ref{sec:four-axis-instantiation}.

\subsection{A Mathematical Unification of Optimizer Updates}
\label{sec:lmo-four-axis}

The universal meta-pipeline of Section~\ref{sec:meta-pipeline} is an operational abstraction, because it only tells us at which step of a single update an optimizer intervenes. The shape of the resulting update direction is exactly what this section formalizes through a complementary mathematical view.

We take the linear minimization oracle, abbreviated LMO, as the central building block, and the reason is geometric. When the same gradient signal is fed into different norm balls, the extremal direction returned by the oracle changes with the shape of the ball, so the constraint geometry and the optimizer behavior become two readings of one object. This single idea is strong enough to place sign updates, spectral orthogonalization, Kronecker-factored preconditioning, curvature weighting, variance reduction, and state compression inside one compact coordinate system, and it lets us compare them without ever switching notation.

Building on this view, we organize every optimizer update along four axes, and we order them so that they follow the actual pipeline of an optimizer. The first axis is the \emph{update domain}, which asks where the update lives, whether in the original parameter space, in the matrix space, in a low-rank subspace, or in some projected coordinate system. The second axis is the \emph{state estimator}, which asks how the effective signal, the momentum, the second-order state, the Gram Hessian, and the projection state are formed from the current gradient together with the accumulated history. The third axis is the \emph{geometry and precondition operator}, which asks how the quantity produced by the state estimator becomes a direction, either through an LMO constraint set or through a Hessian-based preconditioner. The fourth axis is the \emph{finalization wrapper}, which asks how the direction is written back to the parameters, and it absorbs the learning rate, the weight decay, the projection-back step, the routing rule, the fallback, and the refresh schedule. In short, the first axis fixes the space, the second estimates the state, the third generates the direction, and the last commits the update.

\subsubsection{LMO Foundations and Norm-Induced Directions}
\label{sec:lmo-foundations}

\paragraph{Linear minimization oracle.}
Given a convex set $\mathcal{D}$ and an input signal $s$, the LMO is defined as
\begin{equation}
  \operatorname{lmo}_{\mathcal{D}}(s)\in\argmin_{x\in\mathcal{D}}\langle s,x\rangle .
  \label{eq:lmo-general}
\end{equation}
Here $\mathcal{D}$ can be any convex set, such as a simplex, a polytope, an $\ell_1$ ball, a nuclear-norm ball, or, most importantly for this section, a norm ball. Norm balls are exactly the special case that connects the oracle to familiar optimizer directions, and they are the case we develop next.

\paragraph{Norm-constrained LMO.}
When the constraint set is a norm ball of radius $\rho$, written $\mathcal{D}_\rho=\{x:\|x\|\le\rho\}$, the oracle returns the extremal descent direction of that particular geometry. To make this explicit, let $u^\sharp(s)$ denote the steepest-ascent direction on the unit ball,
\begin{equation}
  u^\sharp(s)\in\argmax_{\|u\|\le 1}\langle s,u\rangle ,
\end{equation}
so that the oracle factorizes as
\begin{equation}
  \operatorname{lmo}_{\mathcal{D}_\rho}(s)=-\rho\,u^\sharp(s).
  \label{eq:lmo-norm}
\end{equation}
Now suppose the optimizer takes the descent form $W_{t+1}=W_t-\eta_t\Phi_t(s_t)$. Then the direction operator is exactly
\begin{equation}
  \Phi_t(s_t)=u^\sharp(s_t)=-\frac{1}{\rho}\operatorname{lmo}_{\mathcal{D}_\rho}(s_t),
  \label{eq:lmo-direction-operator}
\end{equation}
which shows that the norm-constrained LMO and steepest descent share one geometric core. In words, the norm ball is what decides which direction counts as steepest. One caveat matters here, because the oracle reports a direction and a boundary point but discards magnitude, so any gradient scale that we wish to keep must be reintroduced through the learning rate, the radius, or a preconditioner.

\paragraph{Unconstrained update-direction view.}
For the rest of the section we adopt the direction-generator view of an unconstrained optimizer,
\begin{equation}
  W_{t+1}=W_t-\eta_t\Phi_t(M_t),
  \label{eq:unconstrained-direction}
\end{equation}
where $W_t\in\mathbb{R}^d$ collects the parameters, $M_t$ is the effective signal handed over by the state estimator, and $\Phi_t$ is the direction operator that the LMO geometry or the preconditioner induces.

\paragraph{Canonical norm geometries.}
Different norm balls give rise to different families of directions, and a few canonical choices already cover most optimizers. The Euclidean ball yields the normalized gradient,
\begin{equation}
  \mathcal{D}_2=\{x:\|x\|_2\le\rho\},
  \quad
  \operatorname{lmo}_{\mathcal{D}_2}(g)=-\rho\frac{g}{\|g\|_2},
  \quad
  \Phi(g)=\frac{g}{\|g\|_2}.
  \label{eq:lmo-euclidean}
\end{equation}
The max-norm ball yields the sign direction, because every vertex of the hypercube is a sign pattern,
\begin{equation}
  \mathcal{D}_\infty=\{x:\|x\|_\infty\le\rho\},
  \quad
  \operatorname{lmo}_{\mathcal{D}_\infty}(g)=-\rho\operatorname{sign}(g),
  \quad
  \Phi(g)=\operatorname{sign}(g).
  \label{eq:lmo-signball}
\end{equation}
The spectral-norm ball yields the matrix polar direction, and writing $M=U\Sigma V^\top$ we obtain
\begin{equation}
  \mathcal{D}_{S_\infty}=\{X:\|X\|_{S_\infty}\le\rho\},
  \quad
  \operatorname{lmo}_{\mathcal{D}_{S_\infty}}(M)=-\rho\,UV^\top,
  \quad
  \Phi(M)=UV^\top.
  \label{eq:lmo-spectral}
\end{equation}
We stress that the SVD here is only a device for computing the polar operator. Muon is the clearest example, because its update domain can stay in the original matrix space with $Q_L=Q_R=I$. Adam fits the same picture once the box is allowed to move with the state. If the state estimator outputs $m_t$ and $v_t$, define the dynamic bound $b_{t,i}=|m_{t,i}|/\sqrt{v_{t,i}}$ and the adaptive box
\begin{equation}
  \mathcal{D}_t^{\operatorname{Adam}}=\{x:|x_i|\le\rho_t b_{t,i},\ \forall i\},
  \quad
  \operatorname{lmo}_{\mathcal{D}_t^{\operatorname{Adam}}}(m_t)=-\rho_t\frac{m_t}{\sqrt{v_t}} .
  \label{eq:lmo-adam-box}
\end{equation}
Adam is therefore a dynamic-boundary LMO over an adaptive constraint set, and that constraint set is itself produced by the Axis~II state estimator. Table~\ref{tab:four-axis-instantiation} collects these correspondences between norms, analysis coordinates, direction operators, and representative optimizers, and it again records that Muon keeps the original update domain even though its direction is computed through an SVD.

\subsubsection{A Four-Axis Decomposition of Optimizer Updates}
\label{sec:four-axis-decomposition}

The reorganization at the heart of this section is easy to state. The LMO reading and the preconditioning reading are two faces of one direction operator $\Phi_t$, so we keep them together inside Axis~III. At the same time, we move momentum, the second moment, variance reduction, and the projection state one step earlier into Axis~II, because all of them are estimated before any direction is formed and all of them feed the operator. With this ordering the update takes a master form,
\begin{align}
  (M_t,H_t,\mathcal{D}_t)
  &=\operatorname{StateEstimator}_t(g_t,\operatorname{State}_{t-1}),
    \label{eq:state-estimator}\\
  D_t &=\Phi_t(M_t;H_t,\mathcal{D}_t),
    \label{eq:geometry-operator}\\
  W_{t+1} &=\operatorname{Finalize}(W_t,D_t),
    \label{eq:finalize}
\end{align}
in which $g_t$ is the current stochastic gradient, $M_t$ is the effective signal, $H_t$ is a Hessian, Gram, Fisher, or second-moment proxy, and $\mathcal{D}_t$ is the LMO constraint set when one is used. Axis~II produces these objects, and only Axis~III turns them into a direction. We describe the four axes one by one below, and because each axis is a largely independent choice, a concrete optimizer follows once we settle on an update domain, a state estimator, a geometry-and-precondition operator, and a finalization wrapper.

\paragraph{Axis~I: Update domain and support.}
This axis, written $\mathcal{X}_t$ or $(Q_L,Q_R)$, fixes the space in which the update is expressed. For SGD, AdamW, and Muon the natural choice is to update directly in the original parameter or matrix space, whereas GaLore deliberately confines the update to a low-rank projected space. Whenever such a projection is present, we write $Z_t=Q_L^\top M_t Q_R$ and then carry the direction back through $D_t=Q_L\,\Phi_t(Z_t;\bar{H}_t,\bar{\mathcal{D}}_t)\,Q_R^\top$, so the direction is first formed in the small space and only afterwards returned to the full space. When there is no genuine subspace, however, no basis should be introduced artificially, and Muon makes the point concrete because it can simply take $Q_L=Q_R=I$. The representative choices are therefore the full space, the matrix space, and a low-rank projected subspace.

\paragraph{Axis~II: State estimator.}
The state estimator, written $\operatorname{StateEstimator}_t$, answers one question. Given the current gradient and the accumulated history, what signal does the optimizer actually hand to the geometry operator? Vanilla SGD keeps nothing, so $M_t=g_t$ and $H_t=I$. Momentum SGD keeps a first-order average, so $M_t=m_t$ with $m_t=\beta m_{t-1}+(1-\beta)g_t$. Adam keeps both a first moment and a second moment,
\begin{equation}
  m_t=\beta_1 m_{t-1}+(1-\beta_1)g_t,
  \quad
  v_t=\beta_2 v_{t-1}+(1-\beta_2)g_t^2,
  \quad
  M_t=m_t,
  \quad
  H_t=\operatorname{diag}(v_t).
\end{equation}
MARS~\cite{yuan2024mars} belongs here as well, even though it is usually described as a variance-reduction method, because in our framework its role is unambiguous. It first turns the raw stochastic gradient into a variance-reduced estimate, and only afterwards maintains the momentum and the second moment from that estimate. Concretely, writing $g_t^{\xi_t}=\nabla f(W_t,\xi_t)$ and $g_{t-1}^{\xi_t}=\nabla f(W_{t-1},\xi_t)$, the corrected gradient is
\begin{equation}
  c_t=g_t^{\xi_t}+\gamma_t\frac{\beta_1}{1-\beta_1}\left(g_t^{\xi_t}-g_{t-1}^{\xi_t}\right),
\end{equation}
and the MARS-AdamW state then follows $m_t=\beta_1 m_{t-1}+(1-\beta_1)c_t$ and $v_t=\beta_2 v_{t-1}+(1-\beta_2)c_t^2$, again with $M_t=m_t$ and $H_t=\operatorname{diag}(v_t)$. This makes the separation clean, because variance reduction in the style of MARS, STORM, or SVRG belongs to Axis~II, whereas the geometry and preconditioning of AdamW, Shampoo, or Muon belong to Axis~III.

\paragraph{Axis~III: Geometry and precondition operator.}
The direction generator, written $\Phi_t$, carries two readings at once, an LMO reading and a preconditioning reading, which is why we name the axis the geometry and precondition operator. From the LMO side it is $\Phi_t(M_t)=-\frac{1}{\rho_t}\operatorname{lmo}_{\mathcal{D}_t}(M_t)$, and from the preconditioning side it is $\Phi_t(M_t)=H_t^{-\alpha}M_t$, where $\alpha$ is an internal exponent of $\Phi_t$. The two readings agree on the familiar cases, since AdamW uses a diagonal second-moment metric, Shampoo uses a Kronecker-factored metric, and Muon uses the Gram Hessian of the current matrix. Muon is worth spelling out, because it shows the equivalence directly. Take $M_t\in\mathbb{R}^{m\times n}$ with $m<n$, and use the left Gram Hessian $H_t=M_t M_t^\top$ in the idealized full-rank case. Then
\begin{equation}
  \Phi_t(M_t)=H_t^{-1/2}M_t=(M_t M_t^\top)^{-1/2}M_t,
  \quad\text{with}\quad
  M_t=U_t\Sigma_t V_t^\top,
  \quad
  \Phi_t(M_t)=U_t V_t^\top,
\end{equation}
so the preconditioned form coincides exactly with the spectral-LMO polar form. Muon is therefore a spectral-norm LMO and a Gram-based preconditioner at the same time, and these are the two faces of Axis~III.

\paragraph{Axis~IV: Finalization and wrapper.}
The last axis, written $\operatorname{Finalize}$, handles everything that happens after a direction has been produced, from the learning rate and the weight decay to the projection-back step, the layer-wise routing, the fallback rule, and the refresh schedule. The decoupled weight decay of AdamW lives here, the projection-back of GaLore lives here, and the unique Muon practice of treating different tensor types with different rules lives here as well.

\paragraph{Relation to the universal meta-pipeline.}
The mathematical framework of this section and the meta-pipeline of Section~\ref{sec:meta-pipeline} describe the same update from two angles, and the two line up step by step. Choosing the update domain $W_t\in\mathcal{X}_t$ is the P1 step and corresponds to Axis~I, estimating the optimizer state $(M_t,H_t,\mathcal{D}_t)=\operatorname{StateEstimator}_t(g_t,\operatorname{State}_{t-1})$ is the P2 step and Axis~II, generating the direction $D_t=\Phi_t(M_t;H_t,\mathcal{D}_t)$ is the P3 step and Axis~III, and finalizing the writeback $W_{t+1}=\operatorname{Finalize}(W_t,D_t)$ is the P4 step and Axis~IV. The ordering is the part worth remembering, because the state estimator always comes first and the geometry operator always comes second, so Axis~II produces the triple $(M_t,H_t,\mathcal{D}_t)$ and Axis~III then consumes that triple to form the direction $\Phi_t(M_t;H_t,\mathcal{D}_t)$.

\subsubsection{Instantiations}
\label{sec:four-axis-instantiation}

We close the section by reading several representative optimizers through the four axes. The main table keeps only short labels, and the detailed formulas follow underneath so that the table stays legible.

\begin{table*}[htbp]
  \centering
  \caption{Optimizer instantiations under the four-axis decomposition
  (Axes~I--IV of Section~\ref{sec:four-axis-decomposition}), spanning all five
  families (T1--T5). Axis~III shows two faces of the same direction operator
  $\Phi_t$, the LMO constraint ball and the preconditioner $H_t$. The LMO
  direction is the steepest descent inside that ball, which is the $\ell_2$ ball,
  the $\ell_\infty$ box, or the metric ball
  $\|\delta\|_{H_t}=\sqrt{\delta^\top H_t\delta}\le 1$ induced by $H_t$. In the Domain
  column, $\mathbb{R}^d$ means all parameters are flattened into $d$ independent
  scalar coordinates and updated element-wise, while
  $\mathbb{R}^{m\times n}$ means the update operates on a whole weight matrix
  and couples its entries, with short tags for factored, INT8, block, row, or
  column variants. Here $g_t$ is the gradient, $m_t,v_t$ the first and second moments,
  $c_t$ the MARS variance-reduced estimate, $s_t$ the AdaBelief variance, $h_t$ a
  diagonal Hessian estimate, $M_t$ the matrix-form momentum, $L_t,R_t$ the
  Shampoo row and column Kronecker factors, $Q_L,Q_R$ their eigenvector matrices,
  $P_t$ the low-rank projection onto the current subspace (learned for GaLore and
  Fira, random for APOLLO, column-orthogonal for Conda) with back-projection
  $P_t^\top$, and $\bar{m}_t,\bar{v}_t,\bar{H}_t$ the quantities formed inside
  $P_t$. The tag +res marks Fira's residual correction in the orthogonal
  complement of $P_t$, VR marks a variance-reduced (MARS) estimate, and matrix
  routing means only matrix parameters receive the matrix update while
  vector-like parameters follow an AdamW-style branch.}
  \label{tab:four-axis-instantiation}
  \scriptsize
  \setlength{\tabcolsep}{1.mm}
  \renewcommand{\arraystretch}{1.3}
  \begin{tabularx}{\textwidth}{
  @{}
  p{0.105\textwidth}
  p{0.105\textwidth}
  Y 
  Y 
  Y 
  Y
  @{}}
    \toprule
    Optimizer & Axis~I: Domain & Axis~II: State estimator
      & Axis~III (LMO) & Axis~III (Precondition) & Axis~IV: Finalization \\
    \midrule
    \rowcolor{famT1!18}\multicolumn{6}{c}{\textbf{\textcolor{famT1!62!black}{T1: Element-wise adaptive moment and scalar control}}} \\
    SGDM
      & $\mathbb{R}^d$ & $m_t$
      & $\ell_2$ ball & $H_t=I$ & LR \\
    Adam, AdamW
      & $\mathbb{R}^d$ & $m_t,v_t$
      & adaptive $\ell_\infty$ & $H_t=\operatorname{diag}(v_t)$
      & LR + decoupled WD \\
    NAdam
      & $\mathbb{R}^d$ & Nesterov $m_t$, $v_t$
      & adaptive $\ell_\infty$ & $H_t=\operatorname{diag}(v_t)$
      & Nesterov + LR + WD \\
    AdaBelief
      & $\mathbb{R}^d$ & $m_t,s_t$
      & adaptive $\ell_\infty$ & $H_t=\operatorname{diag}(s_t)$
      & LR + WD \\
    ADOPT
      & $\mathbb{R}^d$ & ordered, delayed $m_t,v_t$
      & adaptive $\ell_\infty$ & $H_t=\operatorname{diag}(v_t)$
      & LR + WD \\
    Adan
      & $\mathbb{R}^d$ & $m_t$ + grad-diff state
      & adaptive $\ell_\infty$ & $H_t=\operatorname{diag}(v_t)$
      & LR + WD \\
    AdEMAMix
      & $\mathbb{R}^d$ & short/long EMA
      & adaptive $\ell_\infty$ & $H_t=\operatorname{diag}(v_t)$
      & LR + WD \\
    MARS-AdamW
      & $\mathbb{R}^d$ & $c_t,m_t,v_t$
      & adaptive $\ell_\infty$ (VR)
      & $H_t=\operatorname{diag}(v_t)$, $v_t=\operatorname{EMA}(c_t^2)$
      & LR + decoupled WD \\
    RAdam
      & $\mathbb{R}^d$ & rectified $m_t,v_t$
      & adaptive $\ell_\infty$ & $H_t=\operatorname{diag}(v_t)$
      & LR + WD \\
    Prodigy
      & $\mathbb{R}^d$ & $m_t,v_t$ + LR est.\ $d_t$
      & adaptive $\ell_\infty$ & $H_t=\operatorname{diag}(v_t)$
      & automatic LR + WD \\
    \midrule
    \rowcolor{famT2!18}\multicolumn{6}{c}{\textbf{\textcolor{famT2!62!black}{T2: Matrix-level structural methods}}} \\
    Muon
      & $\mathbb{R}^{m\times n}$ & $M_t$
      & spectral (polar) & $H_t=M_t M_t^\top$
      & LR + matrix routing \\
    MARS-Shampoo
      & $\mathbb{R}^{m\times n}$ ($L_t,R_t$) & $c_t,m_t,L_t,R_t$
      & metric ball (VR) & $H_t=L_t^{1/4}\!\otimes\!R_t^{1/4}$
      & LR + damping \\
    Shampoo
      & $\mathbb{R}^{m\times n}$ ($L_t,R_t$) & $m_t,L_t,R_t$
      & metric ball & $H_t=L_t^{1/4}\!\otimes\!R_t^{1/4}$
      & LR + damping \\
    SOAP
      & $Q_L,Q_R$ & $m_t,v_t,Q_L,Q_R$
      & adaptive $\ell_\infty$ in $Q_L,Q_R$
      & $\operatorname{diag}(v_t)$ in $Q_L,Q_R$ & LR + WD \\
    GaLore
      & $P_t$ & $\bar{m}_t,\bar{v}_t$ on $P_t g_t$
      & projected $\ell_\infty$ & $\bar{H}_t=\operatorname{diag}(\bar{v}_t)$
      & $P_t^\top$ back + LR + WD \\
    Fira
      & $P_t$ (+res) & $\bar{m}_t,\bar{v}_t$ + residual
      & projected $\ell_\infty$ (+res)
      & $\bar{H}_t=\operatorname{diag}(\bar{v}_t)$ & $P_t^\top$ back + res + LR \\
    RMNP
      & $\mathbb{R}^{m\times n}$ (row) & $M_t$
      & row-normalized & $H_t=\operatorname{diag}(M_t M_t^\top)$
      & LR + matrix routing \\
    \midrule
    \rowcolor{famT3!18}\multicolumn{6}{c}{\textbf{\textcolor{famT3!62!black}{T3: Discretization and directional quantization}}} \\
    SignSGD
      & $\mathbb{R}^d$ & $g_t$
      & fixed $\ell_\infty$ & $H_t=\operatorname{diag}(|g_t|)$ & LR \\
    Lion
      & $\mathbb{R}^d$ & $m_t$
      & fixed $\ell_\infty$ & $H_t=\operatorname{diag}(|m_t|)$ & LR + WD \\
    MARS-Lion
      & $\mathbb{R}^d$ & $c_t,m_t$
      & fixed $\ell_\infty$ (VR) & $H_t=\operatorname{diag}(|m_t|)$
      & LR + WD \\
    \midrule
    \rowcolor{famT4!18}\multicolumn{6}{c}{\textbf{\textcolor{famT4!62!black}{T4: State compression and structural aggregation}}} \\
    AdaFactor
      & $\mathbb{R}^d$ (factored) & row/col $v_t$ factors
      & adaptive $\ell_\infty$ & factored $\operatorname{diag}(v_t)$
      & LR + factored update \\
    CAME
      & $\mathbb{R}^d$ (factored) & factors + confidence
      & adaptive $\ell_\infty$ & factored $\operatorname{diag}(v_t)$ (+conf.)
      & LR + factored update \\
    Adam-mini
      & $\mathbb{R}^d$ (block) & $m_t,v_t$ (block)
      & block $\ell_\infty$ & block-mean $\operatorname{diag}(v_t)$ & LR + WD \\
    APOLLO
      & $P_t$ (rand.) & $\bar{m}_t,\bar{v}_t$
      & projected $\ell_\infty$ & $\bar{H}_t=\operatorname{diag}(\bar{v}_t)$
      & $P_t^\top$ back + LR \\
    8-bit Adam
      & $\mathbb{R}^d$ (INT8) & $m_t,v_t$ (INT8)
      & adaptive $\ell_\infty$
      & $H_t=\operatorname{diag}(v_t)$ in INT8
      & dequant + LR + WD \\
    Conda
      & $P_t$ (col) & $v_t$ (col)
      & projected $\ell_\infty$ & col-wise $\operatorname{diag}(v_t)$
      & $P_t^\top$ back + LR + WD \\
    \midrule
    \rowcolor{famT5!18}\multicolumn{6}{c}{\textbf{\textcolor{famT5!62!black}{T5: Curvature-aware and geometric regularization}}} \\
    Sophia
      & $\mathbb{R}^d$ & $m_t,h_t$
      & clipped local & $H_t=h_t$ & LR + WD \\
    AdaHessian
      & $\mathbb{R}^d$ & $m_t,h_t$ (Hutch.)
      & metric ball & $H_t=h_t$ & LR + WD \\
    AdamP
      & $\mathbb{R}^d$ & $m_t,v_t$
      & adaptive $\ell_\infty$ & $H_t=\operatorname{diag}(v_t)$
      & radial projection + LR + WD \\
    LAMB
      & $\mathbb{R}^d$ & $m_t,v_t$
      & adaptive $\ell_\infty$ & $H_t=\operatorname{diag}(v_t)$
      & trust ratio + LR + WD \\
    \bottomrule
  \end{tabularx}
\end{table*}

\paragraph{AdamW.}
AdamW estimates two moments in Axis~II, namely $m_t=\beta_1 m_{t-1}+(1-\beta_1)g_t$ together with $v_t=\beta_2 v_{t-1}+(1-\beta_2)g_t^2$. Axis~III then turns these states into a direction through $D_t=\Phi_t(m_t;H_t)=H_t^{-1/2}m_t$ with $H_t=\operatorname{diag}(v_t)$. Read as an LMO this is an adaptive $\ell_\infty$ box, and read as a preconditioner it is a diagonal second-moment metric, which is exactly the two-faced behavior described above.

\paragraph{MARS-AdamW.}
MARS-AdamW shares the Axis~III form of AdamW and differs only in Axis~II. It first builds a corrected gradient $c_t=g_t^{\xi_t}+\gamma_t\frac{\beta_1}{1-\beta_1}(g_t^{\xi_t}-g_{t-1}^{\xi_t})$, then maintains the Adam states from the corrected $c_t$, with $m_t=\operatorname{EMA}(c_t)$ and $v_t=\operatorname{EMA}(c_t^2)$, and only afterwards forms the direction $D_t=\operatorname{diag}(v_t)^{-1/2}m_t$. The variance reduction therefore sits squarely in the state estimator, while the preconditioning remains a later Axis~III step.

\paragraph{Muon.}
In Muon the Axis~II stage produces a momentum matrix $M_t$. The Axis~III reading as an LMO is the spectral-norm ball, so $\operatorname{lmo}_{\mathcal{D}_{S_\infty}}(M_t)=-\rho_t U_t V_t^\top$ with $M_t=U_t\Sigma_t V_t^\top$, which gives the direction operator $\Phi_t(M_t)=U_t V_t^\top$. The reading as a preconditioner instead uses the left Gram Hessian $H_t=M_t M_t^\top$, and then $\Phi_t(M_t)=H_t^{-1/2}M_t$. Both readings describe one update, and we keep the original matrix-space interpretation throughout.

\paragraph{GaLore-Adam.}
GaLore is the clearest case in which Axis~I is nontrivial, because it places the whole update inside a low-rank projected subspace. The construction starts from a pair of projection matrices, $Q_L\in\mathbb{R}^{m\times r}$ and $Q_R\in\mathbb{R}^{n\times r}$, whose columns span the retained subspace and whose rank $r$ is much smaller than the matrix dimensions $m$ and $n$. Given the full-space momentum $M_t\in\mathbb{R}^{m\times n}$, the optimizer first compresses it into the subspace through
\begin{equation}
  \bar{M}_t=Q_L^\top M_t Q_R,
\end{equation}
where $\bar{M}_t\in\mathbb{R}^{r\times r}$ is the projected momentum and the bar notation marks every quantity that lives inside the subspace. The Axis~II stage then runs an ordinary Adam state update on $\bar{M}_t$, maintaining a projected first moment $\bar{m}_t$ and a projected second moment $\bar{v}_t$, where $\bar{m}_t$ plays the role of the momentum and $\bar{v}_t$ plays the role of the coordinate-wise variance estimate, both stored in the small $r\times r$ space so that the memory cost drops from $O(mn)$ to $O(r^2)$. The Axis~III stage forms the direction inside the subspace exactly as AdamW would, namely
\begin{equation}
  \bar{D}_t=\frac{\bar{m}_t}{\sqrt{\bar{v}_t}},
\end{equation}
where the division and the square root act entry by entry, so $\bar{D}_t\in\mathbb{R}^{r\times r}$ is the adaptive direction in the projected coordinates. Finally, the Axis~IV stage lifts this direction back to the original space through
\begin{equation}
  D_t=Q_L\bar{D}_t Q_R^\top,
\end{equation}
where $D_t\in\mathbb{R}^{m\times n}$ is the update actually applied to the weight matrix. Read as an LMO the construction is a projected adaptive box, because the box geometry of Adam is simply expressed in the subspace coordinates, and read as a preconditioner it is a projected diagonal metric, because $\bar{v}_t$ supplies a diagonal scaling that is later rotated back by $Q_L$ and $Q_R$.

\paragraph{SOAP.}
SOAP is the case in which the update domain is a rotated coordinate system. It keeps the two Shampoo factors $L_t=\beta_2 L_{t-1}+(1-\beta_2)G_t G_t^\top$ and $R_t=\beta_2 R_{t-1}+(1-\beta_2)G_t^\top G_t$, where $G_t\in\mathbb{R}^{m\times n}$ is the matrix gradient, and takes their eigenvectors as the orthogonal bases $Q_L$ and $Q_R$ that define Axis~I. These bases are full rank, and they are refreshed only once every $f$ steps, so the preconditioning frequency $f$ is the one extra hyperparameter beyond AdamW. For Axis~II the estimator keeps a first moment in the original space, $M_t=\beta_1 M_{t-1}+(1-\beta_1)G_t$, together with a second moment that lives in the rotated space and is updated every step, $V_t=\beta_2 V_{t-1}+(1-\beta_2)(G_t'\odot G_t')$ with $G_t'=Q_L^\top G_t Q_R$. Axis~III is then the ordinary Adam box read in the eigenbasis. It rotates the momentum through $M_t'=Q_L^\top M_t Q_R$ and forms
\begin{equation}
  N_t'=\frac{M_t'}{\sqrt{V_t}+\epsilon},
\end{equation}
after which Axis~IV rotates the direction back through $N_t=Q_L N_t' Q_R^\top$ and adds the learning rate and decoupled weight decay. The contrast with Shampoo is sharp. Shampoo bakes the curvature into a fixed inverse-root metric, whereas SOAP keeps the basis fixed for $f$ steps and lets the per-step $V_t$ carry the adaptivity. This is why SOAP tolerates a stale basis and stays robust at a low preconditioning frequency.

\paragraph{Summary.}
The section therefore reduces every optimizer to four decisions, namely where it updates, how it estimates the state, how it turns that state into a direction, and how it writes the result back. The most useful consequence is the clean split between the two middle axes. Momentum, the second moment, the projection basis, and the variance reduction of MARS, STORM, or SVRG all belong to Axis~II, whereas the LMO geometry, the diagonal metric of Adam, the polar map of Muon, and the Kronecker metric of Shampoo all belong to Axis~III. Keeping these two apart prevents state estimation from being confused with direction geometry, and at the same time it preserves both the LMO and the preconditioning readings of every method.

\section{Dual-Dimension Taxonomy}
\label{sec:taxonomy}

Section~\ref{sec:unified-theoretical-framework} unifies modern LLM optimizers
from two complementary views.  The universal meta-pipeline asks where an
optimizer intervenes in a single update, whereas the LMO-driven four-axis
decomposition asks how the intervention changes the update direction,
curvature estimate, gradient estimator, and state representation.  This
section turns those two views into a survey taxonomy that can also guide the
benchmark design.  The goal is not to list optimizers chronologically or to
assign informal empirical labels.  The goal is to construct a mechanism
coordinate system that supports literature organization, fair experimental
grouping, and interpretable trade-off analysis.

The taxonomy has two dimensions.  \emph{Dimension A} is a methodological
taxonomy: each optimizer receives one primary mechanism label so that
Section~\ref{sec:optimizer-families} can present T1--T5 as non-overlapping
families.  \emph{Dimension B} is an effect-objective taxonomy: each optimizer
may receive multiple objective labels describing the training properties it
claims to improve or should be evaluated against, such as convergence,
per-step cost, memory, stability, hyperparameter robustness, and
generalization.  The two dimensions answer different questions: what does the
optimizer change, and what outcome is the change intended to improve?

\subsection{Taxonomy Design Principles}
\label{sec:taxonomy-principles}

A single tree taxonomy is not expressive enough for modern LLM optimizers. Lion, for example, can reasonably be described as an Adam alternative, a sign-based optimizer, and a memory-friendly optimizer. All three descriptions are defensible: Lion keeps a first-order momentum structure, replaces continuous-valued directions with a sign map, and removes Adam second-moment state. However, the labels answer different questions.
The sign map is the primary update mechanism, whereas memory saving is an effect of that mechanism.  If Lion is placed only under ``memory-efficient optimizers,'' it is grouped with AdaFactor, 8-bit Adam, Adam-mini, and LOMO, although those methods mainly alter state storage, state sharing, or gradient lifetime. If Lion is copied into every relevant branch, the taxonomy loses its role as a clean survey organization and an experimental grouping rule.

The same ambiguity appears throughout the optimizer landscape.  GaLore
changes the subspace in which gradients and optimizer states live, whereas
Q-GaLore adds low-bit state quantization on top of that subspace mechanism.
SAM can wrap SGD or AdamW, but its defining contribution is a
sharpness-aware perturbation and second gradient evaluation.  LAMB first
performs an Adam-style element-wise update and then applies a layer-wise trust
ratio at writeback.  Cautious AdamW and Cautious Lion inherit different base
optimizers but share the same direction-consistency filter.  These examples
show why the taxonomy must separate primary mechanism, base optimizer,
composable wrapper, and intended effect.

We therefore use three design principles.

\textbf{First, Dimension A uses a single primary mechanism label.}
Each optimizer is assigned to exactly one main family. The assignment is not based on the name of the optimizer or chronology, but on the component whose removal would make the method collapse to a simpler baseline. Direct modifications of Adam-style moments, bias correction, time scales, iterate
averaging, or global step-size adaptation belong to T1.  Matrix routing,
spectral orthogonalization, Kronecker-factored preconditioning, and low-rank
subspace projection belong to T2.  Sign maps and related irreversible
direction discretization belong to T3.  State sharing, compression,
quantization, or elimination belong to T4.  Perturbations, explicit curvature
wrappers, post-update filters, and layer-wise trust-region rules belong to T5.
Composite methods are classified by their incremental contribution and record
secondary tags in the cross-dimension matrix.

\textbf{Second, family boundaries are aligned with the meta-pipeline.}
The five families are not arbitrary names.  T1 mainly modifies element-wise
state evolution in S3 and local scaling in S5.  T2 changes the update space
through S1--S4 matrix routing, matrix transformation, structured state, and
reconstruction.  T3 introduces irreversible direction discretization in S2
and often simplifies S3.  T4 acts on S3/S4 state representation and gradient
lifetime.  T5 acts near S5 through update writeback, post-processing, or
geometric regularization.  Given an update rule, locating its dominant
non-identity pipeline operation is usually enough to assign the method label.

\textbf{Third, Dimension B uses multi-label effect annotations.}
Effect objectives should not be mutually exclusive categories.  A method may
target convergence efficiency and hyperparameter robustness at the same time,
or it may reduce memory while hurting convergence or stability.  An effect
label means that the method claims to improve, is designed to improve, or must
be evaluated on that property.
This separation avoids equating ``designed to save memory'' with ``better
under a fixed hardware budget,'' and avoids assuming that every
sharpness-aware method improves generalization under every protocol.

\subsection{Methodological Taxonomy}
\label{sec:taxonomy-method}

Dimension A organizes the surveyed set of 108 optimizers into five families
and fifteen subclasses.  The five
families correspond to the dominant mechanism that transforms a stochastic
gradient into a parameter update.  The main text keeps the organizing
principles and family definitions that are needed for the survey chapters that
follow, with a compact family-level summary given in
Table~\ref{tab:taxonomy_cross_matrix}.

\textbf{T1: Element-wise adaptive moment and scalar control.}
T1 retains the Adam-style element-wise processing paradigm, but it is broader
than a strict set of Adam variants.  Its core members maintain first moments,
second moments, or related variance surrogates and normalize each parameter
independently.  Its boundary members are scalar first-order predecessors or
outer-control schemes that do not introduce matrix routing, sign
quantization, state compression, or sharpness-aware wrapper gradients.
Differences within the family come from temporal structure, variance
estimation, iterate averaging, corrected decay, and global step-size rules.
In the four-axis view, most T1 methods use the identity analysis basis, a
diagonal Fisher or variance estimate, square-root preconditioning for the
Adam-like core, and full state storage.

\textbf{T2: Matrix-level structural methods.}
T2 treats two-dimensional Transformer weights as matrices rather than as
flattened vectors.  The family includes spectral orthogonalization,
Kronecker-factored preconditioning, and low-rank subspace projection.  These
methods share nontrivial S1 routing and S2 transformation: they either
orthogonalize matrix momentum, maintain row/column curvature factors, or
project gradients into a low-dimensional subspace before updating state.  T2
is the family that most visibly changes Axis~I.  Muon-style methods use a
dynamic SVD or polar basis, Shampoo and SOAP use Kronecker or Fisher eigenbases,
and GaLore changes the effective optimizer-state space through projection.

\textbf{T3: Discretization and directional quantization.}
T3 is defined by sign maps or related direction-discretization operations.
The core idea is to keep coarse direction information while discarding or
quantizing magnitude.  Lion is the representative method: it forms a signed
direction from a gradient-momentum interpolation and keeps only first-order
momentum.  Compared with T1, T3 is not merely a cheaper second-moment
estimator.  It changes the geometry of the update direction.  In the LMO view,
the family is naturally associated with $\ell_\infty$-type direction
selection and weak or absent Hessian preconditioning.  Methods whose primary
increment is a binary or stochastic mask applied to an already computed
update are treated as T5.3 post-update filtering rather than as T3, even when
the mask is motivated by directional agreement.

\textbf{T4: State compression and structural aggregation.}
T4 directly targets the optimizer-state memory bottleneck.  It includes
row/column factorization, low-bit state quantization, block- or layer-level
state sharing, and fused backprop-update schemes.  Its distinction from T2.3
is the object being compressed.  T2.3 first changes the mathematical subspace
of the gradient signal and then runs an optimizer in that subspace.  T4 mainly
changes how the state itself is stored, shared, quantized, or released.  In
the four-axis view, T4 is the most direct instantiation of state compression in Axis~II.

\textbf{T5: Curvature-aware and geometric regularization.}
T5 does not primarily introduce a new element-wise moment estimator or matrix
transform.  Instead, it adds geometric constraints near update finalization.
Sharpness-aware methods change the point at which gradients are evaluated.
Hessian-guided methods replace or augment curvature signals and often add
trust-region clipping.  Post-processing methods centralize, project,
normalize, mask, or sparsify an already computed update.  Layer-wise
trust-region methods rescale updates according to parameter and update norms.
Because many T5 methods are wrappers around a base optimizer, the base
optimizer and the T5 mechanism must be recorded separately.

Boundary cases are resolved by the primary incremental mechanism.  Direct
Adam variants remain T1 even if they claim stability or generalization
benefits, because their main mechanism is still element-wise moment
estimation or step-size control.  Q-GaLore is T4.2 when the focus is the
additional quantized-state mechanism, while base GaLore remains T2.3.  Magma and MGUP are T5.3 because their primary operation is a selective
update mask.  SAM, Cautious optimizers, and
LAMB are classified by wrapper-level contributions in T5 rather than by their
possible AdamW or Lion base optimizers.  This keeps
the main survey families mutually exclusive while leaving room for secondary
tags in the appendix-level optimizer matrix.

\subsection{Objective-Oriented Taxonomy}
\label{sec:taxonomy-objective}

Dimension B describes what an optimizer is intended to improve.  We organize
effect objectives by evaluation cost rather than by method mechanism, because
the feasibility of a benchmark depends on how each objective is measured.  We
use six objectives: O1 convergence efficiency, O2 per-step computational
cost, O3 memory overhead, O4 training stability, O5 hyperparameter
robustness, and O6 generalization quality.  These labels are not mutually
exclusive and should not be read as intrinsic family rankings.  They are
evaluation axes for organizing experiments and interpreting trade-offs.

The six objectives form three measurement layers.

\textbf{Layer 1: directly logged metrics.}
O1--O3 can be obtained from a single training run, timers, profilers, or
analytic byte/FLOP models.  O1 measures how fast loss decreases or a target
quality is reached under a fixed step, token, wall-clock, or compute budget.
O2 measures additional per-step computation relative to AdamW or SGD,
including matrix multiplications, orthogonalization iterations, extra
forward/backward passes, synchronization, and quantization/dequantization.
O3 measures memory from parameters, gradients, optimizer states, temporary
matrices, projection bases, and quantization buffers.

\textbf{Layer 2: single-run derived metrics.}
O4 and the lightweight version of O6 require post-processing of standard
training logs.  O4 measures whether the trajectory is robust to loss spikes,
gradient-norm bursts, overflow, divergence, and incomplete runs.  The
lightweight O6 signal comes from validation loss, train-validation gap, or a
fixed validation protocol during pretraining.

\textbf{Layer 3: cross-configuration metrics.}
O5 and the full version of O6 require multiple training runs or downstream
evaluations.  O5 measures sensitivity to learning rate, weight decay,
momentum, warmup, batch size, and method-specific hyperparameters.  Full O6
measures downstream transfer, out-of-distribution retention, scale transfer,
or other quality metrics beyond pretraining loss.

\begin{table*}[!t]
  \centering
  \caption{Dimension-B effect objectives and measurement sources.}
  \label{tab:effect_objectives}
  \scriptsize
  \setlength{\tabcolsep}{1.5pt}
  \renewcommand{\arraystretch}{1.08}
  \begin{tabularx}{\textwidth}{@{}Y p{0.17\textwidth}p{0.14\textwidth}Y@{}}
    \toprule
    Definition & Data source & Extra cost & Typical outputs \\
    \midrule
    \rowcolor{famO1!18}\multicolumn{4}{c}{\textbf{\textcolor{famO1!62!black}{O1: Convergence Efficiency}}} \\
      Loss reduction and time-to-target under a fixed training budget
      & Train/validation loss logs
      & None
      & Final loss; steps-to-threshold; token efficiency \\
    \rowcolor{famO2!18}\multicolumn{4}{c}{\textbf{\textcolor{famO2!62!black}{O2: Step Cost}}} \\
      Extra per-step computation and synchronization relative to a baseline
      & Timers, FLOP analysis
      & Recorded during training
      & Step time; FLOPs; extra backward count \\
    \rowcolor{famO3!18}\multicolumn{4}{c}{\textbf{\textcolor{famO3!62!black}{O3: Memory}}} \\
      Memory from optimizer states and associated buffers
      & Memory profiler, byte model
      & Recorded during training
      & Peak memory; state and buffer bytes \\
    \rowcolor{famO4!18}\multicolumn{4}{c}{\textbf{\textcolor{famO4!62!black}{O4: Stability}}} \\
      Robustness to spikes, divergence, and gradient fluctuations
      & Loss and gradient-norm curves
      & Offline post-processing
      & Spike rate; gradient CV; divergence rate \\
    \rowcolor{famO5!18}\multicolumn{4}{c}{\textbf{\textcolor{famO5!62!black}{O5: Hyperparameter Robustness}}} \\
      Sensitivity to learning rate, decay, batch size, and other knobs
      & Multiple training runs
      & Search or transfer experiments
      & Usable LR interval; performance variance; tuning burden \\
    \rowcolor{famO6!18}\multicolumn{4}{c}{\textbf{\textcolor{famO6!62!black}{O6: Generalization}}} \\
      Quality beyond the training objective (validation, downstream, OOD, transfer)
      & Validation and downstream evaluation
      & Low for validation, high for full evaluation
      & Validation loss; generalization gap; downstream score \\
    \bottomrule
  \end{tabularx}
\end{table*}

Dimension B must remain neutral before benchmarking.  T2 methods may improve
O1 token efficiency through matrix structure, but their orthogonalization or
Kronecker operations may hurt O2.  T4 methods are designed for O3, but
aggressive compression may degrade O1 or O4.  T5 methods often target O4 and
O6, but SAM-style extra gradient evaluations impose a clear O2 cost.  Thus
effect labels are measurement prompts rather than substitutes for empirical
results.

For the family analyses in Section~\ref{sec:optimizer-families}, each
optimizer or subclass can be annotated by
\begin{equation}
  \operatorname{Effect}(A) \subseteq \{O1,O2,O3,O4,O5,O6\},
  \label{eq:effect-tags}
\end{equation}
where $A$ denotes an optimizer or subclass.  The annotation should distinguish
design intent from experimentally verified effect.  AdaFactor, for instance,
can be marked as O3-targeting under an analytic memory model, but whether it
also improves O1, O4, or O6 under a fixed training protocol must be decided
by the benchmark.

\subsection{Cross-Dimension Analysis}
\label{sec:taxonomy-cross}

The value of the dual taxonomy appears in cross-dimension analysis.
Dimension A gives mechanism labels, while Dimension B gives effect objectives.
Table~\ref{tab:taxonomy_cross_matrix} provides a compact matrix of the
objectives that each method family should emphasize in the benchmark.  The
matrix is a mechanism-informed prior, not a final empirical conclusion.

\begin{table}[!t]
  \centering
  \caption{Compact cross-matrix between method families and effect
  objectives.}
  \label{tab:taxonomy_cross_matrix}
  \small
  \newcommand{\ppos}{\textcolor{green!35!black}{\ensuremath{\boldsymbol{\Uparrow}}}}
  \newcommand{\pos}{\textcolor{green!35!black}{\ensuremath{\boldsymbol{\Uparrow}}}}
  \newcommand{\neu}{\tikz[baseline=-0.55ex]\draw[gray,line width=0.45pt] (0,0) circle (0.62ex);}
  \newcommand{\negv}{\textcolor{red!70!black}{\ensuremath{\boldsymbol{\Downarrow}}}}
  \newcommand{\posneu}{\textcolor{green!65!black}{\ensuremath{\uparrow}}}
  \newcommand{\neupos}{\textcolor{green!65!black}{\ensuremath{\uparrow}}}
  \newcommand{\neuneg}{\textcolor{red!55}{\ensuremath{\downarrow}}}
  \newcommand{\negneu}{\textcolor{red!55}{\ensuremath{\downarrow}}}
  \setlength{\tabcolsep}{5pt}
  \begin{tabularx}{\linewidth}{@{}>{\raggedright\arraybackslash}X*{6}{>{\centering\arraybackslash}m{0.065\linewidth}}@{}}
    \toprule
    Family & O1 & O2 & O3 & O4 & O5 & O6 \\
    \midrule
    T1 Element-wise adaptive moment and scalar control & \ppos & \neu & \negv & \pos & \pos & \neupos \\
    T2 Matrix-level structural methods & \ppos & \negv & \negneu & \pos & \neupos & \neupos \\
    T3 Discretized directions & \posneu & \pos & \pos & \posneu & \neu & \neu \\
    T4 State compression & \neuneg & \neupos & \ppos & \neuneg & \neu & \neuneg \\
    T5 Geometry regularization & \posneu & \negv & \neuneg & \ppos & \pos & \ppos \\
    \bottomrule
  \end{tabularx}
  \vspace{2pt}
  \begin{minipage}{0.98\linewidth}
    \footnotesize \emph{Note.} Dark double-line arrows denote strong priors:
    \ppos{} indicates a favorable target and \negv{} a likely cost. Pale
    single-line arrows denote conditional or protocol-sensitive favorable/cost
    effects. The hollow gray circle \neu{} denotes protocol-dependent
    neutrality. Arrows indicate objective favorability rather than raw metric
    direction.
  \end{minipage}
\end{table}

The symbols in Table~\ref{tab:taxonomy_cross_matrix} should not be read as a
global performance ranking.  A T2 method may be slower per step but still win
in wall-clock time if token efficiency improves enough.  A T4 method may look
worse in a fixed-small-model loss comparison while enabling a larger batch or
model under the same hardware budget.  A T5 method may improve generalization
only at particular data scales, training durations, or evaluation suites.

The cross view also identifies mechanisms that are likely to compose
orthogonally.  Low-rank projection and state quantization can be combined
because T2.3 changes the signal subspace, whereas T4.2 changes the state
representation.  Variance-reduced gradient estimates can be layered onto
AdamW, Lion, or Shampoo because variance reduction (Axis~II) is mostly independent of basis choice
and curvature representation.  Post-update filters can wrap AdamW or Lion
because T5.3 acts after the base direction has been computed.  Layer-wise
trust ratios can similarly wrap an element-wise adaptive update because they
primarily act in S5.

Not every cross-family combination is natural.  Mechanisms that occupy the
same meta-pipeline slot require an explicit ordering and a descent-quality
argument.  Spectral orthogonalization and low-rank projection are both strong
S2 constraints on a matrix signal, so combining them requires specifying whether
projection precedes orthogonalization or vice versa.  LOMO-style streaming
updates can conflict with methods that need global gradient statistics,
delayed basis updates, or a second gradient evaluation.  SAM-style methods
already require extra forward/backward computation, so adding expensive
Kronecker-factored preconditioning may be mathematically valid but practically
unattractive.  These constraints are not implementation details.  They are
composition boundaries that the taxonomy should make explicit.

The full appendix-level matrix should therefore record, for each optimizer,
three fields: the primary Dimension-A label, secondary mechanism tags, and
O1--O6 effect annotations.  The main text keeps only compact summaries such
as Table~\ref{tab:taxonomy_cross_matrix}; the family analyses instantiate the
matrix progressively by describing each subclass's meta-pipeline site,
four-axis coordinates, and expected benchmark signature.  In this way, the
taxonomy is a mechanism-aware
benchmark plan: a new optimizer entering the framework must specify which
pipeline stage it modifies, which four-axis coordinates it changes, and which
effect objectives it claims to improve.

\section{Optimizer Method Families}
\label{sec:optimizer-families}

The five method families below form the mechanism-oriented core of the
survey.  Rather than treating T1--T5 as separate top-level sections, we place
them under one common section and use the same internal template for each
family: meta-pipeline position, LMO-driven four-axis interpretation,
representative methods, and effect-target assessment. This keeps the main progression of the paper compact while preserving the comparison granularity needed for method-level analysis and benchmark design.

\subsection{T1: Element-Wise Adaptive Moment and Scalar Control}
\label{sec:t1}

\subsubsection{Family Overview and Meta-Pipeline Position}
\label{sec:t1-overview}

Element-wise adaptive-moment methods are the default optimizer family for
modern Transformer training.  Their defining property is not the mere presence
of momentum, but the preservation of an element-wise update topology: the
optimizer maintains scalar or coordinate-wise state, normalizes or controls
each parameter independently, and avoids exchanging information across rows,
columns, heads, or layers.  T1 therefore contains the Adam-centered adaptive
moment line together with scalar first-order predecessors and outer-control
schemes whose primary mechanism remains element-wise or scalar.  Methods whose
main contribution is matrix routing, irreversible direction discretization,
state compression, or sharpness-aware wrapper gradients are assigned to T2--T5
instead.

\begin{figure*}[t]
  \centering
  \includegraphics[width=0.75\textwidth]{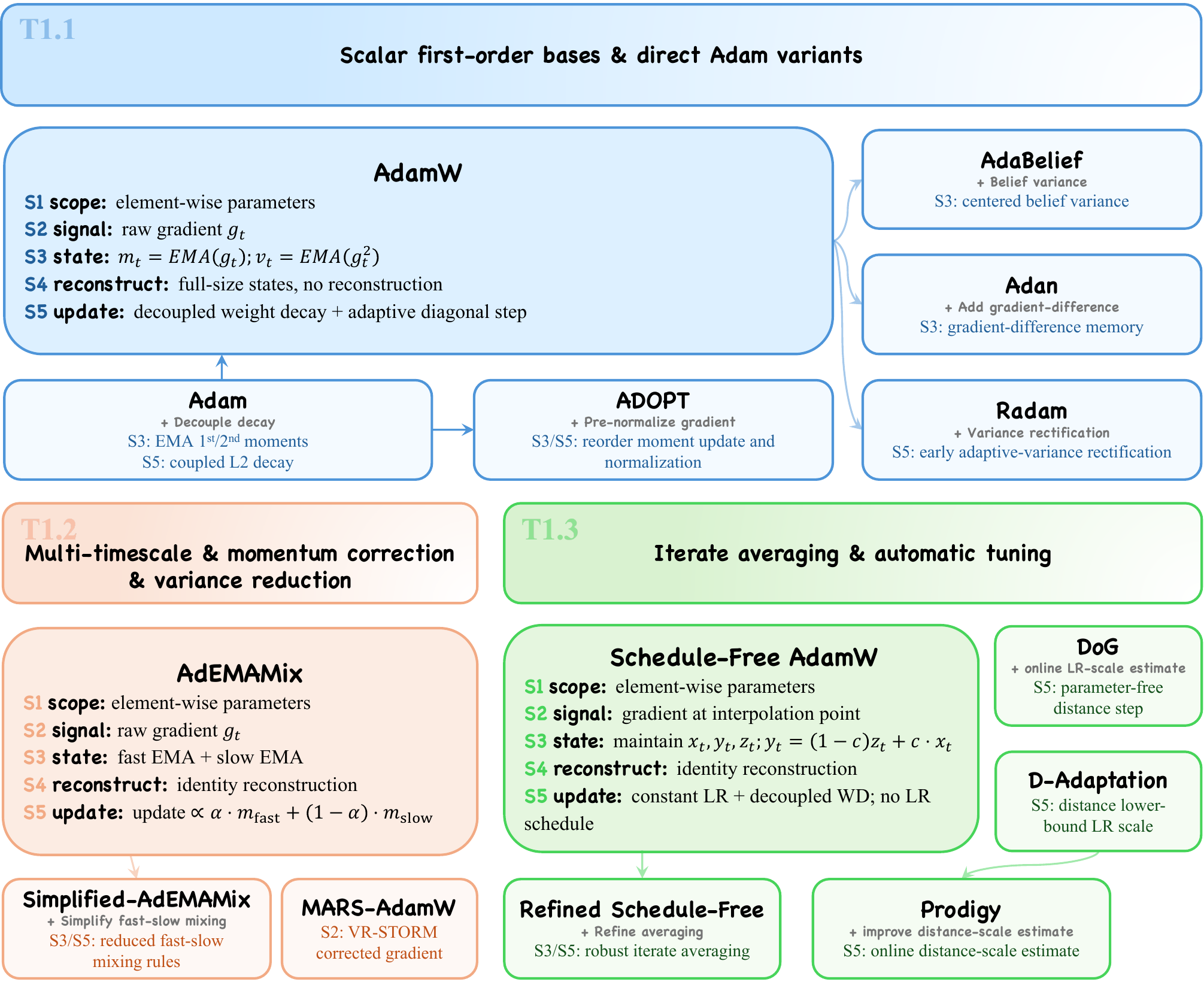}
  \caption{Mechanism overview of T1 element-wise adaptive-moment and scalar
  control methods.  The family is organized by the dominant intervention in
  the update: direct scalar or Adam-style variants, additional temporal or
  estimator-correction channels, and outer iterate or global-scale control.
  Boundary methods are shown only when they preserve scalar or element-wise
  update topology.}
  \label{fig:t1_family_mechanism}
\end{figure*}

\definecolor{hidden-blue}{HTML}{4a90e2}
\definecolor{hidden-black}{HTML}{333333}
\tikzstyle{my-box}=[
  rectangle,
  draw=hidden-black,
  rounded corners,
  text opacity=1,
  minimum height=1.5em,
  minimum width=5em,
  inner sep=2pt,
  align=center,
  fill opacity=.5,
]
\tikzstyle{root}=[
  align=center,
]
\tikzstyle{leaf}=[
  my-box,
  minimum height=1.5em,
  text=black,
  font=\normalsize,
  inner xsep=5pt,
  inner ysep=4pt,
  text width=10.5em,
  fill opacity=1,
]
\tikzstyle{leaf1}=[
  my-box,
  minimum height=1.5em,
  fill=yellow!32,
  text=black,
  font=\normalsize,
  inner xsep=5pt,
  inner ysep=4pt,
  text width=17em,
  align=left,
]
\tikzstyle{leaf2}=[
  my-box,
  minimum height=1.5em,
  fill=hidden-blue!57,
  text=black,
  font=\normalsize,
  inner xsep=5pt,
  inner ysep=4pt,
  text width=17em,
  align=left,
]
\tikzstyle{leaf3}=[
  my-box,
  minimum height=1.5em,
  fill=purple!27,
  text=black,
  font=\normalsize,
  inner xsep=5pt,
  inner ysep=4pt,
  text width=17em,
  align=left,
]

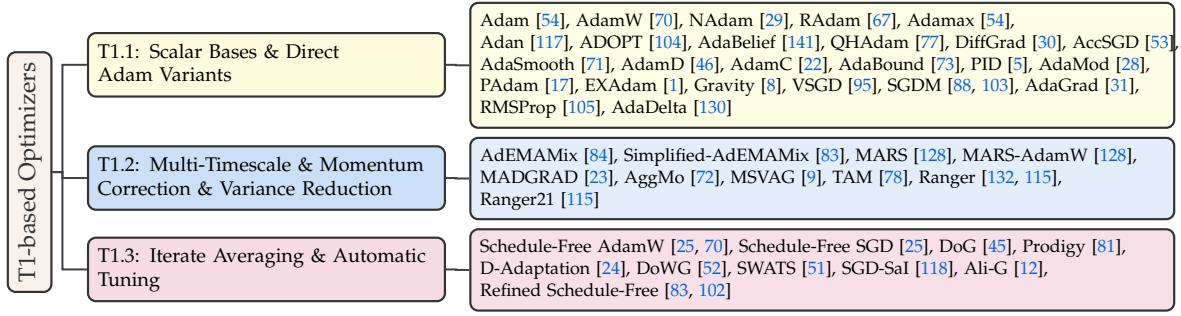
\begin{figure*}[!htbp]
\centering
\resizebox{\textwidth}{!}{
\begin{forest}
forked edges,
for tree={
grow=east,
reversed=true,
anchor=base west,
parent anchor=east,
child anchor=west,
base=left,
font=\large,
rectangle,
draw=hidden-black,
rounded corners,
align=left,
minimum width=4em,
edge+={darkgray, line width=1pt},
s sep=4pt,
inner xsep=5pt,
inner ysep=4pt,
line width=1.1pt,
ver/.style={rotate=90, child anchor=north, parent anchor=south, anchor=center},
},
where level=1{text width=11em, font=\normalsize}{},
where level=2{text width=34.5em, tier=citations, font=\small}{},
[
    T1-based Optimizers,
    ver, fill=brown!10!white, text=hidden-black, root
    [
        T1.1: Scalar Bases \& Direct \\ Adam Variants,
        fill=yellow!32, leaf1
        [{
            Adam~\cite{kingma2015adam},
            AdamW~\cite{loshchilov2019decoupled},
            NAdam~\cite{dozat2016nadam},
            RAdam~\cite{liu2020radam},
            Adamax~\cite{kingma2015adam}, \\
            Adan~\cite{xie2022adan},
            ADOPT~\cite{taniguchi2024adopt},
            AdaBelief~\cite{zhuang2020adabelief},
            QHAdam~\cite{ma2019qhadam},
            DiffGrad~\cite{dubey2019diffgrad},
            AccSGD~\cite{kidambi2018insufficiency}, \\
            AdaSmooth~\cite{lu2023adasmooth},
            AdamD~\cite{john2021adamd},
            AdamC~\cite{defazio2025gradients},
            AdaBound~\cite{luo2019adabound},
            PID~\cite{an2018pid},
            AdaMod~\cite{ding2019adamod}, \\
            PAdam~\cite{chen2018padam},
            EXAdam~\cite{adly2024exadam},
            Gravity~\cite{bahrami2021gravity},
            VSGD~\cite{schaul2013nomore},
            SGDM~\cite{polyak1964some,sutskever2013importance},
            AdaGrad~\cite{duchi2011adaptive}, \\
            RMSProp~\cite{tieleman2012rmsprop},
            AdaDelta~\cite{zeiler2012adadelta}
        }, fill=yellow!12]
    ]
    [
        T1.2: Multi-Timescale \& Momentum \\ Correction \& Variance Reduction,
        fill=hidden-blue!57, leaf2
        [{
            AdEMAMix~\cite{pagliardini2024ademamix},
            Simplified-AdEMAMix~\cite{morwani2025connections},
            MARS~\cite{yuan2024mars},
            MARS-AdamW~\cite{yuan2024mars}, \\MADGRAD~\cite{defazio2022madgrad},
            AggMo~\cite{lucas2018aggmo},
            MSVAG~\cite{balles2018dissecting},
            TAM~\cite{malviya2024torque},
            Ranger~\cite{zhang2019lookahead,wright2021ranger21}, \\
            Ranger21~\cite{wright2021ranger21}
        }, fill=hidden-blue!15]
    ]
    [
        T1.3: Iterate Averaging \& Automatic \\ Tuning,
        fill=purple!27, leaf3
        [{
            Schedule-Free AdamW~\cite{defazio2024road,loshchilov2019decoupled},
            Schedule-Free SGD~\cite{defazio2024road},
            DoG~\cite{ivgi2023dog},
            Prodigy~\cite{mishchenko2023prodigy}, \\
            D-Adaptation~\cite{defazio2023dadaptation},
            DoWG~\cite{khaled2023dowg},
            SWATS~\cite{keskar2017swats},
            SGD-SaI~\cite{xu2024no},
            Ali-G~\cite{berrada2020training}, \\
            Refined Schedule-Free~\cite{morwani2025connections,song2026through}
        }, fill=purple!12]
    ]
  ]
\end{forest}
}
\caption{Taxonomy of element-wise adaptive-moment and scalar-control optimizers.}
\label{tab:t1_family_taxonomy}
\end{figure*}

\paragraph{Subclass structure.}
Figure~\ref{tab:t1_family_taxonomy} gives the complete optimizer membership for
T1.  The three subclasses below are therefore used to interpret the taxonomy
figure, rather than to repeat its entries.  The split is mechanism-based
rather than chronological: each subclass identifies the dominant non-identity
operation added around the AdamW coordinate-wise template.
\begin{enumerate}[label=\textbf{T1.\arabic*},leftmargin=*,itemsep=2pt]
  \item \textbf{Scalar bases and direct Adam variants.}
  This subclass groups optimizers whose main change remains local to the
  coordinate-wise template: the numerator, diagonal scale, bias correction,
  decay finalization, or scalar control law.  It includes both the AdamW line
  and historical scalar or diagonal predecessors, but the latter are used as
  conceptual anchors rather than as evidence that all T1 methods are modern
  large-model optimizers.
  \item \textbf{Multi-timescale, momentum correction, and variance reduction.}
  This subclass changes the signal that enters state evolution.  Its members
  add fast/slow temporal channels, replace the raw mini-batch gradient with a
  corrected or variance-reduced estimator, gate momentum by temporal
  alignment, or bundle such components around an Adam or RAdam core.
  \item \textbf{Iterate averaging and automatic tuning.}
  This subclass leaves the scalar or element-wise base update largely intact
  and instead modifies the outer control system: iterate averaging, global
  learning-rate scale inference, initialization-time scaling, or
  optimizer-phase switching.
\end{enumerate}

For a parameter vector
$\theta_t \in \mathbb{R}^d$ and stochastic gradient $g_t$, the canonical
AdamW update maintains
\begin{align}
  m_t &= \beta_1 m_{t-1} + (1-\beta_1) g_t, \quad v_t = \beta_2 v_{t-1} + (1-\beta_2) g_t \odot g_t, \\
  u_t &= \frac{\hat{m}_t}{\sqrt{\hat{v}_t}+\epsilon}, \quad
  \theta_{t+1} = (1-\eta_t \lambda)\theta_t - \eta_t u_t,
  \label{eq:t1-adamw-template}
\end{align}
where hats denote bias correction and weight decay is decoupled from the
adaptive direction~\citep{kingma2015adam,loshchilov2019decoupled}.  The update
constructs a diagonal preconditioner from the second-moment state and leaves
the parameter topology unchanged.  This coordinate-wise invariance is the
central commonality of T1: individual methods may change the numerator, the
variance proxy, bias correction, decay finalization, temporal state, averaged
iterate, or global step-size scale, but they do not introduce matrix-level
basis changes, low-rank reconstruction, quantized state representation, or
sharpness-aware auxiliary gradients.  Scalar-control boundary methods can be
viewed as setting the adaptive denominator to an identity scale, while
historical adaptive-rate methods provide the diagonal-statistics lineage from
which Adam-style updates emerged.

In the meta-pipeline of Section~\ref{sec:meta-pipeline}, T1 is mostly
identity-like in S1, S2, and S4.  Parameters are routed at their native
coordinate granularity (S1), the stochastic gradient is usually passed directly
into the update rule (S2), and the reconstructed update remains
full-dimensional (S4).  The primary non-identity operation is S3 state
evolution: the optimizer specifies which moving averages, variance proxies,
time scales, cross-moment corrections, variance-reduced estimators, or
averaged iterates are maintained.  S5 supplies the second locus of variation:
AdamW decouples weight decay, RAdam rectifies early adaptive variance, AdamC
adjusts decay--schedule interactions, D-Adaptation and Prodigy infer a global
learning-rate scale, and Schedule-Free methods internalize the role of an
external learning-rate schedule through iterate averaging.

\subsubsection{LMO-Driven Four-Axis Interpretation}
\label{sec:t1-lmo-four-axis}

In the four-axis view, the common T1 coordinate is
\begin{equation}
  \big((I,I),\ \psi,\ \hat{H}_t,\ \alpha,\ \widehat{g}_t,\ \mathrm{Full}\big),
\end{equation}
where the analysis basis is the identity, the curvature estimate is diagonal,
and the state is stored at full coordinate resolution.  AdamW corresponds to
$\hat{H}_t=\operatorname{diag}(\hat{v}_t)$ and
$\alpha=\tfrac{1}{2}$, yielding the square-root preconditioned direction in
Eq.~\eqref{eq:t1-adamw-template}.  Under the Fisher identity discussed in
Section~\ref{sec:prelim-background}, the EMA of $g_t \odot g_t$ can be read as
a diagonal Fisher or Gauss--Newton surrogate.  This makes the family a
diagonal-curvature analogue of the matrix preconditioners in T2, but with no
off-diagonal coupling and no basis change.

The LMO interpretation clarifies what is shared and what is not shared across
T1.  If the adaptive denominator is temporarily fixed, the numerator selects a
coordinate-wise descent pattern.  In the limiting normalized view, a signed
coordinate direction is the LMO solution over an $\ell_\infty$-type ball, and
the diagonal denominator deforms that ball into an anisotropic, coordinate-wise
trust region.  AdamW should therefore not be reduced to either ``sign descent''
or ``Newton descent.''  Its practical update combines an element-wise
directional component, a diagonal curvature scale, bias correction, damping,
and a global step-size schedule.  This is why T1 methods often share the same
Axis~I basis and Axis~III exponent while differing mainly in Axis~II
curvature/variance estimation and gradient-estimation quality, and S5
step-size or regularization control.  Boundary methods with no adaptive
denominator can be viewed as using $\hat{H}_t=I$ and $\alpha=0$; they are
kept in T1 only as scalar first-order predecessors or control schemes, not as
evidence that every T1 method is an adaptive-moment estimator.

This view also separates T1 from neighboring families.  Lion belongs to T3
because its defining operation is an irreversible sign map and it removes the
second-moment state, even though it inherits an Adam-like momentum
interface~\citep{chen2023symbolic}.  AdaFactor belongs to T4 because its
defining contribution is factorized storage of the second-moment state, not a
new full-state moment estimator.  Sophia belongs to T5 because it replaces the
Adam variance proxy with an explicit curvature estimate and clipped
trust-region update~\citep{liu2023sophia}.  T1 is therefore the baseline
coordinate system from which many later optimizers depart.

\subsubsection{Representative Methods}
\label{sec:t1-methods}

Against the subclass map above, the representative methods are organized by
three recurring mechanism patterns: local modifications of AdamW's numerator,
denominator, bias correction, or writeback; additional temporal or
variance-reduced signals; and outer rules for averaging, global scale
inference, or switching.

\paragraph{Scalar Bases and Direct Adam Variants.}
\label{sec:t1-standard-variants}

The first branch is the broadest part of T1.  Its members are best read in
tiers, because some are historical scalar bases, some define the modern Adam
baseline, and some are local corrections to one term of the AdamW template:
at the survey level, they can be summarized as
\begin{equation}
  u_t = \frac{a_t}{b_t+\epsilon}, \quad
  \theta_{t+1} = \theta_t - \eta_t u_t - \eta_t \lambda_t \chi_t(\theta_t),
  \label{eq:t1-direct-variant-template}
\end{equation}
where $a_t$ is a momentum-like numerator, $b_t$ is a diagonal scale or
variance proxy, and $\chi_t$ specifies whether decay is coupled to the
gradient channel or decoupled at writeback.
\begin{itemize}
  \setlength{\itemsep}{2pt}
  \item \textbf{Scalar and diagonal predecessors.}  SGDM introduced a
  heavy-ball momentum buffer for smoother descent
  ~\citep{polyak1964some,sutskever2013importance}; AdaGrad accumulated squared
  gradients to produce coordinate-wise learning rates~\citep{duchi2011adaptive};
  RMSProp replaced AdaGrad's cumulative sum with a running average
  ~\citep{tieleman2012rmsprop}; AdaDelta used ratios of running RMS quantities
  to reduce manual learning-rate dependence~\citep{zeiler2012adadelta}; and
  VSGD estimated local gradient statistics to adapt learning rates
  ~\citep{schaul2013nomore}.  These methods are not all Adam-like, but they
  form the scalar and diagonal-control substrate from which Adam-style methods
  developed.
  \item \textbf{Adam baseline line.}  Adam combined first-moment momentum,
  RMSProp-style second moments, and bias correction into a single adaptive
  method~\citep{kingma2015adam}; Adamax replaces the second-moment denominator
  with an exponentially weighted $\ell_\infty$ norm; and AdamW became the
  practical large-model baseline by decoupling weight decay from the adaptive
  gradient normalizer~\citep{loshchilov2019decoupled}.  In the meta-pipeline,
  AdamW is mainly an S5 change: the direction $u_t$ is computed as in Adam,
  while shrinkage is applied directly to parameters rather than being mixed
  into $g_t$.
  \item \textbf{High-salience direct Adam variants.}  NAdam inserts a
  Nesterov-style lookahead term into Adam's first-moment pathway
  ~\citep{dozat2016nadam}.  RAdam adds a rectification factor for the
  high-variance early adaptive denominator~\citep{liu2020radam}.  AdaBelief
  replaces $g_t^2$ with the centered prediction-error statistic $(g_t-m_t)^2$
  ~\citep{zhuang2020adabelief}.  Adan adds a Nesterov-inspired
  gradient-difference channel using both $g_t$ and $g_t-g_{t-1}$
  ~\citep{xie2022adan}.  ADOPT changes the ordering of second-moment update and
  normalization to obtain guarantees for arbitrary $\beta_2$
  ~\citep{taniguchi2024adopt}.  EXAdam adds adaptive cross-moment debiasing and
  gradient-based acceleration while remaining a direct Adam extension
  ~\citep{adly2024exadam}.
  \item \textbf{Local adaptivity and correction variants.}  QHAdam interpolates
  between immediate and momentum-based directions through quasi-hyperbolic
  weighting~\citep{ma2019qhadam}; AdaMod bounds Adam's adaptive learning rates
  using a momental upper bound~\citep{ding2019adamod}; and PAdam interpolates
  between Adam (or AMSGrad) and SGDM through a partial-adaptivity exponent
  ~\citep{chen2018padam}.  DiffGrad and AdaSmooth adapt learning rates from
  gradient-change or effective-ratio signals
  ~\citep{dubey2019diffgrad,lu2023adasmooth}; AdamD modifies bias correction
  ~\citep{john2021adamd}; and AdamC corrects decay--schedule interactions near
  the end of long-duration training~\citep{defazio2025gradients}.
  \item \textbf{Boundary scalar-control methods.}  AccSGD, PID, and Gravity
  remain in T1.1 only as scalar-control boundary cases
  ~\citep{kidambi2018insufficiency,an2018pid,bahrami2021gravity}.  They help
  explain the control-theoretic and acceleration background of the family, but
  they should not receive the same narrative weight as AdamW, RAdam, AdaBelief,
  Adan, or ADOPT when the target setting is modern LLM training.
\end{itemize}

This tiered reading prevents T1.1 from becoming a flat list of names.  The
central comparison is AdamW versus direct modifications of AdamW's numerator,
denominator, bias correction, or writeback rule; the historical and boundary
methods provide context and ablation anchors.  Several direct variants can be
viewed as substitutions inside Eq.~\eqref{eq:t1-direct-variant-template}; for
example,
\begin{align}
  v_t^{\mathrm{Belief}}
    &= \beta_2 v_{t-1}^{\mathrm{Belief}}
       + (1-\beta_2)(g_t-m_t)\odot(g_t-m_t), \\
  u_t^{\mathrm{RAdam}}
    &= r_t \frac{\hat{m}_t}{\sqrt{\hat{v}_t}+\epsilon}, \quad
  u_t^{\mathrm{PAdam}}
    = \frac{\hat{m}_t}{(\hat{v}_t)^p+\epsilon}.
  \label{eq:t1-direct-substitutions}
\end{align}
These formulas do not make AdaBelief, RAdam, and PAdam the same method; they
show that their primary changes remain local to the diagonal variance proxy,
early-step rectification, or adaptivity exponent.

\paragraph{Multi-Timescale, Momentum Correction, and Variance Reduction.}
\label{sec:t1-multitimescale}

The second branch asks whether a single raw EMA stream is enough.  A fast
momentum reacts to the current mini-batch distribution but is noisy; a slow
momentum is stable but can lag behind regime changes; and a variance-reduced
or alignment-corrected signal can change the quality of the gradient entering
state evolution.  We therefore separate the branch into four mechanism groups:
\begin{equation}
  \widehat{g}_t = \mathcal{C}_t(g_t,g_{t-1},m_{t-1}), \quad
  m_t^{(k)} = \beta_k m_{t-1}^{(k)} + (1-\beta_k)\widehat{g}_t,\quad
  a_t = \sum_{k=1}^{K}\omega_k m_t^{(k)}.
  \label{eq:t1-temporal-template}
\end{equation}
Here $\mathcal{C}_t$ denotes a possible estimator correction and the
$m_t^{(k)}$ streams denote different temporal channels.  Standard AdamW is
the special case $K=1$ and $\widehat{g}_t=g_t$.
\begin{itemize}
  \setlength{\itemsep}{2pt}
  \item \textbf{Multi-time-scale momentum.}  AdEMAMix maintains both a standard
  fast EMA and an older, slower EMA, then mixes them in the update direction
  ~\citep{pagliardini2024ademamix}.  This is especially relevant to LLM
  training, where long horizons can make old gradient information useful
  rather than stale.  Simplified-AdEMAMix preserves the interpretation while
  exposing connections to schedule-free and accelerated SGD variants
  ~\citep{morwani2025connections}.  AggMo is the earlier multi-time-scale
  analogue, aggregating several momentum buffers with different decay rates
  ~\citep{lucas2018aggmo}.
  \item \textbf{Variance-reduced or corrected estimators.}  MARS introduces a
  scaled stochastic recursive momentum estimator that can be paired with AdamW,
  Lion, or Shampoo-style bases~\citep{yuan2024mars}.  In the AdamW instance,
  the signal entering S3 is no longer raw $g_t$ but a gradient-difference
  corrected estimator, so the method changes Axis~II while preserving the
  element-wise base update.  MADGRAD combines AdaGrad-style accumulated
  curvature with a momentumized dual-averaged update~\citep{defazio2022madgrad}.
  MSVAG is more diagnostic than deployment driven: it decomposes Adam into
  sign, magnitude, and variance effects~\citep{balles2018dissecting}.
  \item \textbf{Alignment-aware momentum correction.}  TAM is neither a
  multi-time-scale method nor a STORM-style variance-reduction method.  It
  damps momentum injection according to the alignment between current gradients
  and historical momentum~\citep{malviya2024torque}.  Its role in T1.2 is to show
  that momentum can be corrected by geometry of temporal agreement, not only
  by adding an older EMA or recursive estimator.
  \item \textbf{Composite optimizer bundles.}  Ranger combines RAdam,
  Lookahead, and often gradient centralization; Ranger21 combines AdamW with a
  larger set of stabilization components
  ~\citep{zhang2019lookahead,wright2021ranger21}.  These methods are useful
  implementation bundles, but their components occupy different meta-pipeline
  slots.  We keep them connected to T1 through their Adam or RAdam core and record
  Lookahead-style averaging or post-processing as secondary tags.
\end{itemize}

The common thread is that T1.2 no longer trusts one raw momentum stream as the
only temporal signal.  It either adds more time scales, corrects the estimator
fed into S3, or gates momentum by temporal alignment while leaving routing and
writeback largely element-wise.  In this notation, AdEMAMix primarily changes
$K$ and $\omega_k$; MARS changes $\mathcal{C}_t$ before the AdamW-like state
update; and TAM can be read as inserting an alignment-dependent gate, e.g.,
\begin{equation}
  \widehat{g}_t = \kappa_t g_t, \quad
  \kappa_t = f\!\left(
    \frac{\langle g_t,m_{t-1}\rangle}
         {\|g_t\|\,\|m_{t-1}\|+\epsilon}
  \right),
  \label{eq:t1-alignment-gate}
\end{equation}
where Eq.~\eqref{eq:t1-alignment-gate} is a template for the class of
alignment-aware corrections rather than a replacement for the exact algorithm
in each paper.

\paragraph{Iterate Averaging and Automatic Tuning.}
\label{sec:t1-averaging-autotuning}

The third branch does not primarily change the coordinate-wise denominator.
Instead, it changes the outer control logic around a scalar or element-wise
base optimizer: how iterates are averaged, how the global scale is chosen, or
when the optimizer should change phase.  The subclass has three readable
groups:
\begin{equation}
  y_t = (1-\rho_t)x_t+\rho_t \bar{x}_t, \quad
  x_{t+1}=x_t-\eta_t u_t(y_t), \quad
  \eta_t=\eta_{\mathrm{base}}\,d_t,
  \label{eq:t1-outer-control-template}
\end{equation}
where $y_t$ is an interpolation point, $\bar{x}_t$ is an averaged iterate, and
$d_t$ is an inferred global scale.  Schedule-free methods emphasize the first
two terms; learning-rate-free methods emphasize the online construction of
$d_t$; switching methods change which base rule defines $u_t$.
\begin{itemize}
  \setlength{\itemsep}{2pt}
  \item \textbf{Schedule-free iterate systems.}  Schedule-Free optimization
  replaces an externally prescribed learning-rate schedule with an internal
  iterate-averaging mechanism~\citep{defazio2024road}.  In the AdamW instance,
  the optimizer maintains training, evaluation, and averaged iterates, evaluates
  gradients at an interpolation point, and uses a constant learning-rate
  interface with decoupled weight decay.  Refined Schedule-Free methods further
  analyze language-model trajectories and adjust the method for robustness to
  momentum and large-batch settings~\citep{song2026through}.  Schedule-Free SGD
  is kept in the same subclass as the scalar-base analogue.
  \item \textbf{Learning-rate-free scale inference.}  D-Adaptation estimates a
  lower bound on the distance to the solution online and uses it to set the
  learning-rate scale without line search or extra gradient evaluations
  ~\citep{defazio2023dadaptation}.  Prodigy estimates the same distance scale
  more aggressively and can be used with Adam-like adaptive methods
  ~\citep{mishchenko2023prodigy}.  DoG and DoWG form a related
  parameter-free path, setting dynamic step sizes from distance-over-gradient
  quantities rather than a user-specified learning rate
  ~\citep{ivgi2023dog,khaled2023dowg}.
  \item \textbf{Switching, initialization-time scaling, and global clipping.}
  SWATS starts with Adam and switches to SGD when an estimated SGD learning
  rate becomes stable, aiming to combine Adam's fast early progress with
  SGD-like later generalization~\citep{keskar2017swats}.  SGD-SaI avoids
  Adam-like second-moment state during training and uses an initialization-time
  group signal-to-noise estimate to set learning-rate scales
  ~\citep{xu2024no}.  Ali-G computes adaptive global learning rates
  from interpolation assumptions and clipping~\citep{berrada2020training}.
\end{itemize}

T1.3 therefore records a different kind of contribution from T1.1.  The base
update can still be AdamW-like or scalar first-order, but the claimed gain
comes from reducing schedule design, learning-rate search, or phase-selection
burden.  This is why its benchmark claim should be evaluated under O5
hyperparameter robustness as well as O1 convergence.

T1 is the calibration family for effect assessment.  Its role comes less from
minimal resource use than from the reliability of the AdamW-style control
stack: coordinate-wise normalization, momentum smoothing, decoupled weight
decay, and a mature schedule interface.  This stack is expensive compared with
SGDM because AdamW stores two full-size moment buffers, but it often provides
strong early loss reduction and stable Transformer pretraining under sparse,
heterogeneous, or scale-imbalanced gradients.  For this reason, recent LLM
optimizer studies treat AdamW not as a weak default but as the baseline whose
tuning budget, schedule, weight decay, batch size, and model scale determine
whether a proposed optimizer has actually improved the training procedure.
Zhao et al.~\citep{zhao2025deconstructing} make this point by decomposing
AdamW under controlled substitutions, while the benchmarks of Semenov et
al.~\citep{semenov2025benchmarking} and Wen et al.~\citep{wen2025fantastic}
show that optimizer rankings move with scale and protocol choices.

Table~\ref{tab:t1_effect_assessment} is therefore a mechanism-derived
performance prior rather than an empirical ranking.  Each
entry states what the corresponding T1 mechanism is expected to improve, what
trade-off follows from that mechanism, and where the later benchmark provides
supporting evidence or a useful caution.

\begin{table}[!htbp]
  \centering
  \caption{Mechanism-informed effect assessment for T1 subclasses. The entries are design priors for benchmark planning, not final empirical conclusions.}
  \label{tab:t1_effect_assessment}
  \scriptsize
  \setlength{\tabcolsep}{3pt}
  \renewcommand{\arraystretch}{1.08}
  \begin{tabularx}{\linewidth}{@{}Y Y Y@{}}
    \toprule
    Primary target & Likely cost & Benchmark focus \\
    \midrule
    \rowcolor{famT1!18}\multicolumn{3}{c}{\textbf{\textcolor{famT1!62!black}{T1: Element-wise adaptive moment and scalar control}}} \\
    \rowcolor{famT1!8}\multicolumn{3}{@{}l@{}}{\textbf{\textcolor{famT1!62!black}{T1.1 Scalar bases and direct Adam variants}}} \\
      O1 convergence, O4 stability, O5 default robustness
      & O3 two-state memory (lower for scalar predecessors)
      & AdamW-normalized loss, warmup and $\beta_2$ sensitivity, decay coupling \\
    \rowcolor{famT1!8}\multicolumn{3}{@{}l@{}}{\textbf{\textcolor{famT1!62!black}{T1.2 Multi-timescale and variance reduction}}} \\
      O1 convergence, O4 smoother trajectories
      & Extra S3 state
      & Token efficiency, noise sensitivity, batch-size transfer \\
    \rowcolor{famT1!8}\multicolumn{3}{@{}l@{}}{\textbf{\textcolor{famT1!62!black}{T1.3 Iterate averaging and automatic tuning}}} \\
      O5 hyperparameter robustness (secondary O4)
      & Possible O1 loss when schedules are well tuned
      & LR sweep width, schedule transfer, constant-LR comparison \\
    \bottomrule
  \end{tabularx}
\end{table}

\paragraph{Performance implications.}
The three T1 subclasses produce different performance signatures even when
they share an AdamW-like interface.
\begin{itemize}[leftmargin=*,itemsep=2pt]
  \item \textbf{T1.1: scalar bases and direct Adam variants.}
  Coordinate-wise normalization explains the characteristic T1.1 profile:
  fast early loss reduction, stable behavior under heterogeneous coordinate
  scales, and a forgiving default interface.  The cost is full element-wise
  state for Adam-like methods; scalar predecessors reduce state, but typically
  give up part of the robustness that makes AdamW a strong Transformer
  baseline.  This is the role AdamW plays in
  Sec.~\ref{sec:benchmark-tiered-summary}: it is retained as the default
  reference, and T1 remains the stable baseline family rather than the
  strongest raw-quality family.
  \item \textbf{T1.2: multi-timescale and corrected estimators.}
  Adding temporal channels or corrected estimators mainly improves the quality
  of the signal entering the AdamW geometry.  This predicts smoother
  trajectories and better token efficiency when stochastic gradients are noisy
  or temporally biased, but it also predicts extra state, additional update
  logic, or higher per-step cost.  The Stage~2 analysis in
  Sec.~\ref{sec:benchmark-stage2} follows this pattern: MARS-AdamW is the
  most stable AdamW-style enhancement, but the gain comes with more state and
  compute than plain AdamW.
  \item \textbf{T1.3: averaging and automatic tuning.}
  These methods move the contribution from the local direction to the outer
  control loop.  Their expected advantage is not necessarily a lower
  best-tuned loss, but a wider usable range of learning rates, fewer schedule
  decisions, or smoother phase behavior.  The corresponding risk is that an
  automatic rule may trail a carefully tuned fixed schedule when the latter is
  allowed enough search budget.
\end{itemize}

Across all three subclasses, the central question is not whether Adam-style
adaptivity is useful; it is which components of the AdamW control stack remain
necessary once model scale, batch size, training length, and tuning budget are
fixed.  This framing keeps T1 from becoming a list of Adam variants and makes
it the reference point for the remaining families.
\subsection{T2: Matrix-Level Structural Methods}
\label{sec:t2}

\subsubsection{Family Overview and Meta-Pipeline Position}
\label{sec:t2-overview}

T2 contains optimizers whose primary operation uses the matrix structure of
Transformer parameters.  T1 methods flatten a weight tensor and treat it as a
collection of independent coordinates.  T2 methods instead work directly at the
matrix level, coupling rows, columns, singular directions, Kronecker factors,
or low-rank subspaces.
The typical object is a two-dimensional weight matrix
$W_t\in\mathbb{R}^{m\times n}$ with stochastic gradient
$G_t\in\mathbb{R}^{m\times n}$.  The family is therefore tied to parameter
topology.  Attention projections, MLP projections, and other dense matrix
weights are natural targets, whereas biases, normalization parameters,
embeddings, and sometimes output heads fall back to an element-wise optimizer
such as AdamW or SGD.  This matrix routing is the S1 (parameter scoping and
routing) stage of the meta-pipeline, and by deciding which parameters are
treated as matrices it determines whether matrix transforms such as
orthogonalization or Kronecker-factored preconditioning can be applied at all.

\begin{figure*}[t]
  \centering
  \includegraphics[width=0.75\textwidth]{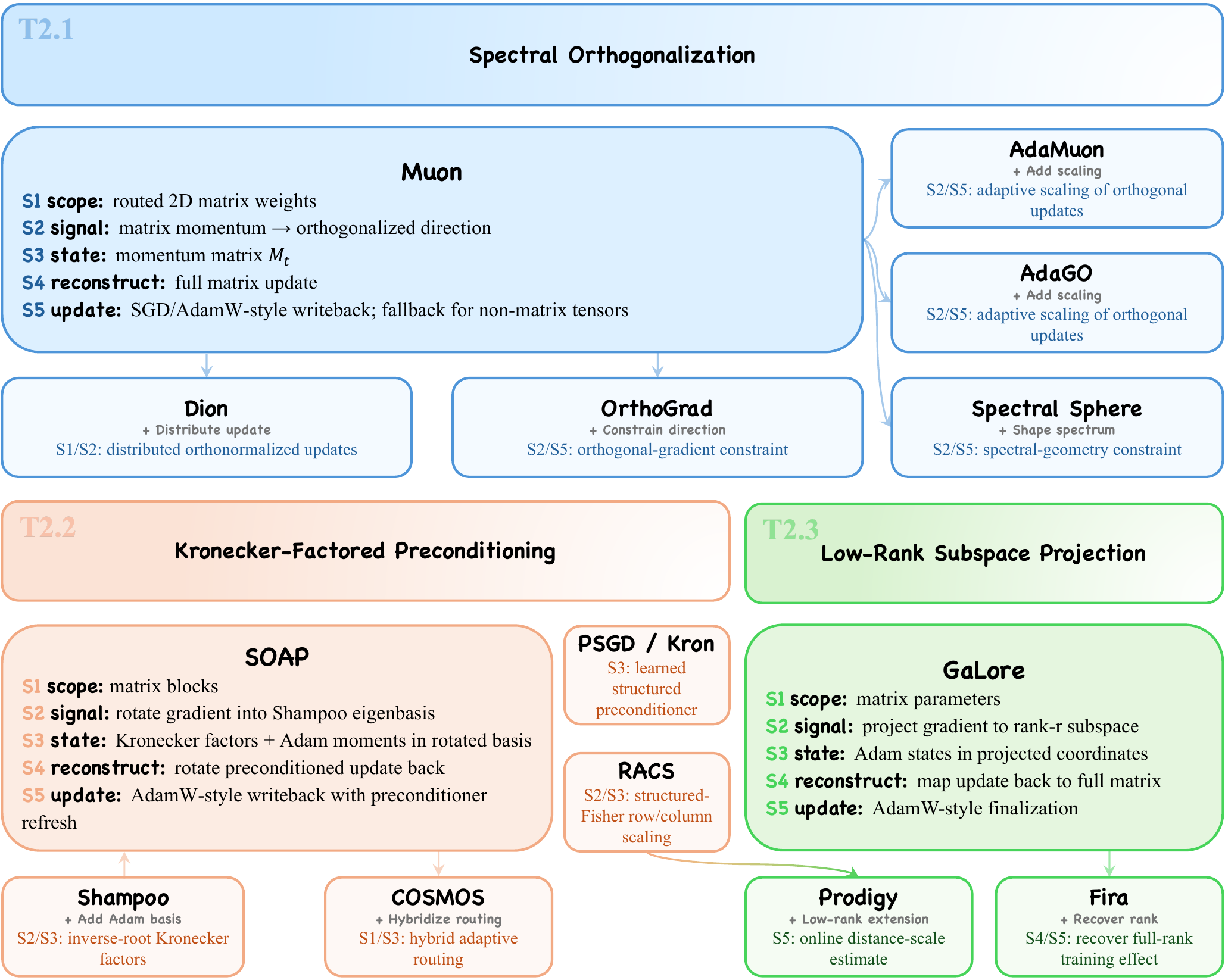}
  \caption{Mechanism schematic for T2 matrix-level structural methods.  The
  schematic summarizes the three matrix routes used in this survey: spectral
  direction selection, Kronecker or structured-Fisher preconditioning, and
  low-rank subspace projection.  It is a taxonomy guide rather than an
  empirical ranking.}
  \label{fig:t2_family_mechanism}
\end{figure*}

\paragraph{Subclass structure and membership.}
Figure~\ref{tab:t2_family_taxonomy} gives the subclass-level optimizer list for T2, and the
three subclasses below specify the corresponding optimizer membership.  The
split is based on the primary matrix operation: spectral direction selection,
structured matrix preconditioning, or subspace projection.
\begin{enumerate}[label=\textbf{T2.\arabic*},leftmargin=*,itemsep=2pt]
  \item \textbf{Spectral orthogonalization.}
  This subclass contains Muon, RMNP~\cite{deng2026rmnp}, MOGA, Dion, AdaMuon, OrthoGrad, AdaGO,
  and Spectral Sphere.  Its members construct an orthogonalized or spectral-geometry update
  from a two-dimensional gradient or momentum object, with Muon as the
  subclass's central representative method for LLMs~\citep{jordan2024muon,pethick2025normlmo,sfyraki2025lions}.
  \item \textbf{Kronecker-factored preconditioning.}
  This subclass contains SOAP, Shampoo, MARS-Shampoo, COSMOS, Kron, PSGD,
  SPlus, and RACS.  Its members estimate row/column curvature, covariance, structured Fisher
  factors, or learned Kronecker-style preconditioners, with Shampoo and SOAP as
  the subclass's core representative methods, around which its mathematical
  framework is built~\citep{gupta2018shampoo,vyas2025soap}.
  \item \textbf{Low-rank subspace projection.}
  This subclass contains GaLore, Fira, and Alice.  Its members project the
  gradient and optimizer state into a low-dimensional subspace, update there,
  and reconstruct a full matrix update, with GaLore as the canonical LLM
  example~\citep{zhao2024galore}.
\end{enumerate}

At the family level, a matrix-structured update can be summarized as
\begin{equation}
  Z_t = B_t^\top G_t C_t,\quad
  \widetilde{\Delta}_t = \Phi_t(Z_t;\mathcal{S}_{t-1}),\quad
  \Delta_t = A_t\widetilde{\Delta}_t D_t^\top,
  \label{eq:t2-matrix-template}
\end{equation}
where $B_t,C_t$ are the left and right analysis bases that rotate or project
the gradient into a new coordinate system, $\Phi_t$ is the direction or
preconditioning rule applied in that coordinate system, $\mathcal{S}_{t-1}$
denotes the structured state, and $A_t,D_t$ map the transformed update back to
the parameter matrix.  Muon uses an SVD or polar basis and flattens the
singular values.  Shampoo and SOAP use Kronecker or Fisher eigenbases.  GaLore
uses projection and reconstruction maps with rank $r\ll\min(m,n)$.  This
template is intentionally broader than any one algorithm.  Its role is to make clear
that T2 changes the geometry in which a matrix update is analyzed before the
usual optimizer writeback is applied.


\definecolor{hidden-blue}{HTML}{4a90e2}
\definecolor{hidden-black}{HTML}{333333}
\tikzstyle{my-box}=[
  rectangle,
  draw=hidden-black,
  rounded corners,
  text opacity=1,
  minimum height=1.5em,
  minimum width=5em,
  inner sep=2pt,
  align=center,
  fill opacity=.5,
]
\tikzstyle{root}=[
  align=center,
]
\tikzstyle{leaf}=[
  my-box,
  minimum height=1.5em,
  text=black,
  font=\normalsize,
  inner xsep=5pt,
  inner ysep=4pt,
  text width=10.5em,
  fill opacity=1,
]
\tikzstyle{leaf1}=[
  my-box,
  minimum height=1.5em,
  fill=yellow!32,
  text=black,
  font=\normalsize,
  inner xsep=5pt,
  inner ysep=4pt,
  text width=17em,
  align=left,
]
\tikzstyle{leaf2}=[
  my-box,
  minimum height=1.5em,
  fill=hidden-blue!57,
  text=black,
  font=\normalsize,
  inner xsep=5pt,
  inner ysep=4pt,
  text width=17em,
  align=left,
]
\tikzstyle{leaf3}=[
  my-box,
  minimum height=1.5em,
  fill=purple!27,
  text=black,
  font=\normalsize,
  inner xsep=5pt,
  inner ysep=4pt,
  text width=17em,
  align=left,
]

\begin{figure*}[!htbp]
\centering
\resizebox{\textwidth}{!}{
\begin{forest}
forked edges,
for tree={
grow=east,
reversed=true,
anchor=base west,
parent anchor=east,
child anchor=west,
base=left,
font=\large,
rectangle,
draw=hidden-black,
rounded corners,
align=left,
minimum width=4em,
edge+={darkgray, line width=1pt},
s sep=4pt,
inner xsep=5pt,
inner ysep=4pt,
line width=1.1pt,
ver/.style={rotate=90, child anchor=north, parent anchor=south, anchor=center},
},
where level=1{text width=11em, font=\normalsize}{},
where level=2{text width=34.5em, tier=citations, font=\small}{},
[
    T2-based Optimizers,
    ver, fill=brown!10!white, text=hidden-black, root
    [
        T2.1: Spectral Orthogonalization,
        fill=yellow!32, leaf1
        [{
            Muon~\cite{jordan2024muon,pethick2025normlmo},
            RMNP~\cite{deng2026rmnp},
            MOGA~\cite{xu2026width},
            Dion~\cite{ahn2025dion},
            AdaMuon~\cite{si2025adamuon},
            OrthoGrad~\cite{hedges2025orthograd}, \\
            AdaGO~\cite{zhang2025adagrad},
            Spectral Sphere~\cite{xie2026controlled}
        }, fill=yellow!12]
    ]
    [
        T2.2: Kronecker-Factored \\ Preconditioning,
        fill=hidden-blue!57, leaf2
        [{
            SOAP~\cite{vyas2025soap},
            Shampoo~\cite{gupta2018shampoo},
            MARS-Shampoo~\cite{yuan2024mars,gupta2018shampoo},
            COSMOS~\cite{liu2025cosmos},
            Kron~\cite{li2017preconditioned}, \\
            PSGD~\cite{li2017preconditioned,li2022black},
            SPlus~\cite{frans2026stable},
            RACS~\cite{gong2025towards}
        }, fill=hidden-blue!15]
    ]
    [
        T2.3: Low-Rank Subspace Projection,
        fill=purple!27, leaf3
        [{
            GaLore~\cite{zhao2024galore},
            Fira~\cite{chen2026fira},
            Alice~\cite{gong2025towards}
        }, fill=purple!12]
    ]
  ]
\end{forest}
}
\caption{Taxonomy of matrix-level structural optimizers.}
\label{tab:t2_family_taxonomy}
\end{figure*}
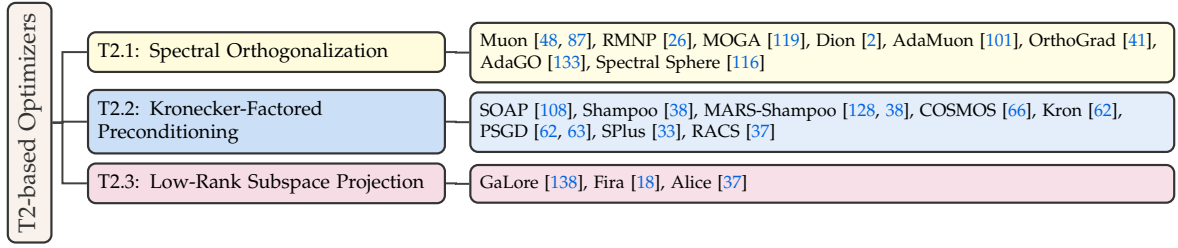

Together, these subclasses make T2 the family where S1 routing, S2
transformation, S3 structured state, and S4 reconstruction all become
first-class design choices.  S5 is usually less distinctive: many T2 methods
still use SGD-style or AdamW-style final writeback after the matrix transform,
possibly with decoupled weight decay, damping, or method-specific scaling.

This definition also fixes the boundary with neighboring families.  Whether a
method belongs to T2 depends on whether its update rule genuinely uses the
matrix geometry.  Conversely, whether a method belongs to T4 likewise depends
on its mechanism.  GaLore does reduce optimizer-state memory, but it does so by
changing the mathematical subspace in which the gradient and Adam state live,
so its primary mechanism is T2.3 (low-rank subspace), and state compression is
only an effect of that choice.  If an additional contribution directly
quantizes, shards, shares, or frees optimizer state, that contribution should
be recorded as a secondary T4 tag.  Likewise,
sharpness-aware wrappers, layer-wise trust ratios, or post-update filters
remain T5 unless the matrix transform itself is the primary new operation.

\subsubsection{LMO-Driven Four-Axis Interpretation}
\label{sec:t2-lmo-four-axis}

The four-axis view in Section~\ref{sec:lmo-four-axis} is especially useful for
T2 because it separates the basis used to analyze a matrix update from the
state used to estimate its scale.  A generic T2 coordinate can be written as
\begin{equation}
  \big((B_t,C_t),\ \psi_t,\ \hat{H}_t,\ \alpha,\ \widehat{G}_t,\ \mathcal{S}_t\big),
\end{equation}
where $B_t$ and $C_t$ are left and right analysis bases, $\psi_t$ is the
matrix direction map, $\hat{H}_t$ is a structured curvature or covariance
proxy when present, and $\mathcal{S}_t$ records whether the optimizer state is
full-rank, factored, or projected.  In T1, $B_t=C_t=I$ and $\mathcal{S}_t$ is
usually full coordinate state.  In T2, the basis itself is part of the method.

For Muon-style spectral-normalization methods, the defining map is a polar or
orthogonalized direction.  With momentum matrix
$M_t=\beta M_{t-1}+(1-\beta)G_t$ and singular value decomposition
$M_t=U_t\Sigma_tV_t^\top$, the update is
\begin{equation}
  W_{t+1}=W_t-\eta_t\,U_tV_t^\top ,
  \label{eq:t2-muon-polar}
\end{equation}
where $U_tV_t^\top$ is usually approximated by a few Newton--Schulz matrix
multiplications rather than an explicit SVD.  This direction is the full polar
factor, so it keeps the singular vectors while flattening every singular value
to one.

Axis~III holds this direction in both the LMO and the preconditioner readings
(Table~\ref{tab:four-axis-instantiation}), and the two agree on the same
$U_tV_t^\top$.  From the LMO side, it is the steepest matrix direction on the
spectral-norm ball,
\begin{equation}
  U_tV_t^\top = \argmax_{\|X\|_{S_\infty}\le 1}\ \langle M_t,X\rangle,
\end{equation}
where $\|\cdot\|_{S_\infty}$ is the spectral norm and
$\langle M_t,X\rangle=\operatorname{tr}(M_t^\top X)$.  By the von Neumann trace
inequality $\langle M_t,X\rangle\le\sum_i\sigma_i(M_t)\sigma_i(X)\le\sum_i\sigma_i(M_t)$,
with equality at $X=U_tV_t^\top$, the steepest direction is exactly the polar
factor.  From the preconditioner side, the same direction is a Gram
preconditioning of the momentum,
\begin{equation}
  \operatorname{msign}(M_t)=(M_tM_t^\top)^{-1/2}M_t=U_tV_t^\top ,
\end{equation}
where $\operatorname{msign}(\cdot)$ is the matrix sign function that sets all
singular values to one, matching the Axis~III metric $H_t=M_tM_t^\top$ in
Table~\ref{tab:four-axis-instantiation}.  The two readings are the two faces of
the same operator $\Phi_t$, and they differ only in whether the geometry is
charged to the direction map $\psi_t$ or to the metric $H_t$.

Kronecker-factored preconditioning occupies a different point in the same
matrix space.  For Shampoo, the matrix case maintains row and column factors
\begin{equation}
  L_t=\beta L_{t-1}+(1-\beta)G_tG_t^\top,\quad
  R_t=\beta R_{t-1}+(1-\beta)G_t^\top G_t,
  \label{eq:t2-shampoo-factors}
\end{equation}
and applies an inverse-root preconditioner,
\begin{equation}
  \Delta_t = L_t^{-1/4}G_tR_t^{-1/4},\quad W_{t+1}=W_t-\eta_t\Delta_t .
  \label{eq:t2-shampoo-update}
\end{equation}
As with Muon, Axis~III holds this update in both the LMO and the preconditioner
readings (Table~\ref{tab:four-axis-instantiation}).  From the preconditioner
side, the row and column factors combine into a Kronecker metric
\begin{equation}
  H_t = L_t^{1/4}\otimes R_t^{1/4},\quad
  \operatorname{vec}(\Delta_t)=H_t^{-1}\operatorname{vec}(G_t),
\end{equation}
whose analysis basis is the eigenbasis of $L_t$ and $R_t$.  The curvature proxy
is a Kronecker-structured matrix that captures coordinate coupling and is richer
than the diagonal vector of Adam.  From the LMO side, the same direction is the
steepest matrix direction on the Kronecker-metric ball
$\{X:\|L_t^{1/4}XR_t^{1/4}\|_F\le\rho\}$, an ellipsoid (Mahalanobis ball)
stretched and rotated by that metric, so Shampoo returns a continuous
preconditioned gradient.
SOAP keeps this basis idea but runs Adam-style coordinate adaptation after
rotating the gradient into the Shampoo eigenbasis, then rotates the update
back~\citep{vyas2025soap}.  It is therefore a bridge between T1 and T2, with a
coordinate-wise adaptive denominator whose coordinates have shifted from raw
parameters to learned matrix-curvature coordinates.

Low-rank projection methods move along Axis~I
(Table~\ref{tab:four-axis-instantiation}), placing the whole update in a
low-dimensional subspace while the curvature estimate largely follows Adam.
GaLore is the representative example, choosing or refreshing a projection basis
$P_t\in\mathbb{R}^{m\times r}$ with $r\ll\min(m,n)$ and projecting the gradient,
\begin{equation}
  \widetilde{G}_t=P_t^\top G_t,
  \label{eq:t2-galore-project}
\end{equation}
running Adam or AdamW state evolution on $\widetilde{G}_t$, and reconstructing a
full update by $\Delta_t=P_t\widetilde{\Delta}_t$.  Because the optimizer state
lives only in the projected low-dimensional coordinates, memory drops
accordingly.  The risk is approximation error.  Useful high-rank gradient
components may be removed unless the basis refresh, rank allocation, or residual
pathway captures them.

\subsubsection{Representative Methods}
\label{sec:t2-methods}

The following discussion follows the T2.1--T2.3 subclass order introduced
above.  Within each subclass, representative methods are grouped by the local
matrix operation they modify rather than by chronology.

\paragraph{Spectral Orthogonalization.}
\label{sec:t2-spectral}

The spectral branch asks whether a matrix update should preserve the
singular-vector frame but discard the singular-value scale.  At the survey
level, its core operation is
\begin{equation}
  M_t=\beta M_{t-1}+(1-\beta)G_t,\quad
  D_t=\operatorname{Polar}(M_t)
      = U_tV_t^\top
      \approx \operatorname{NS}_K(M_t),
  \label{eq:t2-spectral-template}
\end{equation}
where $M_t=U_t\Sigma_tV_t^\top$ and $\operatorname{NS}_K$ denotes a finite
Newton--Schulz approximation to the polar factor.  The resulting update is
usually $W_{t+1}=W_t-\eta_tD_t$ on routed matrix parameters.
\begin{itemize}
  \setlength{\itemsep}{2pt}
  \item \textbf{Canonical orthogonalized-update line.}  Muon is the central
  T2.1 method because it turns the matrix momentum into an orthogonalized
  update rather than a coordinate-wise adaptive update.  In practical LLM
  implementations, Muon is usually applied only to selected two-dimensional
  hidden-layer weights, while embeddings, scalar parameters, normalization
  parameters, and other excluded tensors use an SGD or AdamW-style fallback
  route~\citep{jordan2024muon}.
  \item \textbf{Routing and scaling extensions.}  Dion proposes distributed
  orthonormalized updates and broader parameter handling, making the S1
  routing and systems pathway more explicit~\citep{ahn2025dion}.  AdaMuon and
  AdaGO add adaptive scaling logic to orthogonalized updates, placing them at
  the boundary between T2.1 direction selection and T1-style step-size
  adaptation~\citep{si2025adamuon,zhang2025adagrad}.
  \item \textbf{Spectral-geometry boundary methods.}  OrthoGrad and Spectral
  Sphere are useful boundary points rather than central LLM optimizer
  baselines: the former studies orthogonal-gradient constraints for
  calibration, while the latter studies controlled LLM training under
  spectral-sphere geometry~\citep{hedges2025orthograd,xie2026controlled}.
\end{itemize}

The common thread is a single direction choice, namely to preserve the
singular-vector frame and flatten the singular spectrum.  Whether it
consistently beats AdamW, Shampoo, or SOAP depends on the setting.  The open
empirical question is whether the spectral direction remains beneficial after
matching batch size, warmup, weight decay, fallback optimizer, matrix-only
scope, step time, and tuning budget.  Recent mechanism and fine-tuning studies already warn that
Muon-style conclusions can be protocol-sensitive
~\citep{qu2026muonfinetune,shumaylov2026muonnot}.

\paragraph{Kronecker-Factored Preconditioning.}
\label{sec:t2-kronecker}

The Kronecker branch asks whether a matrix update should be scaled in row and
column curvature coordinates or kept in raw parameter coordinates.  The shared
template is
\begin{align}
  L_t &= \beta L_{t-1}+(1-\beta)G_tG_t^\top,\quad
  R_t = \beta R_{t-1}+(1-\beta)G_t^\top G_t, \\
  \bar{G}_t &= P_{L,t}^\top G_t P_{R,t},\quad
  \Delta_t = P_{L,t}\,\Psi_t(\bar{G}_t;\mathcal{S}_t)\,P_{R,t}^\top,
  \label{eq:t2-kronecker-adaptive-template}
\end{align}
where $P_{L,t}$ and $P_{R,t}$ are eigenvectors or structured bases derived
from row/column statistics.  Shampoo instantiates $\Psi_t$ as inverse-root
preconditioning, whereas SOAP uses Adam-style coordinate adaptation in this
rotated basis.
\begin{itemize}
  \setlength{\itemsep}{2pt}
  \item \textbf{Inverse-root preconditioning base.}  Shampoo is the
  mathematical base of T2.2.  It stores per-mode row and column statistics and
  applies matrix inverse roots~\citep{gupta2018shampoo}, approximating the full
  $(mn)\times(mn)$ curvature matrix with two small factors while keeping more
  coordinate coupling than Adam's per-coordinate diagonal second moment.  For
  matrices, this yields the factors and update in
  Eqs.~\eqref{eq:t2-shampoo-factors}--\eqref{eq:t2-shampoo-update}.
  The benefit is a structured approximation to second-order information; the
  cost is nontrivial state, matrix-root computation, numerical damping, and a
  refresh schedule for the preconditioners.
  \item \textbf{Adam-in-eigenbasis and practical routing.}  SOAP updates the
  Shampoo line by adding Adam-like adaptation in the Shampoo eigenbasis
  ~\citep{vyas2025soap}.  Conceptually, it first chooses the matrix basis from
  Kronecker factors and then applies diagonal adaptivity in that rotated
  coordinate system.  COSMOS extends the practical branch by combining hybrid
  adaptive routing with memory-efficient LLM training goals
  ~\citep{liu2025cosmos}.
  \item \textbf{Learned and structured-Fisher preconditioners.}  PSGD and
  Kron-like instances form a broader learned-preconditioner branch.  The early
  PSGD formulation and later Lie-group preconditioner view learn structured
  preconditioners rather than using Shampoo's fixed inverse-root construction
  ~\citep{li2017preconditioned,li2022black}.  RACS is placed in T2.2 because its
  row-and-column scaled SGD is based on a structured Fisher approximation,
  whereas the low-rank extension in the same source is represented by
  Alice~\citep{gong2025towards}.
\end{itemize}

The common thread is structured matrix scaling.  T2.2 differs from T2.1
because it uses magnitude information in row/column or Fisher coordinates
rather than only flattening a spectrum.  It differs from T2.3 because it does
not make low-rank projection the primary state space.  Because Shampoo and SOAP
tend to improve loss per token while paying a high per-step cost, benchmark
claims for this subclass should report loss at a fixed token budget alongside
loss normalized by wall-clock or step time, so that the per-step overhead is
counted, together with the preconditioner refresh frequency, damping, and
memory overhead that drive this cost.

\paragraph{Low-Rank Subspace Projection.}
\label{sec:t2-low-rank}

The low-rank branch asks whether the optimizer state should live in a
projected matrix subspace.  Its survey-level template is
\begin{equation}
  \widetilde{G}_t=P_t^\top G_t,\quad
  \widetilde{\Delta}_t
    = \operatorname{AdamW}_{\mathrm{subspace}}(\widetilde{G}_t),\quad
  \Delta_t=P_t\widetilde{\Delta}_t+\mathcal{E}_t,
  \label{eq:t2-lowrank-template}
\end{equation}
where $P_t\in\mathbb{R}^{m\times r}$, $r\ll\min(m,n)$, and
$\mathcal{E}_t$ denotes an optional residual or correction pathway.  Plain
GaLore uses the projection-reconstruction route without treating
$\mathcal{E}_t$ as the defining component.  Follow-up methods modify the rank,
refresh, or residual logic.
\begin{itemize}
  \setlength{\itemsep}{2pt}
  \item \textbf{Projected Adam-state base.}  GaLore defines the low-rank
  subspace branch for LLM optimization~\citep{zhao2024galore}.  Its premise is
  that many large-model gradient matrices contain exploitable low-rank
  structure during training.  GaLore periodically constructs a projection
  basis, runs the adaptive update in the projected coordinates, and maps the
  update back to the original matrix.  Because its Adam states live only in this
  low-dimensional subspace, memory is much smaller than for the full matrix.
  \item \textbf{Residual and rank-allocation follow-ups.}  If the rank is too
  small, the projection refresh interval is too long, or the basis misses
  changing gradient directions, a low-rank method may save memory while losing
  high-rank information.  Fira directly addresses this concern by asking
  whether full-rank training quality can be recovered under a low-rank
  constraint, using residual or rank-allocation logic to reduce information
  loss~\citep{chen2026fira}.
  \item \textbf{Structured-Fisher low-rank extensions.}  Alice is another
  low-rank extension, but its source begins from structured Fisher
  approximation and row/column scaling.  Given this structured-Fisher origin, it
  is better described as a low-rank extension on a T2.2 background, whereas
  GaLore is the pure low-rank projection route~\citep{gong2025towards}.
\end{itemize}

The common thread is S2/S3/S4 subspace training, namely projection, subspace
state evolution, and reconstruction.  The memory benefit is an effect of this
subspace choice, while the classification rests on the subspace mechanism
itself.  For benchmarking, T2.3
methods should be compared under explicit memory budgets against full-rank
AdamW, T4 state-compression methods, and T2.2 matrix preconditioners under
matched rank, refresh interval, batch size, and wall-clock accounting.

\subsubsection{Effect-Target Assessment}
\label{sec:t2-effect-assessment}

T2 changes the unit of adaptation from coordinates to matrices.  This shift
can improve the direction itself: spectral methods choose a matrix-level
orientation, Kronecker methods approximate curvature in row--column
coordinates, and low-rank methods restrict state evolution to a cheaper
subspace.  The same shift also explains the family cost.  Matrix operations,
parameter routing, basis refresh, and structured state make T2 more dependent
on layer topology and implementation details than T1.  Table~\ref{tab:t2_effect_assessment}
summarizes these mechanism-derived performance priors.

\begin{table*}[!htbp]
  \centering
  \caption{Mechanism-informed effect assessment for T2 subclasses.  The
  entries are design priors for benchmark planning, not empirical
  conclusions.}
  \label{tab:t2_effect_assessment}
  \scriptsize
  \setlength{\tabcolsep}{3pt}
  \renewcommand{\arraystretch}{1.08}
  \begin{tabularx}{\textwidth}{@{}Y Y Y@{}}
    \toprule
    Primary target & Likely cost or failure mode & Benchmark focus \\
    \midrule
    \rowcolor{famT2!18}\multicolumn{3}{c}{\textbf{\textcolor{famT2!62!black}{T2: Matrix-level structural methods}}} \\
    \rowcolor{famT2!8}\multicolumn{3}{@{}l@{}}{\textbf{\textcolor{famT2!62!black}{T2.1 Spectral orthogonalization}}} \\
      O1 token efficiency, O4 stability
      & O2 extra matrix multiplications, routing sensitivity
      & Matched LR/warmup/batch, matrix-only scope, step-time accounting \\
    \rowcolor{famT2!8}\multicolumn{3}{@{}l@{}}{\textbf{\textcolor{famT2!62!black}{T2.2 Kronecker-factored preconditioning}}} \\
      O1 convergence, O4 conditioning
      & O2 matrix roots, O3 Kronecker state, implementation complexity
      & Preconditioner update frequency, memory and wall-clock overhead, damping \\
    \rowcolor{famT2!8}\multicolumn{3}{@{}l@{}}{\textbf{\textcolor{famT2!62!black}{T2.3 Low-rank subspace projection}}} \\
      O3 memory reduction with possible O1 retention
      & Refresh cost, rank sensitivity, high-rank information loss
      & Rank and interval sweeps, memory budget vs.\ AdamW and T4 \\
    \bottomrule
  \end{tabularx}
\end{table*}

The resulting performance profile is multi-objective.  A spectral method can
improve loss per token while increasing step time; a Kronecker method can
improve conditioning but lose wall-clock advantage when inverse roots are too
frequent; a low-rank method can save memory while requiring enough rank or
refreshes to preserve high-rank gradient information.  The family-level
results in Sec.~\ref{sec:benchmark-family-summary} therefore separate T2 into
several regimes rather than treating it as a single upgrade over AdamW.
The broad optimizer studies by several researchers~\citep{zhao2025deconstructing} and
Semenov et al.~\citep{semenov2025benchmarking} already show that optimizer
rankings move with scale, tuning effort, and training duration. T2 sharpens
that dependence because its gains appear only when the exploited matrix
structure is statistically useful and cheap enough to apply.

\subsection{T3: Discretization and Directional Quantization}
\label{sec:t3}

\subsubsection{Family Overview and Meta-Pipeline Position}
\label{sec:t3-overview}

\begin{wrapfigure}{r}{0.5\textwidth}
  \centering
  \vspace{-1.6em}
  \includegraphics[width=1.0\linewidth]{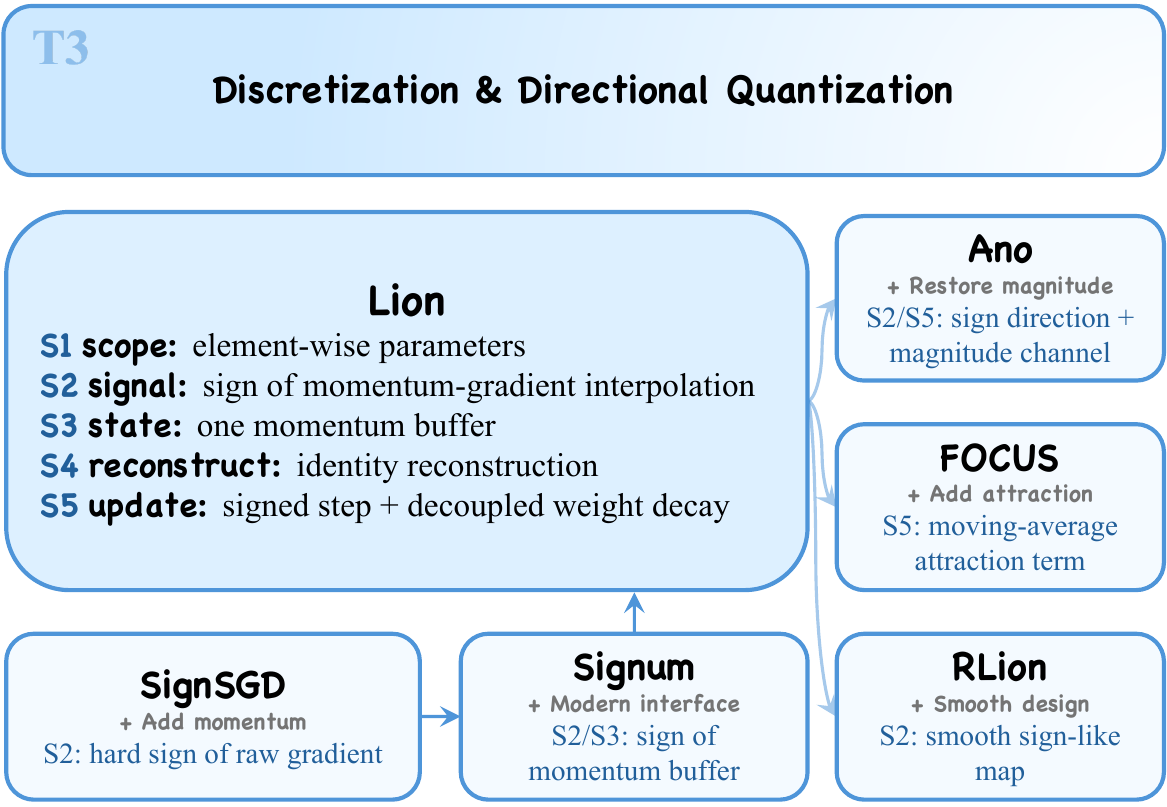}
  \caption{
  Mechanism schematic for T3 discretization and directional quantization. This schematic provides a taxonomy guide rather than an empirical ranking for the compact T3 family, grouping discretization and directional quantization mechanisms by their sign-direction generation methods.
  }
  \vspace{-2.em}
  \label{fig:t3_family_mechanism}
\end{wrapfigure}

T3 contains optimizers whose primary operation discretizes the update
direction.  The characteristic map is
$\operatorname{sign}(\cdot)$ or a close sign-like transformation: it keeps
coordinate-wise orientation while discarding, clipping, or coarsening the
continuous magnitude of the gradient or momentum signal.  This makes T3
different from both T1 and T4.  T1 methods may become sign-like in some
coordinates after adaptive normalization, but their defining mechanism is
still moment estimation and diagonal scaling.  T4 methods may reduce memory,
but their defining mechanism is state storage, sharing, quantization, or
release.  T3 instead changes the geometry of the direction itself.


\definecolor{hidden-blue}{HTML}{4a90e2}
\definecolor{hidden-black}{HTML}{333333}
\tikzstyle{my-box}=[
  rectangle,
  draw=hidden-black,
  rounded corners,
  text opacity=1,
  minimum height=1.5em,
  minimum width=5em,
  inner sep=2pt,
  align=center,
  fill opacity=.5,
]
\tikzstyle{root}=[
  align=center,
]
\tikzstyle{leaf}=[
  my-box,
  minimum height=1.5em,
  text=black,
  font=\normalsize,
  inner xsep=5pt,
  inner ysep=4pt,
  text width=10.5em,
  fill opacity=1,
]
\tikzstyle{leaf1}=[
  my-box,
  minimum height=1.5em,
  fill=yellow!32,
  text=black,
  font=\normalsize,
  inner xsep=5pt,
  inner ysep=4pt,
  text width=15em,
  align=left,
]
\begin{figure*}[!tbp]
\centering
\resizebox{\textwidth}{!}{
\begin{forest}
forked edges,
for tree={
grow=east,
reversed=true,
anchor=base west,
parent anchor=east,
child anchor=west,
base=left,
font=\large,
rectangle,
draw=hidden-black,
rounded corners,
align=left,
minimum width=4em,
edge+={darkgray, line width=1pt},
s sep=4pt,
inner xsep=5pt,
inner ysep=4pt,
line width=1.1pt,
ver/.style={rotate=90, child anchor=north, parent anchor=south, anchor=center},
},
where level=1{text width=11em, font=\normalsize}{},
where level=2{text width=34.5em, align=left, tier=citations, font=\small}{},
[
    T3-based \\ Optimizers,
    fill=brown!10!white, text=hidden-black, root, text width=6em
    [
        T3: Discretization \& Directional \\ Quantization,
        fill=yellow!32, leaf1
        [{
            SignSGD~\cite{bernstein2018signsgd},
            Signum~\cite{bernstein2018signsgd},
            Lion~\cite{chen2023symbolic},
            MARS-Lion~\cite{yuan2024mars,chen2023symbolic},
            RLion~\cite{rong2024rlion},
            FOCUS~\cite{liu2025focus}, \\
            Ano~\cite{kegreisz2025ano}
        }, fill=yellow!12]
    ]
  ]
\end{forest}
}
\caption{Taxonomy of discretized and directionally quantized optimizers.}
\label{tab:t3_family_taxonomy}
\end{figure*}
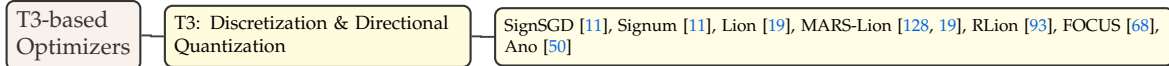

\paragraph{Mechanism-group structure.}
Figure~\ref{tab:t3_family_taxonomy} gives the complete optimizer membership for
T3.  Because the current taxonomy treats T3 as one compact family-level
subclass, the discussion below interprets the taxonomy figure through three
sign-direction routes rather than restating the member list.
\begin{itemize}[leftmargin=*,itemsep=2pt]
  \item \textbf{Pure sign directions.}
  The defining operation is a hard sign map applied to either the raw gradient
  or a momentum buffer before the update is formed.  SignSGD is the canonical
  instance, and its momentum-sign variant is usually discussed as Signum
  ~\citep{bernstein2018signsgd}.
  \item \textbf{Signed momentum.}
  The direction is the sign of an interpolated current-gradient and
  past-momentum signal, with only one persistent momentum buffer.  Lion is the
  central large-model representative of this route~\citep{chen2023symbolic};
  MARS-Lion keeps the same sign map but feeds it a MARS-style corrected signal
  ~\citep{yuan2024mars}.
  \item \textbf{Smooth and hybrid sign directions.}
  These methods preserve a sign-like core but soften discontinuities, add
  trajectory attraction, or reintroduce a limited magnitude channel.  RLion
  smooths the sign decision~\citep{rong2024rlion};
  FOCUS attracts iterates toward a running parameter trajectory
  ~\citep{liu2025focus}; and Ano restores a bounded magnitude channel without
  returning to a full Adam-style second-moment state~\citep{kegreisz2025ano}.
\end{itemize}

At the family level, T3 can be summarized as
\begin{equation}
  z_t = \mathcal{C}_t(g_t,m_{t-1},\theta_t),\quad
  d_t = q_t(z_t),\quad
  \theta_{t+1}
  = (1-\eta_t\lambda)\theta_t
    - \eta_t\big(s_t\odot d_t+a_t\big),
  \label{eq:t3-direction-template}
\end{equation}
where $\mathcal{C}_t$ forms the signal on which the directional decision is
made, $q_t$ is a hard or smooth sign-like map, $s_t$ is a scalar or
coordinate-wise magnitude channel, and $a_t$ is an optional trajectory-control
term.  SignSGD sets $\mathcal{C}_t(g_t)=g_t$, $q_t=\operatorname{sign}$,
$s_t=\mathbf{1}$, and $a_t=0$.  Lion sets
$\mathcal{C}_t=\beta_1m_{t-1}+(1-\beta_1)g_t$ and keeps the hard sign.  Hybrid
variants retain the same directional core while changing $q_t$, $s_t$, or
$a_t$.  Thus the transformed signal in S2 is already a discrete or
near-discrete direction, S3 is usually a first-moment state or no second
moment, S4 is identity, and S5 controls the learning-rate radius, weight
decay, and any added trajectory or scale correction.  Compared with AdamW,
the persistent optimizer state can be reduced because no second-moment buffer
$v_t$ is stored.  This memory benefit is important, but it is an effect of the
sign-direction mechanism rather than the classification rule.

The family boundary is also important for masks and selective updates.  A
method belongs to T3 when the sign-like map directly defines the base update
direction, as in SignSGD, Lion, RLion, FOCUS, and Ano.  By contrast, methods
whose primary increment is a binary or stochastic mask applied to an already
computed AdamW-, Lion-, or Muon-style update are post-update filters.  We
therefore place Magma and MGUP in T5.3 rather than in T3; the
same rule covers cautious and top-$k$ update filters.  This separation keeps
``direction quantization'' distinct from ``selective writeback'' even though
both may use signs or binary decisions internally.

\subsubsection{LMO-Driven Four-Axis Interpretation}
\label{sec:t3-lmo-four-axis}

In the four-axis view, the typical T3 coordinate is
\begin{equation}
  \big((I,I),\ \operatorname{sign},\ I,\ 0,\ \widehat{g}_t,\ \mathcal{S}_t\big),
\end{equation}
where the analysis basis remains the raw coordinate basis, the direction map
is sign-like, the curvature proxy is absent or weak, and
$\mathcal{S}_t$ records whether a first-moment buffer, no buffer, or an
extra magnitude channel is stored.  The LMO explanation comes from the
$\ell_\infty$ ball.  For the linearized problem
\begin{equation}
  \min_{\|d\|_\infty\leq 1}\langle g,d\rangle ,
\end{equation}
the optimizer is $d^*=-\operatorname{sign}(g)$.  The extreme points of the
$\ell_\infty$ ball are coordinate-wise sign patterns, so a sign update is the
steepest linearized step under a fixed maximum coordinate displacement.
This is the coordinate analogue of the spectral sign logic used by Muon in
T2, but with the identity basis rather than an SVD or Kronecker basis.

SignSGD realizes this geometry directly by applying
$d_t=\operatorname{sign}(g_t)$~\citep{bernstein2018signsgd}.  Lion replaces
the raw stochastic gradient with a smoothed interpolation of current gradient
and past momentum before applying the same vertex map.  This modification is
small in algebra but large in practice: the sign decision is made on a
lower-variance temporal signal, while the state remains cheaper than Adam's
two-buffer design.  Recent norm-LMO and stochastic Frank--Wolfe analyses
make this connection explicit by treating sign, normalized, and
orthogonalized updates as different norm-constrained direction choices rather
than unrelated heuristics
~\citep{bernstein2024oldoptimizer,pethick2025normlmo,sfyraki2025lions}.

The same interpretation also clarifies what T3 does not cover.  AdamW may
look sign-like when the adaptive denominator largely cancels gradient
magnitudes, but its update is still controlled by a diagonal curvature or
variance estimate and exponent $\alpha=1/2$.  FOCUS and Ano are closer to T3
because they preserve a sign-direction core while adding trajectory
attraction or an explicit magnitude channel~\citep{liu2025focus,kegreisz2025ano}.
RLion sits on the boundary in the opposite direction: it smooths the hard
sign into a bounded continuous map, so it is best read as a Lion-style
regularization of the same direction geometry rather than as a separate
adaptive-moment family~\citep{rong2024rlion}.

\subsubsection{Representative Methods}
\label{sec:t3-methods}

The following discussion follows the three-part organization introduced above.
Within each group, representative methods are further separated by the local
operation that changes the sign-direction template in
Eq.~\eqref{eq:t3-direction-template}.

\paragraph{Pure Sign Directions.}
\label{sec:t3-sign-predecessors}

The first branch applies a sign map with minimal additional optimizer state.
At the survey level, it can be written as
\begin{equation}
  d_t=\operatorname{sign}(h_t),\quad
  h_t =
  \begin{cases}
    g_t, & \text{pure SignSGD},\\
    \beta h_{t-1}+(1-\beta)g_t, & \text{momentum sign}.
  \end{cases}
  \label{eq:t3-pure-sign-template}
\end{equation}
\begin{itemize}
  \setlength{\itemsep}{2pt}
  \item \textbf{Pure gradient sign.}  SignSGD is the historical anchor for T3.
  Its update transmits and applies only the sign of each stochastic-gradient
  coordinate, and the distributed variant aggregates worker signs through
  majority vote~\citep{bernstein2018signsgd}.  The original motivation was
  partly communication efficiency: a sign vector is a 1-bit message per
  coordinate.  For this taxonomy, the more important point is geometric:
  SignSGD removes continuous magnitude information before the update is formed.
  \item \textbf{Momentum sign.}  Momentum sign methods, often referred to as
  Signum in benchmark and theory discussions, move the sign decision from
  $g_t$ to a smoothed momentum buffer.  This makes the direction less
  sensitive to a single noisy mini-batch while keeping the fixed-magnitude
  update structure.
  \item \textbf{Benchmark role.}  SignSGD should not be evaluated as a modern
  LLM optimizer on historical status alone.  Its communication advantages,
  convergence assumptions, and failure modes were developed before current
  Transformer pretraining protocols.  It remains useful because it isolates
  direction discretization from Adam-style second moments; recent
  language-model optimizer studies use Signum-like variants to decompose how
  much of Adam's behavior comes from signed momentum, layer-wise adaptivity, and
  other control components~\citep{zhao2025deconstructing}.
\end{itemize}

The common thread is the early removal of magnitude.  In the notation of
Eq.~\eqref{eq:t3-direction-template}, pure sign methods set $s_t=\mathbf{1}$
and $a_t=0$; all adaptation must therefore come from the signal $h_t$,
learning-rate schedule, or external routing, not from a diagonal second-moment
denominator.

\paragraph{Signed Momentum.}
\label{sec:t3-lion-line}

The second branch is the modern signed-momentum line.  Lion's compact rule is
\begin{align}
  z_t &= \beta_1 m_{t-1}+(1-\beta_1)g_t, \quad
  d_t = \operatorname{sign}(z_t), \\
  m_t &= \beta_2 m_{t-1}+(1-\beta_2)g_t, \quad
  \theta_{t+1} = (1-\eta_t\lambda)\theta_t-\eta_t d_t .
  \label{eq:t3-lion-update}
\end{align}
\begin{itemize}
  \setlength{\itemsep}{2pt}
  \item \textbf{Search-discovered signed momentum.}  Lion is the central T3
  method for LLM training.  It was discovered through symbolic optimizer
  search and then simplified into Eq.~\eqref{eq:t3-lion-update}
  ~\citep{chen2023symbolic}.  The method retains several practical features of
  AdamW--a momentum interface, decoupled weight decay, and schedule
  compatibility--but it replaces Adam's continuous adaptive direction by the
  sign of a momentum-gradient interpolation.
  \item \textbf{Interface, mechanism, and memory effect.}  Lion is often
  described as an AdamW alternative, a sign optimizer, and a memory-friendlier
  optimizer.  In this taxonomy, only the second description is primary.  The
  AdamW interface is inherited scaffolding, and the memory saving follows from
  removing $v_t$.
  \item \textbf{Protocol-sensitive tuning.}  Because $\|d_t\|_\infty=1$, the
  learning rate directly sets the largest coordinate movement, and
  gradient-magnitude outliers cannot enlarge individual update coordinates.
  This can improve robustness in some noisy regimes, but it can also make
  learning rate, weight decay, warmup, and schedule choices sharper than in
  AdamW.  Recent LLM studies include Lion as an important baseline, but also
  show that rankings change with model scale, token budget, batch size, and
  tuning effort
  ~\citep{zhao2025deconstructing,semenov2025benchmarking,wen2025fantastic,schlotthauer2025budget}.
  \item \textbf{Cross-family wrappers.}  MARS-Lion changes the
  gradient-estimation axis by feeding a variance-reduced signal into a
  Lion-style sign map; it does not create a new T3 subclass.  Cautious Lion
  applies a consistency filter after the base direction is computed, so its
  primary added mechanism is T5.3 post-update filtering.
\end{itemize}

The common thread is signed momentum under an AdamW-like training interface.
Lion therefore functions as the central comparison point for T3: pure sign
methods show what is lost without a temporal buffer, while hybrid variants
test how much smoothing or magnitude information can be restored without
leaving the sign-direction geometry.

\paragraph{Smooth and Hybrid Sign Directions.}
\label{sec:t3-hybrid-sign}

The third branch tests which parts of the hard-sign template can be softened
or supplemented.  A broad template is
\begin{equation}
  d_t=q_\tau(z_t),\quad
  \theta_{t+1}
  =(1-\eta_t\lambda)\theta_t
  -\eta_t\big(s_t\odot d_t+\rho_t(\theta_t-\bar{\theta}_t)\big),
  \label{eq:t3-hybrid-sign-template}
\end{equation}
where $q_\tau$ is a smooth or bounded sign-like map, $s_t$ may restore a
limited magnitude channel, and $\bar{\theta}_t$ is an optional moving-average
trajectory reference.
\begin{itemize}
  \setlength{\itemsep}{2pt}
  \item \textbf{Smooth sign maps.}  RLion replaces the discontinuous sign
  function with a bounded smooth alternative based on an $\arctan$-style
  transformation~\citep{rong2024rlion}.  This keeps the update in the same
  sign-like direction family while reducing abrupt coordinate flips near zero.
  Its evidence is mainly outside LLM pretraining, so it is a boundary method
  rather than a coequal anchor with Lion.
  \item \textbf{Trajectory attraction around sign directions.}  FOCUS starts
  from the observation that sign-momentum methods can move aggressively in
  noisy, flat, or low-curvature regions, then adds attraction toward a moving
  average of parameters to concentrate the trajectory~\citep{liu2025focus}.
  The update still uses a sign-direction core, but S5 gains an additional
  trajectory-control term.
  \item \textbf{Magnitude side channels.}  Ano separates direction from
  magnitude, using momentum to smooth the direction while using instantaneous
  gradient magnitudes to set step sizes in noisy landscapes
  ~\citep{kegreisz2025ano}.  It is best read as a sign-direction plus
  magnitude-channel hybrid, not as a return to full Adam-style second-moment
  adaptivity.
\end{itemize}

These methods show that T3 is not limited to hard binary updates.  They also
make the boundary sharper: methods dominated by update masks or writeback
filters are better handled in T5.3, even when they use directional agreement
signals, whereas RLion, FOCUS, and Ano still define the base direction through
a sign-like route.

\subsubsection{Effect-Target Assessment}
\label{sec:t3-effect-assessment}

T3 replaces continuous update magnitudes with sign-like directions.  This
creates a compact performance profile: the update is cheap, the persistent
state can be smaller than AdamW, and the largest coordinate displacement is
controlled directly by the learning rate.  The same discretization removes
amplitude information before the step is formed.  SignSGD exposes this limit
in its purest form~\citep{bernstein2018signsgd}; Lion makes the route more
practical by applying the sign map to a smoothed momentum-gradient signal
~\citep{chen2023symbolic}.  Table~\ref{tab:t3_effect_assessment} summarizes
the mechanism-derived performance priors.

\begin{table}[!htbp]
  \centering
  \caption{Mechanism-informed effect assessment for T3 methods. The entries
  are design priors for benchmark planning, not empirical conclusions.}
  \label{tab:t3_effect_assessment}
  \scriptsize
  \setlength{\tabcolsep}{3pt}
  \renewcommand{\arraystretch}{1.08}
  \begin{tabularx}{\linewidth}{@{}Y Y Y@{}}
    \toprule
    Primary target & Likely cost or failure mode & Benchmark focus \\
    \midrule
    \rowcolor{famT3!18}\multicolumn{3}{c}{\textbf{\textcolor{famT3!62!black}{T3: Discretization and directional quantization}}} \\
    \rowcolor{famT3!8}\multicolumn{3}{@{}l@{}}{\textbf{\textcolor{famT3!62!black}{SignSGD / Signum}}} \\
      O2 communication simplicity, O4 outlier robustness
      & Magnitude loss, sign-noise sensitivity
      & Pure vs.\ momentum sign vs.\ AdamW under matched settings \\
    \rowcolor{famT3!8}\multicolumn{3}{@{}l@{}}{\textbf{\textcolor{famT3!62!black}{Lion}}} \\
      O3 low state, O2 cheap updates
      & LR, WD, and warmup become protocol-sensitive
      & LR/WD/schedule sweeps against strong AdamW baselines \\
    \rowcolor{famT3!8}\multicolumn{3}{@{}l@{}}{\textbf{\textcolor{famT3!62!black}{RLion / FOCUS / Ano}}} \\
      O4 stability around sign directions
      & Extra knobs, limited LLM evidence
      & Source domains, extra-knob tuning, per-component ablations \\
    \bottomrule
  \end{tabularx}
\end{table}

The first expected gain is resource efficiency rather than peak quality.  Lion
can reduce optimizer state by removing $v_t$, and sign updates avoid expensive
direction construction, but neither fact guarantees a better loss trajectory.
Fixed-magnitude updates can be tolerant to local magnitude outliers while
remaining sensitive to learning rate, warmup, and weight decay.  This
distinction matches the benchmark narrative in
Sec.~\ref{sec:benchmark-family-summary}: T3 is efficient and locally tolerant
to learning-rate perturbations, yet its tuned perplexity is generally weaker
than the best T1 and T2 alternatives.

\subsection{T4: State Compression and Structural Aggregation}
\label{sec:t4}

\subsubsection{Family Overview and Meta-Pipeline Position}
\label{sec:t4-overview}

T4 contains optimizers whose primary contribution is to reduce the memory
footprint, granularity, precision, or lifetime of optimizer state.  For a
model with $d$ trainable parameters, AdamW stores at least two full-size state
tensors, $m_t$ and $v_t$, in addition to parameters, gradients, activations,
and distributed-training buffers.  This cost is significant in
LLM training, because optimizer state competes directly with model size, sequence
length, batch size, activation rematerialization, sharding, and communication
buffers.  The central
issue is which optimizer histories must be retained, how faithfully they must be
represented, and when they can be reconstructed, broadcast, dequantized, or
discarded.

\begin{figure}[t]
  \centering
  \includegraphics[width=0.75\linewidth]{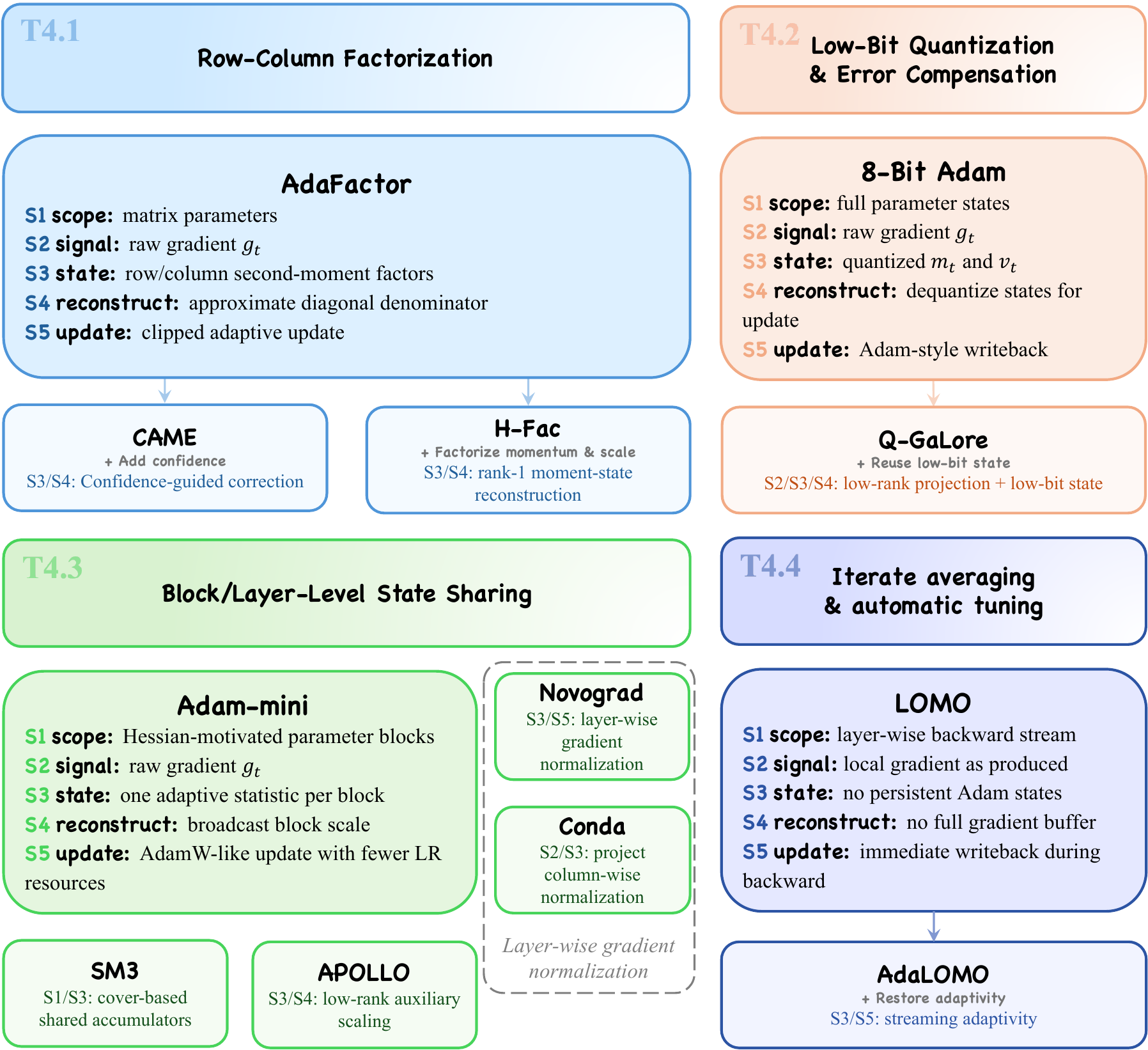}
  \caption{\textbf{Mechanism schematic for T4 state compression and structural
aggregation.} The schematic organizes the family by the form of memory
reduction: factored second-moment storage, low-bit state representation,
shared adaptive statistics, and streaming gradient consumption.}
  \label{fig:t4_family_mechanism}
\end{figure}

\paragraph{Subclass structure and membership.}
Figure~\ref{tab:t4_family_taxonomy} gives the subclass-level optimizer list for T4, and the
four subclasses below specify the corresponding optimizer membership.  The
split follows the state representation being changed.
\begin{enumerate}[label=\textbf{T4.\arabic*},leftmargin=*,itemsep=2pt]
  \item \textbf{Row-column factorization.}
  This subclass contains AdaFactor and CAME.  Its members replace a full
  matrix second-moment tensor with row and column summaries, then reconstruct
  an approximate denominator only when the update is formed
  ~\citep{shazeer2018adafactor,luo2023came}.
  \item \textbf{Low-bit quantization and error compensation.}
  This subclass contains 8-bit Adam and Q-GaLore.  Its members store optimizer
  statistics, projection objects, or related accumulated information in
  low-bit form and rely on block-wise scaling, stochastic rounding, or
  compensation logic to control long-horizon numerical error
  ~\citep{dettmers20218,zhang2024q}.
  \item \textbf{Block- and layer-level state sharing.}
  This subclass contains Adam-mini, APOLLO, SM3, Conda, and NovoGrad.  Its
  members reduce the number or dimensionality of adaptive statistics by
  sharing them across blocks, covers, columns, layers, or low-rank auxiliary
  structures~\citep{zhang2025adam,zhu2025apollo,anil2019memory,wang2025conda,ginsburg2019stochastic}.
  \item \textbf{Fused backprop-update.}
  This subclass contains LOMO and AdaLOMO.  Its members change the optimizer
  pipeline so that gradients are consumed during backpropagation as soon as
  they are produced
  ~\citep{lv2024full,lv2024adalomo}.
\end{enumerate}

At the family level, a compressed-state adaptive update can be summarized as
\begin{equation}
  \mathcal{Z}_t = \mathcal{C}_t(\mathcal{S}_t),\quad
  \widehat{\mathcal{S}}_t = \mathcal{R}_t(\mathcal{Z}_t),\quad
  \theta_{t+1}
  =
  \theta_t-\eta_t\,
  \mathcal{U}_t\big(g_t,\widehat{\mathcal{S}}_t\big),
  \label{eq:t4-state-compression-template}
\end{equation}
where $\mathcal{S}_t$ denotes the full optimizer state that an AdamW-like
method would store, $\mathcal{C}_t$ is a compression, sharing, quantization,
or lifetime-shortening map, $\mathcal{R}_t$ reconstructs or exposes the
usable state at update time, and $\mathcal{U}_t$ is the base update rule.
Full-state AdamW is the special case in which $\mathcal{C}_t$ and
$\mathcal{R}_t$ are identities.  T4 begins when at least one of them is
nontrivial.

In the meta-pipeline, T4 primarily modifies S3 and S4.  The state maintained
by S3 may be factored, quantized, shared, reduced to a low-dimensional
auxiliary object, or eliminated.  S4 then reconstructs, broadcasts,
dequantizes, or streams the required quantity into the update.  T4.3 also
uses S1 because the grouping of parameters into blocks, covers, columns, or
layers determines which coordinates share statistics.  T4.4 changes S5 as
well because the writeback occurs during the backward pass, before
all gradients have been materialized.

This definition separates T4 from nearby families.  GaLore belongs to T2.3
because it changes the mathematical subspace in which gradients and Adam
states are updated.  Q-GaLore is discussed in T4.2 only for the additional
quantization component on top of that low-rank mechanism
~\citep{zhao2024galore,zhang2024q}.  Lion uses less memory than AdamW,
but its primary mechanism is a sign-direction map, so it remains T3.  ZeRO,
activation checkpointing, CPU/NVMe offloading, and parameter-efficient
fine-tuning can be combined with T4 optimizers, but they are systems or
training-regime techniques in this taxonomy.  T4 is reserved for methods that change the optimizer's own state
representation or gradient-consumption pipeline.


\definecolor{hidden-blue}{HTML}{4a90e2}
\definecolor{hidden-black}{HTML}{333333}
\tikzstyle{my-box}=[
  rectangle,
  draw=hidden-black,
  rounded corners,
  text opacity=1,
  minimum height=1.5em,
  minimum width=5em,
  inner sep=2pt,
  align=center,
  fill opacity=.5,
]
\tikzstyle{root}=[
  align=center,
]
\tikzstyle{leaf}=[
  my-box,
  minimum height=1.5em,
  text=black,
  font=\normalsize,
  inner xsep=5pt,
  inner ysep=4pt,
  text width=10.5em,
  fill opacity=1,
]
\tikzstyle{leaf1}=[
  my-box,
  minimum height=1.5em,
  fill=yellow!32,
  text=black,
  font=\normalsize,
  inner xsep=5pt,
  inner ysep=4pt,
  text width=17em,
  align=left,
]
\tikzstyle{leaf2}=[
  my-box,
  minimum height=1.5em,
  fill=hidden-blue!57,
  text=black,
  font=\normalsize,
  inner xsep=5pt,
  inner ysep=4pt,
  text width=17em,
  align=left,
]
\tikzstyle{leaf3}=[
  my-box,
  minimum height=1.5em,
  fill=purple!27,
  text=black,
  font=\normalsize,
  inner xsep=5pt,
  inner ysep=4pt,
  text width=17em,
  align=left,
]
\tikzstyle{leaf4}=[
  my-box,
  minimum height=1.5em,
  fill=teal!30,
  text=black,
  font=\normalsize,
  inner xsep=5pt,
  inner ysep=4pt,
  text width=17em,
  align=left,
]

\begin{figure*}[!htbp]
\centering
\resizebox{\textwidth}{!}{
\begin{forest}
forked edges,
for tree={
grow=east,
reversed=true,
anchor=base west,
parent anchor=east,
child anchor=west,
base=left,
font=\large,
rectangle,
draw=hidden-black,
rounded corners,
align=left,
minimum width=4em,
edge+={darkgray, line width=1pt},
s sep=4pt,
inner xsep=5pt,
inner ysep=4pt,
line width=1.1pt,
ver/.style={rotate=90, child anchor=north, parent anchor=south, anchor=center},
},
where level=1{text width=11em, font=\normalsize}{},
where level=2{text width=34.5em, align=left, tier=citations, font=\small}{},
[
    T4-based Optimizers,
    ver, fill=brown!10!white, text=hidden-black, root
    [
        T4.1: Row-Column Factorization,
        fill=yellow!32, leaf1
        [{
            AdaFactor~\cite{shazeer2018adafactor},
            CAME~\cite{luo2023came}
        }, fill=yellow!12]
    ]
    [
        T4.2: Low-Bit Quantization \& Error \\
        Compensation,
        fill=hidden-blue!57, leaf2
        [{
            8-bit Adam~\cite{dettmers20218},
            Q-GaLore~\cite{zhang2024q}
        }, fill=hidden-blue!15]
    ]
    [
        T4.3: Block/Layer-Level State \\ Sharing,
        fill=purple!27, leaf3
        [{
            Adam-mini~\cite{zhang2025adam},
            APOLLO~\cite{zhu2025apollo},
            BlockwiseLR~\cite{wang2025sharpness},
            Conda~\cite{wang2025conda},
            NovoGrad~\cite{ginsburg2019stochastic},\\
            SM3~\cite{anil2019memory},
            SAC~\cite{li2026sac},
            SGG~\cite{li2025taming}
        }, fill=purple!12]
    ]
    [
        T4.4: Fused Backprop-Update,
        fill=teal!30, leaf4
        [{
            LOMO~\cite{lv2024full},
            AdaLOMO~\cite{lv2024adalomo}
        }, fill=teal!12]
    ]
  ]
\end{forest}
}
\caption{Taxonomy of state-compressed and structurally aggregated optimizers.}
\label{tab:t4_family_taxonomy}
\end{figure*}
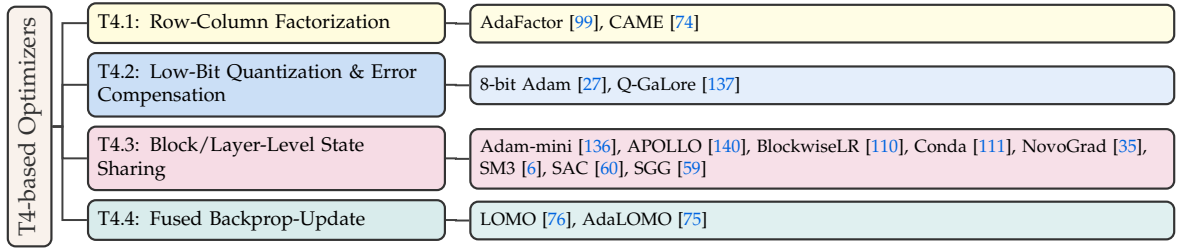

\subsubsection{LMO-Driven Four-Axis Interpretation}
\label{sec:t4-lmo-four-axis}

In the four-axis notation of Section~\ref{sec:lmo-four-axis}, T4 is the most
direct instantiation of state compression in Axis~II.  A generic adaptive update can be written as
\begin{equation}
  \theta_{t+1}
  =
  \theta_t
  - \eta_t\,
  \frac{\widehat{m}_t}{\sqrt{\mathcal{R}(\mathcal{C}(v_t))}+\epsilon},
  \label{eq:t4-compressed-denominator}
\end{equation}
where $\mathcal{C}$ compresses the second-moment state and $\mathcal{R}$
reconstructs, broadcasts, or dequantizes it for the update.  In full AdamW,
$\mathcal{C}$ and $\mathcal{R}$ are both identities.  In T4, at least one of
them is nontrivial.  The geometry of the base direction can remain
coordinate-wise Adam-like, and the innovation is the representation of the
statistics that control the denominator.

For matrix parameters $W_t\in\mathbb{R}^{m\times n}$, a factored method keeps
row and column statistics, for example
\begin{equation}
  r_{i,t}=\beta r_{i,t-1}+(1-\beta)\frac{1}{n}\sum_{j=1}^{n}G_{ij,t}^2,
  \quad
  c_{j,t}=\beta c_{j,t-1}+(1-\beta)\frac{1}{m}\sum_{i=1}^{m}G_{ij,t}^2,
  \label{eq:t4-row-column-states}
\end{equation}
and reconstructs an approximate matrix denominator such as
$\widehat{V}_{ij,t}\propto r_{i,t}c_{j,t}$.  This lowers state from
$O(mn)$ to $O(m+n)$, but it assumes that row and column summaries preserve
enough information about the diagonal curvature proxy.  For block sharing,
the compression map is instead a partition
$\{B_k\}_{k=1}^{K}$ with shared statistics
\begin{equation}
  s_{k,t} = \beta s_{k,t-1}
  +(1-\beta)\frac{1}{|B_k|}\sum_{i\in B_k}g_{i,t}^2,
  \quad i\in B_k,
  \label{eq:t4-block-sharing}
\end{equation}
which is then broadcast to all coordinates in the block.  The compression
quality depends on whether the partition aligns with the structure of the
loss landscape.

Low-bit methods preserve the same state shape but change the value
representation.  With a $b$-bit block quantizer $Q_b$, the stored state is
$Q_b(m_t)$ or $Q_b(v_t)$, and the update
uses $\operatorname{dequant}(Q_b(\cdot))$ at computation time.  The key design
choices are block size, dynamic range tracking, stochastic rounding, and
special handling of sensitive tensors such as embeddings.  Fused
backprop-update methods change a different part of Axis~II, namely that the state may not
be stored at all.  Gradients are consumed as soon as they are produced during
backpropagation, which changes the lifetime of $g_t$ and restricts operations
that require global gradient visibility.

\subsubsection{Representative Methods}
\label{sec:t4-methods}

\noindent
The following discussion follows the T4.1--T4.4 subclass order introduced
above.  Within each subclass, representative methods are grouped by the local
state operation they modify.

\paragraph{Row-Column Factorization.}
\label{sec:t4-factorized}

The first T4 branch replaces a dense matrix-valued second-moment state by two
marginal summaries.  For $G_t\in\mathbb{R}^{m\times n}$, the representative
template is
\begin{align}
  R_t &= \beta R_{t-1}+(1-\beta)\operatorname{rowmean}(G_t\odot G_t), \quad
  C_t = \beta C_{t-1}+(1-\beta)\operatorname{colmean}(G_t\odot G_t), \\
  \widehat{V}_{ij,t} &=\frac{R_{i,t}C_{j,t}}{\operatorname{mean}(R_t)},\quad
  \Delta_{ij,t}=\frac{\widehat{m}_{ij,t}}{\sqrt{\widehat{V}_{ij,t}}+\epsilon}.
  \label{eq:t4-row-column-template}
\end{align}
\begin{itemize}
  \setlength{\itemsep}{2pt}
  \item \textbf{AdaFactor base.}
  AdaFactor is the mathematical anchor of T4.1.  It preserves the
  Adam and RMSProp intuition that recent squared
  gradients provide a useful scale estimate, but replaces the full
  $O(mn)$ matrix state with $O(m+n)$ row and column accumulators
  ~\citep{shazeer2018adafactor}.  It reconstructs a cheaper approximation to a
  diagonal denominator while staying at the diagonal, coordinate-wise scale level.
  \item \textbf{Confidence-guided correction.}  CAME starts from the same
  factored-state premise and adds confidence-guided correction to damp or
  reweight entries where the factorized surrogate is unreliable
  ~\citep{luo2023came}.  Its role is to show the second step in compressed
  optimization.  Once state is approximated, the optimizer also needs a way to
  detect and control compression error.
  \item \textbf{Boundary to shared accumulators.}  SM3 is adjacent because a
  matrix cover can resemble row-column accumulators, but its defining
  abstraction is a general cover-based shared state~\citep{anil2019memory}.  We therefore
  discuss SM3 in T4.3.
\end{itemize}

The common thread is an approximation to a diagonal second-moment tensor.  The
descent geometry itself is unchanged.  The main benchmark question is therefore
whether the row and column marginals preserve enough curvature heterogeneity
for the target Transformer block, sequence length, and clipping rule.

\paragraph{Low-Bit Quantization and Error Compensation.}
\label{sec:t4-low-bit}

The second branch preserves the shape of the base optimizer state but changes
the numerical representation in which that state is stored.  A block-wise
quantized state can be written schematically as
\begin{equation}
  z_{B,t}=Q_b(x_{B,t};a_{B,t}),\quad
  \tilde{x}_{B,t}=\operatorname{dequant}(z_{B,t};a_{B,t}),\quad
  \Delta_t=\mathcal{U}_{\mathrm{Adam}}\big(g_t,\tilde{m}_t,\tilde{v}_t\big),
  \label{eq:t4-low-bit-template}
\end{equation}
where $x_{B,t}$ is a state block, $a_{B,t}$ records the local dynamic range,
and $Q_b$ is a $b$-bit quantizer.
\begin{itemize}
  \setlength{\itemsep}{2pt}
  \item \textbf{Block-wise dynamic quantization.}  8-bit Adam is the central
  T4.2 method.  It identifies optimizer statistics as compressible objects and
  stores them using block-wise dynamic quantization while maintaining an
  Adam-like update interface~\citep{dettmers20218}.  The block-wise range is
  essential, because a single global scale would be too coarse for tensors whose
  coordinates have heterogeneous magnitudes.
  \item \textbf{Quantized low-rank combinations.}  Q-GaLore is the clearest
  cross-family member.  Base GaLore is T2.3 because it projects gradients and
  Adam states into a low-rank subspace.  Q-GaLore adds low-bit projection and
  weight representations, including INT4 projection matrices and stochastic
  rounding for accumulated information~\citep{zhao2024galore,zhang2024q}.
  It is therefore cited here only when the quantization component is under
  discussion.
  \item \textbf{Evaluation consequences.}  Low-bit state storage preserves the
  base optimizer geometry, so an 8-bit Adam update is still Adam-like in
  direction and scaling.  Its possible advantage is O3 memory and sometimes
  bandwidth/cache behavior, while its risks are quantization noise, block-size
  sensitivity, dequantization overhead, and precision-sensitive tensors such
  as embeddings.
\end{itemize}

The defining taxonomy feature of T4.2 is the low-bit representation of
optimizer history.  Dynamic quantization, stochastic rounding, and compensation
mechanisms are implementation-specific tools for making low-bit state usable.

\paragraph{Block- and Layer-Level State Sharing.}
\label{sec:t4-structure-aware}

The third branch reduces the number of adaptive statistics by sharing them
over a partition or structural cover.  The generic form is
\begin{equation}
  s_{k,t}
  =
  \beta s_{k,t-1}
  +(1-\beta)\mathcal{A}_{i\in B_k}(g_{i,t}^2),\quad
  \Delta_{i,t}
  =
  \frac{\widehat{m}_{i,t}}{\sqrt{s_{k(i),t}}+\epsilon},
  \label{eq:t4-shared-state-template}
\end{equation}
where $B_k$ is a block, cover element, column, layer, or auxiliary state
group, $\mathcal{A}$ is an aggregation rule, and $k(i)$ maps coordinate $i$
to the statistic that will be broadcast back to it.
\begin{itemize}
  \setlength{\itemsep}{2pt}
  \item \textbf{Transformer block sharing.}  Adam-mini is the central
  LLM-oriented T4.3 method.  It argues that Adam's coordinate-wise denominator
  contains more distinct learning-rate resources than many Transformer blocks
  require, partitions parameters using Hessian-motivated structure, and
  assigns shared adaptive scales to the resulting blocks
  ~\citep{zhang2025adam}.  In meta-pipeline terms, S1 selects the blocks,
  S3 maintains one statistic per block, and S4 broadcasts that statistic back
  to the block coordinates. Similar designs are appeared in Blockwise LR~\cite{wang2025sharpness}, SGG~\cite{li2025taming},
  and SAC~\cite{li2026sac}, which rescales the learning rate of network blocks
  or the certain groups of parameters to obtain better parameter update controls.
  \item \textbf{Cover and layer extremes.}  SM3 uses tensor covers to maintain
  shared accumulators, so a matrix may be covered by row and column sets while
  higher-order tensors use more general covers~\citep{anil2019memory}. NovoGrad
  moves to a coarser layer-wise statistic or gradient normalization, keeping
  less second-moment state than Adam while retaining momentum and weight-decay
  controls~\citep{ginsburg2019stochastic}.  These methods provide historical
  anchors for the aggregation continuum.
  \item \textbf{Low-dimensional and column-wise aggregation.}  APOLLO
  approximates AdamW's learning-rate adaptation with an auxiliary low-rank
  state based on random projection, aiming for an AdamW-like interface under
  SGD-like memory~\citep{zhu2025apollo}.  Conda uses column-wise second-moment
  normalization after projecting updates into an orthogonal subspace
  ~\citep{wang2025conda}.  Conda remains in T4.3 because its stored adaptive
  statistics are column-level and coarser than coordinate-level, but it is a
  boundary member with a secondary T2 tag.
\end{itemize}

The shared-state branch is controlled by granularity.  Fine blocks preserve
more heterogeneity but save less memory.  Layers or coarse covers save more
state but risk underfitting heterogeneous curvature.  Recent APOLLO and Conda
claims should be read as source-reported preprint evidence until their status
and LLM-scale comparison protocols are finalized.

\paragraph{Fused Backprop-Update.}
\label{sec:t4-fused}

The fourth branch changes the lifetime of gradients, reaching deeper than the
shape or precision of a persistent state tensor.  A fused update can be
schematized as
\begin{equation}
  g_t^{(l)}=\operatorname{backward}_l(\theta_t),\quad
  \theta_{t+1}^{(l)}
  =
  \theta_t^{(l)}-\eta_t\,\mathcal{U}^{(l)}(g_t^{(l)},\mathcal{z}_t^{(l)}),
  \quad
  \operatorname{free}(g_t^{(l)}),
  \label{eq:t4-fused-template}
\end{equation}
where layer $l$ is updated while its gradient is still local to the backward
pass and before a full-model gradient buffer is materialized.
\begin{itemize}
  \setlength{\itemsep}{2pt}
  \item \textbf{Streaming state-free update.}  LOMO defines T4.4.  It updates
  parameters as soon as each layer's gradient becomes available during
  backpropagation, shortening gradient lifetime and eliminating persistent
  Adam-style optimizer states~\citep{lv2024full}.  This is an optimizer-level
  pipeline change because it determines when gradients are consumed and which
  global operations remain feasible.
  \item \textbf{Streaming adaptivity.}  AdaLOMO keeps the fused low-memory
  pipeline but reintroduces adaptive learning-rate information through
  compressed second-moment estimation and grouped update normalization
  ~\citep{lv2024adalomo}.  Its primary label remains T4.4 because the fused
  backprop-update interface is the defining operation, while the factored adaptive
  state is a secondary T4.1-like component.
  \item \textbf{Composition limits.}  Streaming updates can conflict with
  global gradient clipping, delayed low-rank basis refresh, SAM-style second
  gradients, and matrix preconditioners that require complete gradient
  statistics.  T4.4 therefore has stronger composition constraints than
  ordinary state compression.
\end{itemize}

This branch illustrates the most aggressive T4 trade-off.  Removing state and
shortening gradient lifetime can make full-parameter fine-tuning feasible
under tight memory budgets, but the price is reduced access to global gradient
information.  Later streaming methods often reintroduce a small amount of
structured state to recover stability or convergence behavior.

\subsubsection{Effect-Target Assessment}
\label{sec:t4-effect-assessment}

T4 methods have the clearest primary objective among the method families, namely O3 memory reduction.  The evaluation challenge is that
O3 improvements change the feasible training regime.  A method can look worse
than AdamW at the same model and batch size but become useful when it enables
a larger model, longer sequence, or larger batch on fixed hardware.  Conversely,
a memory-efficient optimizer can lose its advantage if compression slows each
step or requires more tokens to reach the same loss.  Table~\ref{tab:t4_effect_assessment}
summarizes the mechanism-informed benchmark prior.

\begin{table*}[b]
  \centering
  \caption{Mechanism-informed effect assessment for T4 subclasses.  The
  entries are design priors for benchmark planning, not empirical
  conclusions.}
  \label{tab:t4_effect_assessment}
  \scriptsize
  \setlength{\tabcolsep}{3pt}
  \renewcommand{\arraystretch}{1.08}
  \begin{tabularx}{\textwidth}{@{}Y Y Y@{}}
    \toprule
    Primary target & Likely cost or failure mode & Benchmark focus \\
    \midrule
    \rowcolor{famT4!18}\multicolumn{3}{c}{\textbf{\textcolor{famT4!62!black}{T4: State compression and structural aggregation}}} \\
    \rowcolor{famT4!8}\multicolumn{3}{@{}l@{}}{\textbf{\textcolor{famT4!62!black}{T4.1 Row-column factorization}}} \\
      O3 second-moment memory reduction
      & Factorization error, possible O4 instability
      & AdamW vs.\ AdaFactor vs.\ CAME at matched clipping and memory \\
    \rowcolor{famT4!8}\multicolumn{3}{@{}l@{}}{\textbf{\textcolor{famT4!62!black}{T4.2 Low-bit quantization and error compensation}}} \\
      O3 state memory, sometimes O2 bandwidth
      & Quantization noise, block-size sensitivity, sensitive embeddings
      & Precision, block size, rounding, convergence at matched memory \\
    \rowcolor{famT4!8}\multicolumn{3}{@{}l@{}}{\textbf{\textcolor{famT4!62!black}{T4.3 Block/layer-level state sharing}}} \\
      O3 fewer adaptive statistics
      & Bad grouping underfits curvature, hurting O1/O4
      & Grouping granularity, partition choice, LR/WD robustness \\
    \rowcolor{famT4!8}\multicolumn{3}{@{}l@{}}{\textbf{\textcolor{famT4!62!black}{T4.4 Fused backprop-update}}} \\
      O3 peak memory via shorter gradient lifetime
      & Incompatible with global-gradient operations, possible convergence gap
      & Peak memory, buffer lifetime, clipping and sharding compatibility \\
    \bottomrule
  \end{tabularx}
\end{table*}

The main benchmark rule is to report both absolute performance and
memory-normalized performance.  If AdamW fits comfortably, a compressed method
must justify any approximation error or overhead through convergence,
stability, or tuning benefits.  If AdamW does not fit, the relevant baseline
may be a smaller model, gradient accumulation, offloading, sharding, or
parameter-efficient fine-tuning, and full-state AdamW alone is then no longer
the right comparison.  T4
evaluation should therefore include peak memory, persistent optimizer state,
gradient-buffer lifetime, step time, token-normalized loss, wall-clock loss,
and sensitivity to learning rate, weight decay, clipping, and quantization
granularity.  Recent optimizer and memory-efficient pretraining studies make
this point explicit.  Optimizer ranking is protocol dependent, and memory
savings are only useful when the saved memory is converted into a better
training configuration~\citep{zhao2025deconstructing,semenov2025benchmarking,wen2025fantastic,glentis2025scalable}.

\subsection{T5: Curvature-Aware and Geometric Regularization}
\label{sec:t5}

\subsubsection{Family Overview and Meta-Pipeline Position}
\label{sec:t5-overview}
T5 contains optimizers whose primary contribution is to impose a curvature or
geometry constraint on the effective gradient, the candidate update, or the
final parameter writeback.  The family is broader than explicit second-order
optimization.  Some members estimate diagonal Hessian information, whereas
others keep the base optimizer unchanged and act only through perturbations,
projections, masks, norm controls, or layer-wise trust ratios.  T5 therefore
asks when a proposed step should be
accepted, rescaled, clipped, filtered, or re-evaluated because of local
sharpness, curvature, direction consistency, parameter scale, or update
geometry.

\begin{figure}[t]
  \centering
  \includegraphics[width=0.75\linewidth]{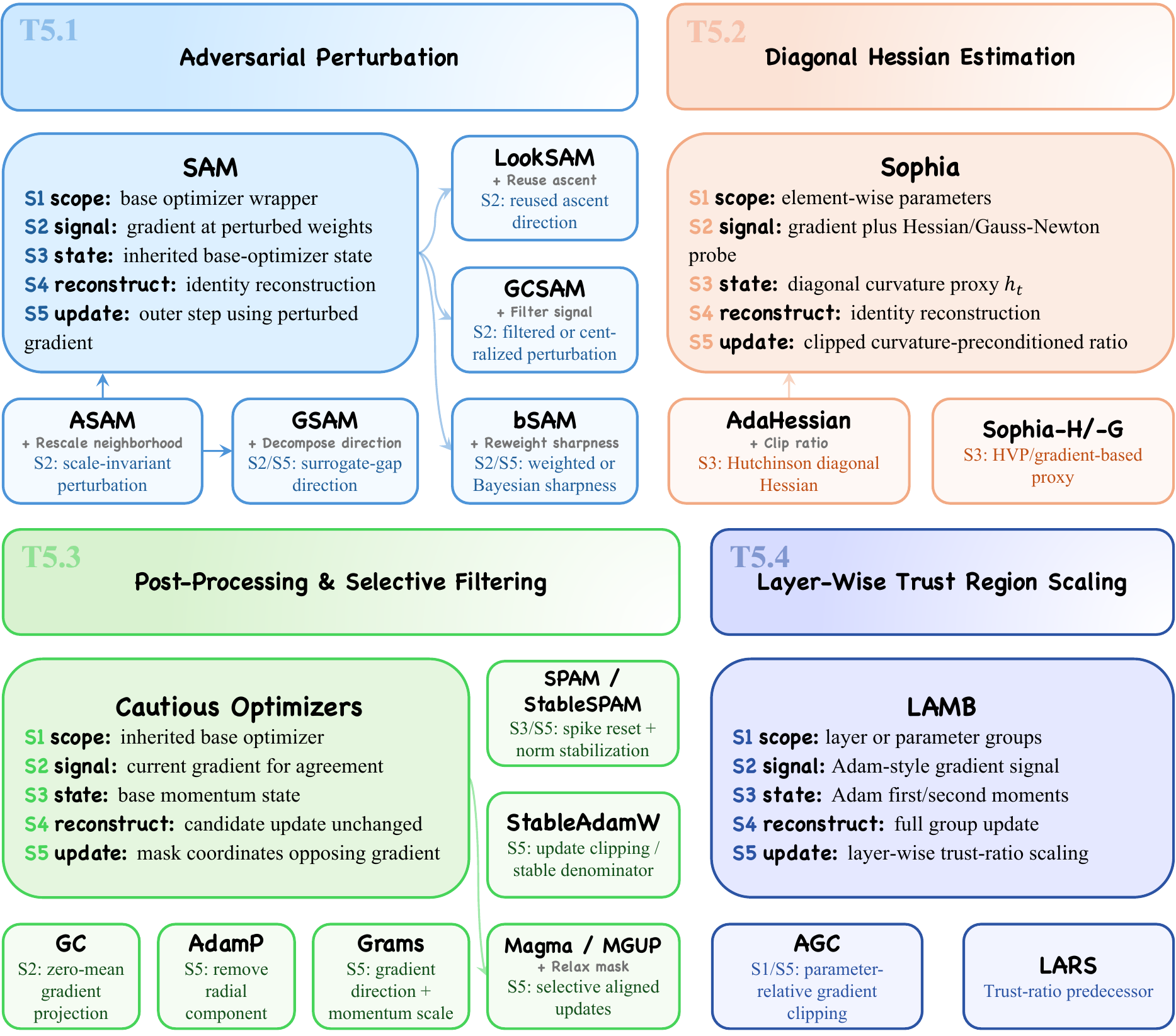}
  \caption{Mechanism schematic for T5 curvature-aware and geometric
  regularization methods.  The schematic organizes the family by where the
  geometric intervention enters the update: perturbed-gradient evaluation,
  diagonal-curvature estimation, post-update filtering, or layer-wise
  trust-region scaling.}
  \label{fig:t5_family_mechanism}
\end{figure}

\definecolor{hidden-blue}{HTML}{4a90e2}
\definecolor{hidden-black}{HTML}{333333}
\tikzstyle{my-box}=[
  rectangle,
  draw=hidden-black,
  rounded corners,
  text opacity=1,
  minimum height=1.5em,
  minimum width=5em,
  inner sep=2pt,
  align=center,
  fill opacity=.5,
]
\tikzstyle{root}=[
  align=center,
]
\tikzstyle{leaf}=[
  my-box,
  minimum height=1.5em,
  text=black,
  font=\normalsize,
  inner xsep=5pt,
  inner ysep=4pt,
  text width=10.5em,
  fill opacity=1,
]
\tikzstyle{leaf1}=[
  my-box,
  minimum height=1.5em,
  fill=yellow!32,
  text=black,
  font=\normalsize,
  inner xsep=5pt,
  inner ysep=4pt,
  text width=17em,
  align=left,
]
\tikzstyle{leaf2}=[
  my-box,
  minimum height=1.5em,
  fill=hidden-blue!57,
  text=black,
  font=\normalsize,
  inner xsep=5pt,
  inner ysep=4pt,
  text width=17em,
  align=left,
]
\tikzstyle{leaf3}=[
  my-box,
  minimum height=1.5em,
  fill=purple!27,
  text=black,
  font=\normalsize,
  inner xsep=5pt,
  inner ysep=4pt,
  text width=17em,
  align=left,
]
\tikzstyle{leaf4}=[
  my-box,
  minimum height=1.5em,
  fill=teal!30,
  text=black,
  font=\normalsize,
  inner xsep=5pt,
  inner ysep=4pt,
  text width=17em,
  align=left,
]

\begin{figure*}[!htbp]
\centering
\resizebox{\textwidth}{!}{
\begin{forest}
forked edges,
for tree={
grow=east,
reversed=true,
anchor=base west,
parent anchor=east,
child anchor=west,
base=left,
font=\large,
rectangle,
draw=hidden-black,
rounded corners,
align=left,
minimum width=4em,
edge+={darkgray, line width=1pt},
s sep=4pt,
inner xsep=5pt,
inner ysep=4pt,
line width=1.1pt,
ver/.style={rotate=90, child anchor=north, parent anchor=south, anchor=center},
},
where level=1{text width=11em, font=\normalsize}{},
where level=2{text width=34.5em, align=left, tier=citations, font=\small}{},
[
    T5-based Optimizers,
    ver, fill=brown!10!white, text=hidden-black, root
    [
        T5.1: Adversarial Perturbation \\ (SAM Family),
        fill=yellow!32, leaf1
        [{
            SAM~\cite{foret2021sharpness},
            ASAM~\cite{kwon2021asam},
            GSAM~\cite{zhuang2022surrogate},
            WSAM~\cite{yue2023sharpness},
            bSAM~\cite{mollenhoff2022sam},
            LookSAM~\cite{liu2022towards}, \\
            FriendlySAM~\cite{li2024friendly},
            GCSAM~\cite{hassan2025gcsam}
        }, fill=yellow!12]
    ]
    [
        T5.2: Diagonal Hessian Estimation,
        fill=hidden-blue!57, leaf2
        [{
            Sophia~\cite{liu2023sophia},
            AdaHessian~\cite{yao2021adahessian}
        }, fill=hidden-blue!15]
    ]
    [
        T5.3: Update Post-Processing \\ \& Selective Filtering,
        fill=purple!27, leaf3
        [{
            GC~\cite{yong2020gradient},
            AdamP~\cite{heo2020adamp},
            StableAdamW~\cite{wortsman2023stable},
            Norm Loss~\cite{wortsman2023stable},
            Stable WD~\cite{wortsman2023stable}, \\
            Grams~\cite{cao2024grams},
            C-AdamW~\cite{liang2024cautious},
            C-Lion~\cite{liang2024cautious},
            SPAM~\cite{huang2025spam},
            StableSPAM~\cite{huanggradientstabilizer},
            Magma~\cite{joo2026surprising}, \\
            MGUP~\cite{chang2026mgup}
        }, fill=purple!12]
    ]
    [
        T5.4: Layer-wise Trust Region Scaling,
        fill=teal!30, leaf4
        [{
            LAMB~\cite{you2019large},
            AGC~\cite{brock2021high}
        }, fill=teal!12]
    ]
  ]
\end{forest}
}
\caption{Taxonomy of curvature-aware and geometry-regularized optimizers.
The prefix C- denotes Cautious.}
\label{tab:t5_family_taxonomy}
\end{figure*}
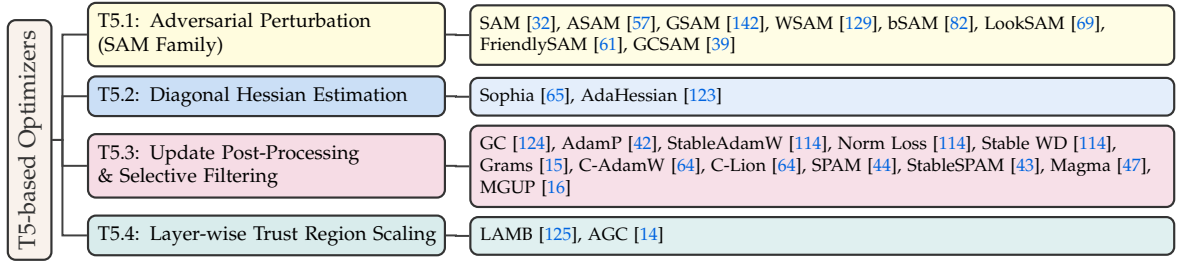

\paragraph{Subclass structure.}
Figure~\ref{tab:t5_family_taxonomy} gives the complete optimizer membership for
T5.  The four subclasses below interpret that taxonomy figure by the dominant
geometric operation.
\begin{enumerate}[label=\textbf{T5.\arabic*},leftmargin=*,itemsep=2pt]
  \item \textbf{Adversarial perturbation (SAM family).}
  SAM replaces the ordinary gradient by a gradient measured at a local
  adversarial perturbation~\citep{foret2021sharpness}.  ASAM changes the
  perturbation geometry~\citep{kwon2021asam}.  Later variants modify the
  sharpness objective or its probabilistic interpretation
  ~\citep{zhuang2022surrogate,yue2023sharpness,mollenhoff2022sam}, reduce perturbation
  frequency~\citep{liu2022towards}, separate stochastic-noise components in
  the perturbation~\citep{li2024friendly}, or combine SAM with gradient
  centralization~\citep{hassan2025gcsam}.
  \item \textbf{Diagonal Hessian estimation.}
  Sophia maintains a diagonal Hessian or Gauss--Newton proxy and uses clipping
  to form an LLM-oriented trust-region-like update~\citep{liu2023sophia}.
  AdaHessian provides the earlier randomized diagonal-Hessian route
  ~\citep{yao2021adahessian}.
  \item \textbf{Update post-processing and selective filtering.}
  This subclass centralizes, projects, normalizes, masks, stabilizes, or
  selectively amplifies a gradient or candidate update after the base direction
  has been defined.  Gradient Centralization and AdamP are projection-style
  instances~\citep{yong2020gradient,heo2020adamp}, StableAdamW and its norm or
  decay components stabilize AdamW-style writeback~\citep{wortsman2023stable},
  Cautious Optimizers instantiate direction-consistency masks
  ~\citep{liang2024cautious}, Grams separates direction and adaptive
  magnitude~\citep{cao2024grams}, SPAM and StableSPAM target spike-aware or
  norm-stabilized writeback
  ~\citep{huang2025spam,huanggradientstabilizer}, Magma and
  MGUP occupy the selective-update and momentum-gradient-alignment corner of
  the subclass~\citep{joo2026surprising,chang2026mgup}.
  \item \textbf{Layer-wise trust-region scaling.}
  LAMB compares update and parameter norms to rescale Adam-style layer updates
  ~\citep{you2019large}, whereas AGC clips gradients relative to the receiving
  parameter group~\citep{brock2021high}.
\end{enumerate}

At the family level, a T5 method can be summarized as
\begin{align}
  \tilde{g}_t
  &=
  \mathcal{P}_t(g_t,\theta_t),\quad
  c_t =
  \mathcal{H}_t(\tilde{g}_t,\mathcal{S}_t), \\
  \Delta_t^{\mathrm{base}}
  &=
  \mathcal{U}_{\mathrm{base}}(\tilde{g}_t,\mathcal{S}_t,c_t),\quad
  \Delta_t =
  \mathcal{G}_t(\Delta_t^{\mathrm{base}},\tilde{g}_t,\theta_t,c_t), \\
  \theta_{t+1}
  &=
  \theta_t-\eta_t\Delta_t .
  \label{eq:t5-family-template}
\end{align}
Here $\mathcal{P}_t$ denotes the effective-gradient map, which is non-identity for SAM-style perturbations and some filtering methods, $c_t$ is an optional curvature or scale signal, $\mathcal{U}_{\mathrm{base}}$ is the underlying optimizer update, and $\mathcal{G}_t$ is the T5 geometry operator.  Ordinary AdamW with no wrapper is the identity case
$\mathcal{P}_t=I$, $\mathcal{H}_t=\varnothing$, and $\mathcal{G}_t=I$.

In the meta-pipeline, T5 primarily acts near S5, but different subclasses
enter through different interfaces.  T5.1 reaches S5 through an S0/S2
perturbed-gradient evaluation.  T5.2 modifies S3 by storing a curvature proxy
and modifies S5 by clipping the resulting ratio.  T5.3 uses S2 or S5 to filter
the gradient or candidate update.  T5.4 uses S1 to define layer or
parameter-group routes and S5 to apply group-level trust ratios.  S4 is
usually identity-like because the update remains full-dimensional unless the
method is primarily a T2 subspace method or a T4 state-compression method.

This definition fixes the boundary with the previous families.  A direct
AdamW variant remains T1 even if it improves stability, because its decisive
change is still the element-wise moment formula.  Muon, Shampoo, and GaLore
remain T2 because they change the matrix basis or subspace before the update
is formed.  Lion remains T3 because its sign map defines the base direction.
AdaFactor, 8-bit Adam, and Adam-mini remain T4 because they primarily alter
state representation.  By contrast, SAM can wrap SGD or AdamW, Cautious Lion
can wrap Lion, and LAMB can wrap an Adam-style update.  Their primary labels
come from the wrapper-level geometry added around the base optimizer.

\subsubsection{LMO-Driven Four-Axis Interpretation}
\label{sec:t5-lmo-four-axis}

T5 exposes the boundary between the LMO direction coordinate and the
post-direction geometry of an optimizer.  The LMO view explains how a base
direction is selected under a norm constraint.  T5 asks whether that direction
should be evaluated at a different point, clipped by curvature, filtered by
alignment, or rescaled by parameter-group geometry before it is written back.
In four-axis notation, the base optimizer first supplies
\begin{equation}
  \big((B_{\mathrm{a}},B_{\mathrm{u}}),\,
  \psi_{\mathrm{base}},\,\widehat{H}_t,\,
  \alpha,\,\widehat{g}_t,\,\mathcal{R}\big),
\end{equation}
and T5 appends an outer operator to the resulting step.  For methods whose
effective gradient is unchanged, Eq.~\eqref{eq:t5-family-template} reduces to
\begin{equation}
\begin{aligned}
  \Delta_t^{\mathrm{base}}
  = \mathcal{U}_{\mathrm{base}}(g_t,\mathcal{S}_t),
   \quad \Delta_t=\mathcal{G}_t\!\left(\Delta_t^{\mathrm{base}}, g_t, \theta_t\right), \quad
  \theta_{t+1}=\theta_t-\eta_t\Delta_t .
\end{aligned}
\label{eq:t5-geometry-wrapper}
\end{equation}
Here $\mathcal{U}_{\mathrm{base}}$ may be SGD, AdamW, Lion, Muon, or another
optimizer.  For most T5.3 and T5.4 methods, the base optimizer's four-axis
coordinates remain intact, and the T5 mechanism is the additional S5 map
$\mathcal{G}_t$.

SAM is different because it changes the point at which the gradient is
measured.  Its local objective can be written as
\begin{equation}
\begin{aligned}
  \min_{\theta}\ \max_{\|\epsilon\|\leq \rho}
  \mathcal{L}(\theta+\epsilon),
  \quad
  \epsilon_t
  \approx
  \rho\,\frac{g_t}{\|g_t\|_2}.
\end{aligned}
\label{eq:t5-sam-objective}
\end{equation}
The update then uses $\nabla \mathcal{L}(\theta_t+\epsilon_t)$ in place of the
ordinary mini-batch gradient.  This makes SAM an S0/S5 wrapper, where the inner
ascent direction is a local geometric probe, while the outer step can still
be executed by SGD, AdamW, or another base optimizer.  The price is the
second forward-backward pass needed to evaluate the perturbed gradient.

Diagonal-curvature methods fit more directly into Axes~II and~III.  Sophia-H
uses a stochastic Hessian-vector product to estimate a diagonal curvature
proxy, while Sophia-G uses a cheaper gradient-based proxy.  Both use clipping
to avoid unconstrained Newton-like steps~\citep{liu2023sophia}.  In schematic
form,
\begin{equation}
\begin{aligned}
  h_t=\beta_2 h_{t-1}+(1-\beta_2)\widehat{\operatorname{diag}}(H_t), \quad
  \Delta_t=\operatorname{clip}\!\left(\frac{m_t}{\gamma h_t+\epsilon}, -1, 1\right).
\end{aligned}
\label{eq:t5-sophia-clipping}
\end{equation}
AdaHessian is the earlier diagonal-Hessian route, using Hutchinson-style
estimation and Hessian-powered adaptive scaling~\citep{yao2021adahessian}.
Compared with AdamW, T5.2 therefore changes both the curvature estimator and
the final direction function.

Post-processing and layer-wise trust-region methods are simpler in the
four-axis view but important in practice.  A cautious filter can be expressed
as
\begin{equation}
  \mathcal{G}^{\mathrm{cautious}}(\Delta,g)
  =
  \Delta \odot \mathbf{1}[\Delta\odot g > 0],
\label{eq:t5-cautious-filter}
\end{equation}
up to the sign convention used for $\Delta$.  The key property is that only
coordinates whose proposed update agrees with the instantaneous gradient are
kept~\citep{liang2024cautious}.  A LAMB-style trust ratio instead rescales
the whole layer:
\begin{equation}
\begin{aligned}
  \Delta_t^{(l)}
  &=
  \phi_t^{(l)}\Delta_{t,\mathrm{Adam}}^{(l)},
  \quad
  \phi_t^{(l)}
  =
  \frac{\|\theta_t^{(l)}\|_2}
       {\|\Delta_{t,\mathrm{Adam}}^{(l)}\|_2+\epsilon}.
\end{aligned}
\label{eq:t5-lamb-ratio}
\end{equation}
Both operations illustrate the same principle, namely that T5 often leaves the base
direction geometry untouched, then constrains which parts of that direction
are written into the parameters and at what group-level scale.

\subsubsection{Representative Methods}
\label{sec:t5-methods}

\noindent
The following discussion follows the T5.1--T5.4 subclass order introduced
above.  Within each subclass, representative methods are grouped by the local
geometry operation they modify.

\paragraph{Adversarial Perturbation (SAM Family).}
\label{sec:t5-adversarial-perturbation}

The first branch changes the gradient evaluation point.  At the survey level,
SAM-style methods can be written as
\begin{equation}
\begin{aligned}
  \epsilon_t=\mathcal{A}_t(g_t,\theta_t;\rho), \quad
  \tilde{g}_t=\nabla_{\theta}\mathcal{L}(\theta_t+\epsilon_t), \quad
  \Delta_t=\mathcal{U}_{\mathrm{base}}(\tilde{g}_t,\mathcal{S}_t),
\end{aligned}
\label{eq:t5-adversarial-template}
\end{equation}
where $\mathcal{A}_t$ specifies the perturbation rule and
$\mathcal{U}_{\mathrm{base}}$ can be SGD, AdamW, or another first-order base
optimizer.
\begin{itemize}
  \setlength{\itemsep}{2pt}
  \item \textbf{Min--max anchor.}  SAM instantiates
  $\mathcal{A}_t$ by the approximate adversarial perturbation in
  Eq.~\eqref{eq:t5-sam-objective}, then updates with the perturbed gradient
  ~\citep{foret2021sharpness}.  This directly connects optimization to
  flatness-motivated generalization, but it also changes the cost model, because
  without approximation or reuse, one step requires two sequential
  forward-backward passes.
  \item \textbf{Perturbation geometry and sharpness objective.}  ASAM changes
  the neighborhood geometry to reduce sensitivity to parameter rescaling
  ~\citep{kwon2021asam}.  GSAM introduces a surrogate-gap objective and
  direction decomposition to balance loss decrease and sharpness reduction
  ~\citep{zhuang2022surrogate}.  WSAM treats weighted sharpness as a regularization
  term~\citep{yue2023sharpness}.  bSAM links the SAM objective to a Bayesian
  relaxation and motivates Adam-like uncertainty-aware variants
  ~\citep{mollenhoff2022sam}.
  \item \textbf{Perturbation frequency and signal composition.}  LookSAM
  targets the extra-gradient bottleneck by periodically refreshing or reusing
  the ascent direction~\citep{liu2022towards}.  FriendlySAM separates
  full-gradient and stochastic-gradient-noise components in the perturbation
  signal~\citep{li2024friendly}.  GCSAM combines the perturbation route
  with gradient centralization before the ascent step
  ~\citep{hassan2025gcsam}.
\end{itemize}

The common thread is perturbed-gradient evaluation.  Claims about this branch should therefore be read through the
matched-compute lens: improvements under a fixed number of tokens do not
automatically imply improvements under a fixed wall-clock or accelerator
budget.

\paragraph{Diagonal Hessian Estimation.}
\label{sec:t5-diagonal-hessian}

The second branch replaces AdamW's squared-gradient denominator with a more
explicit diagonal curvature proxy.  A compact template is
\begin{align}
  q_t
  &\approx
  \widehat{\operatorname{diag}}(H_t)
  \quad\text{or}\quad
  \widehat{\operatorname{diag}}(G_t^{\mathrm{GN}}),\\
  h_t
  &=
  \beta_2 h_{t-1}+(1-\beta_2)q_t,\quad
  \Delta_t =
  \Pi_{\gamma}\!\left(\frac{m_t}{h_t+\epsilon}\right),
  \label{eq:t5-diagonal-curvature-template}
\end{align}
where $\Pi_{\gamma}$ denotes a clipping or trust-region map that prevents
small or noisy curvature estimates from producing unbounded Newton-like
coordinates.
\begin{itemize}
  \setlength{\itemsep}{2pt}
  \item \textbf{LLM-oriented clipped curvature.}  Sophia is the main
  representative method for LLM pretraining.  Sophia-H estimates curvature
  with a stochastic Hessian-vector product, whereas Sophia-G uses a cheaper
  gradient-based proxy.  Both combine the proxy with a clipped coordinate-wise
  update~\citep{liu2023sophia}.
  \item \textbf{Diagonal-Hessian predecessor.}  AdaHessian uses randomized
  Hessian-vector products to estimate diagonal curvature and smooths that
  estimate before applying Hessian-powered adaptive scaling
  ~\citep{yao2021adahessian}.  It establishes the diagonal-Hessian route, but
  its target regime predates current LLM pretraining protocols.
\end{itemize}

T5.2 is adjacent to both T1 and T2 but distinct from each.  It differs from T1
because its denominator is intended to approximate Hessian or Gauss--Newton
curvature.  It differs from T2 because it
remains diagonal.  Benchmark reports should therefore expose Hessian-estimation
cadence, HVP batch, Sophia-H versus Sophia-G choice, clipping parameters, and
mixed-precision behavior.

\paragraph{Update Post-Processing and Selective Filtering.}
\label{sec:t5-post-processing}

The third branch acts after a base direction has been formed.  The taxonomy
figure lists methods whose surface forms differ substantially, but the common
mechanism is an additional operator between the base optimizer and parameter
writeback.  At this level, the branch can be summarized as
\begin{equation}
\begin{aligned}
  \bar{\Delta}_t=\Pi_t(\Delta_t^{\mathrm{base}},g_t,\theta_t),\quad
  M_t=\mathcal{M}_t(\bar{\Delta}_t,g_t,\theta_t,\mathcal{S}_t),\quad
  \Delta_t=M_t\odot\bar{\Delta}_t ,
\end{aligned}
\label{eq:t5-filter-template}
\end{equation}
where $\Pi_t$ may centralize, project, normalize, or stabilize the candidate
update, and $M_t$ is a coordinate mask or selection weight.  The all-ones mask
recovers projection-only methods.  This template deliberately separates
post-processing from base-direction construction, because $\Delta_t^{\mathrm{base}}$
may come from AdamW, Lion, Muon, or another optimizer, while the T5.3
contribution is the subsequent geometric test, projection, norm control, or
selection policy.
\begin{itemize}
  \setlength{\itemsep}{2pt}
  \item \textbf{Projection and radial control.}  Gradient Centralization
  subtracts the mean from gradient vectors or convolutional kernels, placing
  the update in a zero-mean subspace that can be attached to SGD, Adam, or
  related optimizers~\citep{yong2020gradient}.  AdamP addresses a different
  geometric failure mode by removing radial components for
  scale-invariant weights, so that accumulated momentum does not mainly increase parameter
  norms~\citep{heo2020adamp}.  Both methods leave the base optimizer
  recognizable, and their contribution is the projection $\Pi_t$ before writeback.
  \item \textbf{Stabilized writeback and norm control.}  StableAdamW analyzes
  loss spikes caused by unstable AdamW-style update magnitudes and motivates
  stabilized denominators or AdamW--AdaFactor hybrids~\citep{wortsman2023stable}.
  The Norm Loss and Stable WD entries in Figure~\ref{tab:t5_family_taxonomy}
  should be read as related S5 regularization components from the same
  stability-oriented design space.  Their
  role in this taxonomy is to expose norm and decay control as writeback
  mechanisms that can change stability without changing the moment estimator.
  \item \textbf{Direction consistency and magnitude separation.}  Cautious
  Optimizers apply the mask in Eq.~\eqref{eq:t5-cautious-filter} to a proposed
  update, keeping coordinates only when the update and instantaneous gradient
  agree under the paper's sign convention~\citep{liang2024cautious}.  This is
  why Cautious AdamW (C-AdamW) and Cautious Lion (C-Lion) share the same T5.3
  label even though their base directions belong to T1 and T3, respectively.  Grams makes a complementary
  separation, where the current gradient supplies direction, while momentum mainly
  supplies adaptive magnitude scaling~\citep{cao2024grams}.
  \item \textbf{Spike-aware and selective-update policies.}  SPAM identifies
  gradient spikes in LLM training as a stability bottleneck and combines
  spike-aware clipping, momentum reset, and sparse momentum behavior
  ~\citep{huang2025spam}.  StableSPAM broadens this idea into norm-stabilized
  writeback, keeping the instantaneous gradient direction while replacing
  unstable update magnitudes with running norm statistics
  ~\citep{huanggradientstabilizer}.  Magma demonstrates that random or
  momentum-aligned masks can be effective when applied to adaptive optimizer
  updates~\citep{joo2026surprising}, whereas MGUP assigns larger steps to
  momentum-gradient-aligned coordinates and smaller nonzero steps elsewhere
  ~\citep{chang2026mgup}.
\end{itemize}

The common thread is that T5.3 filters operate on gradients or candidate
updates after the base optimizer's core direction has been selected.  This
also fixes the boundary with T3.  Sign methods define the base direction
itself, whereas T5.3 methods accept, reject, rescale, or regularize an already
computed direction.  Consequently, a Lion wrapper with a cautious mask belongs
to T5.3 as a wrapper-level contribution, while Lion itself remains T3.

\paragraph{Layer-Wise Trust-Region Scaling.}
\label{sec:t5-layerwise-trust-region}

The fourth branch uses parameter-group geometry to determine the step scale.
For a layer or parameter group $l$, the representative form is
\begin{align}
  r_t^{(l)}
  &=
  \frac{\|\theta_t^{(l)}\|_2}
       {\|\Delta_{t,\mathrm{base}}^{(l)}\|_2+\epsilon},\quad
  \Delta_t^{(l)}
  =
  \Psi(r_t^{(l)})\,\Delta_{t,\mathrm{base}}^{(l)}, \\
  g_t^{(l)}
  &\leftarrow
  g_t^{(l)}
  \min\!\left(
  1,\frac{\tau\|\theta_t^{(l)}\|_2}{\|g_t^{(l)}\|_2+\epsilon}
  \right),
  \label{eq:t5-layerwise-template}
\end{align}
where the first line is a trust-ratio update and the second line is an
adaptive clipping rule.
\begin{itemize}
  \setlength{\itemsep}{2pt}
  \item \textbf{Trust-ratio update.}  LAMB first forms an Adam-style update and
  then rescales it by the layer-wise ratio in Eq.~\eqref{eq:t5-lamb-ratio}
  ~\citep{you2019large}.  Its contribution is
  the S1/S5 layer-wise grouping and trust-ratio writeback used for large-batch
  training.
  \item \textbf{Parameter-relative clipping.}  Adaptive Gradient Clipping
  clips a layer or unit when the gradient norm is too large relative to the
  parameter norm~\citep{brock2021high}.  Unlike fixed global clipping, the
  threshold scales with the receiving parameter group.
\end{itemize}

Layer-wise trust-region methods are syntactically easy to combine with
Adam-style bases, but their behavior depends on parameter-group conventions.
Benchmarks should state which tensors are excluded, how zero or near-zero
norms are handled, whether trust ratios are clipped, and whether the ratio is
applied before or after weight decay and global clipping.

\subsubsection{Effect-Target Assessment}
\label{sec:t5-effect-assessment}

T5 acts on the geometry of the candidate step rather than on the base
optimizer alone.  The expected benefit is therefore stability or
generalization: SAM-style perturbations seek flatter neighborhoods, diagonal
curvature methods clip steps by local curvature, post-update filters suppress
harmful components, and layer-wise trust ratios control scale mismatch across
parameter groups.  Each mechanism also adds a protocol-dependent cost.  SAM
spends extra forward-backward work~\citep{foret2021sharpness}; Sophia-style
updates rely on curvature proxies and clipping choices~\citep{liu2023sophia};
LAMB-style trust ratios depend on layer norms and excluded tensors
~\citep{you2019large}.  Table~\ref{tab:t5_effect_assessment} summarizes the
mechanism-derived performance priors.

\begin{table*}[b]
  \centering
  \caption{Mechanism-informed effect assessment for T5 subclasses.  The
  entries are design priors for benchmark planning, not empirical
  conclusions.}
  \label{tab:t5_effect_assessment}
  \scriptsize
  \setlength{\tabcolsep}{3pt}
  \renewcommand{\arraystretch}{1.08}
  \begin{tabularx}{\textwidth}{@{}Y Y Y@{}}
    \toprule
    Primary target & Likely cost or failure mode & Benchmark focus \\
    \midrule
    \rowcolor{famT5!18}\multicolumn{3}{c}{\textbf{\textcolor{famT5!62!black}{T5: Curvature-aware and geometric regularization}}} \\
    \rowcolor{famT5!8}\multicolumn{3}{@{}l@{}}{\textbf{\textcolor{famT5!62!black}{T5.1 Adversarial perturbation (SAM family)}}} \\
      O6 generalization, O4 flatness
      & Extra forward-backward pass, radius sensitivity, wall-clock loss
      & Matched-token vs.\ matched-compute curves, radius and frequency, downstream evaluation \\
    \rowcolor{famT5!8}\multicolumn{3}{@{}l@{}}{\textbf{\textcolor{famT5!62!black}{T5.2 Diagonal Hessian estimation}}} \\
      O1 convergence, O4 stability via clipped curvature
      & HVP overhead, clipping sensitivity, stale curvature
      & Estimation cadence, Sophia-H vs.\ Sophia-G, clipping rule \\
    \rowcolor{famT5!8}\multicolumn{3}{@{}l@{}}{\textbf{\textcolor{famT5!62!black}{T5.3 Update post-processing and selective filtering}}} \\
      O4 stability, O5 tuning width
      & May drop useful descent components, extra hyperparameters
      & Same base with/without filter, kept-coordinate ratio, spike rate \\
    \rowcolor{famT5!8}\multicolumn{3}{@{}l@{}}{\textbf{\textcolor{famT5!62!black}{T5.4 Layer-wise trust-region scaling}}} \\
      O4/O5 under large batch and heterogeneous layers
      & Trust-ratio instability at small norms, exclusion sensitivity
      & Layer-wise norms, excluded tensors, clipping order, large-batch scaling \\
    \bottomrule
  \end{tabularx}
\end{table*}

The distinction between mechanism and observed effect is especially important
for this family.  A method belongs to T5 because it imposes a geometric
constraint; it need not improve generalization in every benchmark.  Conversely,
a T1 or T2 optimizer can be stable without becoming T5.  Budgeted optimizer
studies such as Schlotthauer et al.~\citep{schlotthauer2025budget} make this
point concrete: extra computation can change the apparent ranking if the
comparison is made at fixed steps rather than fixed compute.  The benchmark
summary in Sec.~\ref{sec:benchmark-family-summary} is consistent with this
caution: T5 methods are competitive in limited regimes, but they do not show a
uniform advantage under the current LLM benchmark.
\section{Benchmark Study}
\label{sec:benchmark}

This section presents a mechanism-aware benchmark of representative optimizers
for LLM pretraining. Rather than treating the benchmark as a standalone
leaderboard, we use the framework developed in the preceding sections---the
universal meta-pipeline, the LMO-driven four-axis decomposition, and the
dual-dimension taxonomy---to organize and interpret the empirical results.
Each experiment is linked to a concrete framework question: which optimizer
families perform well under controlled short-context pretraining, which
quality--efficiency trade-offs emerge across optimization quality, runtime,
and memory cost (O1--O3), whether short-context advantages transfer to
long-context and cross-architecture settings (O6), and how training dynamics
and hyperparameter perturbations reveal additional stability and robustness
properties (O4--O5). We further use Muon as a representative
matrix-structured optimizer to study how its sub-operations compose across
scale and architecture.

Compared with existing LLM-pretraining optimizer benchmarks
(e.g., \emph{Fantastic Pretraining Optimizers}~\cite{wen2025fantastic},
\emph{Benchmarking Optimizers for LLM Pretraining}~\cite{semenov2025benchmarking}),
our benchmark emphasizes three aspects. First, results are analyzed not only at
the optimizer level, but also at the family and axis levels, so that empirical
patterns can be mapped back to the methodological taxonomy. Second, we evaluate
optimizers under a multi-objective protocol: final quality is measured by
perplexity, Stage~2 additionally reports downstream Commonsense Reasoning
average accuracy (CS Avg.), efficiency is measured by per-step runtime and
optimizer-state memory, stability is analyzed through gradient-norm dynamics,
robustness is probed by learning-rate perturbation, and generalization is
evaluated by cross-scenario transfer. Third, we conduct a mechanistic ablation
of Muon to examine how matrix-structured optimizer
components---orthogonalization, scaling, momentum placement, and
finalization---interact. In this way, the benchmark both evaluates practical
optimizer choices and provides empirical feedback on the proposed framework.

\subsection{Experimental Setup}
\label{sec:benchmark-setup}

\paragraph{Optimizer selection and taxonomy coverage.}
We instantiate the taxonomy by selecting 24 representative optimizers spanning
all five methodological families (T1--T5) and the major regions of the
four-axis coordinate space. Their four-axis coordinates are listed in
Table~\ref{tab:four-axis-instantiation} (Section~\ref{sec:four-axis-decomposition}),
which places each optimizer in its family and records its update domain, state
estimator, geometry-and-precondition operator, and finalization. This makes
explicit that our optimizer set is a systematic coverage of the classification
space.

\paragraph{Training protocol.}
Our evaluation comprises two complementary stages.
Stage~1 sweeps all optimizers on C4~\cite{raffel2020exploring} under the
LLaMA architecture at sequence length 256, across four model scales---60M,
130M, 350M, and 1B (trained for 10k, 20k, 60k, and 100k steps,
respectively)---using final C4 validation perplexity as the quality metric.
Stage~2 transfers the stronger Stage-1
optimizers to the higher-quality FineWeb-Edu
corpus~\cite{penedo2024finewebdatasetsdecantingweb,karpathy2024finewebedu100b}
with long 32k sequences,
at 340M and 1B ($\sim$30k steps), and across four
architectures---Transformer++~\cite{yang2023gated}
(standard attention) and three linear-attention variants (Gated
DeltaNet~\cite{yang2025gated},
DeltaNet~\cite{yang2024parallelizing}, GLA~\cite{yang2023gated})---to probe portability across dataset, sequence length, and
architecture. We weight the 350M and 1B results most heavily, as small-scale
rankings are unreliable predictors of behavior at scale. To ensure that
observed differences genuinely stem from the optimizers themselves, we adopt a
strict \emph{controlled-variable} principle: within each
stage, only optimizer-related hyperparameters are tuned (\texttt{betas},
\texttt{eps}, \texttt{lr}, and method-specific knobs such as APOLLO's projection
rank and projection interval~$T$), while all architectural, data, and schedule
settings are held fixed across optimizers.
 
\paragraph{Stage-dependent regularization.}
The two S5 (Update Finalization) operations of the meta-pipeline---decoupled
weight decay and gradient clipping---are treated differently across the two
stages, by design. In Stage~1 both are \emph{disabled}, so that the comparison
isolates the S2/S3 machinery that distinguishes one family from another, and
separates it from a generic finalization-stage regularizer. In Stage~2
both are \emph{enabled} identically for all optimizers, to measure
generalization under a realistic, production-style recipe. This also lets us
distinguish whether an optimizer's advantage is intrinsic to its S2/S3 design
(surviving into the regularized regime) or merely an artifact of the
unregularized Stage-1 setting.

\paragraph{Evaluation metrics (mapped to O1--O6).}
\textbf{O1 (optimization quality):} C4 validation PPL in Stage~1, while
Stage~2 uses WikiText~\cite{merity2016pointer} test PPL and downstream Commonsense Reasoning average
accuracy (CS Avg.). CS Avg. is evaluated with lm-eval-harness~\cite{eval-harness} and
computed over ten tasks/splits: ARC-Easy, ARC-Challenge, HellaSwag,
WinoGrande, PIQA, OpenBookQA, BoolQ, COPA, LAMBADA-OpenAI, and
SciQ~\cite{clark2018think,zellers2019hellaswag,sakaguchi2021winogrande,bisk2020piqa,mihaylov2018can,clark2019boolq,roemmele2011choice,paperno-EtAl:2016:P16-1,welbl2017crowdsourcing}. \textbf{O2 (per-step compute):} isolated wall-clock
optimizer runtime per step ($T$, ms). \textbf{O3 (memory):} isolated
optimizer-state memory (Mem, GB), excluding model parameters, activations, and
gradients. \textbf{O4 (training stability):} long-context gradient-norm
dynamics, mainly post-warmup GNormCV, with spike and NaN/Inf diagnostics.
\textbf{O5 (hyperparameter robustness):} learning-rate perturbation around the
tuned learning rate, using $1/5$ and $5\times$ perturbations. \textbf{O6
(generalization):} cross-scenario transfer across dataset, context length,
scale, and architecture, summarized by rank stability and sequence-length
sensitivity. These measurements support the optimizer-level and family-level
O1--O6 assessment below.
\subsection{Results and Analysis}
\label{sec:benchmark-main-results}

\subsubsection{Stage~1: Broad Screening on C4}
\label{sec:benchmark-stage1}

We first conduct a broad screening of all 24 optimizers on C4 under the
LLaMA architecture. This stage is designed as a controlled short-context
comparison, where the goal is to evaluate three directly measured quantities:
validation perplexity (PPL), optimizer-state memory (Mem), and per-step
optimizer runtime (T). These three metrics correspond to optimization quality
(O1), per-step compute (O2), and memory efficiency (O3). 

Table~\ref{tab:c4_llama_full_screening} reports the results at 60M, 130M,
350M, and 1B. Optimizers are grouped by methodological family (T1--T5) and
sorted by 1B PPL within each family. This layout allows us to compare both
individual optimizers and family-level tendencies without turning the table
into a flat leaderboard.

\begin{table*}[!t]
\centering
\caption{\textbf{Stage-1 screening on C4 (LLaMA, seq.\ 256).} C4 validation PPL,
optimizer-state memory (Mem), and per-step runtime (T) at four scales; lower is better.
Grouped by family, sorted by 1B PPL.}
\label{tab:c4_llama_full_screening}
\scriptsize
\setlength{\tabcolsep}{0.8mm}
\resizebox{1.\linewidth}{!}{
\begin{tabular}{ll|ccc|ccc|ccc|ccc}
\toprule
\multirow{3}{*}{\textbf{Optimizer}} & \multirow{3}{*}{\textbf{Venue}}
& \multicolumn{3}{c|}{\textbf{60M}} & \multicolumn{3}{c|}{\textbf{130M}}& \multicolumn{3}{c|}{\textbf{350M}} & \multicolumn{3}{c}{\textbf{1B}} \\
\cmidrule(lr){3-5}\cmidrule(lr){6-8}\cmidrule(lr){9-11}\cmidrule(lr){12-14}
 & & \makecell{\textbf{PPL}\\$\downarrow$} & \makecell{\textbf{Mem}\\GB$\downarrow$} & \makecell{\textbf{T}\\ms$\downarrow$} & \makecell{\textbf{PPL}\\$\downarrow$} & \makecell{\textbf{Mem}\\GB$\downarrow$} & \makecell{\textbf{T}\\ms$\downarrow$} & \makecell{\textbf{PPL}\\$\downarrow$} & \makecell{\textbf{Mem}\\GB$\downarrow$} & \makecell{\textbf{T}\\ms$\downarrow$} & \makecell{\textbf{PPL}\\$\downarrow$} & \makecell{\textbf{Mem}\\GB$\downarrow$} & \makecell{\textbf{T}\\ms$\downarrow$} \\
\midrule
\rowcolor{famT1!18}\multicolumn{14}{c}{\textbf{\textcolor{famT1!62!black}{\textit{T1: Element-wise adaptive moment and scalar control}}}} \\
Adan & TPAMI'24 & 30.25 & 0.433 & 2.32 & 22.84 & 1.000 & 4.72 & 17.29 & 2.742 & 12.06 & 14.35 & 9.977 & 39.67 \\
RAdam & ICLR'20 & 30.12 & 0.217 & 1.53 & 23.22 & 0.500 & 3.07 & 17.34 & 1.371 & 7.64 & 14.47 & 4.989 & 23.79 \\
\rowcolor{basebg}
AdamW & ICLR'19 & 30.08 & 0.217 & \cellcolor{bestbg}\textbf{1.14} & 23.18 & 0.500 & \cellcolor{bestbg}\textbf{2.31} & 17.78 & 1.371 & 5.97 & 14.48 & 4.989 & 18.62 \\
NAdam & ICLR'16 & 33.72 & 0.217 & 3.45 & 24.51 & 0.500 & 4.93 & 17.90 & 1.371 & 9.96 & 14.67 & 4.989 & 20.91 \\
MARS-AdamW & ICML'25 & 30.01 & 0.325 & 7.62 & 22.86 & 0.750 & 11.05 & 16.95 & 2.057 & 22.12 & 14.90 & 7.483 & 34.70 \\
Prodigy & ICML'24 & 33.44 & 0.433 & 8.36 & 24.13 & 1.000 & 12.29 & 18.27 & 2.742 & 24.30 & 15.61 & 9.977 & 36.78 \\
AdaBelief & NeurIPS'20 & 30.08 & 0.433 & 5.76 & 23.45 & 1.000 & 8.55 & 17.61 & 2.742 & 19.10 & 16.79 & 9.977 & 55.48 \\
\midrule
\rowcolor{famT2!18}\multicolumn{14}{c}{\textbf{\textcolor{famT2!62!black}{\textit{T2: Matrix-level structural methods}}}} \\
MARS-Shampoo & ICML'25 & 30.03 & 0.325 & 26.27 & 22.56 & 0.750 & 37.94 & 16.82 & 2.057 & 78.71 & \cellcolor{bestbg}13.72 & 7.483 & 513.7 \\
Muon & arXiv'25 & \cellcolor{bestbg}\textbf{28.26} & 0.109 & 21.01 & \cellcolor{bestbg}\textbf{21.81} & 0.250 & 30.48 & \cellcolor{bestbg}16.60 & 0.686 & 61.66 & \cellcolor{bestbg}13.72 & 2.495 & 379.0 \\
RMNP & ICML'26 & 29.88 & 0.109 & 3.26 & \cellcolor{bestbg}22.54 & 0.250 & 4.63 & 16.85 & 0.686 & 9.32 & 13.87 & 2.495 & 16.94 \\
SOAP & ICLR'25 & \cellcolor{bestbg}29.47 & \cellcolor{worstbg}0.731 & \cellcolor{worstbg}50.58 & 22.67 & \cellcolor{worstbg}2.214 & \cellcolor{worstbg}110.4 & 17.14 & \cellcolor{worstbg}7.465 & \cellcolor{worstbg}302.5 & 14.04 & \cellcolor{worstbg}29.299 & \cellcolor{worstbg}1371.5 \\
GaLore & ICML'24 & 34.56 & 0.062 & 4.21 & 25.32 & 0.199 & 5.88 & 19.18 & 0.426 & 11.85 & 14.29 & 0.790 & 15.29 \\
Shampoo & ICML'18 & 30.22 & 0.217 & 22.36 & 22.56 & 0.500 & 33.27 & 17.03 & 1.371 & 66.05 & 14.29 & 4.989 & 389.4 \\
\midrule
\rowcolor{famT3!18}\multicolumn{14}{c}{\textbf{\textcolor{famT3!62!black}{\textit{T3: Discretization and directional quantization}}}} \\
MARS-Lion & ICML'25 & 32.41 & 0.325 & 5.72 & 25.68 & 0.750 & 8.49 & 18.78 & 2.057 & 17.11 & 15.73 & 7.483 & 24.77 \\
Lion & NeurIPS'23 & 35.94 & 0.109 & 2.07 & 25.56 & 0.250 & 3.01 & \cellcolor{worstbg}19.30 & 0.686 & \cellcolor{bestbg}\textbf{5.80} & \cellcolor{worstbg}17.02 & 2.494 & \cellcolor{bestbg}\textbf{12.48} \\
\midrule
\rowcolor{famT4!18}\multicolumn{14}{c}{\textbf{\textcolor{famT4!62!black}{\textit{T4: State compression and structural aggregation}}}} \\
APOLLO & MLSys'25 & 30.86 & 0.062 & 8.62 & 22.74 & 0.149 & 12.65 & \cellcolor{bestbg}\textbf{16.43} & 0.426 & 26.21 & \cellcolor{bestbg}\textbf{13.53} & 0.790 & 28.65 \\
Conda & arXiv'25 & \cellcolor{bestbg}28.65 & 0.245 & 4.88 & \cellcolor{bestbg}21.91 & 0.595 & 7.11 & \cellcolor{bestbg}16.45 & 1.703 & 13.90 & 14.25 & 6.317 & 62.33 \\
8-bit Adam & ICLR'22 & 30.46 & 0.110 & 4.11 & 23.30 & 0.254 & 7.27 & 17.67 & 0.697 & 16.89 & 14.53 & 2.534 & 42.38 \\
CAME & ACL'23 & 31.40 & 0.218 & 14.99 & 23.79 & 0.502 & 21.76 & 17.60 & 1.376 & 44.89 & 14.53 & 4.997 & 87.46 \\
AdaFactor & ICML'18 & 30.00 & \cellcolor{bestbg}\textbf{0.001} & 9.90 & 22.94 & \cellcolor{bestbg}\textbf{0.002} & 14.63 & 17.85 & \cellcolor{bestbg}\textbf{0.003} & 29.70 & 14.92 & \cellcolor{bestbg}\textbf{0.004} & 56.46 \\
Adam-mini & ICLR'25 & 30.50 & 0.109 & 5.68 & 23.62 & 0.251 & 8.31 & 18.12 & 0.686 & 16.68 & 15.51 & 2.495 & 20.81 \\
\midrule
\rowcolor{famT5!18}\multicolumn{14}{c}{\textbf{\textcolor{famT5!62!black}{\textit{T5: Curvature-aware and geometric regularization}}}} \\
AdamP & ICLR'21 & 30.21 & 0.217 & 12.82 & 23.07 & 0.500 & 19.13 & 17.39 & 1.371 & 39.98 & 14.57 & 4.989 & 64.69 \\
LAMB & ICLR'20 & 30.03 & 0.217 & 9.14 & 23.40 & 0.500 & 13.17 & 17.25 & 1.371 & 26.62 & 16.09 & 4.989 & 44.18 \\
Sophia & ICLR'24 & \cellcolor{worstbg}36.27 & 0.217 & 3.92 & \cellcolor{worstbg}25.76 & 0.500 & 5.66 & 18.86 & 1.371 & 11.06 & 16.45 & 4.989 & 20.05 \\
\bottomrule
\end{tabular}}
\end{table*}

\paragraph{Quality.}
At the 1B scale, the best PPL is achieved by APOLLO (13.53), followed by a
group of matrix-structured methods, including MARS-Shampoo (13.72), Muon
(13.72), and RMNP (13.87). This shows that the strongest short-context
optimization quality is distributed across multiple families, because state-compressed
optimization and matrix-structured optimization both produce highly competitive
points. At smaller scales, however, the ranking is less stable. Muon leads at
60M and 130M, whereas APOLLO becomes strongest only at 350M and 1B. Therefore,
Stage~1 already suggests that optimizer quality is scale-dependent and should be
evaluated across multiple scales.

\paragraph{Runtime.}
The runtime column shows a different pattern. Lion is the fastest method at
350M and 1B, while AdamW remains one of the cheapest practical baselines.
Several quality-oriented matrix methods are much more expensive: SOAP, Muon,
MARS-Shampoo, and Shampoo\footnote{Our Shampoo implementation follows Meta's
\href{https://github.com/facebookresearch/optimizers}{\texttt{optimizers}}
library, incorporating the fix discussed in
\href{https://github.com/facebookresearch/optimizers/issues/265\#issuecomment-4668270192}{this issue}.}
incur substantial per-step overhead due to
matrix-level transformations or preconditioning. RMNP is the main exception
among matrix-structured optimizers, achieving strong PPL while keeping
runtime close to lightweight methods. Thus, Stage~1 reveals a clear separation
between heavy matrix-structured optimizers and lightweight approximations.

\paragraph{Optimizer-state memory.}
The lowest-memory region is mostly occupied by state-compressed or
subspace-based methods. AdaFactor has the
lowest optimizer-state memory at all scales, while APOLLO and GaLore also use
far less memory than AdamW at the 1B scale. However, the lowest memory does not
necessarily imply the best quality. AdaFactor is extremely memory-efficient
but only moderate in PPL. APOLLO is the most attractive Stage~1 memory point
because it combines low memory with the best 1B PPL, but whether this advantage
transfers beyond short-context C4 training must be tested separately.

\begin{takeaway}
\noindent\textbf{Stage~1 takeaway.}
\emph{Short-context screening reveals different objective frontiers across
T-families, and a universal winner does not exist.} T4 shows the strongest advantage
when memory efficiency is prioritized, T2 contributes the main high-quality
and structured-update candidates, and T1 remains the most reliable balanced
reference. Even in the same C4 short-context setting, different families
therefore occupy different favorable regions in quality, runtime, and memory.
Optimizer choice is consequently objective-dependent.
\end{takeaway}

\subsubsection{Stage~1 Pareto Analysis over PPL, Runtime, and Memory}
\label{sec:benchmark-pareto}

The Stage~1 results are inherently multi-objective. A method with lower PPL
may become unattractive if it requires substantially higher runtime or
optimizer-state memory, whereas a highly efficient method may be unusable if
its quality degradation is too large. We therefore analyze the short-context
C4 results through Pareto frontiers over the three directly measured
objectives, namely optimization quality (O1), per-step compute (O2), and
optimizer-state memory (O3).

\paragraph{PPL--runtime and PPL--memory frontiers.}
We plot two Pareto views at the 1B scale: PPL versus per-step optimizer
runtime, and PPL versus optimizer-state memory. An optimizer is Pareto-optimal
if no other optimizer achieves both lower PPL and lower cost along the
corresponding efficiency axis.

\begin{figure*}[t]
  \centering
  \includegraphics[width=1.0\textwidth]{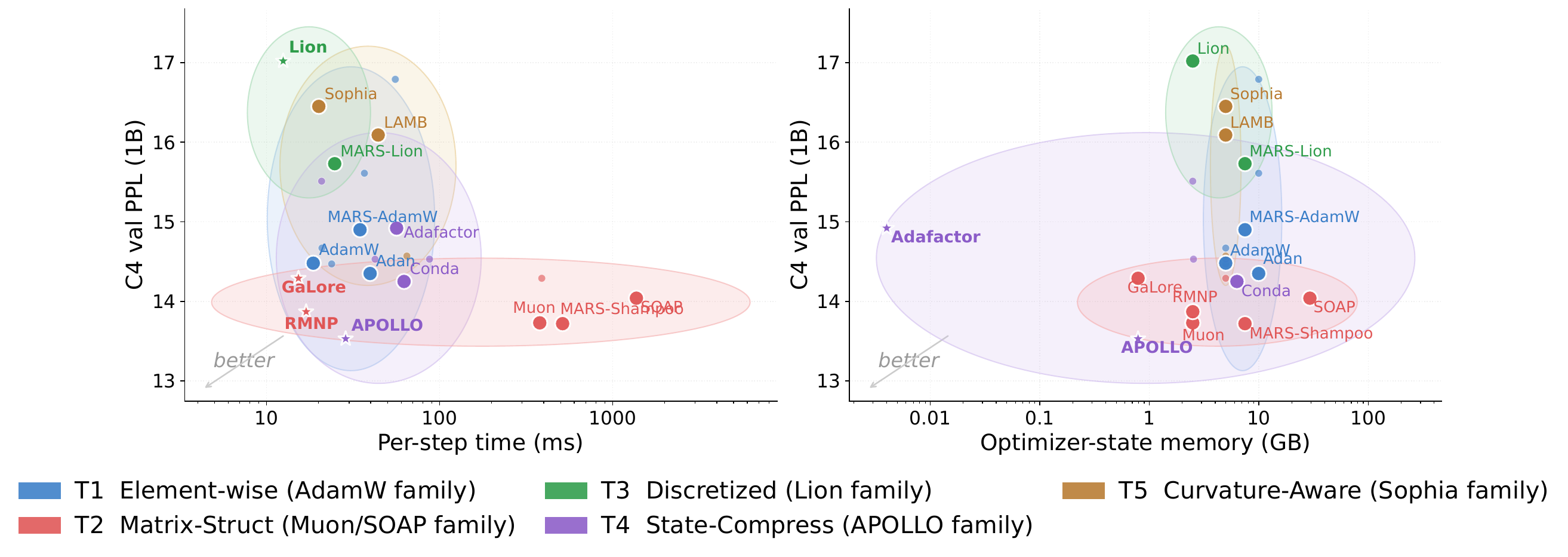}
  \caption{\textbf{Stage-1 Pareto frontiers (1B).} PPL vs.\ per-step runtime (left) and
vs.\ optimizer-state memory (right); lower-left is better. Colors denote families,
stars mark frontier members.}
  \label{fig:pareto_1b}
\end{figure*}

Figure~\ref{fig:pareto_1b} shows that the runtime frontier is not occupied by
the heaviest quality-oriented methods. Lion is the fastest point, but its PPL
is too high to make it a strong overall choice. GaLore is also efficient in
runtime, but its quality remains behind the best optimizers. RMNP occupies the
most important middle region, where it is much faster than heavy matrix-structured
methods while still achieving strong PPL. APOLLO appears as the best
short-context quality point and remains non-dominated in runtime.

This pattern separates matrix-structured methods into two regimes. SOAP, Muon,
MARS-Shampoo, and Shampoo improve quality through matrix-level operations or
preconditioning, but they pay substantial runtime overhead. RMNP, in contrast,
retains a large part of the matrix-structured quality benefit while staying
close to the efficient frontier. It therefore represents the clearest
quality--runtime compromise among matrix-structured methods in Stage~1.

The memory frontier is more concentrated. AdaFactor achieves the smallest
optimizer-state memory, but its PPL is only moderate. APOLLO is the strongest
memory-frontier point in Stage~1, achieving the best 1B PPL while using much
less optimizer-state memory than AdamW. This result demonstrates the potential
of state-compressed optimization under short-context training. However, it
should be read as a local Stage~1 optimum.

\begin{figure*}[t]
  \centering
  \includegraphics[width=1.0\textwidth]{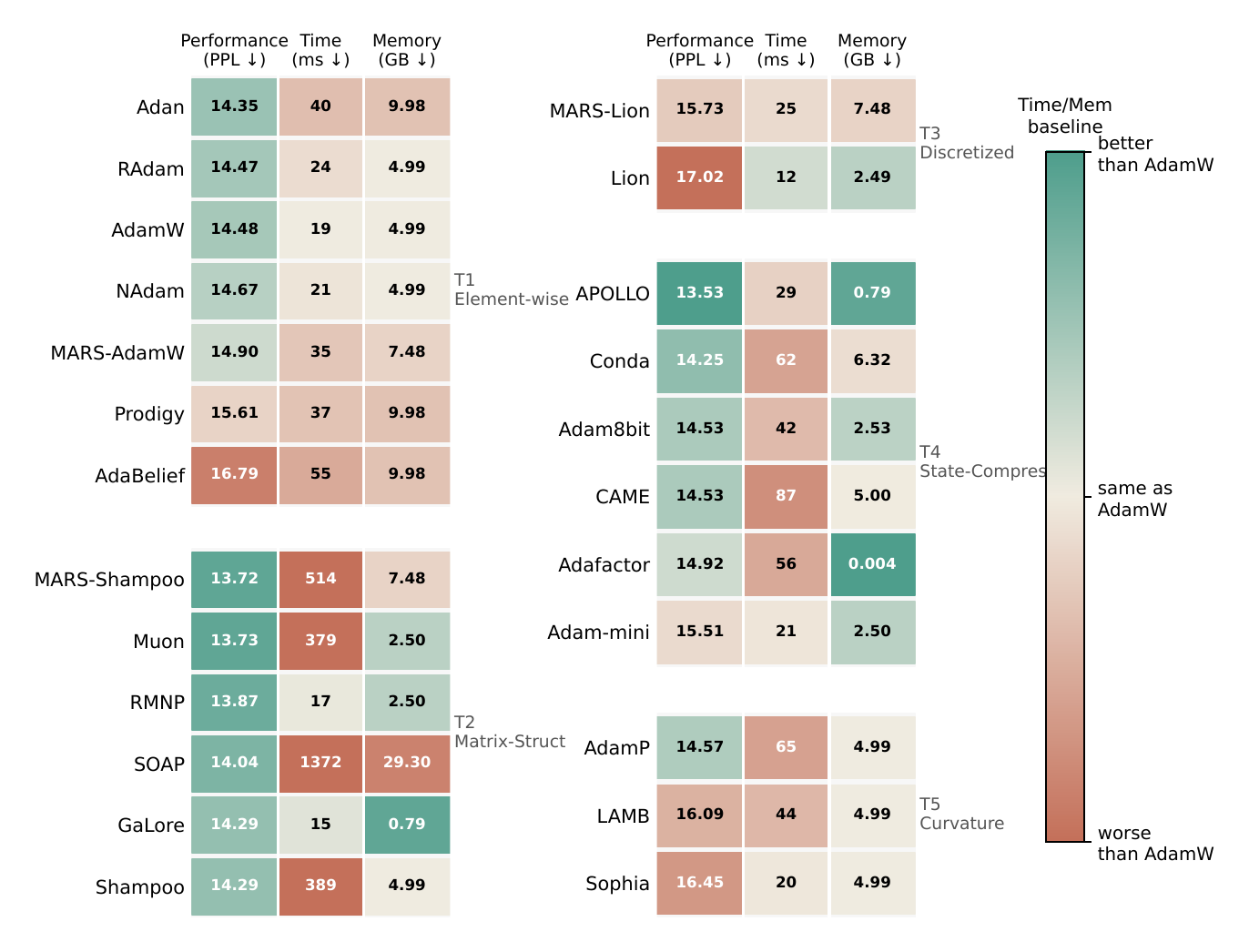}
  \caption{\textbf{Optimizer-level heatmap of the three Stage-1 metrics (1B).} Rows are grouped
by family, and columns are C4 PPL, runtime, and optimizer-state memory. Green is favorable,
red unfavorable.}
  \label{fig:optimizer_heatmap}
\end{figure*}

\paragraph{Optimizer-level heatmap.}
To complement the Pareto plots, Fig.~\ref{fig:optimizer_heatmap} summarizes
the three Stage~1 metrics at the 1B scale for all 24 optimizers. This figure
keeps the analysis at the optimizer level and shows that even within the same
family, methods can occupy very different quality--efficiency regimes.

The heatmap highlights substantial within-family variation. RMNP and SOAP both
belong to the matrix-structured family, but RMNP is a balanced method whereas
SOAP is a quality-oriented but expensive outlier. APOLLO and AdaFactor both
belong to state-compressed optimization, but APOLLO achieves much better PPL
whereas AdaFactor is more extreme in memory reduction. Similarly, AdamW and
MARS-AdamW belong to the same element-wise family, yet they differ in the
quality--runtime balance. Therefore, family labels are useful for organizing
the benchmark, but optimizer selection still requires metric-level analysis.

\begin{takeaway}
\noindent\textbf{Stage~1 Pareto takeaway.}
The Pareto frontier should be read as each T-family's best attainable
trade-off. T4 is most competitive when
optimizer-state memory is the main constraint, lightweight T2 variants are
more attractive when runtime matters, and T1 provides the stable reference
region when neither cost dimension dominates. Heavier T2 methods can improve
quality, but their benefit must be evaluated together with the runtime and
memory cost needed to obtain it.
\end{takeaway}

\subsubsection{Stage~2: Generalization across Data, Context Length, and Architectures}
\label{sec:benchmark-stage2}

We migrate the stronger optimizers from Stage~1 to a substantially more
demanding regime: the higher-quality FineWeb-Edu corpus, a 32k context length,
two model scales (340M and 1B), and four architectures---Transformer++
(standard attention) together with three linear-attention variants (Gated
DeltaNet, DeltaNet, GLA). This setting changes three factors at once relative
to Stage~1: the data distribution, the sequence length, and the parameter
topology induced by the token-mixing architecture.

\begin{figure*}[!t]
\centering
\includegraphics[width=\textwidth]{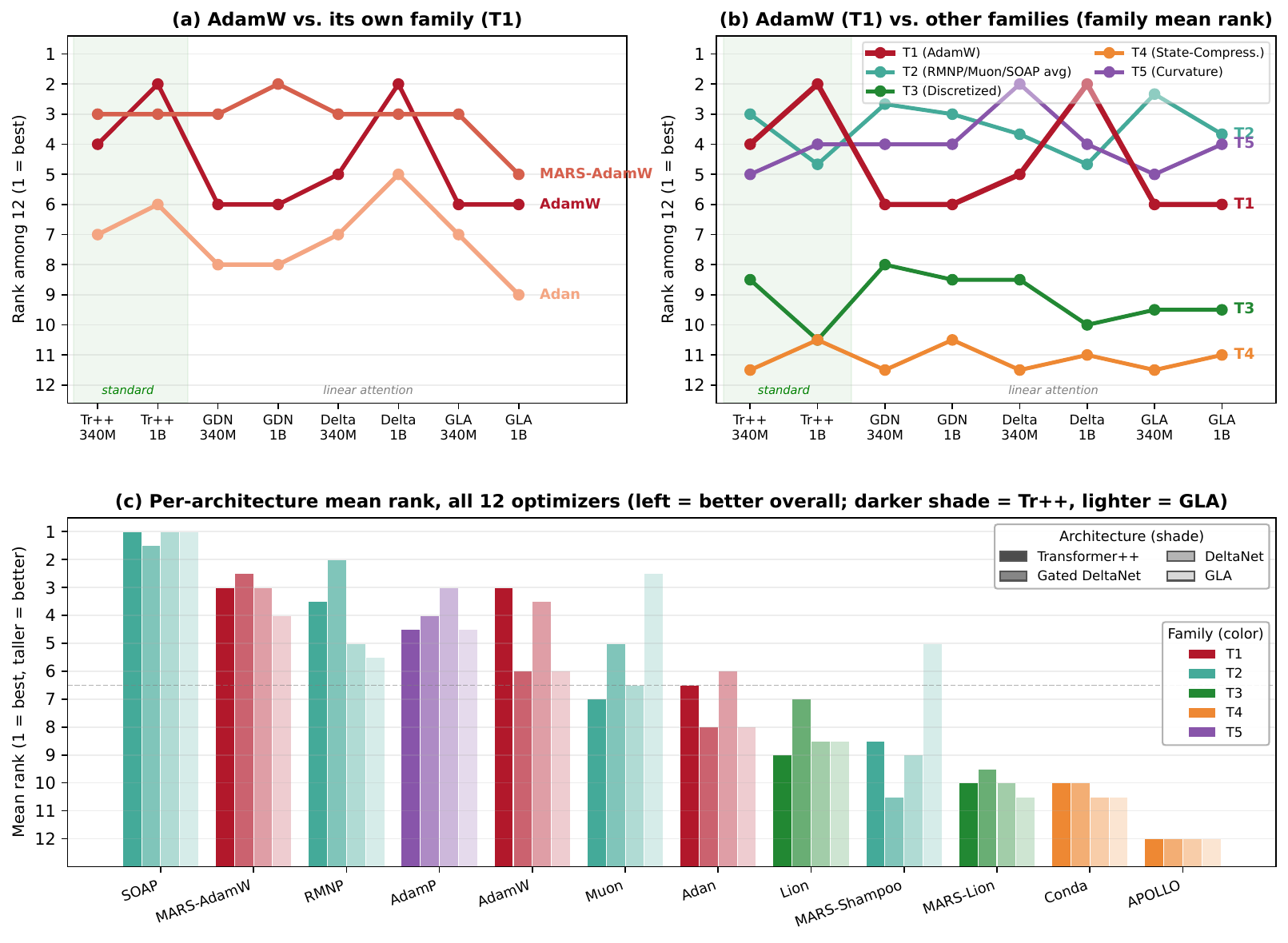}
\caption{\textbf{Cross-scenario rank stability (FineWeb-Edu, 32k).} Absolute ranks among
all twelve optimizers per scenario. (a) AdamW vs.\ its T1 family; (b) AdamW vs.\ other
families' mean rank; (c) per-architecture mean rank per optimizer (color = family,
hatch = architecture).}
\label{fig:rank_stability}
\end{figure*}

Because different architectures have different modeling capacity, absolute
perplexity is incomparable across them. Comparing raw PPL between, say,
Transformer++ and GLA would conflate the optimizer with the architecture. We
therefore analyze \emph{within-architecture ranking} and its stability across
architectures. For each architecture-scale pair, optimizers are ranked by
WikiText test PPL, and an optimizer that holds a similar position everywhere is
broadly generalizable, whereas one whose position swings sharply is
architecture-sensitive. This rank-based view is the direct empirical proxy for
the generalization objective (O6). To check whether the PPL-based picture is
consistent with downstream behavior, we additionally report the average
Commonsense Reasoning score (CS Avg.) over lm-eval-harness tasks for every
scenario. Table~\ref{tab:stage2_cross_arch} gives both PPL and CS Avg. across
the eight scenarios. Figure~\ref{fig:rank_stability} visualizes the resulting
within-architecture PPL ranks.

\begin{table*}[!t]
\centering
\caption{\textbf{Stage-2 cross-architecture generalization (FineWeb-Edu, 32k).} Left block: WikiText test PPL (lower is better). Right block: downstream commonsense-reasoning accuracy (CS Avg, \%, average over ten lm-eval-harness tasks; higher is better). Both at 340M and 1B across four architectures. Per-scenario best in green. Absolute PPL is not comparable across architectures; rows are ordered by PPL-based cross-scenario stability. Gray row = AdamW. Note that the best PPL is almost always SOAP, whereas the best CS is spread across SOAP, MARS-AdamW, RMNP, and Muon.}
\label{tab:stage2_cross_arch}
\scriptsize
\setlength{\tabcolsep}{0.6mm}
\renewcommand{\arraystretch}{1.05}
\resizebox{1.\linewidth}{!}{
\begin{tabular}{l|cccccccc|cccccccc}
\toprule
\multirow{4}{*}{\textbf{Optimizer}}
 & \multicolumn{8}{c|}{\textbf{WikiText PPL} ($\downarrow$)}
 & \multicolumn{8}{c}{\textbf{Commonsense Reasoning Avg} ($\uparrow$)} \\
\cmidrule(lr){2-9}\cmidrule(lr){10-17}
 & \multicolumn{2}{c}{Tr++} & \multicolumn{2}{c}{GDN} & \multicolumn{2}{c}{Delta} & \multicolumn{2}{c|}{GLA}
 & \multicolumn{2}{c}{Tr++} & \multicolumn{2}{c}{GDN} & \multicolumn{2}{c}{Delta} & \multicolumn{2}{c}{GLA} \\
\cmidrule(lr){2-3}\cmidrule(lr){4-5}\cmidrule(lr){6-7}\cmidrule(lr){8-9}
\cmidrule(lr){10-11}\cmidrule(lr){12-13}\cmidrule(lr){14-15}\cmidrule(lr){16-17}
 & 340M & 1B & 340M & 1B & 340M & 1B & 340M & 1B
 & 340M & 1B & 340M & 1B & 340M & 1B & 340M & 1B \\
\midrule
\rowcolor{famT1!18}\multicolumn{17}{c}{\textbf{\textcolor{famT1!62!black}{\textit{T1: Element-wise adaptive moment and scalar control}}}} \\
MARS-AdamW
 & 24.57 & 18.94 & 24.17 & 20.04 & 26.79 & 20.67 & 28.28 & 21.89
 & 52.50 & 57.46 & \cellcolor{bestbg}\textbf{54.91} & \cellcolor{bestbg}\textbf{58.18} & 51.69 & 56.80 & 51.24 & 55.71 \\
\rowcolor{basebg}
AdamW
 & 24.62 & 18.90 & 24.47 & 20.33 & 27.16 & 20.66 & 28.67 & 22.06
 & 52.28 & 56.55 & 53.67 & 57.01 & 51.74 & 55.56 & 51.06 & 56.69 \\
Adan
 & 25.55 & 19.41 & 24.78 & 20.55 & 27.28 & 20.88 & 29.00 & 22.51
 & 52.48 & 57.21 & 52.83 & 57.93 & 51.78 & 56.50 & 51.01 & 55.07 \\
\midrule
\rowcolor{famT2!18}\multicolumn{17}{c}{\textbf{\textcolor{famT2!62!black}{\textit{T2: Matrix-level structural methods}}}} \\
SOAP
 & \cellcolor{bestbg}\textbf{23.90} & \cellcolor{bestbg}\textbf{18.72} & 23.85 & \cellcolor{bestbg}\textbf{19.86} & \cellcolor{bestbg}\textbf{26.02} & \cellcolor{bestbg}\textbf{20.38} & \cellcolor{bestbg}\textbf{27.04} & \cellcolor{bestbg}\textbf{20.62}
 & \cellcolor{bestbg}\textbf{53.75} & \cellcolor{bestbg}\textbf{57.71} & 54.77 & 57.22 & 52.60 & 56.49 & 52.21 & \cellcolor{bestbg}\textbf{57.57} \\
RMNP
 & 24.37 & 19.40 & \cellcolor{bestbg}\textbf{23.65} & 20.26 & 26.80 & 21.06 & 28.60 & 22.23
 & 53.35 & 57.12 & 54.45 & 57.30 & \cellcolor{bestbg}\textbf{53.25} & \cellcolor{bestbg}\textbf{57.32} & 50.72 & 55.79 \\
Muon
 & 25.05 & 19.86 & 24.34 & 20.32 & 27.18 & 21.18 & 27.47 & 21.54
 & 53.25 & 56.36 & 54.45 & 57.20 & 52.00 & 56.65 & \cellcolor{bestbg}\textbf{52.50} & 56.85 \\
MARS-Shampoo
 & 26.43 & 19.74 & 25.99 & 24.87 & 28.26 & 21.25 & 29.20 & 21.53
 & 51.96 & 57.30 & 53.37 & 57.58 & 51.43 & 56.74 & 52.01 & 57.33 \\
\midrule
\rowcolor{famT3!18}\multicolumn{17}{c}{\textbf{\textcolor{famT3!62!black}{\textit{T3: Discretization and directional quantization}}}} \\
Lion
 & 26.02 & 20.26 & 24.76 & 20.38 & 28.20 & 21.44 & 29.47 & 22.40
 & 51.07 & 55.22 & 53.24 & 55.74 & 49.96 & 54.22 & 50.14 & 53.96 \\
MARS-Lion
 & 26.20 & 21.17 & 25.24 & 22.20 & 28.25 & 22.72 & 29.67 & 23.79
 & 51.61 & 54.51 & 52.96 & 55.50 & 50.94 & 53.65 & 50.69 & 53.91 \\
\midrule
\rowcolor{famT4!18}\multicolumn{17}{c}{\textbf{\textcolor{famT4!62!black}{\textit{T4: State compression and structural aggregation}}}} \\
Conda
 & 28.30 & 19.86 & 26.11 & 21.07 & 29.09 & 21.75 & 37.38 & 22.89
 & 51.61 & 57.24 & 53.45 & 57.18 & 51.46 & 56.10 & 48.28 & 54.95 \\
APOLLO
 & 34.08 & 25.29 & 30.36 & 29.29 & 34.73 & 25.58 & 37.75 & 27.78
 & 48.19 & 53.61 & 50.92 & 53.73 & 49.04 & 53.88 & 48.38 & 52.33 \\
\midrule
\rowcolor{famT5!18}\multicolumn{17}{c}{\textbf{\textcolor{famT5!62!black}{\textit{T5: Curvature-aware and geometric regularization}}}} \\
AdamP
 & 24.68 & 19.04 & 24.32 & 20.29 & 26.77 & 20.68 & 28.66 & 21.86
 & 51.69 & 56.82 & 53.82 & 57.07 & 51.53 & 56.73 & 51.14 & 55.31 \\
\bottomrule
\end{tabular}}
\end{table*}

\paragraph{Rank-stability visualization.}
Figure~\ref{fig:rank_stability} complements Table~\ref{tab:stage2_cross_arch}
by visualizing how optimizer ranks change across the eight architecture-scale
scenarios. Panel~(a) compares AdamW with other element-wise adaptive methods,
showing whether improvements within the AdamW family are stable. Panel~(b)
aggregates optimizers by family to reveal the broader ranking bands across
architectures. Panel~(c) returns to the optimizer level and summarizes each
method's per-architecture mean rank, making architecture sensitivity visible
within a single optimizer. Together, the table and figure separate three
phenomena: consistently transferable methods, methods that are strong only on
specific architectures, and methods that systematically fail to transfer.

\paragraph{SOAP is the most stable cross-scenario optimizer.}
SOAP (T2) holds the top PPL position in seven of the eight scenarios and never
leaves the top two, giving it an almost flat top line in
Figure~\ref{fig:rank_stability}(b). Its advantage is therefore highly stable
across both architecture and scale, which suggests that, in this benchmark,
SOAP's Kronecker or Fisher basis preconditioning transfers across both
standard-attention and linear-attention architectures. We treat SOAP as the
strongest quality-oriented generalization baseline on perplexity, while noting
that its lead becomes less exclusive under CS Avg. and that its runtime and
memory costs remain substantial in Stage~1.

\paragraph{MARS-AdamW is the most stable AdamW-style enhancement.}
Within the element-wise family, MARS-AdamW is consistently ahead of plain AdamW
and far ahead of Adan, as panel~(a) shows. It also attains the best CS Avg. on
both Gated DeltaNet columns of Table~\ref{tab:stage2_cross_arch}. This pattern
is consistent with the framework interpretation that the STORM-style
variance-reduced estimator (Axis~II) can improve the gradient signal while
largely preserving the AdamW update geometry. This improvement has a cost.
MARS-AdamW introduces additional state and computation, so its runtime and
optimizer-state memory exceed AdamW's in Stage~1. We therefore read it as a
stable quality-oriented AdamW extension.

\paragraph{Muon shows architecture-dependent behavior.}
Muon sits in the middle of the field on standard attention but moves toward the
top on GLA, where it also attains the best CS Avg. at 340M. Panel~(c) makes the
pattern explicit. Among its four architecture-level bars, the GLA bar is the
best while the Transformer++ and DeltaNet bars are worse. This trajectory
suggests that Muon's spectral/polar direction geometry (Axis~III) interacts
with the parameter topology of the target architecture. It also motivates the
cross-architecture ablation below, where we test whether Muon's component-wise
gains remain stackable outside the standard Transformer setting.

\paragraph{State-compressed methods generalize worst.}
The two T4 members sit at the bottom of the table. APOLLO is last or near-last
in every one of the eight scenarios under both PPL and CS Avg., despite having
been the single best optimizer in the Stage-1 short-context screening. This
reversal---from short-context champion to long-context worst---is the sharpest
generalization failure in the benchmark, and we isolate its cause next.

\paragraph{CS Avg. corroborates the PPL ranking, with informative exceptions.}
The CS Avg. results in Table~\ref{tab:stage2_cross_arch} largely track the PPL
ordering, but the two views are not identical, and the mismatches are
informative. Agreement is strongest at the extremes. APOLLO is last or
near-last under CS Avg. in every scenario, confirming that its long-context
collapse is not a PPL-only artifact, while SOAP remains in the top tier. The
divergences are equally telling. SOAP's near-exclusive lead on PPL does not
fully carry over to CS Avg., where MARS-AdamW tops both Gated DeltaNet columns, RMNP
tops both DeltaNet columns, and on Gated-DeltaNet-340M the best PPL and the
best CS Avg. belong to different optimizers. In short, even when downstream
average accuracy is considered alongside perplexity, the best optimizer remains
scenario-dependent.

\begin{takeaway}
\noindent\textbf{Stage~2 takeaway.}
\emph{Long-context generalization is a stress test for T-families, and whether
an optimizer generalizes is ultimately determined by the specific optimizer
mechanism.} T1 and T2 form the strongest
quality bands in the long-context setting, while T4 exposes the clearest
transfer boundary from short to long context. At the same time, variation
within the same family remains substantial. One representative may transfer
well, whereas another can be sensitive to architecture. Generalization should
therefore be understood as an interaction between optimizer mechanism and
target model topology.
\end{takeaway}

\subsubsection{Scenario Sensitivity of Memory-Efficient Optimizers}
\label{sec:benchmark-mem-eff-sensitivity}

The Stage-1 screening showed that an aggressive state-compression method,
APOLLO, can be the single best optimizer under short-context training, yet
Stage~2 showed it generalizing worst under long context. Short-context
performance alone therefore overestimates the practical value of such methods.
To isolate this effect as much as possible, we conduct a controlled
sequence-length ablation that changes exactly one variable: we fix the
dataset (C4) and the LLaMA architecture, and vary only the context length
from 256 to 32k (Table~\ref{tab:seqlen_numbers}). The short-context points
come from the Stage-1 LLaMA setting at sequence length 256, and the
long-context points from the same LLaMA architecture on C4 extended to a
32k context length. We compare AdamW (the element-wise baseline),
Lion (a sign-based efficient method, T3), and APOLLO (state compression, T4).
The question is whether an optimizer's degradation from short to long context
tracks the AdamW baseline or grows disproportionately.

\paragraph{Finding and framework attribution.}
The result is unambiguous from Table~\ref{tab:seqlen_numbers}. At 256 tokens, APOLLO is the best optimizer in the table (13.53). At 32k, it collapses to 35.40, a degradation of $+21.87$ PPL, roughly three times the AdamW baseline increase of $+7.39$. It is the only clear outlier. Lion, by contrast, degrades by $+6.29$, slightly \emph{less} than AdamW.

The framework explains the contrast through Axis~II. APOLLO compresses its optimizer state by random projection into a low-dimensional subspace.

\begin{wraptable}{r}{0.35\linewidth}
\centering
\vspace{-2.0em}
\small
\setlength{\tabcolsep}{6pt}
\renewcommand{\arraystretch}{1.1}
\caption{\textbf{Sequence-length effect.}
PPL at 256 vs.\ 32k context, lower is better.}
\label{tab:seqlen_numbers}
\resizebox{1.\linewidth}{!}{
\begin{tabular}{lccc}
\toprule
\textbf{Optimizer} & \textbf{256} & \textbf{32k} & \textbf{$\Delta$} \\
\midrule
\rowcolor{worstbg}
APOLLO & 13.53 & 35.40 & $+21.87$ \\
Muon   & 13.72 & 22.54 & $+8.81$  \\
SOAP   & 14.04 & 21.62 & $+7.58$  \\
\rowcolor{basebg}
AdamW  & 14.48 & 21.87 & $+7.39$  \\
Lion   & 17.02 & 23.31 & $+6.29$  \\
\bottomrule
\end{tabular}
}
\vspace{-1.0em}
\end{wraptable}

Under long context, the gradient's effective rank rises and its structure becomes
more complex, so a fixed low-dimensional projection discards proportionally
more information, and the compression that was nearly lossless at short context
becomes lossy at long context. Lion (T3) does not compress state and only
discretizes the direction (Axis~III), so its degradation tracks AdamW. This turns an isolated
empirical surprise into a mechanistic statement, namely that the benefit of aggressive
state compression (Axis~II) is bounded by the rank and structure of the
gradient statistics, and that boundary becomes visible when the context length
grows.

\paragraph{A note on matched conditions.}
The short-context points are taken from the Stage-1 C4 configuration
(LLaMA, sequence length 256), and the long-context points from the same
LLaMA architecture on C4 extended to a 32k context length. The architecture
and dataset are therefore held fixed and context length is the only varied
factor. Where training budget
or token count differs between the two configurations, the comparison should
be read as a qualitative indication of the degradation \emph{trend}, and the
reported ranking of degradations (APOLLO $\gg$ Muon
$\approx$ SOAP $\approx$ AdamW $>$ Lion) is robust to these differences.

\begin{takeaway}
\noindent\textbf{Sequence-length takeaway.}
\emph{T4 is constrained by the structure and effective rank of long-context
gradients.} A T4 optimizer can look highly competitive under short-context
training, where compressed states may still preserve the dominant gradient
information. As the context length increases, however, the gradient statistics
become more complex and the same compression can become lossy. In contrast,
T3 methods reduce update complexity without directly compressing optimizer
state, making them less exposed to this particular sequence-length failure
mode. Short-context quality alone therefore cannot certify the long-context
reliability of T4.
\end{takeaway}

\subsubsection{Optimizer Robustness Across Vision Backbones}
Unlike the language-model experiments where context length is the primary
source of variation, the CIFAR100 benchmark fixes the dataset and training
protocol while varying the backbone from a CNN (ResNet50~\cite{he2016deep}) to a Vision
Transformer (DeiT-S~\cite{touvron2021training}) and a MetaFormer
architecture~\cite{yu2022metaformer} (CAFormer-S12~\cite{yu2023metaformer}). This
setting allows us to isolate whether optimizer improvements transfer across
architectural families or are tied to specific gradient structures. The image classification experiments on CIFAR100 follow the settings of BOCB~\cite{li2024unveiling}.

\paragraph{No optimizer dominates all vision backbones.}

Table~\ref{tab:cifar100_results} shows that the best optimizer changes across
architectures. AdaBelief achieves the highest accuracy of 80.53\% on ResNet50,
Muon is the strongest optimizer on DeiT-S, achieving 77.38\%, and Adan obtains the best
result of 84.89\% on CAFormer-S12. The identity of the top-performing optimizer
therefore depends on the backbone, suggesting that optimizer quality is closely
coupled to the specific architecture.

\paragraph{Matrix-structured optimizers favor Transformer architectures.}

The most notable trend is the performance of Muon. Compared with AdamW, Muon
improves DeiT-S by more than five percentage points (77.38\% \textit{vs.} 72.15\%) while
providing only modest gains on ResNet50. Within the four-axis framework, Muon
modifies both the update domain (Axis~I) and direction construction
(Axis~III), preserving matrix-level gradient structure and treating
parameters as correlated wholes. This design appears particularly well aligned with
the correlated gradients of attention layers, explaining its clear advantage on
Transformer-style backbones.

\paragraph{Adaptive moment methods remain the most robust family.}
Although methods such as Muon achieve the highest peak performance, AdamW,
Adan, and AdaBelief consistently remain competitive across all three
architectures. In contrast, some efficiency-oriented methods exhibit much
stronger backbone sensitivity. The most extreme example is MARS-Lion, whose
accuracy drops to 33.70 on DeiT-S despite achieving reasonable results on
ResNet50 and CAFormer-S12. This suggests that aggressive approximations in
state estimation or direction construction can interact poorly with the more
complex gradient geometry of Transformer architectures.

\begin{takeaway}
\noindent\textbf{Vision-backbone takeaway.}
Optimizer performance is strongly architecture-dependent. Matrix-structured
methods such as Muon achieve substantial gains on Transformer-style models,
while adaptive-moment methods remain the most robust across CNNs,
Transformers, and MetaFormers. The large performance variance observed for
methods such as MARS-Lion further indicates that single-backbone evaluations
can substantially overestimate optimizer generality.
\end{takeaway}
\begin{wraptable}{r}{0.5\linewidth}
\centering
\vspace{-1.4em}
\scriptsize
\caption{CIFAR100 Top-1 Accuracy (\%) across CNN, ViT and MetaFormer architectures.}
\label{tab:cifar100_results}
\setlength{\tabcolsep}{2.mm}
\resizebox{1.\linewidth}{!}{
\begin{tabular}{lccc}
\toprule
\textbf{Optimizer} &
\textbf{ResNet50} &
\textbf{DeiT-S} &
\textbf{CAFormer-S12} \\
\midrule
\rowcolor{famT1!18}\multicolumn{4}{c}{\textbf{\textcolor{famT1!62!black}{\textit{T1: Element-wise adaptive moment and scalar control}}}} \\
AdaGrad     & 73.30 & 67.24 & 38.09 \\
AdaDelta    & 75.07 & 65.44 & 82.08 \\
RMSProp     & 74.25 & 70.71 & 81.83 \\
Adam        & 74.55 & 71.04 & 82.18 \\
\rowcolor{basebg}
AdamW       & 75.56 & 72.15 & 83.60 \\
Adamax       & 75.21 & 73.31 & 82.50 \\
NAdam       & 74.82 & 72.75 & 82.83 \\
RAdam       & 75.19 & 72.41 & 82.35 \\
AdaBelief   & \cellcolor{bestbg}\textbf{80.53} & 70.66 & 83.56 \\
Adan        & 77.08 & 76.33 & \cellcolor{bestbg}\textbf{84.89} \\
AdaBound    & 78.11 & 68.59 & 82.38 \\
NovoGrad    & 79.36 & 73.13 & 82.98 \\
MARS-AdamW  & 74.19 & 71.57 & 80.48 \\
\midrule

\rowcolor{famT2!18}\multicolumn{4}{c}{\textbf{\textcolor{famT2!62!black}{\textit{T2: Matrix-level structural methods}}}} \\
RMNP     & 73.39 & 71.83 & 82.44 \\
Muon     & 75.25 & \cellcolor{bestbg}\textbf{77.38} & 84.43 \\
GaLore   & 73.53 & 70.88 & 82.19 \\
MOGA     & 63.20 & 62.48 & 79.00 \\
\midrule

\rowcolor{famT3!18}\multicolumn{4}{c}{\textbf{\textcolor{famT3!62!black}{\textit{T3: Discretization and directional quantization}}}} \\
MARS-Lion   & 71.53 & 33.70 & 77.02 \\
Lion      & 75.28 & 74.57 & 79.59 \\
\midrule

\rowcolor{famT4!18}\multicolumn{4}{c}{\textbf{\textcolor{famT4!62!black}{\textit{T4: State compression and structural aggregation}}}} \\
AdaFactor     & 75.41  & 74.02  & 82.36 \\
APOLLO  & 74.09 & 71.24 & 82.00 \\
CAME          & 66.62 & 71.05 & 81.83 \\
Conda         & 73.87 & 70.76 & 82.45 \\
\midrule

\rowcolor{famT5!18}\multicolumn{4}{c}{\textbf{\textcolor{famT5!62!black}{\textit{T5: Curvature-aware and geometric regularization}}}} \\
LAMB      & 77.19 & 75.39 & 83.74 \\
AdamP     & 78.17 & 71.55 & 83.40 \\
Sophia    & 75.19 & 71.47 & 82.96 \\
\bottomrule
\end{tabular}
}
\vspace{-2em}
\end{wraptable}

\subsubsection{Auxiliary Stability Analysis from Gradient-Norm Dynamics}
\label{sec:benchmark-o4-stability}

The preceding analyses cover optimization quality (O1), runtime (O2), memory
(O3), and cross-scenario transfer (O6). To evaluate training stability (O4),
we analyze the gradient-norm trajectories logged during the FineWeb-Edu
long-context runs. Gradient-norm spikes and large gradient-scale fluctuations
are increasingly used as direct indicators of training instability in modern
LLM pretraining, including gradient-norm statistics, spike diagnostics, and the
coefficient of variation (CV) of raw gradient norms
\cite{huanggradientstabilizer,bae2026affine,wang2025adagc,kumar2025zclip}.

For each run, we compute the coefficient of variation of the gradient norm,
\begin{equation}
\mathrm{GNormCV}=\frac{\mathrm{std}(\|g_t\|)}{\mathrm{mean}(\|g_t\|)} ,
\end{equation}
together with auxiliary diagnostics: clipping rate, spike rate, maximum
consecutive spike length, tail trend, and NaN/Inf count. We adopt GNormCV as
the primary O4 statistic for three reasons. First, it is scale-free and thus
comparable across optimizers whose absolute gradient norms differ by orders of
magnitude. Second, it captures \emph{relative} volatility while remaining decoupled from absolute
magnitude, which suits cross-architecture and cross-scale comparison. Third, it
exposes \emph{soft instability}, meaning that a run may complete without any NaN/Inf event
yet still exhibit highly volatile gradient dynamics. The remaining diagnostics
serve to localize the source and persistence of that instability.

\begin{figure*}[!t]
  \centering
  \includegraphics[width=\textwidth]{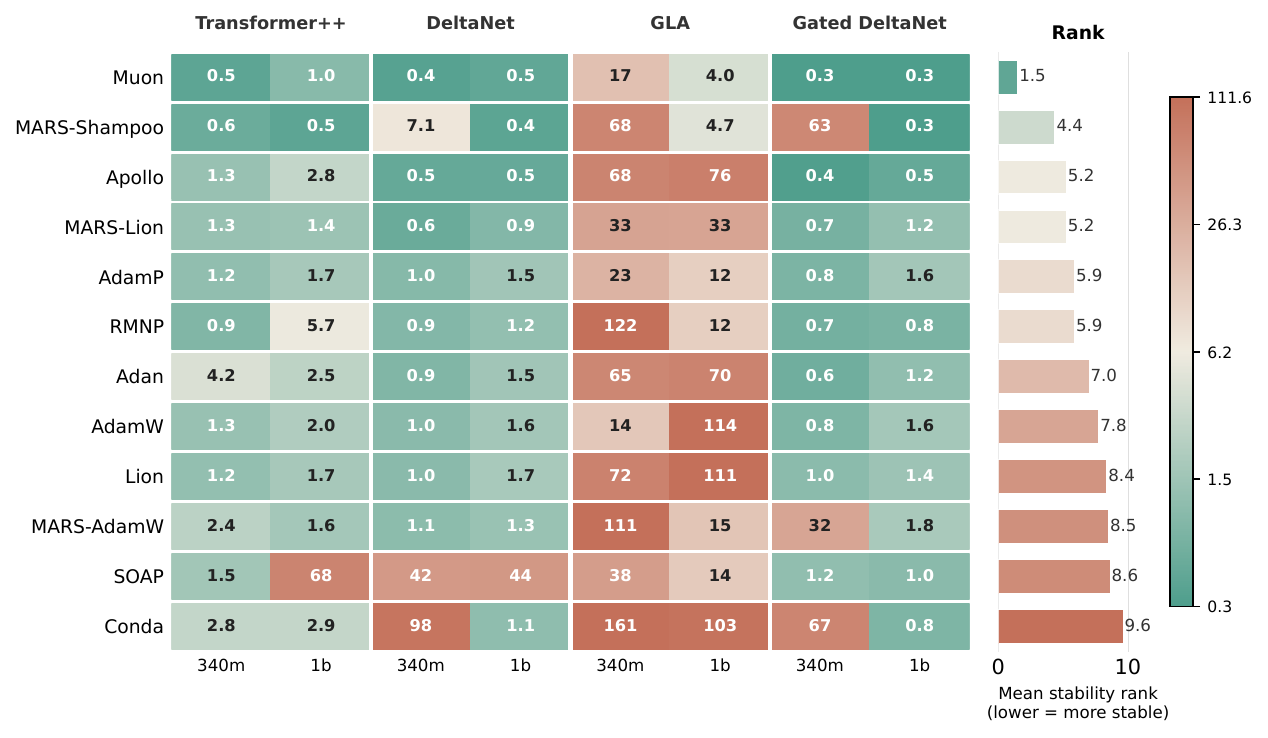}
  \caption{\textbf{Auxiliary O4 stability analysis from gradient-norm dynamics across architectures.}
  Each cell shows the coefficient of variation of the gradient norm (GNormCV)
  for one optimizer in one architecture-scale scenario of the FineWeb-Edu
  long-context benchmark. Lower GNormCV indicates smoother relative
  gradient-norm dynamics and thus better auxiliary stability. The bar chart on
  the right reports the mean stability rank across the eight scenarios (lower
  is better).}
  \label{fig:o4_gnorm_cross_arch}
\end{figure*}

Figure~\ref{fig:o4_gnorm_cross_arch} shows that O4 gives a different view from
final PPL. All statistics are computed on \emph{pre-clip} gradient norms, so
that GNormCV reflects each optimizer's own raw dynamics. Muon attains the best aggregate stability rank, placing first or
second in every one of the eight scenarios. This is consistent with its
spectral orthogonalization, which decouples the update scale from gradient
magnitude and so damps relative gradient-norm fluctuation. RMNP, AdamP, and
MARS-AdamW form a moderate band, and AdamW is a reliable reference on the
standard- and DeltaNet-style architectures---though, as shown below, no
optimizer is immune to architecture-induced volatility.

\paragraph{Completion status does not certify stability.}
Every summarized run finishes with zero NaN/Inf events, yet GNormCV spans more
than two orders of magnitude, from $\sim$0.32 to $>$160. The conventional
``did it diverge?'' criterion would rate all of these runs equally stable, whereas
GNormCV instead exposes the soft instability they hide. The auxiliary
diagnostics pinpoint its source. For SOAP and Conda---and, on two
architectures, MARS-Shampoo---the large GNormCV is driven by \emph{rare
single-step bursts}. In the affected scenarios the
gradient-norm std is large (e.g.\ SOAP on Transformer++/1B reaches
std~$\approx$~52 at mean~$\approx$~0.76) while the spike rate stays below
$10^{-3}$, the maximum consecutive spike length is~1, and the tail trend is
near zero ($|$slope$|\!\sim\!10^{-6}$). These are isolated outlier steps that a
NaN/Inf check misses but GNormCV makes visible.

\paragraph{Volatility is architecture-driven, and its form matters.}
Stability is strongly architecture-dependent. On Transformer++, DeltaNet, and
Gated DeltaNet, most optimizers keep GNormCV in the $0.3$--$2$ range, and the
matrix methods Muon and MARS-Shampoo are often the smoothest. GLA is the clear
exception: nearly every optimizer jumps to GNormCV of $10$--$160$, including
AdamW ($113.7$ at 1B) and RMNP ($121.8$ at 340M). Crucially, these GLA blow-ups
again share the single-step-spike signature---spike rate $\sim\!10^{-4}$ and
maximum consecutive spike~$=$~1---with the tail trend staying flat, so the
linear-attention topology amplifies \emph{occasional extreme steps} while the
overall training trajectory still converges stably. This mirrors the Stage-2 rank-stability
finding, namely that both generalization and stability are shaped by the interaction
between optimizer geometry and model architecture.

\begin{takeaway}
\noindent\textbf{O4 takeaway.}
\emph{Training stability must be measured from gradient dynamics, not inferred
from whether a run finishes.} T2 can improve gradient-norm smoothness when its
matrix-level direction construction regularizes the update geometry, but the
same family can also contain methods that produce rare burst-like spikes under
certain architectures. Since all tuned runs may finish without NaN/Inf while
showing very different GNormCV profiles, stability is a mechanism-level
property that must be evaluated directly and cannot be inferred from final PPL
or T-family membership.
\end{takeaway}

\subsubsection{Auxiliary Learning-Rate Perturbation Robustness}
\label{sec:benchmark-o5-lr-robustness}

We next probe hyperparameter robustness (O5) through a local learning-rate
perturbation test. While O1--O3 are measurable from a single tuned run, O5 asks
whether an optimizer remains usable when the learning rate departs from its
tuned value---a practical concern in LLM pretraining, where exhaustive retuning
is often too expensive when scaling the model, changing the dataset, or
transferring to a new architecture.

The test is conducted on the FineWeb-Edu Gated~DeltaNet/340M setting. For each
optimizer $o$, let $\eta_o^\star$ denote its tuned learning rate, and evaluate
three points, $0.2\eta_o^\star$, $\eta_o^\star$, and $5\eta_o^\star$. The
sensitivity score $s_{\mathrm{LR}}$ is the larger relative WikiText-PPL increase
of the two perturbed runs over the tuned run, reported as a percentage. A
smaller $s_{\mathrm{LR}}$ means a flatter local response and stronger
robustness. A perturbed run that diverges or produces an invalid loss is
assigned the maximal sensitivity (treated as the most sensitive case), so that
instability is penalized and made explicit. We stress
that this is a deliberately \emph{local} diagnostic over three points around the
tuned value. The $5\times$ point is merely an aggressive local stress test.

\begin{figure*}[!t]
\centering
\includegraphics[width=1.0\textwidth]{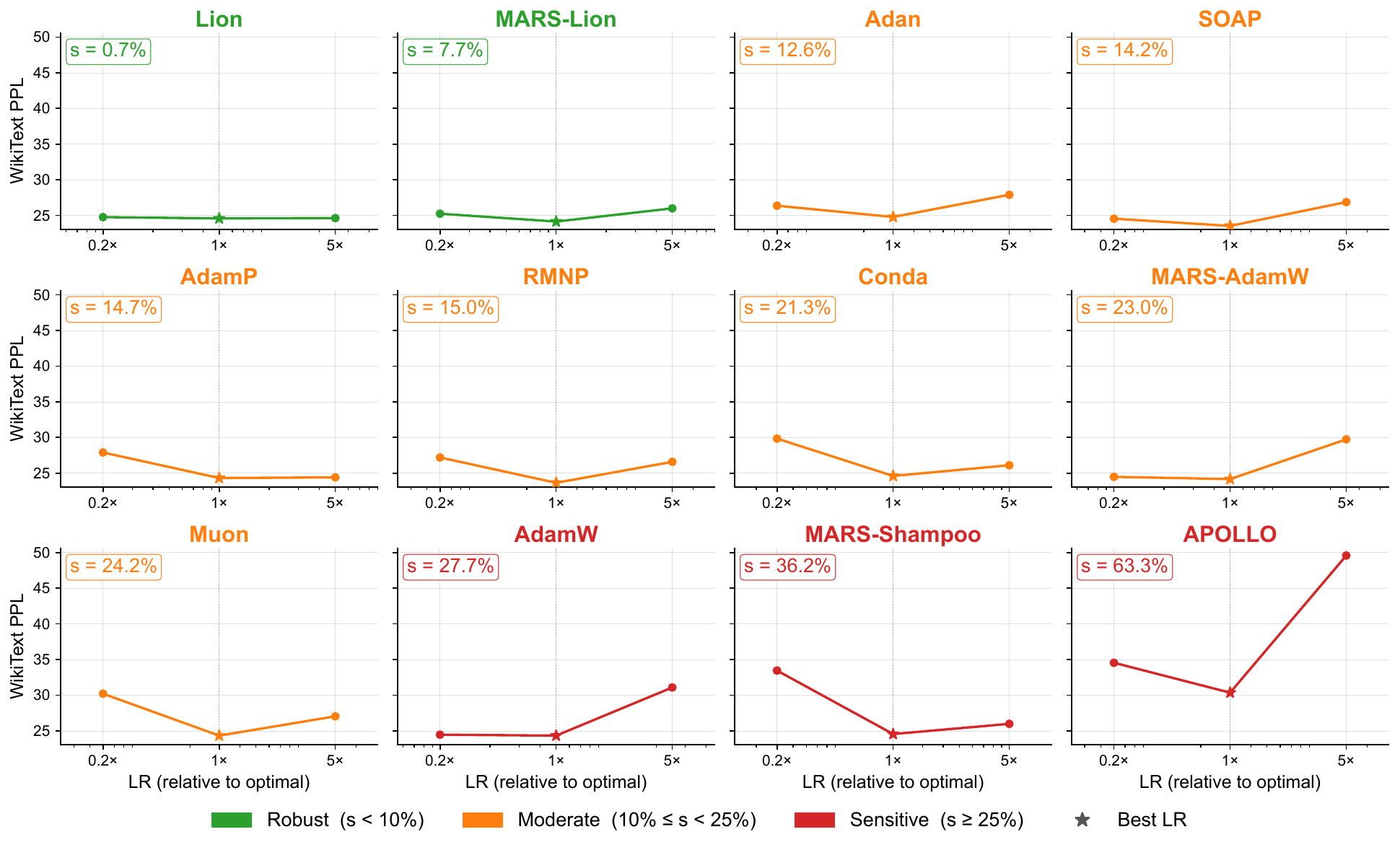}
\caption{\textbf{Auxiliary learning-rate perturbation robustness.}
Each panel shows WikiText PPL under $0.2\times$, $1\times$, and $5\times$
the tuned learning rate. The star marks the tuned learning rate. The
sensitivity score $s_{\mathrm{LR}}$ is shown as $s$ in each panel title and
measures the worst relative PPL degradation from the tuned point, where lower is
more robust. Green, orange, and red titles indicate robust, moderate, and
sensitive learning-rate response regimes.}
\label{fig:o5_lr_robustness}
\end{figure*}

Figure~\ref{fig:o5_lr_robustness} shows that tuned quality and learning-rate
robustness are distinct. Lion and MARS-Lion have the flattest local response
curves ($s_{\mathrm{LR}}=0.7\%$ and $7.7\%$), consistent with sign-direction
methods bounding the per-coordinate step by the learning rate and so absorbing
moderate LR misspecification. We read this as local tolerance. The response is flat in part because the fixed-magnitude update keeps
quality weak across a wide LR range (cf.\ Lion's Stage-1/Stage-2 PPL), so a flat
curve here should not be equated with a strong optimizer. Adan, SOAP, AdamP,
RMNP, Conda, MARS-AdamW, and Muon form a moderate-sensitivity band, where their
degradation stays bounded but shows visible, direction-dependent dependence on
the perturbation.

The sensitive group exposes a different failure mode. AdamW, MARS-Shampoo, and
APOLLO reach reasonable tuned quality, but their PPL rises sharply under at
least one perturbed LR. For AdamW the sensitivity is concentrated on the
\emph{over-large} ($5\times$) side, i.e.\ it
remains well-behaved when the LR is reduced but degrades when pushed past its
tuned value---so its placement here reflects the aggressive upper stress point.
APOLLO is the extreme case, with its
response rising steeply at $5\times$, yielding the largest $s_{\mathrm{LR}}$.
This is consistent with APOLLO's broader profile in the benchmark---strong
short-context quality and low memory, yet poor long-context transfer and high
sensitivity---reinforcing that favorable tuned quality or memory says nothing
about behavior under imperfect hyperparameter transfer.

\begin{takeaway}
\noindent\textbf{O5 takeaway.}
\emph{Learning-rate robustness is not implied by tuned quality.} T3 tends to
be locally tolerant to learning-rate perturbation because its direction-bounded
updates limit the effect of moderate LR misspecification, but this tolerance
can coexist with weaker tuned quality. T1 and T2 usually show a more moderate
sensitivity profile, while T4 can be fragile when the tuned learning rate is
transferred imperfectly. Thus, a T-family's best quality or memory profile
does not necessarily imply robustness under hyperparameter mismatch.
\end{takeaway}

\subsubsection{Family-Level Objective Summary}
\label{sec:benchmark-family-summary}

\begin{figure}[!t]
\centering
\includegraphics[width=\linewidth]{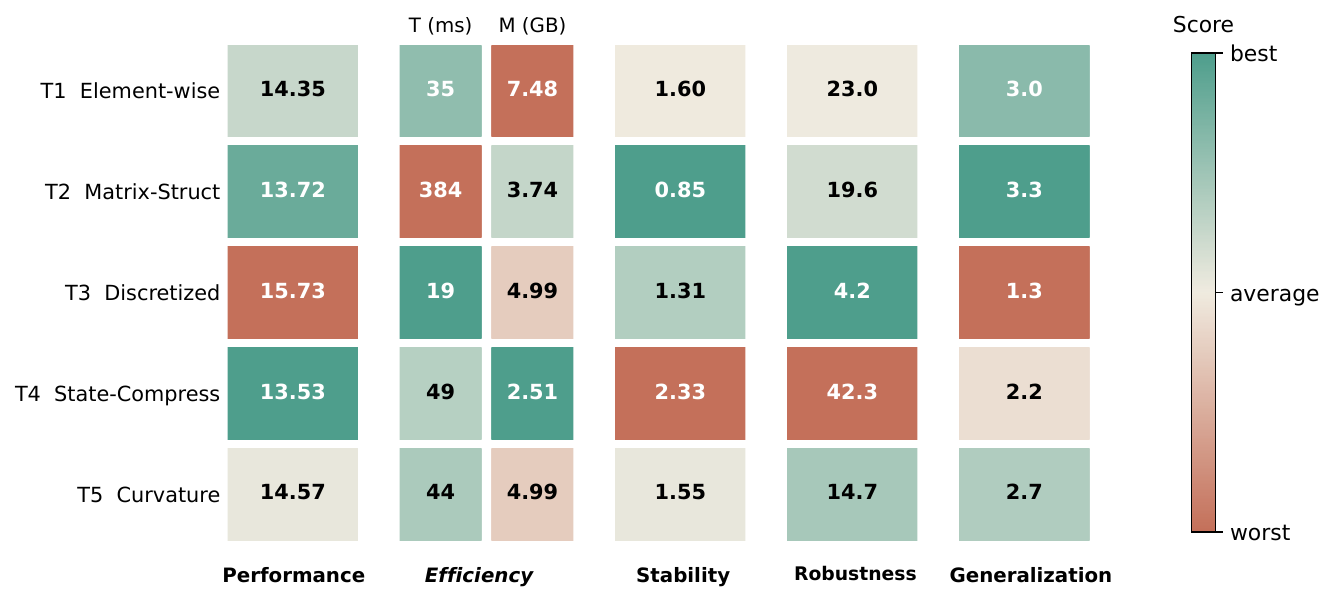}
\caption{\textbf{Family-level objective summary.}
Per-family profile over $O1$--$O6$. $O1$--$O3$ are directly measured from
Stage~1; $O4$ is measured from gradient-norm stability; $O5$ is probed by an
auxiliary learning-rate perturbation test; $O6$ is measured through Stage~2
rank stability and sequence-length sensitivity. Green indicates a more
favorable profile and red indicates a less favorable profile.}
\label{fig:family_heatmap}
\end{figure}

After Stage~1 analyzes the directly measured quality--efficiency trade-offs,
Stage~2 evaluates cross-scenario transfer, and the auxiliary analyses cover
training stability and learning-rate perturbation robustness, we summarize the
results at the optimizer-family level. This summary connects the empirical
benchmark back to the objective-oriented taxonomy. Unlike the Stage~1 Pareto
plots, which only cover $O1$--$O3$, the family-level view in
Figure~\ref{fig:family_heatmap} summarizes the broader objective profile, namely
quality, runtime, memory, training stability, hyperparameter robustness, and
generalization.

Figure~\ref{fig:family_heatmap} consolidates the preceding evidence. $T1$
methods form the stable baseline. They are not the strongest in raw PPL, but
they remain moderate in runtime and memory, transfer reasonably across
architectures, and provide a reliable reference for stability and robustness
comparisons. $T2$ methods provide strong matrix-structured quality and
generalization potential, but they are internally heterogeneous, where SOAP gives
the strongest long-context quality at high cost, RMNP provides the most
balanced quality--efficiency trade-off, and Muon shows the strongest observed
gradient-norm stability profile while remaining architecture-sensitive. $T3$
methods are efficient and locally robust to learning-rate perturbation, but
their tuned PPL is generally weaker. $T4$ methods achieve attractive
short-context memory efficiency, but their Stage~2 behavior reveals a clear
generalization boundary. APOLLO is the most representative case, where a
favorable short-context quality--memory trade-off does not transfer reliably
to long-context training. $T5$ methods remain competitive only in limited
regimes and do not provide a consistent advantage under the current benchmark.

The family-level summary reinforces the main message of the benchmark, namely that
optimizer selection is not a single-axis ranking problem. Different families
occupy different regions of the $O1$--$O6$ objective space, and an optimizer
that is strong in one objective can be weak in another. Quality, cost,
stability, robustness, and transfer must therefore be evaluated jointly. In
particular, $O5$ should be interpreted as a local learning-rate robustness
diagnostic whose scope is limited to local perturbations of the learning rate.

\begin{wraptable}{r}{0.45\linewidth}
\centering
\vspace{-1.5em}
\caption{\textbf{Tiered classification of the benchmarked optimizers.}}
\label{tab:optimizer_tiers}
\setlength{\tabcolsep}{1.0mm}
\renewcommand{\arraystretch}{1.15}
    \resizebox{1.\linewidth}{!}{
    \begin{tabular}{ll}
    \toprule
    \textbf{Tier} & \textbf{Optimizers} \\
    \midrule
    \textbf{Tier I} & Muon, RMNP, AdamW \\
    
    \textbf{Tier II} & MARS-Lion, MARS-Shampoo, APOLLO\\
                     & Conda, AdamP, MARS-AdamW, SOAP, \\
                     & Adan, Lion \\
    
    \textbf{Tier III} & RAdam, NAdam, Prodigy, AdaBelief, \\
                      & GaLore, Shampoo, 8-bit Adam, CAME, \\
                      & AdaFactor, Adam-mini, LAMB, Sophia \\
    \bottomrule
    \end{tabular}
    }
\vspace{-2.0em}
\end{wraptable}
\subsubsection{Tiered Optimizer Summary}
\label{sec:benchmark-tiered-summary}

Combining the Stage~1 short-context screening, the Stage~2 long-context
generalization study, and the sequence-length sensitivity analysis, we organize
the evaluated optimizers into three tiers. These three tiers are a grouping
from three perspectives that summarizes the overall empirical evidence, namely
short-context quality--efficiency behavior, cross-scenario transfer, and
mechanistic value for further analysis. Interpreting them as a strict global
leaderboard would be misleading.

\paragraph{Tier~I: primary candidates.}
Tier~I contains the optimizers that are most important for practical baselines
and subsequent mechanistic analysis. AdamW is retained as the default reference
because it is stable, efficient, and widely used, even though other methods surpass it in some settings. RMNP is selected as the most balanced
matrix-structured optimizer, because it achieves strong quality while avoiding the
prohibitive runtime and memory cost of heavier matrix methods. Muon is included
because it combines strong Stage~1 performance with a clean matrix-structured
mechanism that can be decomposed into orthogonalization, learning-rate scaling,
momentum placement, and operator-order choices. Although Muon shows
architecture-dependent behavior in Stage~2, this sensitivity is itself
mechanistically informative and motivates the ablation study below.

\paragraph{Tier~II: scenario-dependent methods.}
Tier~II contains optimizers with clear strengths but stronger dependence on
the training regime, context length, architecture, or efficiency budget. SOAP
achieves the strongest long-context cross-scenario quality, but its runtime and
optimizer-state memory costs make it less attractive as a default choice.
APOLLO is the best short-context quality--memory point in Stage~1, but its
long-context collapse shows that aggressive state compression does not transfer
automatically. MARS-AdamW and AdamP are stable in Stage~2, yet they do not
dominate the Stage~1 quality--efficiency frontier. MARS-Shampoo, Conda, Adan,
Lion, and MARS-Lion similarly provide useful comparison points, but their
advantages are limited by cost, transferability, or final quality. We therefore
treat this tier as a set of important conditional baselines whose applicability
depends on the specific training scenario.

\paragraph{Tier~III: diagnostic failure cases.}
Tier~III contains methods that are weak, dominated, or insufficiently robust
under the current benchmark protocol. Some methods retain isolated advantages, where
AdaFactor has extremely low optimizer-state memory, GaLore has low runtime and
subspace compression, and 8-bit Adam reduces state precision. However, these
advantages do not yield a consistently favorable trade-off across quality,
efficiency, and transfer. Other methods, such as Sophia, LAMB, Prodigy, RAdam,
NAdam, and AdaBelief, do not improve sufficiently over stronger baselines in
the evaluated regimes. Their value is mainly diagnostic, because they help identify
failure modes associated with direction discretization, noisy curvature
estimation, aggressive state compression, or insufficient scaling behavior.

\paragraph{Selection guidance.}
The tiered summary leads to a constraint-driven view of optimizer selection.
For general-purpose LLM pretraining, AdamW should remain the default reference, because although it is not the
strongest method in every metric, it is stable, inexpensive,
and consistently interpretable across settings. When a better balance between
quality and efficiency is required, RMNP is the most attractive alternative in
this benchmark. It preserves much of the benefit of matrix-structured
optimization while avoiding the extreme runtime and memory overhead of heavier
matrix methods.

When final quality is the dominant objective and compute or optimizer-state
memory is not the bottleneck, SOAP provides the strongest long-context quality
reference. Its Stage~2 rank stability suggests that it transfers well across
the tested architectures, but its high cost makes it better suited as a
quality ceiling. Muon occupies a different
role, namely that it is a strong and interpretable matrix-structured optimizer, but its
architecture-dependent behavior suggests that it should be used with awareness
of the target model topology.

For memory-constrained short-context training, APOLLO and AdaFactor are useful
candidates, but they should be separated carefully. AdaFactor is the safer
low-memory baseline with moderate quality, whereas APOLLO is a high-reward but
high-risk compressed-state method. It is excellent in the Stage~1
quality--memory trade-off, yet its advantage does not survive the long-context
setting. For low-cost exploratory runs, Lion can reduce per-step overhead, but
the quality gap should be expected. Overall, the benchmark does not identify a
universal best optimizer. It shows that optimizer choice should be made by
matching the method's dominant strength to the binding constraint of the
training regime: stability, quality, runtime, memory, or cross-scenario
transfer.

\begin{takeaway}
\noindent\textbf{Tiered-summary takeaway.}
Optimizer selection should be matched to the dominant constraint of the
training regime. T1 is the safest reference family. T2 provides the strongest
quality or stability ceiling when matrix-level cost is acceptable, while lightweight
T2 variants offer a more practical balance between quality and efficiency.
T3 is useful when runtime or local LR tolerance matters more than peak quality.
T4 is attractive mainly under short-context memory pressure, but requires
additional long-context validation. Overall, the benchmark supports
mechanism-aware optimizer selection.
\end{takeaway}
\subsection{Mechanistic Ablation of Muon}
\label{sec:benchmark-muon-ablation}
The benchmark identifies Muon as a strong and mechanistically clean
matrix-structured optimizer. Muon is commonly understood as a momentum-based
matrix optimizer whose key operation is Newton--Schulz (NS)
orthogonalization~\cite{jordan2024muon}. However, this description leaves open
three questions: whether NS is truly indispensable, which auxiliary operations
provide additional gains, and whether the sub-operations can be reordered
freely. We therefore use Muon as a controlled case study for the meta-pipeline
view.

\subsubsection{Single-scene decomposition on C4-350M}
\label{sec}

\begin{figure*}[!t]
  \centering
  \includegraphics[width=1.0\textwidth]{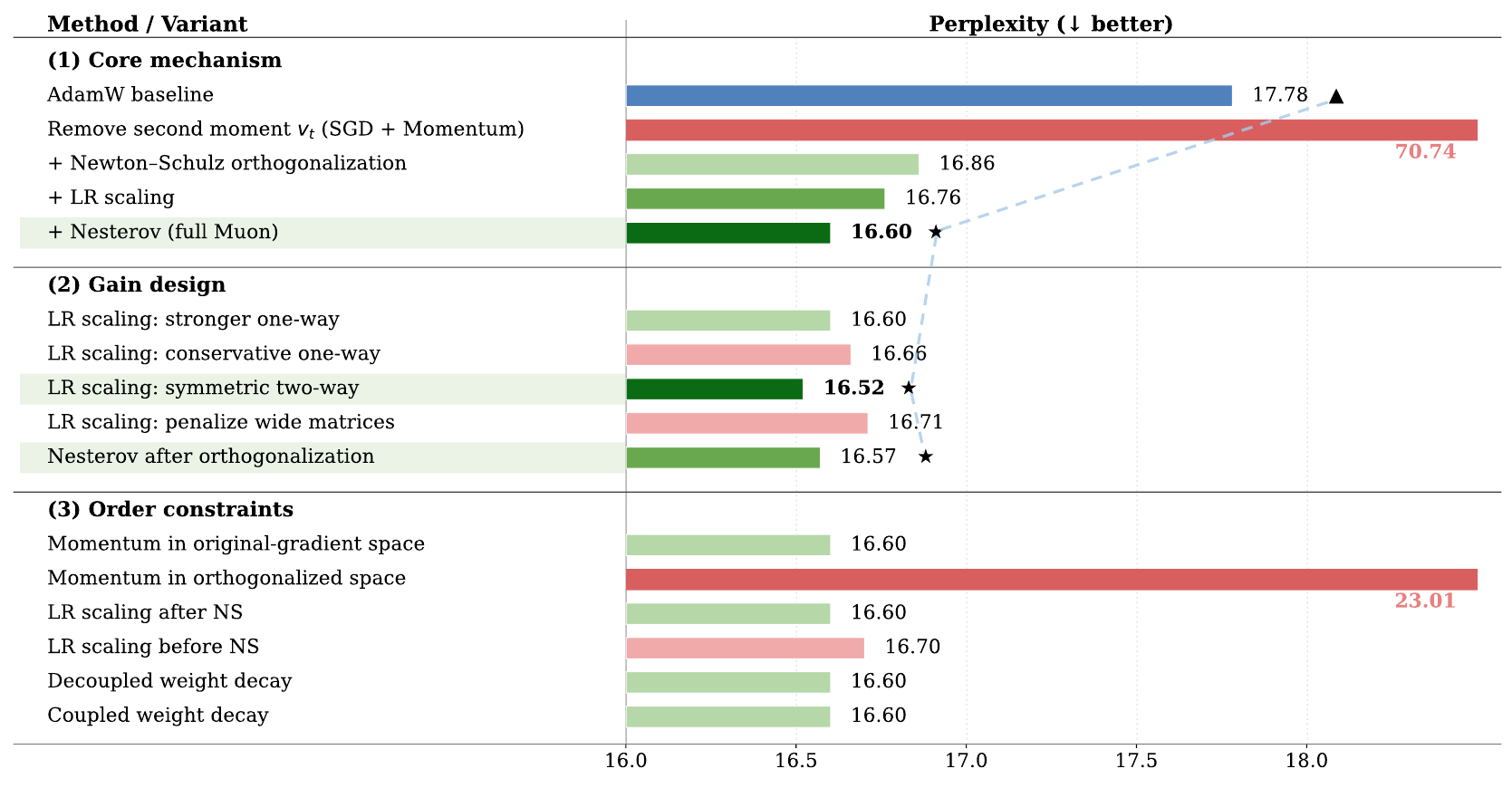}
  \caption{\textbf{Mechanistic ablation of Muon (C4, 350M).} Decomposition into core
operations, gain operations, and operator-order constraints, with bars showing absolute PPL (lower
is better). The 70.74 bar is truncated off-scale.}

  \label{fig:muon_ablation_c4_350m}
\end{figure*}

We first perform a controlled decomposition on C4-LLaMA at the 350M scale,
matching the Stage~1 screening setting. Figure~\ref{fig:muon_ablation_c4_350m}
summarizes the results as absolute validation PPL, where lower is better. The
three blocks correspond to core mechanism recovery, gain design, and
operator-order constraints.

\paragraph{NS orthogonalization is the core mechanism.}
The first block shows that simply removing AdamW's second moment is
catastrophic, since PPL increases from 17.78 to 70.74. Adding NS orthogonalization
immediately recovers performance to 16.86, already better than AdamW. This
identifies NS as Muon's core replacement for AdamW-style diagonal adaptive
scaling.

\paragraph{Scaling and Nesterov are secondary gains.}
Once NS orthogonalization is present, LR scaling and Nesterov correction
provide smaller but consistent improvements. LR scaling improves PPL from
16.86 to 16.76, and adding Nesterov gives the full Muon result of 16.60.
Thus, these operations refine the orthogonalized direction, but they are not
responsible for the main recovery from the failed no-$v_t$ variant.

\paragraph{The best gains are geometry-aware.}
The second block shows that the form of the gain operation matters. Symmetric
two-way LR scaling gives the best single-scene result, reaching 16.52, whereas
penalizing wide matrices worsens PPL to 16.71. Post-NS Nesterov also improves
over the standard ordering, reaching 16.57. These results suggest that gain
operations should be designed around the geometry of the NS-orthogonalized
direction.

\paragraph{Operator order is constrained.}
The third block shows that Muon's operations are not freely permutable.
Accumulating momentum in the orthogonalized space degrades PPL to 23.01,
whereas accumulating momentum in the original-gradient space keeps PPL at
16.60. Similarly, applying LR scaling before NS worsens PPL to 16.70,
indicating that spectral normalization can absorb or distort the intended
scale correction. Weight-decay ordering is less important in this specific
C4-350M setting, where coupled and decoupled variants are nearly equivalent.

\begin{takeaway}
\noindent\textbf{Single-scene Muon takeaway.}
Muon's main improvement comes from NS orthogonalization, which replaces
AdamW's diagonal second-moment scaling with a matrix-level direction. LR
scaling and Nesterov further refine this direction, but only when placed after
the appropriate upstream operations. The ablation therefore supports the
meta-pipeline view: Muon changes only a few stages, yet the location and order
of those changes are essential.
\end{takeaway}

\subsubsection{Cross-scale and cross-architecture validation}
\label{sec:benchmark-muon-cross-validation}

The single-scene ablation identifies the core and gain operations, but it does
not tell us whether the gain operations transfer. We therefore validate the two
best gains---symmetric two-way LR scaling and post-NS Nesterov---across scale
and architecture. Table~\ref{tab:muon_cross_validation} compares standard
Muon, each gain alone, and their combination on C4-LLaMA at 350M and 1B, as
well as on FineWeb-Edu 32k with Gated DeltaNet at 340M.

\begin{table}[t]
\centering
\vspace{-0.25em}
\caption{\textbf{Cross-scale and Cross-Architecture Validation for Gain Operations in Muon.} Standard Muon, symmetric two-way LR scaling, post-NS Nesterov, and their combination; lower PPL is better. Gains stack on the standard Transformer but not on Gated DeltaNet.}
\label{tab:muon_cross_validation}
\small
\setlength{\tabcolsep}{6pt}
\renewcommand{\arraystretch}{1.14}
\resizebox{0.95\linewidth}{!}{
\begin{tabular}{lccccc}
\toprule
\textbf{Scenario}
& \makecell{\textbf{Standard}\\\textbf{Muon}}
& \makecell{\textbf{Symmetric}\\\textbf{LR Scaling}}
& \makecell{\textbf{Post-NS}\\\textbf{Nesterov}}
& \makecell{\textbf{Both}\\\textbf{combined}}
& \textbf{Best config.} \\
\midrule
\multicolumn{6}{l}{\textit{Standard Transformer: gains are stackable}} \\
C4-LLaMA, 350M
& 16.60
& 16.52
& 16.57
& \bestnum{16.51}
& Both combined \\
C4-LLaMA, 1B
& 13.72
& 13.64
& 13.64
& \bestnum{13.58}
& Both combined \\
\midrule
\multicolumn{6}{l}{\textit{Linear attention: stacking effect disappears}} \\
\rowcolor{warnbg!35}
FineWeb-Edu 32k, GDN-340M
& 24.26
& \bestnum{24.02}
& 24.12
& 24.12
& Symmetric LR Scaling \\
\bottomrule
\end{tabular}}

\end{table}
On standard Transformer models, the two gains are transferable and partially
stackable. At 350M, symmetric LR scaling improves PPL from 16.60 to 16.52,
post-NS Nesterov reaches 16.57, and combining both gives 16.51. At 1B, both
individual gains reach 13.64, and the combined variant further improves to
13.58. This indicates that the two gains act through mostly separable
mechanisms on standard Transformer architectures.

On Gated DeltaNet, the behavior is different. Both gains remain useful
individually, with symmetric LR scaling improving PPL from 24.26 to 24.02, and
post-NS Nesterov reaching 24.12. However, their combination no longer improves
over the best single gain, remaining at 24.12. This suggests that the
orthogonality of optimizer components is architecture-dependent, meaning that operations
that are separable on standard Transformers can interact on linear-attention
parameter topologies.

\begin{takeaway}
\noindent\textbf{Cross-validation takeaway.}
The Muon gains transfer across scale on standard Transformers, but lose
additivity on Gated DeltaNet. Thus, LR scaling and post-NS Nesterov are useful
refinements, but their interaction depends on the target architecture. This
explains Muon's role in the benchmark: it is not a universally best optimizer,
but it is strong, interpretable, and exposes where matrix-structured optimizer
design meets model topology.
\end{takeaway}
\section{Discussion}
\label{sec:discussion}

This paper was organized around three questions about where an optimizer acts,
why its update takes a particular geometric form, and what training objective it
is meant to improve. We answered them through three layers. The first is the
meta-pipeline of Section~\ref{sec:meta-pipeline}, the second is the four-axis
decomposition of Section~\ref{sec:four-axis-decomposition}, and the third is
the dual-dimension taxonomy of Section~\ref{sec:taxonomy}. All three are
grounded in the benchmark of Section~\ref{sec:benchmark}. This section steps back from those instruments and asks what they collectively do and
do not establish. We first delimit the scope
of the present study (Section~\ref{sec:discussion-limitations}), then use the
framework to extract technique-level lessons about benefit, composition, and
conflict (Section~\ref{sec:discussion-compositions}), and finally
distill the open directions that follow (Section~\ref{sec:discussion-open-problems}).

\subsection{Limitations}
\label{sec:discussion-limitations}

The introduction framed optimizer selection as a jointly empirical,
mathematical, and engineering problem. A modern training run is constrained by more than whether an update reduces the
loss. Compute, optimizer-state memory, tuning budget, and task diversity all
bound it. The literature is also hard to navigate because mechanisms are named
locally, empirical claims are protocol-sensitive, and mathematical accounts are
fragmented
(Section~\ref{sec:intro-background}). Our framework targets exactly these gaps,
and the benchmark supplies controlled evidence for them. The same framing,
however, makes the boundaries of the current study explicit.

\paragraph{Empirical scope.}
The empirical study substantiates the framework primarily on language-model
pretraining, covering four model scales from 60M to 1B parameters, four
architectures spanning standard and linear attention, and context lengths from
256 to 32k tokens. In addition, the benchmark includes a CIFAR100 vision-backbone study across
ResNet50, DeiT-S, and CAFormer-S12. This study tests whether optimizer behavior
transfers across architecture families within image classification.
Thus, claims about \emph{domain-dependent crossings} are supported by the
language-modeling regime and by a targeted CIFAR100 cross-architecture check.
In future work we will add validation on ImageNet-scale vision and additional
modalities, extending these into a full cross-modality conclusion.

\paragraph{Mechanism attribution.}
Every effect-target assessment in Section~\ref{sec:benchmark} is deliberately a
mechanism-informed prior. The mechanistic explanations we give are also \emph{qualitative}. They cover
APOLLO's rank-bounded compression, Muon's interaction with architecture topology,
and MARS's variance reduction on the state estimator while preserving base
geometry. We can say \emph{why} an optimizer plausibly behaves as observed. In future work
we will develop metrics that quantify these explanations, measuring quantities
such as the effective
rank of the gradient as context grows, the staleness of a preconditioning basis,
and the fraction of a technique's benefit that is intrinsic to the algorithm
versus
induced by the tuning protocol. Such interpretability metrics must be designed
and validated carefully, and we will provide a quantitative interpretability
evaluation in a subsequent version of this work. For future work, we plan to add
heavy-tailed self-regularization indices as training-quality metrics and to
evaluate optimizers on downstream commonsense reasoning benchmarks.

\paragraph{Protocol dependence.}
Our conclusions are tied to a strict controlled-variable protocol, in which only
optimizer-related hyperparameters are tuned while architecture, data, and
schedule are held fixed. This isolates optimizer mechanism, but it also makes
several findings protocol-relative. Many apparent
improvements over AdamW shrink or disappear once the baseline is retuned. The
learning-rate robustness objective (O5) is reported as a \emph{local} diagnostic.
The long-context comparison is
matched on dataset and architecture family but not on token budget, so it
certifies a degradation \emph{trend}.
A few memory-efficient methods are drawn from recent preprints whose broader
LLM-scale behavior is still being established. Their results should therefore
be interpreted under the present protocol and remain preliminary for the
method class.

\paragraph{Method coverage.}
We instantiate 24 representative optimizers chosen to cover the major regions of
the four-axis space (Table~\ref{tab:four-axis-instantiation}), but the surveyed literature
exceeds one hundred methods, and not every axis combination is realized by an
existing optimizer or exercised by our benchmark. Consequently, family-level statements should be read as conclusions about the
\emph{tested} instances and protocol. This caution is especially important for the claim that
curvature-aware and geometry-regularized methods (T5) are competitive only in
limited regimes under the current benchmark. These boundaries motivate the
technique-level reading and the open directions that follow.

\subsection{Technique-Level Lessons}
\label{sec:discussion-compositions}

A central payoff of the framework is that it shifts attention from
\emph{which optimizer is the best} to a more tractable set of technique-level
questions. Which mechanisms carry the gains? Which ones can be stacked? Which
ones interfere?
The governing principle, established by the meta-pipeline and the cross-dimension
analysis (Section~\ref{sec:taxonomy-cross}), is one of \emph{locality}.
Mechanisms that act on different pipeline stages or different axes tend to
compose, whereas mechanisms that contend for the same slot require an explicit
ordering and may conflict.

\paragraph{Benefit carriers.}
Under the tested protocol, the strongest gains are associated with
\emph{geometry-sensitive direction maps} (Axis~III) and with the
\emph{structured state} that feeds them (Axis~II). Scalar control tweaks alone
contribute little. The Muon ablation makes this concrete. Removing the
diagonal second moment alone collapses quality (PPL $17.78\!\rightarrow\!70.74$),
whereas replacing it with Newton--Schulz orthogonalization recovers and then
surpasses the AdamW baseline ($\rightarrow\!16.86$). This identifies spectral
orthogonalization as the indispensable core.
Kronecker-factored preconditioning is another strong carrier. SOAP attains
the strongest transferable quality profile in Stage~2, with a mean rank near the
top across architectures. This comes at the cost of the highest per-step runtime
and the highest memory use.
Variance reduction in the style of MARS is a genuine, reusable improvement to the
state estimator. It improves the gradient signal while preserving the base update
geometry, which makes MARS-AdamW the most stable AdamW-style enhancement in
Stage~2. The gain comes with extra state and computation, so MARS is better read
as a quality-oriented extension. Finally, simple
memory techniques that \emph{preserve} geometry, such as row/column
factorization (AdaFactor) and low-bit state (8-bit Adam), deliver
reliable memory savings without changing the descent direction, which is why they
are safe defaults when memory is the binding constraint.

\paragraph{Limited returns.}
Conversely, several techniques are marginal in the tested regime. Small
element-wise refinements of the AdamW template land in the weakest tier for
tuned perplexity. These include early-variance rectification (RAdam),
Nesterov lookahead (NAdam), and prediction-error variance (AdaBelief). Automatic
step-size tuning (Prodigy) lands in the same tier. This is consistent with
their stated targets of robustness or convenience.
Diagonal curvature proxies and geometric wrappers (Sophia, LAMB) are likewise
weak or only situationally competitive here. They add cost through an extra
Hessian-vector product, clipping or trust-ratio parameters, or a second gradient
evaluation, yet they do not deliver a consistent quality return under matched
tuning. These mechanisms still have value. Their advantage is simply
conditional, and the benchmark is meant to expose that condition.

\paragraph{Compatible compositions.}
Locality predicts, and the survey confirms, several orthogonal compositions.
(i)~A variance-reduced signal (Axis~II) can feed essentially any base direction,
because it is largely independent of basis and curvature choice. This is exactly
the pattern behind the MARS-AdamW, MARS-Lion, and MARS-Shampoo constructions.
(ii)~Low-rank projection (T2.3, acting on the signal subspace across S2 and
S4) and state quantization (T4.2, acting on the state representation across S3
and S4) modify disjoint objects and combine cleanly, as in Q-GaLore.
(iii)~Post-update filters (T5.3) wrap a base optimizer because they act
\emph{after} the core direction is chosen. This yields the cautious variants of
AdamW and Lion from a single mechanism. (iv)~Layer-wise trust ratios (T5.4)
act at finalization (S5) and can therefore wrap an element-wise adaptive update
without touching its direction. (v)~Even within a single method, the Muon
ablation shows that smaller gain operations can act through partially separable
mechanisms. In Muon, symmetric
two-way learning-rate scaling and post-orthogonalization Nesterov momentum stack
across model scales on the standard Transformer.

\paragraph{Conflicts and limits.}
The same locality principle predicts where stacking fails. (i)~Two strong S2
matrix constraints, spectral orthogonalization and low-rank projection, contend
for the same slot. Combining them is therefore not free. The
result depends on whether projection precedes orthogonalization or vice versa,
and the two orders yield different directions and state footprints.
(ii)~Fused backprop-update streaming (LOMO-style, T4.4) shortens the gradient's
lifetime and thereby conflicts with
any mechanism that needs complete gradient statistics, including global
clipping, delayed low-rank basis refresh, SAM-style second gradients, or matrix
preconditioners.
(iii)~Some compositions are valid but impractical. Stacking SAM's double
forward-backward on top of expensive Kronecker-factored preconditioning multiplies cost
for little expected return. (iv)~Most importantly, two conflicts are
\emph{empirical and quantitative}. Aggressive state compression is
\emph{rank-bounded}. APOLLO is the single best optimizer at 256 tokens but the
worst at 32k. Its degradation is
$+21.87$ PPL, which is roughly three times the AdamW baseline degradation. The
reason is that a fixed low-dimensional projection discards proportionally more
information as the gradient's effective rank rises
with context (Section~\ref{sec:benchmark-mem-eff-sensitivity},
Table~\ref{tab:seqlen_numbers}). Muon's spectral geometry is also
\emph{architecture-conditional}. The gain
operations that stack on the standard Transformer no longer improve beyond the best single gain on Gated DeltaNet.
The operator order matters as well. Accumulating momentum in the
orthogonalized space hurts, and so does scaling the
learning rate before orthogonalization
(Section~\ref{sec:benchmark-muon-ablation},
Table~\ref{tab:muon_cross_validation}).

\paragraph{Design takeaway.}
Read together, these results recommend treating the four-axis coordinate of a
method as a practical compatibility test. Techniques that occupy different axes
or pipeline stages are good candidates for composition. Techniques that share a
slot demand an explicit ordering. Benefits that depend on gradient rank or
architecture topology, especially compression and spectral geometry, must be
validated in the target regime, and their transfer must be confirmed.
The implication is that optimizer geometry must be matched to model structure
and training dynamics. This is the sense in which no single optimizer dominates
the multi-objective frontier. The unit of design is a \emph{composition} chosen
against a specific budget, context length, and architecture.

\subsection{Open Problems and Future Directions}
\label{sec:discussion-open-problems}

The limitations and the technique-level reading point to a coherent agenda for
LLM optimizer design and evaluation.

\paragraph{Diagnostics and adaptive compression.}
The most direct gap is the absence of measures that turn our qualitative
mechanism attributions into testable quantities. Useful examples include an
effective-rank estimate that predicts when compression will become lossy, a
basis-staleness measure that predicts when a Kronecker or low-rank preconditioner
should refresh, and an intrinsic-versus-protocol decomposition that separates
genuine algorithmic gains from tuning artifacts. Such metrics would convert the
present coordinate system into a predictive diagnostic, and they point to a
concrete algorithmic target. The APOLLO collapse shows the payoff of an
effective-rank estimate, namely compression schemes whose retained rank or projection
adapts to the rising effective rank of the gradient as context length grows, so
that short-context memory wins do not become long-context quality failures.

\paragraph{Architecture-aware geometry and transfer.}
Rankings cross across dataset, context length, and architecture in ways that no
single-setting comparison anticipates, and the two clearest mechanisms behind
those crossings, compression versus gradient rank and spectral geometry versus
attention topology, are understood only post hoc. This motivates a predictive
account of \emph{when} a given geometry transfers, tied to measurable properties
of the model and data, which would let practitioners select an optimizer without
an exhaustive sweep. The same handle suggests a design principle, namely to choose
optimizer geometry from measurable properties of the architecture and the
evolving training dynamics. RMNP~\citep{deng2026rmnp} is one example. Under row-wise
block-diagonal dominance in the layer-wise Hessian and diagonal dominance in the
momentum Gram matrix, its row normalization is a structure-aware specialization
of Muon-style preconditioning,
and future optimizers should co-design the preconditioner with the architecture
and the statistics that emerge during training.

\paragraph{Multi-objective selection.}
Because the O1--O6 objectives genuinely trade off and no method is uniformly
best, optimizer choice is a multi-objective problem that the field still treats
as a single-number leaderboard. Cost-aware comparison (token efficiency versus
wall-clock and memory), Pareto-aware selection, and optimizer designs that target
a specified region of the frontier are all open.

\paragraph{Compositional search.}
The framework identifies compatible and conflicting compositions by hand.
Automating this would mean searching axis-compatible combinations, resolving
same-slot ordering, and co-designing variance reduction, direction geometry, and
state compression. It also includes the under-explored question of how to combine
two S2 constraints without sacrificing either.

\paragraph{Cost-effective curvature.}
Methods that use curvature across T2 and T5 show the same cost tension.
Sophia follows a T5 diagonal-curvature route, while Shampoo and SOAP follow T2
Kronecker-basis routes. They deliver either inconsistent gains or strong gains
at prohibitive cost. Cheaper and more stable curvature estimation would help
determine when these mechanisms become default tools and when they remain
specialist choices. The same is true for a clear characterization of the regimes
in which curvature information is worth its overhead.

\paragraph{Matched protocols.}
Finally, the protocol sensitivity documented throughout argues for standardized,
matched-budget evaluation. Useful examples include full learning-rate-transfer
studies, matched-token long-context
comparisons, and reporting conventions that make the controlled-variable
assumptions explicit. This would make it easier to separate intrinsic
algorithmic progress from tuning and protocol effects.

\section{Conclusion}
\label{sec:conclusion}

Optimizer selection for large-scale model training has become a system-level,
multi-objective decision, yet the field offers more than one hundred methods
described in incompatible vocabularies and supported by protocol-sensitive
evidence. This paper set out to make this body of work navigable and, ultimately,
actionable. We treated every optimizer as a structured transformation of a
stochastic training signal and showed that modern methods are sparse
modifications of a shared five-stage meta-pipeline. We then used norm-constrained linear minimization oracles to place sign,
spectral-orthogonalized,
Kronecker-factored preconditioned, and projected updates inside a single four-axis
decomposition. Finally, we organized the resulting mechanisms into a
dual-dimension taxonomy that records both where a method acts and which training
objective it targets. A controlled benchmark over language-model pretraining
then turned this framework from a descriptive device into an evaluative one. The benchmark spans four scales from 60M to 1B parameters, four architectures,
and context lengths from 256 to 32k tokens, with a targeted CIFAR100
vision-backbone study used to probe architecture-dependent optimizer behavior
beyond language modeling.

Our main conclusion is simple. There is no universal best optimizer, and the
practical question is which mechanism's strength matches the binding constraint
of the training regime. Under the
tested protocol, the strongest quality gains are associated with
geometry-sensitive direction maps and the structured state that feeds them.
Scalar refinements of the adaptive-moment template alone contribute little. Most
element-wise variants of AdamW do not survive a retuned baseline. Two empirical
regularities sharpen this picture and should be treated as design constraints.
Aggressive state compression is rank-bounded and
degrades sharply as the gradient's effective rank rises with context length.
Spectral matrix geometry is architecture-conditional, so its gains must be
validated on the target topology, and their transfer must be confirmed.
Composability, in turn, is governed by locality on the four axes. Techniques on
different axes or pipeline stages stack, while techniques that contend for the
same slot require an explicit ordering. The unit of design is therefore a
composition chosen against a budget.

These findings translate into concrete guidance. Across the benchmarked
methods, the mainstream choices and the regime each one fits
(Table~\ref{tab:optimizer_tiers}) are the following.
\begin{itemize}
  \item \textbf{AdamW} is the default reference for general-purpose
  pretraining. It is not the strongest on any single metric, but it is stable,
  inexpensive, interpretable, and the baseline every other choice should be
  measured against.
  \item \textbf{RMNP}~\citep{deng2026rmnp} is better read as a
  practical, architecture-aware step beyond standard Muon-style preconditioning.
  It targets a better quality--efficiency balance by
  using architecture-induced row-wise structure. When the relevant curvature and
  momentum statistics are diagonally dominant in a row-wise sense, a Muon-style
  matrix direction can be implemented through a cheaper row-normalized form.
  \item \textbf{SOAP} is the quality ceiling for long-context training and the
  most transferable method across architectures. It is too costly to be a
  default, but it is useful when final quality dominates and compute and memory
  are not the bottleneck.
  \item \textbf{Muon} is a stable, mechanistically transparent
  matrix method. It should be used with awareness of the target model topology.
  \item \textbf{AdaFactor} is the safe low-memory baseline at moderate quality.
  \textbf{APOLLO} is a high-reward, high-risk compressed-state method that excels
  at short context but collapses at long context, so its short-context wins do
  not certify it.
  \item \textbf{Lion} is a cheap exploratory option with an expected quality gap.
  Curvature-aware and geometry-regularized methods such as Sophia and LAMB remain
  situational under the current protocol.
\end{itemize}
In short, the decision rule is to identify the single binding constraint of a
run, such as stability, quality, runtime, memory, or cross-scenario transfer,
and select the method whose dominant strength addresses it. Starting from AdamW,
move to RMNP, SOAP, or a memory-efficient method only when a specific constraint
demands it.

Beyond these recommendations, the framework's main contribution to the
community is an operational coordinate system. Any optimizer, including methods
that have not yet been proposed, can be located by its pipeline stages and its
four-axis coordinate. Its composability with other techniques can be predicted from that
location, and competing methods can be compared under
explicit mechanism and objective assumptions.
This gives practitioners a shared vocabulary and a constraint-driven decision
procedure. For method designers, it maps crowded regions, unexplored
compositions, and empirical boundaries that any new method must respect, such as
rank-bounded compression and architecture-conditional geometry. A further
lesson, developed in Section~\ref{sec:discussion-open-problems}, is that
optimizer geometry should be coupled to the model architecture and training
dynamics, with RMNP as one concrete example. The
same coordinate system marks where this study should grow next. One direction is to develop
quantitative interpretability metrics that turn the present qualitative mechanism
attributions into predictive diagnostics. A
second direction is to extend the study beyond language modeling, building on the
foundation established here.


\clearpage
\newpage
\bibliographystyle{plainnat}
\setcitestyle{numbers}
\bibliography{ref}

@String(AAAI = {AAAI})

@article{kingma2015adam,
  title={Adam: A method for stochastic optimization},
  author={Kingma, Diederik P and Ba, Jimmy},
  journal={arXiv preprint arXiv:1412.6980},
  year={2014}
}

@article{loshchilov2019decoupled,
  title={Decoupled weight decay regularization},
  author={Loshchilov, Ilya and Hutter, Frank},
  journal={arXiv preprint arXiv:1711.05101},
  year={2017}
}

@article{vaswani2017attention,
  title={Attention is all you need},
  author={Vaswani, Ashish and Shazeer, Noam and Parmar, Niki and Uszkoreit, Jakob and Jones, Llion and Gomez, Aidan N and Kaiser, {\L}ukasz and Polosukhin, Illia},
  journal={Advances in neural information processing systems},
  volume={30},
  year={2017}
}

@article{chen2023symbolic,
  title={Symbolic discovery of optimization algorithms},
  author={Chen, Xiangning and Liang, Chen and Huang, Da and Real, Esteban and Wang, Kaiyuan and Pham, Hieu and Dong, Xuanyi and Luong, Thang and Hsieh, Cho-Jui and Lu, Yifeng and others},
  journal={Advances in neural information processing systems},
  volume={36},
  pages={49205--49233},
  year={2023}
}

@inproceedings{gupta2018shampoo,
  title={Shampoo: Preconditioned stochastic tensor optimization},
  author={Gupta, Vineet and Koren, Tomer and Singer, Yoram},
  booktitle={International Conference on Machine Learning},
  pages={1842--1850},
  year={2018},
  organization={PMLR}
}

@article{zhao2024galore,
  title={Galore: Memory-efficient llm training by gradient low-rank projection},
  author={Zhao, Jiawei and Zhang, Zhenyu and Chen, Beidi and Wang, Zhangyang and Anandkumar, Anima and Tian, Yuandong},
  journal={arXiv preprint arXiv:2403.03507},
  year={2024}
}

@article{foret2021sharpness,
  title={Sharpness-aware minimization for efficiently improving generalization},
  author={Foret, Pierre and Kleiner, Ariel and Mobahi, Hossein and Neyshabur, Behnam},
  journal={arXiv preprint arXiv:2010.01412},
  year={2020}
}

@article{duchi2011adaptive,
  title={Adaptive subgradient methods for online learning and stochastic optimization.},
  author={Duchi, John and Hazan, Elad and Singer, Yoram},
  journal={Journal of machine learning research},
  volume={12},
  number={7},
  year={2011}
}

@incollection{tieleman2012rmsprop,
  title={Divide the gradient by a running average of its recent magnitude. coursera: Neural networks for machine learning},
  author={Tieleman, Tijmen and Hinton, Geoffrey},
  booktitle={Technical report},
  year={2017},
  publisher={University of Toronto}
}

@article{amari1998natural,
  title={Natural gradient works efficiently in learning},
  author={Amari, Shun-Ichi},
  journal={Neural computation},
  volume={10},
  number={2},
  pages={251--276},
  year={1998},
  publisher={MIT Press}
}

@inproceedings{liu2023sophia,
  title={Sophia: A scalable stochastic second-order optimizer for language model pre-training},
  author={Liu, Hong and Li, Zhiyuan and Hall, David and Liang, Percy and Ma, Tengyu},
  booktitle={International Conference on Learning Representations},
  volume={2024},
  pages={1621--1650},
  year={2024}
}

@article{zhao2025deconstructing,
  title={Deconstructing what makes a good optimizer for language models, 2025},
  author={Zhao, Rosie and Morwani, Depen and Brandfonbrener, David and Vyas, Nikhil and Kakade, Sham},
  journal={URL https://arxiv. org/abs/2407.07972}
}

@article{semenov2025benchmarking,
  title={Benchmarking optimizers for large language model pretraining},
  author={Semenov, Andrei and Pagliardini, Matteo and Jaggi, Martin},
  journal={arXiv preprint arXiv:2509.01440},
  year={2025}
}

@article{wen2025fantastic,
  title={Fantastic pretraining optimizers and where to find them},
  author={Wen, Kaiyue and Hall, David and Ma, Tengyu and Liang, Percy},
  journal={arXiv preprint arXiv:2509.02046},
  year={2025}
}

@article{schlotthauer2025budget,
  title={Pre-Training LLMs on a budget: A comparison of three optimizers},
  author={Schlotthauer, Joel and Kroos, Christian and Hinze, Chris and Hangya, Viktor and Hahn, Luzian and K{\"u}ch, Fabian},
  journal={arXiv preprint arXiv:2507.08472},
  year={2025}
}

@article{bernstein2024oldoptimizer,
  title={Old optimizer, new norm: An anthology},
  author={Bernstein, Jeremy and Newhouse, Laker},
  journal={arXiv preprint arXiv:2409.20325},
  year={2024}
}

@article{pethick2025normlmo,
  title={Training deep learning models with norm-constrained lmos},
  author={Pethick, Thomas and Xie, Wanyun and Antonakopoulos, Kimon and Zhu, Zhenyu and Silveti-Falls, Antonio and Cevher, Volkan},
  journal={arXiv preprint arXiv:2502.07529},
  year={2025}
}

@article{qu2026muonfinetune,
  title={Can Muon Fine-tune Adam-Pretrained Models?},
  author={Qu, Xingyu and Huang, Peigeng and Horvath, Samuel},
  journal={arXiv preprint arXiv:2605.10468},
  year={2026}
}

@article{shumaylov2026muonnot,
  title={Muon is Not That Special: Random or Inverted Spectra Work Just as Well},
  author={Shumaylov, Zakhar and Da Costa, Natha{\"e}l and Zaika, Peter and Mucs{\'a}nyi, B{\'a}lint and Massucco, Alex and Gelberg, Yoav and Sch{\"o}nlieb, Carola-Bibiane and Gal, Yarin and Hennig, Philipp},
  journal={arXiv preprint arXiv:2605.11181},
  year={2026}
}

@article{zhang2026evolution,
  title={Evolution of Optimization Methods: Algorithms, Scenarios, and Evaluations},
  author={Zhang, Tong and Zhang, Jiangning and Xue, Zhucun and Jiang, Juntao and Xu, Yicheng and Xu, Chengming and Hu, Teng and Xie, Xingyu and Hu, Xiaobin and Wang, Yabiao and others},
  journal={arXiv preprint arXiv:2604.12968},
  year={2026}
}

@article{altinel2025development,
  title={Development of Deep Learning Optimizers: Approaches, Concepts, and Update Rules},
  author={Alt{\i}nel, Do{\u{g}}ay},
  journal={arXiv preprint arXiv:2509.18396},
  year={2025}
}

@article{ranganath2026navigating,
  title={Navigating LLM Valley: From AdamW to Memory-Efficient and Matrix-Based Optimizers},
  author={Ranganath, Aditya},
  journal={arXiv preprint arXiv:2605.09176},
  year={2026}
}

@article{glentis2025scalable,
  title={Scalable parameter and memory efficient pretraining for llm: Recent algorithmic advances and benchmarking},
  author={Glentis, Athanasios and Li, Jiaxiang and Shang, Qiulin and Han, Andi and Tsaknakis, Ioannis and Wei, Quan and Hong, Mingyi},
  journal={arXiv preprint arXiv:2505.22922},
  year={2025}
}

@article{sfyraki2025lions,
  title={Lions and muons: Optimization via stochastic frank-wolfe},
  author={Sfyraki, Maria-Eleni and Wang, Jun-Kun},
  journal={arXiv preprint arXiv:2506.04192},
  year={2025}
}

@article{zhang2024zerothorder,
  title={Revisiting zeroth-order optimization for memory-efficient llm fine-tuning: A benchmark},
  author={Zhang, Yihua and Li, Pingzhi and Hong, Junyuan and Li, Jiaxiang and Zhang, Yimeng and Zheng, Wenqing and Chen, Pin-Yu and Lee, Jason D and Yin, Wotao and Hong, Mingyi and others},
  journal={arXiv preprint arXiv:2402.11592},
  year={2024}
}

@misc{dozat2016nadam,
  title={Incorporating nesterov momentum into adam},
  author={Dozat, Timothy},
  year={2016}
}

@article{liu2020radam,
  title={On the variance of the adaptive learning rate and beyond},
  author={Liu, Liyuan and Jiang, Haoming and He, Pengcheng and Chen, Weizhu and Liu, Xiaodong and Gao, Jianfeng and Han, Jiawei},
  journal={arXiv preprint arXiv:1908.03265},
  year={2019}
}

@article{zhuang2020adabelief,
  title={Adabelief optimizer: Adapting stepsizes by the belief in observed gradients},
  author={Zhuang, Juntang and Tang, Tommy and Ding, Yifan and Tatikonda, Sekhar C and Dvornek, Nicha and Papademetris, Xenophon and Duncan, James},
  journal={Advances in neural information processing systems},
  volume={33},
  pages={18795--18806},
  year={2020}
}

@article{xie2022adan,
  title={Adan: Adaptive nesterov momentum algorithm for faster optimizing deep models},
  author={Xie, Xingyu and Zhou, Pan and Li, Huan and Lin, Zhouchen and Yan, Shuicheng},
  journal={IEEE Transactions on Pattern Analysis and Machine Intelligence},
  volume={46},
  number={12},
  pages={9508--9520},
  year={2024},
  publisher={IEEE}
}

@article{taniguchi2024adopt,
  title={{ADOPT}: Modified {Adam} Can Converge with Any {$\beta_2$} with the Optimal Rate},
  author={Taniguchi, Shohei and Harada, Keno and Minegishi, Gouki and Oshima, Yuta and Jeong, Seong Cheol and Nagahara, Go and Iiyama, Tomoshi and Suzuki, Masahiro and Iwasawa, Yusuke and Matsuo, Yutaka},
  journal={Advances in Neural Information Processing Systems},
  volume={37},
  pages={72438--72474},
  year={2024}
}

@article{ma2019qhadam,
  title={Quasi-hyperbolic momentum and Adam for deep learning},
  author={Ma, Jerry and Yarats, Denis},
  journal={arXiv preprint arXiv:1810.06801},
  year={2018}
}

@article{ding2019adamod,
  title={An adaptive and momental bound method for stochastic learning},
  author={Ding, Jianbang and Ren, Xuancheng and Luo, Ruixuan and Sun, Xu},
  journal={arXiv preprint arXiv:1910.12249},
  year={2019}
}

@article{chen2018padam,
  title={Closing the generalization gap of adaptive gradient methods in training deep neural networks},
  author={Chen, Jinghui and Zhou, Dongruo and Tang, Yiqi and Yang, Ziyan and Cao, Yuan and Gu, Quanquan},
  journal={arXiv preprint arXiv:1806.06763},
  year={2018}
}

@article{luo2019adabound,
  title={Adaptive gradient methods with dynamic bound of learning rate},
  author={Luo, Liangchen and Xiong, Yuanhao and Liu, Yan and Sun, Xu},
  journal={arXiv preprint arXiv:1902.09843},
  year={2019}
}

@inproceeding{pagliardini2024ademamix,
  title={The ademamix optimizer: Better, faster, older},
  author={Pagliardini, Matteo and Ablin, Pierre and Grangier, David},
  booktitle={International Conference on Learning Representations},
  volume={2025},
  pages={64715--64757},
  year={2025}
}

@article{yuan2024mars,
  title={Mars: Unleashing the power of variance reduction for training large models},
  author={Yuan, Huizhuo and Liu, Yifeng and Wu, Shuang and Zhou, Xun and Gu, Quanquan},
  journal={arXiv preprint arXiv:2411.10438},
  year={2024}
}

@article{defazio2022madgrad,
  title={A momentumized, adaptive, dual averaged gradient method},
  author={Defazio, Aaron and Jelassi, Samy},
  journal={Journal of Machine Learning Research},
  volume={23},
  number={144},
  pages={1--34},
  year={2022}
}

@article{lucas2018aggmo,
  title={Aggregated momentum: Stability through passive damping},
  author={Lucas, James and Sun, Shengyang and Zemel, Richard and Grosse, Roger},
  journal={arXiv preprint arXiv:1804.00325},
  year={2018}
}

@article{zhang2019lookahead,
  title={Lookahead optimizer: k steps forward, 1 step back},
  author={Zhang, Michael and Lucas, James and Ba, Jimmy and Hinton, Geoffrey E},
  journal={Advances in neural information processing systems},
  volume={32},
  year={2019}
}

@article{wright2021ranger21,
  title={Ranger21: a synergistic deep learning optimizer},
  author={Wright, Less and Demeure, Nestor},
  journal={arXiv preprint arXiv:2106.13731},
  year={2021}
}

@article{defazio2024road,
  title={The road less scheduled},
  author={Defazio, Aaron and Yang, Xingyu and Mehta, Harsh and Mishchenko, Konstantin and Khaled, Ahmed and Cutkosky, Ashok},
  journal={Advances in Neural Information Processing Systems},
  volume={37},
  pages={9974--10007},
  year={2024}
}

@inproceedings{defazio2023dadaptation,
  title={Learning-rate-free learning by d-adaptation},
  author={Defazio, Aaron and Mishchenko, Konstantin},
  booktitle={International conference on machine learning},
  pages={7449--7479},
  year={2023},
  organization={PMLR}
}

@article{mishchenko2023prodigy,
  title={Prodigy: An expeditiously adaptive parameter-free learner},
  author={Mishchenko, Konstantin and Defazio, Aaron},
  journal={arXiv preprint arXiv:2306.06101},
  year={2023}
}

@inproceeding{ivgi2023dog,
  title={DoG is SGD’s best friend: A parameter-free dynamic step size schedule},
  author={Ivgi, Maor and Hinder, Oliver and Carmon, Yair},
  booktitle={International conference on machine learning},
  pages={14465--14499},
  year={2023},
  organization={PMLR}
}

@article{khaled2023dowg,
  title={Dowg unleashed: An efficient universal parameter-free gradient descent method},
  author={Khaled, Ahmed and Mishchenko, Konstantin and Jin, Chi},
  journal={Advances in Neural Information Processing Systems},
  volume={36},
  pages={6748--6769},
  year={2023}
}

@article{keskar2017swats,
  title={Improving generalization performance by switching from adam to sgd},
  author={Keskar, Nitish Shirish and Socher, Richard},
  journal={arXiv preprint arXiv:1712.07628},
  year={2017}
}

@article{zeiler2012adadelta,
  title={Adadelta: an adaptive learning rate method},
  author={Zeiler, Matthew D},
  journal={arXiv preprint arXiv:1212.5701},
  year={2012}
}

@article{polyak1964some,
  title={Some methods of speeding up the convergence of iteration methods},
  author={Polyak, Boris T},
  journal={Ussr computational mathematics and mathematical physics},
  volume={4},
  number={5},
  pages={1--17},
  year={1964},
  publisher={Elsevier}
}

@inproceedings{sutskever2013importance,
  title={On the importance of initialization and momentum in deep learning},
  author={Sutskever, Ilya and Martens, James and Dahl, George and Hinton, Geoffrey},
  booktitle={International conference on machine learning},
  pages={1139--1147},
  year={2013},
  organization={pmlr}
}

@inproceedings{schaul2013nomore,
  title={No more pesky learning rates},
  author={Schaul, Tom and Zhang, Sixin and LeCun, Yann},
  booktitle={International conference on machine learning},
  pages={343--351},
  year={2013},
  organization={PMLR}
}

@inproceeding{kidambi2018insufficiency,
  title={On the insufficiency of existing momentum schemes for stochastic optimization},
  author={Kidambi, Rahul and Netrapalli, Praneeth and Jain, Prateek and Kakade, Sham},
  booktitle={2018 Information Theory and Applications Workshop (ITA)},
  pages={1--9},
  year={2018},
  organization={IEEE}
}

@article{dubey2019diffgrad,
  title={diffGrad: an optimization method for convolutional neural networks},
  author={Dubey, Shiv Ram and Chakraborty, Soumendu and Roy, Swalpa Kumar and Mukherjee, Snehasis and Singh, Satish Kumar and Chaudhuri, Bidyut Baran},
  journal={IEEE transactions on neural networks and learning systems},
  volume={31},
  number={11},
  pages={4500--4511},
  year={2019},
  publisher={IEEE}
}

@incollection{lu2023adasmooth,
  title={AdaSmooth: an adaptive learning rate method based on effective ratio},
  author={Lu, Jun},
  booktitle={Sentiment Analysis and Deep Learning: Proceedings of ICSADL 2022},
  pages={273--293},
  year={2023},
  publisher={Springer}
}

@article{john2021adamd,
  title={AdamD: Improved bias-correction in Adam},
  author={John, John St},
  journal={arXiv preprint arXiv:2110.10828},
  year={2021}
}

@article{defazio2025gradients,
  title={Why gradients rapidly increase near the end of training},
  author={Defazio, Aaron},
  journal={arXiv preprint arXiv:2506.02285},
  year={2025}
}

@inproceedings{an2018pid,
  title={A PID controller approach for stochastic optimization of deep networks},
  author={An, Wangpeng and Wang, Haoqian and Sun, Qingyun and Xu, Jun and Dai, Qionghai and Zhang, Lei},
  booktitle={Proceedings of the IEEE conference on computer vision and pattern recognition},
  pages={8522--8531},
  year={2018}
}

@article{bahrami2021gravity,
  title={Gravity optimizer: a kinematic approach on optimization in deep learning},
  author={Bahrami, Dariush and Zadeh, Sadegh Pouriyan},
  journal={arXiv preprint arXiv:2101.09192},
  year={2021}
}

@inproceedings{balles2018dissecting,
  title={Dissecting adam: The sign, magnitude and variance of stochastic gradients},
  author={Balles, Lukas and Hennig, Philipp},
  booktitle={International Conference on Machine Learning},
  pages={404--413},
  year={2018},
  organization={PMLR}
}

@article{morwani2025connections,
  title={Connections between schedule-free optimizers, ademamix, and accelerated sgd variants},
  author={Morwani, Depen and Vyas, Nikhil and Zhang, Hanlin and Kakade, Sham},
  journal={arXiv preprint arXiv:2502.02431},
  year={2025}
}

@article{malviya2024torque,
  title={Torque-Aware Momentum},
  author={Malviya, Pranshu and Mordido, Goncalo and Baratin, Aristide and Harikandeh, Reza Babanezhad and Dziugaite, Gintare Karolina and Pascanu, Razvan and Chandar, Sarath},
  journal={arXiv preprint arXiv:2412.18790},
  year={2024}
}

@article{adly2024exadam,
  title={EXAdam: The Power of Adaptive Cross-Moments},
  author={Adly, Ahmed M},
  journal={arXiv preprint arXiv:2412.20302},
  year={2024}
}

@inproceedings{berrada2020training,
  title={Training neural networks for and by interpolation},
  author={Berrada, Leonard and Zisserman, Andrew and Kumar, M Pawan},
  booktitle={International conference on machine learning},
  pages={799--809},
  year={2020},
  organization={PMLR}
}

@article{xu2024no,
  title={No more adam: Learning rate scaling at initialization is all you need},
  author={Xu, Minghao and Xiang, Lichuan and Cai, Xu and Wen, Hongkai},
  journal={arXiv preprint arXiv:2412.11768},
  year={2024}
}

@article{song2026through,
  title={Through the river: Understanding the benefit of schedule-free methods for language model training},
  author={Song, Minhak and Baek, Beomhan and Ahn, Kwangjun and Yun, Chulhee},
  journal={Advances in Neural Information Processing Systems},
  volume={38},
  pages={127524--127555},
  year={2026}
}

@misc{jordan2024muon,
  title={Muon: An Optimizer for Hidden Layers in Neural Networks},
  author={Jordan, Keller and Jin, Yuchen and Boza, Vlado and You, Jiacheng and Cesista, Franz and Newhouse, Laker and Bernstein, Jeremy},
  year={2024},
  howpublished={\url{https://kellerjordan.github.io/posts/muon/}}
}

@article{ahn2025dion,
  title={Dion: Distributed orthonormalized updates},
  author={Ahn, Kwangjun and Xu, Byron and Abreu, Natalie and Fan, Ying and Magakyan, Gagik and Sharma, Pratyusha and Zhan, Zheng and Langford, John},
  journal={arXiv preprint arXiv:2504.05295},
  year={2025}
}

@article{si2025adamuon,
  title={Adamuon: Adaptive muon optimizer},
  author={Si, Chongjie and Zhang, Debing and Shen, Wei},
  journal={arXiv preprint arXiv:2507.11005},
  year={2025}
}

@article{zhang2025adagrad,
  title={Adagrad meets muon: Adaptive stepsizes for orthogonal updates},
  author={Zhang, Minxin and Liu, Yuxuan and Schaeffer, Hayden},
  journal={arXiv preprint arXiv:2509.02981},
  year={2025}
}

@article{hedges2025orthograd,
  title={OrthoGrad Improves Neural Calibration},
  author={Hedges, C Evans},
  journal={arXiv preprint arXiv:2506.04487},
  year={2025}
}

@article{xie2026controlled,
  title={Controlled llm training on spectral sphere},
  author={Xie, Tian and Luo, Haoming and Tang, Haoyu and Hu, Yiwen and Liu, Jason Klein and Ren, Qingnan and Wang, Yang and Zhao, Wayne Xin and Yan, Rui and Su, Bing and others},
  journal={arXiv preprint arXiv:2601.08393},
  year={2026}
}

@inproceedings{vyas2025soap,
  title={SOAP: Improving and stabilizing shampoo using adam for language modeling},
  author={Vyas, Nikhil and Morwani, Depen and Zhao, Rosie and Shapira, Itai and Brandfonbrener, David and Janson, Lucas and Kakade, Sham},
  booktitle={International Conference on Learning Representations},
  volume={2025},
  pages={93423--93444},
  year={2025}
}

@article{liu2025cosmos,
  title={Cosmos: A hybrid adaptive optimizer for memory-efficient training of llms},
  author={Liu, Liming and Xu, Zhenghao and Zhang, Zixuan and Kang, Hao and Li, Zichong and Liang, Chen and Chen, Weizhu and Zhao, Tuo},
  journal={arXiv preprint arXiv:2502.17410},
  year={2025}
}

@article{li2017preconditioned,
  title={Preconditioned stochastic gradient descent},
  author={Li, Xi-Lin},
  journal={IEEE transactions on neural networks and learning systems},
  volume={29},
  number={5},
  pages={1454--1466},
  year={2017},
  publisher={IEEE}
}

@article{li2022black,
  title={Black box lie group preconditioners for sgd},
  author={Li, Xilin},
  journal={arXiv preprint arXiv:2211.04422},
  year={2022}
}

@article{chen2026fira,
  title={Fira: Can we achieve full-rank training of llms under low-rank constraint?},
  author={Chen, Xi and Feng, Kaituo and Li, Changsheng and Lai, Xunhao and Yue, Xiangyu and Yuan, Ye and Wang, Guoren},
  journal={Advances in Neural Information Processing Systems},
  volume={38},
  pages={120680--120712},
  year={2026}
}

@article{gong2025towards,
  title={Towards efficient optimizer design for llm via structured fisher approximation with a low-rank extension},
  author={Gong, Wenbo and Scetbon, Meyer and Ma, Chao and Meeds, Edward},
  journal={arXiv preprint arXiv:2502.07752},
  year={2025}
}

@inproceedings{bernstein2018signsgd,
  title={signSGD: Compressed optimisation for non-convex problems},
  author={Bernstein, Jeremy and Wang, Yu-Xiang and Azizzadenesheli, Kamyar and Anandkumar, Animashree},
  booktitle={International conference on machine learning},
  pages={560--569},
  year={2018},
  organization={PMLR}
}

@inproceeding{rong2024rlion,
  title={RLion: A Refined Lion Optimizer for Deep Learning},
  author={Rong, Jian and Ma, ChenHao and Zhang, QingHui and Cao, Yong},
  year={2024}
}

@article{liu2025focus,
  title={Focus: First order concentrated updating scheme},
  author={Liu, Yizhou and Liu, Ziming and Gore, Jeff},
  journal={arXiv preprint arXiv:2501.12243},
  year={2025}
}

@article{kegreisz2025ano,
  title={ANO: Faster is Better in Noisy Landscape},
  author={Kegreisz, Adrien},
  journal={arXiv preprint arXiv:2508.18258},
  year={2025}
}

@inproceedings{shazeer2018adafactor,
  title={Adafactor: Adaptive learning rates with sublinear memory cost},
  author={Shazeer, Noam and Stern, Mitchell},
  booktitle={International conference on machine learning},
  pages={4596--4604},
  year={2018},
  organization={PMLR}
}

@inproceedings{luo2023came,
  title={Came: Confidence-guided adaptive memory efficient optimization},
  author={Luo, Yang and Ren, Xiaozhe and Zheng, Zangwei and Jiang, Zhuo and Jiang, Xin and You, Yang},
  booktitle={Proceedings of the 61st Annual Meeting of the Association for Computational Linguistics (Volume 1: Long Papers)},
  pages={4442--4453},
  year={2023}
}

@article{dettmers20218,
  title={8-bit optimizers via block-wise quantization},
  author={Dettmers, Tim and Lewis, Mike and Shleifer, Sam and Zettlemoyer, Luke},
  journal={arXiv preprint arXiv:2110.02861},
  year={2021}
}

@article{zhang2024q,
  title={Q-galore: Quantized galore with int4 projection and layer-adaptive low-rank gradients},
  author={Zhang, Zhenyu and Jaiswal, Ajay and Yin, Lu and Liu, Shiwei and Zhao, Jiawei and Tian, Yuandong and Wang, Zhangyang},
  journal={arXiv preprint arXiv:2407.08296},
  year={2024}
}

@inproceedings{zhang2025adam,
  title={Adam-mini: Use fewer learning rates to gain more},
  author={Zhang, Yushun and Chen, Congliang and Li, Ziniu and Ding, Tian and Wu, Chenwei and Kingma, Diederik Durk and Ye, Yinyu and Luo, Zhi-Quan and Sun, Ruoyu},
  booktitle={International Conference on Learning Representations},
  volume={2025},
  pages={28033--28063},
  year={2025}
}

@article{zhu2025apollo,
  title={Apollo: Sgd-like memory, adamw-level performance},
  author={Zhu, Hanqing and Zhang, Zhenyu and Cong, Wenyan and Liu, Xi and Park, Sem and Chandra, Vikas and Long, Bo and Pan, David Z and Wang, Zhangyang and Lee, Jinwon},
  journal={Proceedings of Machine Learning and Systems},
  volume={7},
  year={2025}
}

@inproceedings{wang2025sharpness,
  title={The Sharpness Disparity Principle in Transformers for Accelerating Language Model Pre-Training},
  author={Wang, Jinbo and Wang, Mingze and Zhou, Zhanpeng and Yan, Junchi and Wu, Lei and others},
  booktitle={International Conference on Machine Learning},
  pages={64859--64879},
  year={2025},
  organization={PMLR}
}

@misc{li2026sac,
title={{SAC}: Adaptive Learning Rate Scaling with Architectural Constraints},
author={Siyuan Li and Juanxi Tian and Zedong Wang and Anna Wang and Xin Jin and Chang Yu and Ruoyu Sun and Cheng Tan},
year={2026},
url={https://openreview.net/forum?id=EB92tITeNq}
}

@article{li2025taming,
  title={Taming LLMs by Scaling Learning Rates with Gradient Grouping},
  author={Li, Siyuan and Tian, Juanxi and Wang, Zedong and Jin, Xin and Liu, Zicheng and Zhang, Wentao and Xu, Dan},
  journal={arXiv preprint arXiv:2506.01049},
  year={2025}
}

@article{anil2019memory,
  title={Memory efficient adaptive optimization},
  author={Anil, Rohan and Gupta, Vineet and Koren, Tomer and Singer, Yoram},
  journal={Advances in Neural Information Processing Systems},
  volume={32},
  year={2019}
}

@article{wang2025conda,
  title={Conda: Column-Normalized Adam for Training Large Language Models Faster},
  author={Wang, Junjie and Zhou, Pan and Dong, Yiming and Li, Huan and Li, Jia and Zhou, Xun and Lao, Qicheng and Fang, Cong and Lin, Zhouchen},
  journal={arXiv preprint arXiv:2509.24218},
  year={2025}
}

@article{ginsburg2019stochastic,
  title={Stochastic gradient methods with layer-wise adaptive moments for training of deep networks},
  author={Ginsburg, Boris and Castonguay, Patrice and Hrinchuk, Oleksii and Kuchaiev, Oleksii and Lavrukhin, Vitaly and Leary, Ryan and Li, Jason and Nguyen, Huyen and Zhang, Yang and Cohen, Jonathan M},
  journal={arXiv preprint arXiv:1905.11286},
  year={2019}
}

@inproceedings{lv2024full,
  title={Full parameter fine-tuning for large language models with limited resources},
  author={Lv, Kai and Yang, Yuqing and Liu, Tengxiao and Guo, Qipeng and Qiu, Xipeng},
  booktitle={Proceedings of the 62nd Annual Meeting of the Association for Computational Linguistics (Volume 1: Long Papers)},
  pages={8187--8198},
  year={2024}
}

@inproceedings{lv2024adalomo,
  title={Adalomo: Low-memory optimization with adaptive learning rate},
  author={Lv, Kai and Yan, Hang and Guo, Qipeng and Lv, Haijun and Qiu, Xipeng},
  booktitle={Findings of the Association for Computational Linguistics: ACL 2024},
  pages={12486--12502},
  year={2024}
}

@inproceedings{kwon2021asam,
  title={Asam: Adaptive sharpness-aware minimization for scale-invariant learning of deep neural networks},
  author={Kwon, Jungmin and Kim, Jeongseop and Park, Hyunseo and Choi, In Kwon},
  booktitle={International conference on machine learning},
  pages={5905--5914},
  year={2021},
  organization={PMLR}
}

@article{zhuang2022surrogate,
  title={Surrogate gap minimization improves sharpness-aware training},
  author={Zhuang, Juntang and Gong, Boqing and Yuan, Liangzhe and Cui, Yin and Adam, Hartwig and Dvornek, Nicha and Tatikonda, Sekhar and Duncan, James and Liu, Ting},
  journal={arXiv preprint arXiv:2203.08065},
  year={2022}
}

@inproceedings{yue2023sharpness,
  title={Sharpness-aware minimization revisited: Weighted sharpness as a regularization term},
  author={Yue, Yun and Jiang, Jiadi and Ye, Zhiling and Gao, Ning and Liu, Yongchao and Zhang, Ke},
  booktitle={Proceedings of the 29th ACM SIGKDD Conference on Knowledge Discovery and Data Mining},
  pages={3185--3194},
  year={2023}
}

@article{mollenhoff2022sam,
  title={Sam as an optimal relaxation of bayes},
  author={M{\"o}llenhoff, Thomas and Khan, Mohammad Emtiyaz},
  journal={arXiv preprint arXiv:2210.01620},
  year={2022}
}

@inproceedings{liu2022towards,
  title={Towards efficient and scalable sharpness-aware minimization},
  author={Liu, Yong and Mai, Siqi and Chen, Xiangning and Hsieh, Cho-Jui and You, Yang},
  booktitle={Proceedings of the IEEE/CVF Conference on Computer Vision and Pattern Recognition},
  pages={12360--12370},
  year={2022}
}

@inproceedings{li2024friendly,
  title={Friendly sharpness-aware minimization},
  author={Li, Tao and Zhou, Pan and He, Zhengbao and Cheng, Xinwen and Huang, Xiaolin},
  booktitle={Proceedings of the IEEE/CVF conference on computer vision and pattern recognition},
  pages={5631--5640},
  year={2024}
}

@inproceedings{yao2021adahessian,
  title={Adahessian: An adaptive second order optimizer for machine learning},
  author={Yao, Zhewei and Gholami, Amir and Shen, Sheng and Mustafa, Mustafa and Keutzer, Kurt and Mahoney, Michael},
  booktitle={proceedings of the AAAI conference on artificial intelligence},
  volume={35},
  number={12},
  pages={10665--10673},
  year={2021}
}

@inproceedings{yong2020gradient,
  title={Gradient centralization: A new optimization technique for deep neural networks},
  author={Yong, Hongwei and Huang, Jianqiang and Hua, Xiansheng and Zhang, Lei},
  booktitle={European Conference on Computer Vision},
  pages={635--652},
  year={2020},
  organization={Springer}
}

@article{heo2020adamp,
  title={Adamp: Slowing down the slowdown for momentum optimizers on scale-invariant weights},
  author={Heo, Byeongho and Chun, Sanghyuk and Oh, Seong Joon and Han, Dongyoon and Yun, Sangdoo and Kim, Gyuwan and Uh, Youngjung and Ha, Jung-Woo},
  journal={arXiv preprint arXiv:2006.08217},
  year={2020}
}

@article{wortsman2023stable,
  title={Stable and low-precision training for large-scale vision-language models},
  author={Wortsman, Mitchell and Dettmers, Tim and Zettlemoyer, Luke and Morcos, Ari and Farhadi, Ali and Schmidt, Ludwig},
  journal={Advances in Neural Information Processing Systems},
  volume={36},
  pages={10271--10298},
  year={2023}
}

@article{cao2024grams,
  title={Grams: Gradient descent with adaptive momentum scaling},
  author={Cao, Yang and Li, Xiaoyu and Song, Zhao},
  journal={arXiv preprint arXiv:2412.17107},
  year={2024}
}

@article{liang2024cautious,
  title={Cautious optimizers: Improving training with one line of code},
  author={Liang, Kaizhao and Chen, Lizhang and Liu, Bo and Liu, Qiang},
  journal={arXiv preprint arXiv:2411.16085},
  year={2024}
}

@article{huang2025spam,
  title={SPAM: Spike-aware adam with momentum reset for stable LLM training},
  author={Huang, Tianjin and Zhu, Ziquan and Jin, Gaojie and Liu, Lu and Wang, Zhangyang and Liu, Shiwei},
  journal={arXiv preprint arXiv:2501.06842},
  year={2025}
}

@inproceeding{huanggradientstabilizer,
  title={GradientStabilizer: Fix the Norm, Not the Gradient},
  author={Huang, Tianjin and Wang, Zhangyang and Hu, Haotian and Zhang, Zhenyu and Jin, Gaojie and Li, Xiang and Shen, Li and Shang, Jiaxing and Chen, Tianlong and Li, Ke and others},
  booktitle={Forty-third International Conference on Machine Learning}
}

@article{joo2026surprising,
  title={On Surprising Effectiveness of Masking Updates in Adaptive Optimizers},
  author={Joo, Taejong and Xia, Wenhan and Kim, Cheolmin and Zhang, Ming and Ie, Eugene},
  journal={arXiv preprint arXiv:2602.15322},
  year={2026}
}

@article{chang2026mgup,
  title={MGUP: A momentum-gradient alignment update policy for stochastic optimization},
  author={Chang, Da and Yuan, Ganzhao},
  journal={Advances in Neural Information Processing Systems},
  volume={38},
  pages={20488--20537},
  year={2026}
}

@article{you2019large,
  title={Large batch optimization for deep learning: Training bert in 76 minutes},
  author={You, Yang and Li, Jing and Reddi, Sashank and Hseu, Jonathan and Kumar, Sanjiv and Bhojanapalli, Srinadh and Song, Xiaodan and Demmel, James and Keutzer, Kurt and Hsieh, Cho-Jui},
  journal={arXiv preprint arXiv:1904.00962},
  year={2019}
}

@inproceedings{brock2021high,
  title={High-performance large-scale image recognition without normalization},
  author={Brock, Andy and De, Soham and Smith, Samuel L and Simonyan, Karen},
  booktitle={International conference on machine learning},
  pages={1059--1071},
  year={2021},
  organization={PMLR}
}

@article{li2024unveiling,
  title={Unveiling the backbone-optimizer coupling bias in visual representation learning},
  author={Li, Siyuan and Tian, Juanxi and Wang, Zedong and Zhang, Luyuan and Liu, Zicheng and Jin, Weiyang and Liu, Yang and Sun, Baigui and Li, Stan Z},
  journal={arXiv preprint arXiv:2410.06373},
  year={2024}
}

@article{bae2026affine,
  title={Affine-Scaled Attention: Towards Flexible and Stable Transformer Attention},
  author={Bae, Jeongin and Park, Baeseong and Park, Gunho and Kim, Minsub and Lee, Joonhyung and Yoo, Junhee and Woo, Sunghyeon and Ryu, Jiwon and Kwon, Se Jung and Lee, Dongsoo},
  journal={arXiv preprint arXiv:2602.23057},
  year={2026}
}

@article{wang2025adagc,
  title={Adagc: Improving training stability for large language model pretraining},
  author={Wang, Guoxia and Li, Shuai and Chen, Congliang and Zeng, Jinle and Yang, Jiabin and Yu, Dianhai and Ma, Yanjun and Shen, Li},
  journal={arXiv preprint arXiv:2502.11034},
  year={2025}
}

@article{kumar2025zclip,
  title={Zclip: Adaptive spike mitigation for llm pre-training},
  author={Kumar, Abhay and Owen, Louis and Chowdhury, Nilabhra Roy and G{\"u}ra, Fabian},
  journal={arXiv preprint arXiv:2504.02507},
  year={2025}
}

@article{deng2026rmnp,
  title={RMNP: Row-momentum normalized preconditioning for scalable matrix-based optimization},
  author={Deng, Shenyang and Ouyang, Zhuoli and Pang, Tianyu and Liu, Zihang and Jin, Ruochen and Yu, Shuhua and Yang, Yaoqing},
  journal={arXiv preprint arXiv:2603.20527},
  year={2026}
}

@article{xu2026width,
  title={On the width scaling of neural optimizers under matrix operator norms i: Row/column normalization and hyperparameter transfer},
  author={Xu, Ruihan and Li, Jiajin and Lu, Yiping},
  journal={arXiv preprint arXiv:2603.09952},
  year={2026}
}

@article{frans2026stable,
  title={A stable whitening optimizer for efficient neural network training},
  author={Frans, Kevin and Levine, Sergey and Abbeel, Pieter},
  journal={Advances in Neural Information Processing Systems},
  volume={38},
  pages={174086--174110},
  year={2026}
}

@inproceedings{clark2019boolq,
  title = {{BoolQ}: Exploring the Surprising Difficulty of Natural Yes/No Questions},
  author = {Clark, Christopher and Lee, Kenton and Chang, Ming-Wei and Kwiatkowski, Tom and Collins, Michael and Toutanova, Kristina},
  booktitle = {Proceedings of the 2019 Conference of the North American Chapter of the Association for Computational Linguistics: Human Language Technologies},
  pages = {2924--2936},
  year = {2019},
  publisher = {Association for Computational Linguistics},
  doi = {10.18653/v1/N19-1300}
}

@inproceedings{bisk2020piqa,
  title = {{PIQA}: Reasoning about Physical Commonsense in Natural Language},
  author = {Bisk, Yonatan and Zellers, Rowan and Le Bras, Ronan and Gao, Jianfeng and Choi, Yejin},
  booktitle = {Proceedings of the AAAI Conference on Artificial Intelligence},
  volume = {34},
  number = {05},
  pages = {7432--7439},
  year = {2020},
  doi = {10.1609/aaai.v34i05.6239}
}

@inproceedings{zellers2019hellaswag,
  title = {{HellaSwag}: Can a Machine Really Finish Your Sentence?},
  author = {Zellers, Rowan and Holtzman, Ari and Bisk, Yonatan and Farhadi, Ali and Choi, Yejin},
  booktitle = {Proceedings of the 57th Annual Meeting of the Association for Computational Linguistics},
  pages = {4791--4800},
  year = {2019},
  publisher = {Association for Computational Linguistics},
  doi = {10.18653/v1/P19-1472}
}

@article{sakaguchi2021winogrande,
  title={Winogrande: An adversarial winograd schema challenge at scale},
  author={Sakaguchi, Keisuke and Bras, Ronan Le and Bhagavatula, Chandra and Choi, Yejin},
  journal={Communications of the ACM},
  volume={64},
  number={9},
  pages={99--106},
  year={2021},
  publisher={ACM New York, NY, USA}
}

@article{clark2018think,
  title={Think you have solved question answering? try arc, the ai2 reasoning challenge},
  author={Clark, Peter and Cowhey, Isaac and Etzioni, Oren and Khot, Tushar and Sabharwal, Ashish and Schoenick, Carissa and Tafjord, Oyvind},
  journal={arXiv preprint arXiv:1803.05457},
  year={2018}
}

@inproceedings{mihaylov2018can,
  title={Can a suit of armor conduct electricity? a new dataset for open book question answering},
  author={Mihaylov, Todor and Clark, Peter and Khot, Tushar and Sabharwal, Ashish},
  booktitle={Proceedings of the 2018 conference on empirical methods in natural language processing},
  pages={2381--2391},
  year={2018}
}

@misc{eval-harness,
  author       = {Gao, Leo and Tow, Jonathan and Abbasi, Baber and Biderman, Stella and Black, Sid and DiPofi, Anthony and Foster, Charles and Golding, Laurence and Hsu, Jeffrey and Le Noac'h, Alain and Li, Haonan and McDonell, Kyle and Muennighoff, Niklas and Ociepa, Chris and Phang, Jason and Reynolds, Laria and Schoelkopf, Hailey and Skowron, Aviya and Sutawika, Lintang and Tang, Eric and Thite, Anish and Wang, Ben and Wang, Kevin and Zou, Andy},
  title        = {The Language Model Evaluation Harness},
  month        = 07,
  year         = 2024,
  publisher    = {Zenodo},
  version      = {v0.4.3},
  doi          = {10.5281/zenodo.12608602},
  url          = {https://zenodo.org/records/12608602}
}

@article{hassan2025gcsam,
  title={GCSAM: Gradient Centralized Sharpness Aware Minimization},
  author={Hassan, Mohamed and Vakanski, Aleksandar and Zhang, Boyu and Xian, Min},
  journal={IEEE Access},
  year={2025},
  publisher={IEEE}
}

@article{raffel2020exploring,
  title={Exploring the limits of transfer learning with a unified text-to-text transformer},
  author={Raffel, Colin and Shazeer, Noam and Roberts, Adam and Lee, Katherine and Narang, Sharan and Matena, Michael and Zhou, Yanqi and Li, Wei and Liu, Peter J},
  journal={Journal of machine learning research},
  volume={21},
  number={140},
  pages={1--67},
  year={2020}
}

@misc{penedo2024finewebdatasetsdecantingweb,
      title={The FineWeb Datasets: Decanting the Web for the Finest Text Data at Scale},
      author={Guilherme Penedo and Hynek Kydlíček and Loubna Ben allal and Anton Lozhkov and Margaret Mitchell and Colin Raffel and Leandro Von Werra and Thomas Wolf},
      year={2024},
      eprint={2406.17557},
      archivePrefix={arXiv},
      primaryClass={cs.CL},
      url={https://arxiv.org/abs/2406.17557},
}

@misc{karpathy2024finewebedu100b,
  title={FineWeb-Edu-100B-Shuffle},
  author={Karpathy, Andrej},
  year={2024},
  howpublished={\url{https://huggingface.co/datasets/karpathy/fineweb-edu-100b-shuffle}},
}

@inproceedings{roemmele2011choice,
  title={Choice of plausible alternatives: An evaluation of commonsense causal reasoning},
  author={Roemmele, Melissa and Bejan, Cosmin Adrian and Gordon, Andrew S},
  booktitle={2011 AAAI Spring Symposium Series},
  year={2011},
  url={https://people.ict.usc.edu/~gordon/publications/AAAI-SPRING11A.PDF},
}

@inproceedings{yang2025gated,
  title={Gated delta networks: Improving mamba2 with delta rule},
  author={Yang, Songlin and Kautz, Jan and Hatamizadeh, Ali},
  booktitle={International Conference on Learning Representations},
  volume={2025},
  pages={29687--29707},
  year={2025}
}

@article{yang2024parallelizing,
  title={Parallelizing linear transformers with the delta rule over sequence length},
  author={Yang, Songlin and Wang, Bailin and Zhang, Yu and Shen, Yikang and Kim, Yoon},
  journal={Advances in neural information processing systems},
  volume={37},
  pages={115491--115522},
  year={2024}
}

@article{merity2016pointer,
  title={Pointer sentinel mixture models},
  author={Merity, Stephen and Xiong, Caiming and Bradbury, James and Socher, Richard},
  journal={arXiv preprint arXiv:1609.07843},
  year={2016}
}

@article{yang2023gated,
  title={Gated linear attention transformers with hardware-efficient training},
  author={Yang, Songlin and Wang, Bailin and Shen, Yikang and Panda, Rameswar and Kim, Yoon},
  journal={arXiv preprint arXiv:2312.06635},
  year={2023}
}

@InProceedings{paperno-EtAl:2016:P16-1,
  author    = {Paperno, Denis  and  Kruszewski, Germ\'{a}n  and  Lazaridou,
Angeliki  and  Pham, Ngoc Quan  and  Bernardi, Raffaella  and  Pezzelle,
Sandro  and  Baroni, Marco  and  Boleda, Gemma  and  Fernandez, Raquel},
  title     = {The {LAMBADA} dataset: Word prediction requiring a broad
discourse context},
  booktitle = {Proceedings of the 54th Annual Meeting of the Association for
Computational Linguistics (Volume 1: Long Papers)},
  month     = {August},
  year      = {2016},
  address   = {Berlin, Germany},
  publisher = {Association for Computational Linguistics},
  pages     = {1525--1534},
  url       = {http://www.aclweb.org/anthology/P16-1144}
}

@inproceedings{welbl2017crowdsourcing,
  title={Crowdsourcing multiple choice science questions},
  author={Welbl, Johannes and Liu, Nelson F and Gardner, Matt},
  booktitle={Proceedings of the 3rd Workshop on Noisy User-generated Text},
  pages={94--106},
  year={2017}
}

@inproceeding{krizhevsky2009learning,
  title={Learning multiple layers of features from tiny images},
  author={Krizhevsky, Alex and Hinton, Geoffrey and others},
  year={2009},
  publisher={Toronto, ON, Canada}
}

@inproceedings{he2016deep,
  title={Deep residual learning for image recognition},
  author={He, Kaiming and Zhang, Xiangyu and Ren, Shaoqing and Sun, Jian},
  booktitle={Proceedings of the IEEE conference on computer vision and pattern recognition},
  pages={770--778},
  year={2016}
}

@inproceedings{touvron2021training,
  title={Training data-efficient image transformers \& distillation through attention},
  author={Touvron, Hugo and Cord, Matthieu and Douze, Matthijs and Massa, Francisco and Sablayrolles, Alexandre and J{\'e}gou, Herv{\'e}},
  booktitle={International conference on machine learning},
  pages={10347--10357},
  year={2021},
  organization={PMLR}
}

@article{yu2023metaformer,
  title={Metaformer baselines for vision},
  author={Yu, Weihao and Si, Chenyang and Zhou, Pan and Luo, Mi and Zhou, Yichen and Feng, Jiashi and Yan, Shuicheng and Wang, Xinchao},
  journal={IEEE Transactions on Pattern Analysis and Machine Intelligence},
  volume={46},
  number={2},
  pages={896--912},
  year={2023},
  publisher={IEEE}
}

@inproceedings{yu2022metaformer,
  title={Metaformer is actually what you need for vision},
  author={Yu, Weihao and Luo, Mi and Zhou, Pan and Si, Chenyang and Zhou, Yichen and Wang, Xinchao and Feng, Jiashi and Yan, Shuicheng},
  booktitle={Proceedings of the IEEE/CVF conference on computer vision and pattern recognition},
  pages={10819--10829},
  year={2022}
}

\clearpage
\newpage

\beginappendix

\section{Additional Experimental Hyperparameter Configurations}
\label{app:hparam-configs}

To improve the reproducibility of the benchmark results, this appendix reports
the main optimizer hyperparameters used in the experiments. The main text
reports the resulting perplexity, downstream performance, per-step optimizer
runtime, and optimizer-state memory. In contrast, this appendix focuses on the
training configurations, including learning rates, momentum coefficients,
numerical stability constants, and method-specific auxiliary parameters.

\subsection{Stage-1 C4-LLaMA Short-Context Screening}
\label{app:stage1-hparams}

Table~\ref{tab:appendix_c4_screen_hparams} summarizes the tuned hyperparameter
configurations used in the Stage-1 C4-LLaMA short-context screening
experiments. For each optimizer, we report the momentum coefficients,
numerical stability constant, and the selected learning rate at each model
scale. Method-specific auxiliary settings, such as projection rank and
projection interval, are summarized in the table note.

\begin{table}[!htbp]
\centering
\caption{Hyperparameter configurations for the C4-LLaMA short-context Stage-1 screening experiments.}
\label{tab:appendix_c4_screen_hparams}
\scriptsize
\setlength{\tabcolsep}{0.8em}
\renewcommand{\arraystretch}{1.2}
\resizebox{\linewidth}{!}{
\begin{tabular}{lcccccccc}
\toprule
\textbf{Optimizer}
& $\boldsymbol{\beta_1}$
& $\boldsymbol{\beta_2}$
& $\boldsymbol{\beta_3}$
& $\boldsymbol{\epsilon}$
& \textbf{60M lr}
& \textbf{130M lr}
& \textbf{350M lr}
& \textbf{1B lr} \\
\midrule
\rowcolor{famT1!18}\multicolumn{9}{c}{\textbf{\textcolor{famT1!62!black}{\textit{T1: Element-wise adaptive moment and scalar control}}}} \\
Adan         & 0.9  & 0.92  & 0.99 & $1\times10^{-8}$  & 0.003  & 0.003  & 0.003  & 0.001  \\
RAdam        & 0.9  & 0.99  & $-$   & $1\times10^{-8}$  & 0.003  & 0.001  & 0.001  & 0.0005 \\
AdamW        & 0.9  & 0.99  & $-$   & $1\times10^{-8}$  & 0.003  & 0.001  & 0.001  & 0.0005 \\
NAdam        & 0.9  & 0.99  & $-$   & $1\times10^{-8}$  & 0.001  & 0.001  & 0.001  & 0.0005 \\
MARS-AdamW   & 0.95 & 0.99  & $-$   & $1\times10^{-8}$  & 0.005  & 0.005  & 0.005  & 0.001  \\
Prodigy      & 0.9  & 0.95  & $-$   & $1\times10^{-8}$  & 0.5    & 1      & 2      & 2      \\
AdaBelief    & 0.9  & 0.999 & $-$   & $1\times10^{-16}$/$10^{-12}$ & 0.003 & 0.001 & 0.001 & 0.001 \\
\midrule
\rowcolor{famT2!18}\multicolumn{9}{c}{\textbf{\textcolor{famT2!62!black}{\textit{T2: Matrix-level structural methods}}}} \\
MARS-Shampoo & 0.95 & 0.99  & $-$   & $1\times10^{-8}$  & 0.05   & 0.03   & 0.02   & 0.01   \\
Muon         & 0.9  & 0.95  & $-$   & $1\times10^{-8}$  & 0.02   & 0.01   & 0.006  & 0.006  \\
RMNP         & 0.9/0.95 & 0.95 & $-$ & $1\times10^{-8}$  & 0.01   & 0.01   & 0.01   & 0.01   \\
SOAP         & 0.9  & 0.95  & $-$   & $1\times10^{-8}$  & 0.003  & 0.002  & 0.001  & 0.0005 \\
GaLore       & 0.9  & 0.98  & $-$   & $1\times10^{-6}$  & 0.03   & 0.03   & 0.01   & 0.01   \\
Shampoo      & 0.9  & 0.999 & $-$   & $1\times10^{-8}$  & 0.05   & 0.05   & 0.02   & 0.01   \\
\midrule
\rowcolor{famT3!18}\multicolumn{9}{c}{\textbf{\textcolor{famT3!62!black}{\textit{T3: Discretization and directional quantization}}}} \\
MARS-Lion    & 0.9  & 0.98  & $-$   & $1\times10^{-8}$  & 0.0005 & 0.0002 & 0.0002 & 0.0002 \\
Lion         & 0.9  & 0.98  & $-$   & $-$                 & 0.0002 & 0.0002 & 0.0002 & 0.0001 \\
\midrule
\rowcolor{famT4!18}\multicolumn{9}{c}{\textbf{\textcolor{famT4!62!black}{\textit{T4: State compression and structural aggregation}}}} \\
APOLLO       & 0.9  & 0.99  & $-$   & $1\times10^{-6}$  & 0.02   & 0.01   & 0.01   & 0.01   \\
Conda        & 0.9  & 0.99  & $-$   & $1\times10^{-8}$  & 0.01   & 0.01   & 0.01   & 0.0005 \\
8-bit Adam     & 0.9  & 0.99  & $-$   & $1\times10^{-8}$  & 0.003  & 0.001  & 0.0005 & 0.0005 \\
CAME         & 0.9  & 0.999 & $-$   & $1\times10^{-6}$  & 0.001  & 0.0005 & 0.0005 & 0.0003 \\
AdaFactor    & 0.9  & $-$    & $-$   & $-$                 & 0.002  & 0.002  & 0.001  & 0.0005 \\
Adam-mini    & 0.9  & 0.99  & $-$   & $1\times10^{-8}$  & 0.003  & 0.001  & 0.0005 & 0.0003 \\
\midrule
\rowcolor{famT5!18}\multicolumn{9}{c}{\textbf{\textcolor{famT5!62!black}{\textit{T5: Curvature-aware and geometric regularization}}}} \\
AdamP        & 0.9  & 0.98  & $-$   & $1\times10^{-8}$  & 0.005  & 0.001  & 0.001  & 0.0005 \\
LAMB         & 0.9  & 0.99  & $-$   & $1\times10^{-6}$  & 0.005  & 0.003  & 0.001  & 0.001  \\
Sophia       & 0.9  & 0.99  & $-$   & $1\times10^{-8}$  & 0.0002 & 0.0002 & 0.0002 & 0.0001 \\
\bottomrule
\end{tabular}}

\vspace{2pt}
{\fontsize{8pt}{8pt}\selectfont
\raggedright
\textbf{Note:} $-$ indicates that the corresponding hyperparameter is not used or that no explicit numerical stability constant is set. The learning rates are the final configurations used for each optimizer and model scale in the C4-LLaMA short-context Stage-1 screening experiments. AdaBelief uses $\epsilon=1\times10^{-16}$ for 60M, 130M, and 350M, and $\epsilon=1\times10^{-12}$ for 1B. RMNP uses $\beta_1=\beta_2=0.95$ at 1B and applies an Adam-style auxiliary branch to non-matrix parameters. The projection ranks of APOLLO are 128/192/256/256, with projection interval 200. The ranks of Conda are 128/192/256/256, with projection intervals 2000/2000/2000/200. The ranks of GaLore are 128/256/256/1024, with projection intervals 200/500/500/200.
\par
}
\end{table}

\subsection{Stage-2 FineWeb-Edu 32k Long-Context Experiments}
\label{app:stage2-fwe-hparams}

Tables~\ref{tab:appendix_fwe_stage2_combined_expanded} report the best-run hyperparameter
configurations for the Stage-2 FineWeb-Edu 32k experiments at 340M and 1B,
respectively. The tables list the learning rate, momentum coefficients,
numerical stability constant, and optimizer-specific auxiliary parameters
encoded in the log filenames.

\begin{table}[!htbp]
\centering
\caption{Hyperparameter configurations for the FineWeb-Edu 32k Stage-2 experiments at model parameters of 340M and 1B.}
\scriptsize
\setlength{\tabcolsep}{1.2em}
\renewcommand{\arraystretch}{1.03}
\resizebox{1.\linewidth}{!}{
\begin{tabular}{l ccccc | ccccc}
\toprule
\multirow{2}{*}{\textbf{Optimizer}}
& \multicolumn{5}{c}{\textbf{340M}} & \multicolumn{5}{c}{\textbf{1B}} \\
\cmidrule(r){2-6} \cmidrule(l){7-11}
& \textbf{lr} & $\boldsymbol{\beta_1}$ & $\boldsymbol{\beta_2}$ & $\boldsymbol{\beta_3}$ & $\boldsymbol{\epsilon}$
& \textbf{lr} & $\boldsymbol{\beta_1}$ & $\boldsymbol{\beta_2}$ & $\boldsymbol{\beta_3}$ & $\boldsymbol{\epsilon}$ \\
\midrule
\rowcolor{basebg}
\multicolumn{11}{l}{\textit{Transformer++}} \\
AdamW        & 1e-3 & 0.9  & 0.99 & $-$   & 1e-15 & 1e-3 & 0.9  & 0.99 & $-$   & 1e-15 \\
AdamP        & 1e-3 & 0.9  & 0.98 & $-$   & 1e-15 & 1e-3 & 0.9  & 0.98 & $-$   & 1e-15 \\
Adan         & 3e-3 & 0.9  & 0.92 & 0.99 & 1e-8  & 3e-3 & 0.9  & 0.92 & 0.99 & 1e-8  \\
Lion         & 3e-4 & 0.9  & 0.99 & $-$   & $-$    & 3e-4 & 0.9  & 0.99 & $-$   & $-$    \\
MARS-AdamW   & 3e-3 & 0.95 & 0.99 & $-$   & 1e-8  & 3e-3 & 0.95 & 0.99 & $-$   & 1e-8  \\
MARS-Lion    & 2e-4 & 0.9  & 0.98 & $-$   & 1e-12 & 1e-4 & 0.9  & 0.98 & $-$   & 1e-12 \\
MARS-Shampoo & 1e-2 & 0.95 & 0.99 & $-$   & 1e-12 & 5e-3 & 0.95 & 0.99 & $-$   & 1e-12 \\
Muon         & 5e-3 & 0.9  & 0.95 & $-$   & 1e-15 & 3e-3 & 0.9  & 0.95 & $-$   & 1e-15 \\
RMNP         & 3e-3 & 0.9  & 0.99 & $-$   & 1e-15 & 3e-3 & 0.9  & 0.99 & $-$   & 1e-15 \\
SOAP         & 3e-3 & 0.9  & 0.95 & $-$   & 1e-15 & 3e-3 & 0.9  & 0.95 & $-$   & 1e-15 \\
APOLLO       & 3e-3 & 0.9  & 0.99 & $-$   & 1e-12 & 3e-3 & 0.9  & 0.99 & $-$   & 1e-12 \\
Conda        & 1e-2 & 0.9  & 0.99 & $-$   & 1e-12 & 5e-3 & 0.9  & 0.99 & $-$   & 1e-12 \\
\midrule
\rowcolor{basebg}
\multicolumn{11}{l}{\textit{GLA}} \\
AdamW        & 1e-3 & 0.9  & 0.99 & $-$   & 1e-15 & 1e-3 & 0.9  & 0.99 & $-$   & 1e-15 \\
AdamP        & 1e-3 & 0.9  & 0.98 & $-$   & 1e-15 & 1e-3 & 0.9  & 0.98 & $-$   & 1e-15 \\
Adan         & 3e-3 & 0.9  & 0.92 & 0.99 & 1e-8  & 3e-3 & 0.9  & 0.92 & 0.99 & 1e-8  \\
Lion         & 3e-4 & 0.9  & 0.99 & $-$   & $-$    & 3e-4 & 0.9  & 0.99 & $-$   & $-$    \\
MARS-AdamW   & 3e-3 & 0.95 & 0.99 & $-$   & 1e-8  & 3e-3 & 0.95 & 0.99 & $-$   & 1e-8  \\
MARS-Lion    & 2e-4 & 0.9  & 0.98 & $-$   & 1e-12 & 1e-4 & 0.9  & 0.98 & $-$   & 1e-12 \\
MARS-Shampoo & 1e-2 & 0.95 & 0.99 & $-$   & 1e-12 & 5e-3 & 0.95 & 0.99 & $-$   & 1e-12 \\
Muon         & 5e-3 & 0.9  & 0.95 & $-$   & 1e-15 & 3e-3 & 0.9  & 0.95 & $-$   & 1e-15 \\
RMNP         & 3e-3 & 0.9  & 0.99 & $-$   & 1e-15 & 3e-3 & 0.9  & 0.99 & $-$   & 1e-15 \\
SOAP         & 3e-3 & 0.9  & 0.95 & $-$   & 1e-15 & 3e-3 & 0.9  & 0.95 & $-$   & 1e-15 \\
APOLLO       & 3e-3 & 0.9  & 0.99 & $-$   & 1e-12 & 3e-3 & 0.9  & 0.99 & $-$   & 1e-12 \\
Conda        & 1e-2 & 0.9  & 0.99 & $-$   & 1e-12 & 5e-3 & 0.9  & 0.99 & $-$   & 1e-12 \\
\midrule
\rowcolor{basebg}
\multicolumn{11}{l}{\textit{DeltaNet}} \\
AdamW        & 1e-3 & 0.9  & 0.99 & $-$   & 1e-15 & 1e-3 & 0.9  & 0.99 & $-$   & 1e-15 \\
AdamP        & 1e-3 & 0.9  & 0.98 & $-$   & 1e-15 & 1e-3 & 0.9  & 0.98 & $-$   & 1e-15 \\
Adan         & 3e-3 & 0.9  & 0.92 & 0.99 & 1e-8  & 3e-3 & 0.9  & 0.92 & 0.99 & 1e-8  \\
Lion         & 3e-3 & 0.9  & 0.99 & $-$   & $-$    & 3e-3 & 0.9  & 0.99 & $-$   & $-$    \\
MARS-AdamW   & 3e-3 & 0.95 & 0.99 & $-$   & 1e-8  & 3e-3 & 0.95 & 0.99 & $-$   & 1e-8  \\
MARS-Lion    & 2e-4 & 0.9  & 0.98 & $-$   & 1e-12 & 1e-4 & 0.9  & 0.98 & $-$   & 1e-12 \\
MARS-Shampoo & 1e-2 & 0.95 & 0.99 & $-$   & 1e-12 & 5e-3 & 0.95 & 0.99 & $-$   & 1e-12 \\
Muon         & 5e-3 & 0.9  & 0.95 & $-$   & 1e-15 & 3e-3 & 0.9  & 0.95 & $-$   & 1e-15 \\
RMNP         & 3e-3 & 0.9  & 0.99 & $-$   & 1e-15 & 3e-3 & 0.9  & 0.99 & $-$   & 1e-15 \\
SOAP         & 3e-3 & 0.9  & 0.95 & $-$   & 1e-15 & 3e-3 & 0.9  & 0.95 & $-$   & 1e-15 \\
APOLLO       & 3e-3 & 0.9  & 0.99 & $-$   & 1e-12 & 3e-3 & 0.9  & 0.99 & $-$   & 1e-12 \\
Conda        & 1e-2 & 0.9  & 0.99 & $-$   & 1e-12 & 5e-3 & 0.9  & 0.99 & $-$   & 1e-12 \\
\midrule
\rowcolor{basebg}
\multicolumn{11}{l}{\textit{Gated DeltaNet}} \\
AdamW        & 1e-3 & 0.9  & 0.99 & $-$   & 1e-8  & 1e-3 & 0.9  & 0.99 & $-$   & 1e-8  \\
AdamP        & 1e-3 & 0.9  & 0.98 & $-$   & 1e-15 & 1e-3 & 0.9  & 0.98 & $-$   & 1e-15 \\
Adan         & 3e-3 & 0.9  & 0.92 & 0.99 & 1e-8  & 3e-3 & 0.9  & 0.92 & 0.99 & 1e-8  \\
Lion         & 1e-4 & 0.9  & 0.99 & $-$   & 1e-8  & 1e-4 & 0.9  & 0.99 & $-$   & 1e-8  \\
MARS-AdamW   & 5e-3 & 0.95 & 0.99 & $-$   & 1e-15 & 3e-3 & 0.95 & 0.99 & $-$   & 1e-15 \\
MARS-Lion    & 2e-4 & 0.9  & 0.98 & $-$   & 1e-12 & 1e-4 & 0.9  & 0.98 & $-$   & 1e-12 \\
MARS-Shampoo & 1e-2 & 0.95 & 0.99 & $-$   & 1e-12 & 5e-3 & 0.95 & 0.99 & $-$   & 1e-12 \\
Muon         & 3e-3 & 0.9  & 0.99 & $-$   & 1e-15 & 3e-3 & 0.9  & 0.99 & $-$   & 1e-15 \\
RMNP         & 3e-3 & 0.9  & 0.99 & $-$   & 1e-15 & 3e-3 & 0.9  & 0.99 & $-$   & 1e-15 \\
SOAP         & 3e-3 & 0.9  & 0.95 & $-$   & 1e-15 & 1e-3 & 0.9  & 0.95 & $-$   & 1e-15 \\
APOLLO       & 3e-3 & 0.9  & 0.99 & $-$   & 1e-12 & 3e-3 & 0.9  & 0.99 & $-$   & 1e-12 \\
Conda        & 1e-2 & 0.9  & 0.99 & $-$   & 1e-12 & 5e-3 & 0.9  & 0.99 & $-$   & 1e-12 \\
\bottomrule
\end{tabular}}
\label{tab:appendix_fwe_stage2_combined_expanded}
\end{table}

\section{Detailed Stage-2 Commonsense Reasoning Results}
\label{app:stage2-cs-details}
Tables~\ref{tab:appendix_stage2_cs_transformerpp_340m}--\ref{tab:appendix_stage2_cs_gated_deltanet_1b}
further report the detailed per-task Commonsense Reasoning results for each
architecture and model scale. CS Avg. is the average over ARC-Easy,
ARC-Challenge, HellaSwag, WinoGrande, PIQA, OpenBookQA, BoolQ,
COPA~\cite{roemmele2011choice}, LAMBADA-OpenAI~\cite{paperno-EtAl:2016:P16-1}, and SciQ~\cite{welbl2017crowdsourcing}.

\begin{table}[t]
\centering
\caption{Detailed Stage-2 Commonsense Reasoning results for Transformer++ at 340M.}
\label{tab:appendix_stage2_cs_transformerpp_340m}
\tiny
\setlength{\tabcolsep}{2.1pt}
\renewcommand{\arraystretch}{0.96}
\resizebox{\linewidth}{!}{
\begin{tabular}{lcccccccccccc}
\toprule
\textbf{Optimizer} & \textbf{PPL} & \textbf{CS Avg.} & \textbf{ARC-E} & \textbf{ARC-C} & \textbf{HellaSwag} & \textbf{WinoGrande} & \textbf{PIQA} & \textbf{OBQA} & \textbf{BoolQ} & \textbf{COPA} & \textbf{LAMBADA} & \textbf{SciQ} \\
\midrule
SOAP & \cellcolor{bestbg}\textbf{23.90} & \cellcolor{bestbg}\textbf{53.75} & \cellcolor{bestbg}\textbf{54.92} & \cellcolor{bestbg}\textbf{31.31} & \cellcolor{bestbg}\textbf{44.42} & 54.30 & \cellcolor{bestbg}\textbf{67.79} & \cellcolor{bestbg}\textbf{36.80} & 61.28 & 71.00 & \cellcolor{bestbg}\textbf{38.56} & 77.10 \\
RMNP & 24.37 & 53.35 & 54.71 & 29.52 & 43.35 & 54.22 & 66.65 & 34.80 & 60.58 & 74.00 & 38.00 & 77.70 \\
MARS-AdamW & 24.57 & 52.50 & 51.89 & 28.50 & 42.97 & 51.85 & 66.65 & 34.40 & 60.61 & 73.00 & 36.87 & 78.30 \\
\rowcolor{basebg}
AdamW & 24.62 & 52.28 & 52.44 & 29.52 & 42.09 & 54.54 & 66.97 & 35.20 & \cellcolor{worstbg}58.78 & 69.00 & 35.71 & 78.50 \\
AdamP & 24.68 & 51.69 & 50.80 & 30.12 & 42.65 & 53.04 & 66.65 & 32.80 & 59.48 & 70.00 & 35.92 & 75.40 \\
Muon & 25.05 & 53.25 & 53.20 & 29.95 & 42.84 & \cellcolor{bestbg}\textbf{55.64} & 66.21 & 35.60 & 61.50 & 72.00 & 36.19 & \cellcolor{bestbg}\textbf{79.40} \\
Adan & 25.55 & 52.48 & 51.39 & 29.52 & 42.30 & 53.12 & 66.27 & \cellcolor{worstbg}31.60 & 60.61 & \cellcolor{bestbg}\textbf{75.00} & 37.88 & 77.10 \\
Lion & 26.02 & 51.07 & 51.43 & 28.07 & 40.09 & 51.54 & 65.45 & 33.60 & 60.86 & 69.00 & 34.47 & 76.20 \\
MARS-Lion & 26.20 & 51.61 & 50.93 & 27.30 & 40.27 & \cellcolor{worstbg}51.07 & 65.94 & 35.60 & 60.89 & 72.00 & 34.23 & 77.90 \\
MARS-Shampoo & 26.43 & 51.96 & 52.10 & 28.67 & 42.95 & 54.06 & 66.21 & 32.00 & \cellcolor{bestbg}\textbf{61.77} & 69.00 & 36.46 & 76.40 \\
Conda & 28.30 & 51.61 & 52.65 & 28.41 & 39.77 & 54.54 & 65.51 & 35.20 & 60.58 & \cellcolor{worstbg}67.00 & 33.44 & 79.00 \\
APOLLO & \cellcolor{worstbg}34.08 & \cellcolor{worstbg}48.19 & \cellcolor{worstbg}47.52 & \cellcolor{worstbg}26.96 & \cellcolor{worstbg}37.52 & 51.14 & \cellcolor{worstbg}64.47 & \cellcolor{worstbg}31.60 & 60.03 & \cellcolor{worstbg}67.00 & \cellcolor{worstbg}24.02 & \cellcolor{worstbg}71.60 \\
\bottomrule
\end{tabular}
}
\end{table}

\begin{table}[t]
\centering
\caption{Detailed Stage-2 Commonsense Reasoning results for Transformer++ at 1B.}
\label{tab:appendix_stage2_cs_transformerpp_1b}
\tiny
\setlength{\tabcolsep}{2.1pt}
\renewcommand{\arraystretch}{0.96}
\resizebox{\linewidth}{!}{
\begin{tabular}{lcccccccccccc}
\toprule
\textbf{Optimizer} & \textbf{PPL} & \textbf{CS Avg.} & \textbf{ARC-E} & \textbf{ARC-C} & \textbf{HellaSwag} & \textbf{WinoGrande} & \textbf{PIQA} & \textbf{OBQA} & \textbf{BoolQ} & \textbf{COPA} & \textbf{LAMBADA} & \textbf{SciQ} \\
\midrule
SOAP & \cellcolor{bestbg}\textbf{18.72} & \cellcolor{bestbg}\textbf{57.71} & 60.14 & 34.64 & \cellcolor{bestbg}\textbf{53.21} & 58.01 & 70.89 & 37.40 & 61.87 & 72.00 & \cellcolor{bestbg}\textbf{45.39} & 83.50 \\
\rowcolor{basebg}
AdamW & 18.90 & 56.55 & 59.85 & 34.64 & 51.21 & 55.33 & 70.35 & 36.00 & 61.83 & \cellcolor{worstbg}69.00 & 44.05 & 83.20 \\
MARS-AdamW & 18.94 & 57.46 & \cellcolor{bestbg}\textbf{61.24} & \cellcolor{bestbg}\textbf{35.84} & 52.55 & \cellcolor{bestbg}\textbf{58.41} & 70.51 & 37.20 & 59.24 & 70.00 & 44.50 & \cellcolor{bestbg}\textbf{85.10} \\
AdamP & 19.04 & 56.82 & 60.77 & 35.41 & 51.26 & 55.01 & 70.78 & \cellcolor{bestbg}\textbf{38.40} & 59.60 & 70.00 & 42.71 & 84.20 \\
RMNP & 19.40 & 57.12 & 59.97 & 35.24 & 51.53 & 56.27 & 70.35 & 36.20 & 61.74 & 72.00 & 44.19 & 83.70 \\
Adan & 19.41 & 57.21 & 59.81 & 34.39 & 51.61 & 56.83 & 70.51 & 37.20 & 61.07 & 73.00 & 43.97 & 83.70 \\
MARS-Shampoo & 19.74 & 57.30 & 59.09 & 33.87 & 51.89 & 57.22 & 70.73 & 38.20 & \cellcolor{bestbg}\textbf{62.26} & 71.00 & 44.28 & 84.40 \\
Conda & 19.86 & 57.24 & 59.01 & 33.79 & 51.32 & 57.93 & 70.29 & 36.80 & 62.02 & \cellcolor{bestbg}\textbf{76.00} & 42.69 & 82.50 \\
Muon & 19.86 & 56.36 & 59.26 & 32.76 & 50.31 & 54.85 & \cellcolor{bestbg}\textbf{71.71} & 37.40 & 59.85 & 70.00 & 43.04 & 84.40 \\
Lion & 20.26 & 55.22 & 57.70 & 32.00 & 48.13 & 53.35 & 69.31 & 37.60 & 61.10 & 72.00 & 40.15 & 80.90 \\
MARS-Lion & 21.17 & 54.51 & 56.02 & 31.23 & 47.04 & 53.91 & 69.80 & 37.60 & \cellcolor{worstbg}57.61 & 72.00 & 39.96 & 79.90 \\
APOLLO & \cellcolor{worstbg}25.29 & \cellcolor{worstbg}53.61 & \cellcolor{worstbg}54.67 & \cellcolor{worstbg}30.89 & \cellcolor{worstbg}45.46 & \cellcolor{worstbg}52.49 & \cellcolor{worstbg}68.82 & \cellcolor{worstbg}35.80 & 60.83 & 75.00 & \cellcolor{worstbg}33.13 & \cellcolor{worstbg}79.00 \\
\bottomrule
\end{tabular}}
\end{table}

\begin{table}[t]
\centering
\caption{Detailed Stage-2 Commonsense Reasoning results for GLA at 340M.}
\label{tab:appendix_stage2_cs_gla_340m}
\tiny
\setlength{\tabcolsep}{2.1pt}
\renewcommand{\arraystretch}{0.96}
\resizebox{\linewidth}{!}{
\begin{tabular}{lcccccccccccc}
\toprule
\textbf{Optimizer} & \textbf{PPL} & \textbf{CS Avg.} & \textbf{ARC-E} & \textbf{ARC-C} & \textbf{HellaSwag} & \textbf{WinoGrande} & \textbf{PIQA} & \textbf{OBQA} & \textbf{BoolQ} & \textbf{COPA} & \textbf{LAMBADA} & \textbf{SciQ} \\
\midrule
SOAP & \cellcolor{bestbg}\textbf{27.04} & 52.21 & 52.99 & \cellcolor{bestbg}\textbf{29.27} & 43.11 & \cellcolor{bestbg}\textbf{52.80} & 66.54 & 33.80 & 60.61 & 67.00 & \cellcolor{bestbg}\textbf{36.60} & \cellcolor{bestbg}\textbf{79.40} \\
Muon & 27.47 & \cellcolor{bestbg}\textbf{52.50} & 52.69 & 29.01 & \cellcolor{bestbg}\textbf{43.32} & 52.41 & 66.81 & 34.80 & \cellcolor{bestbg}\textbf{62.23} & \cellcolor{bestbg}\textbf{72.00} & 35.14 & 76.60 \\
MARS-AdamW & 28.28 & 51.24 & 53.37 & 28.41 & 41.55 & 50.59 & 66.92 & 34.00 & 60.12 & 66.00 & 34.89 & 76.50 \\
RMNP & 28.60 & 50.72 & 50.55 & 27.73 & 40.77 & 51.38 & 66.21 & 33.80 & 60.40 & 66.00 & 34.48 & 75.90 \\
AdamP & 28.66 & 51.14 & 52.53 & 27.82 & 40.81 & 50.12 & \cellcolor{bestbg}\textbf{67.41} & 33.60 & 60.83 & 70.00 & 34.04 & 74.30 \\
\rowcolor{basebg}
AdamW & 28.67 & 51.06 & 53.07 & 29.01 & 40.13 & 51.78 & 66.76 & 33.80 & 59.33 & 67.00 & 34.29 & 75.40 \\
Adan & 29.00 & 51.01 & 52.19 & 27.22 & 40.55 & 50.83 & 66.27 & 34.80 & 60.00 & 69.00 & 33.07 & 76.20 \\
MARS-Shampoo & 29.20 & 52.01 & \cellcolor{bestbg}\textbf{53.75} & 29.10 & 42.44 & 52.33 & 66.70 & \cellcolor{bestbg}\textbf{35.40} & 61.35 & 67.00 & 34.60 & 77.40 \\
Lion & 29.47 & 50.14 & 50.13 & 27.56 & 39.54 & 51.78 & 65.61 & \cellcolor{worstbg}32.20 & 59.57 & 70.00 & 33.46 & \cellcolor{worstbg}71.60 \\
MARS-Lion & 29.67 & 50.69 & 52.15 & 27.47 & 39.56 & 52.25 & 65.61 & 34.40 & 60.52 & 67.00 & 32.04 & 75.90 \\
Conda & 37.38 & \cellcolor{worstbg}48.28 & \cellcolor{worstbg}47.10 & \cellcolor{worstbg}26.54 & \cellcolor{worstbg}35.42 & 50.28 & \cellcolor{worstbg}62.95 & 33.20 & 61.59 & \cellcolor{worstbg}65.00 & 27.81 & 72.90 \\
APOLLO & \cellcolor{worstbg}37.75 & 48.38 & 48.48 & 29.01 & 38.45 & \cellcolor{worstbg}49.49 & 64.36 & 32.40 & \cellcolor{worstbg}58.87 & 66.00 & \cellcolor{worstbg}24.28 & 72.50 \\
\bottomrule
\end{tabular}}
\end{table}

\begin{table}[t]
\centering
\caption{Detailed Stage-2 Commonsense Reasoning results for GLA at 1B.}
\label{tab:appendix_stage2_cs_gla_1b}
\tiny
\setlength{\tabcolsep}{2.1pt}
\renewcommand{\arraystretch}{0.96}
\resizebox{\linewidth}{!}{
\begin{tabular}{lcccccccccccc}
\toprule
\textbf{Optimizer} & \textbf{PPL} & \textbf{CS Avg.} & \textbf{ARC-E} & \textbf{ARC-C} & \textbf{HellaSwag} & \textbf{WinoGrande} & \textbf{PIQA} & \textbf{OBQA} & \textbf{BoolQ} & \textbf{COPA} & \textbf{LAMBADA} & \textbf{SciQ} \\
\midrule
SOAP & \cellcolor{bestbg}\textbf{20.62} & \cellcolor{bestbg}\textbf{57.57} & 59.72 & 34.39 & \cellcolor{bestbg}\textbf{53.06} & \cellcolor{bestbg}\textbf{58.88} & 69.80 & 38.20 & 60.55 & 73.00 & 44.77 & 83.30 \\
MARS-Shampoo & 21.53 & 57.33 & 60.14 & \cellcolor{bestbg}\textbf{35.07} & 51.13 & 55.80 & \cellcolor{bestbg}\textbf{70.84} & 37.00 & 61.71 & 72.00 & \cellcolor{bestbg}\textbf{45.18} & 84.40 \\
Muon & 21.54 & 56.85 & 60.61 & 33.79 & 50.48 & 54.06 & 70.73 & 36.60 & 58.84 & \cellcolor{bestbg}\textbf{76.00} & 42.19 & \cellcolor{bestbg}\textbf{85.20} \\
AdamP & 21.86 & 55.31 & 60.02 & 32.94 & 49.19 & 54.70 & 69.21 & 38.40 & 59.51 & 69.00 & 39.96 & 80.20 \\
MARS-AdamW & 21.89 & 55.71 & 58.50 & 33.87 & 49.81 & 54.85 & 69.91 & 36.80 & 58.10 & 75.00 & 40.05 & 80.20 \\
\rowcolor{basebg}
AdamW & 22.06 & 56.69 & \cellcolor{bestbg}\textbf{61.15} & 33.11 & 48.47 & 54.54 & 69.59 & \cellcolor{bestbg}\textbf{39.80} & \cellcolor{bestbg}\textbf{61.90} & \cellcolor{bestbg}\textbf{76.00} & 40.54 & 81.80 \\
RMNP & 22.23 & 55.79 & 59.39 & 32.76 & 49.36 & 53.91 & 69.80 & 36.80 & 59.57 & 71.00 & 42.44 & 82.90 \\
Lion & 22.40 & 53.96 & 57.15 & 31.91 & 47.10 & 53.35 & 69.70 & 36.40 & 58.17 & 70.00 & 38.29 & 77.50 \\
Adan & 22.51 & 55.07 & 58.38 & 31.66 & 48.45 & 53.83 & 69.48 & \cellcolor{worstbg}35.40 & 61.65 & 70.00 & 39.34 & 82.50 \\
Conda & 22.89 & 54.95 & 57.15 & 32.08 & 48.92 & 53.51 & 69.64 & 37.40 & 61.22 & 71.00 & 39.18 & 79.40 \\
MARS-Lion & 23.79 & 53.91 & 55.60 & \cellcolor{worstbg}31.31 & \cellcolor{worstbg}45.72 & 53.91 & 69.48 & 36.00 & 61.80 & 70.00 & 38.46 & \cellcolor{worstbg}76.80 \\
APOLLO & \cellcolor{worstbg}27.78 & \cellcolor{worstbg}52.33 & \cellcolor{worstbg}53.54 & 31.48 & 45.89 & \cellcolor{worstbg}52.49 & \cellcolor{worstbg}67.52 & \cellcolor{worstbg}35.20 & \cellcolor{worstbg}57.89 & \cellcolor{worstbg}66.00 & \cellcolor{worstbg}34.39 & 78.90 \\
\bottomrule
\end{tabular}}
\end{table}

\begin{table}[t]
\centering
\caption{Detailed Stage-2 Commonsense Reasoning results for DeltaNet at 340M.}
\label{tab:appendix_stage2_cs_deltanet_340m}
\tiny
\setlength{\tabcolsep}{2.1pt}
\renewcommand{\arraystretch}{0.96}
\resizebox{\linewidth}{!}{
\begin{tabular}{lcccccccccccc}
\toprule
\textbf{Optimizer} & \textbf{PPL} & \textbf{CS Avg.} & \textbf{ARC-E} & \textbf{ARC-C} & \textbf{HellaSwag} & \textbf{WinoGrande} & \textbf{PIQA} & \textbf{OBQA} & \textbf{BoolQ} & \textbf{COPA} & \textbf{LAMBADA} & \textbf{SciQ} \\
\midrule
SOAP & \cellcolor{bestbg}\textbf{26.02} & 52.60 & 54.55 & 28.58 & \cellcolor{bestbg}\textbf{42.70} & 52.80 & 67.19 & \cellcolor{bestbg}\textbf{36.80} & 59.60 & 69.00 & \cellcolor{bestbg}\textbf{36.50} & 78.30 \\
AdamP & 26.77 & 51.53 & 52.69 & 27.22 & 41.30 & 53.12 & 66.81 & \cellcolor{worstbg}32.00 & 61.31 & 67.00 & 36.15 & 77.70 \\
MARS-AdamW & 26.79 & 51.69 & 53.75 & 28.41 & 41.72 & 52.64 & \cellcolor{bestbg}\textbf{67.52} & 33.00 & 59.66 & 67.00 & 35.11 & 78.10 \\
RMNP & 26.80 & \cellcolor{bestbg}\textbf{53.25} & 55.01 & \cellcolor{bestbg}\textbf{30.20} & 41.71 & 53.04 & 66.87 & 34.80 & 60.24 & \cellcolor{bestbg}\textbf{74.00} & 35.75 & \cellcolor{bestbg}\textbf{80.90} \\
\rowcolor{basebg}
AdamW & 27.16 & 51.74 & 51.60 & 29.18 & 40.66 & 52.57 & 66.43 & 34.80 & \cellcolor{bestbg}\textbf{62.14} & 70.00 & 33.92 & 76.10 \\
Muon & 27.18 & 52.00 & \cellcolor{bestbg}\textbf{55.22} & 29.27 & 42.25 & 52.41 & 66.87 & 33.40 & 61.13 & 67.00 & 33.77 & 78.70 \\
Adan & 27.28 & 51.78 & 52.31 & 29.78 & 41.39 & 52.88 & 67.41 & 34.60 & \cellcolor{worstbg}57.55 & 70.00 & 33.53 & 78.30 \\
Lion & 28.20 & 49.96 & \cellcolor{worstbg}49.83 & \cellcolor{worstbg}26.28 & 39.13 & 52.01 & 65.94 & 32.40 & 59.69 & \cellcolor{worstbg}66.00 & 32.29 & 76.00 \\
MARS-Lion & 28.25 & 50.94 & 50.42 & 28.92 & 39.87 & 52.09 & \cellcolor{worstbg}64.64 & 32.80 & 61.71 & 67.00 & 34.17 & 77.80 \\
MARS-Shampoo & 28.26 & 51.43 & 52.31 & 28.67 & 41.19 & 52.72 & 66.38 & 33.80 & 58.93 & 70.00 & 32.25 & 78.00 \\
Conda & 29.09 & 51.46 & 53.03 & 28.41 & 40.89 & \cellcolor{bestbg}\textbf{54.30} & 64.80 & 34.00 & 59.24 & 70.00 & 32.49 & 77.40 \\
APOLLO & \cellcolor{worstbg}34.73 & \cellcolor{worstbg}49.04 & 50.55 & 27.65 & \cellcolor{worstbg}39.02 & \cellcolor{worstbg}50.36 & 64.47 & 32.20 & 58.84 & 70.00 & \cellcolor{worstbg}24.24 & \cellcolor{worstbg}73.10 \\
\bottomrule
\end{tabular}}
\end{table}

\begin{table}[t]
\centering
\caption{Detailed Stage-2 Commonsense Reasoning results for DeltaNet at 1B.}
\label{tab:appendix_stage2_cs_deltanet_1b}
\tiny
\setlength{\tabcolsep}{2.1pt}
\renewcommand{\arraystretch}{0.96}
\resizebox{\linewidth}{!}{
\begin{tabular}{lcccccccccccc}
\toprule
\textbf{Optimizer} & \textbf{PPL} & \textbf{CS Avg.} & \textbf{ARC-E} & \textbf{ARC-C} & \textbf{HellaSwag} & \textbf{WinoGrande} & \textbf{PIQA} & \textbf{OBQA} & \textbf{BoolQ} & \textbf{COPA} & \textbf{LAMBADA} & \textbf{SciQ} \\
\midrule
SOAP & \cellcolor{bestbg}\textbf{20.38} & 56.49 & 59.55 & 33.45 & \cellcolor{bestbg}\textbf{51.44} & 55.49 & \cellcolor{bestbg}\textbf{71.33} & 38.20 & \cellcolor{worstbg}55.02 & 74.00 & \cellcolor{bestbg}\textbf{43.78} & 82.60 \\
\rowcolor{basebg}
AdamW & 20.66 & 55.56 & 59.51 & 33.62 & 49.71 & \cellcolor{worstbg}52.72 & 70.62 & 36.60 & 59.33 & \cellcolor{worstbg}69.00 & 41.94 & 82.60 \\
MARS-AdamW & 20.67 & 56.80 & 61.53 & 33.19 & 50.78 & 55.09 & 70.51 & 39.40 & 60.76 & 73.00 & 42.05 & 81.70 \\
AdamP & 20.68 & 56.73 & 61.41 & 34.56 & 49.74 & 53.51 & 71.16 & 38.40 & 60.49 & 73.00 & 42.98 & 82.00 \\
Adan & 20.88 & 56.50 & 58.75 & 33.70 & 50.72 & 54.85 & 70.19 & 36.60 & 61.50 & 73.00 & 41.35 & 84.30 \\
RMNP & 21.06 & \cellcolor{bestbg}\textbf{57.32} & \cellcolor{bestbg}\textbf{61.74} & 34.47 & 50.04 & \cellcolor{bestbg}\textbf{56.75} & 70.57 & 36.60 & 61.68 & 74.00 & 41.99 & \cellcolor{bestbg}\textbf{85.40} \\
Muon & 21.18 & 56.65 & 59.30 & 33.19 & 50.30 & 54.85 & 70.19 & 37.00 & 61.25 & \cellcolor{bestbg}\textbf{76.00} & 41.98 & 82.40 \\
MARS-Shampoo & 21.25 & 56.74 & 59.85 & 32.42 & 50.62 & 56.59 & 70.84 & \cellcolor{bestbg}\textbf{39.60} & \cellcolor{bestbg}\textbf{62.35} & 71.00 & 41.80 & 82.30 \\
Lion & 21.44 & 54.22 & 58.08 & 31.91 & 47.38 & 53.75 & \cellcolor{worstbg}68.72 & 36.80 & 55.17 & 72.00 & 39.38 & 79.00 \\
Conda & 21.75 & 56.10 & 57.95 & \cellcolor{bestbg}\textbf{34.64} & 49.86 & 54.38 & 70.13 & 36.60 & 61.62 & 73.00 & 42.01 & 80.80 \\
MARS-Lion & 22.72 & \cellcolor{worstbg}53.65 & \cellcolor{worstbg}55.26 & 32.00 & \cellcolor{worstbg}45.63 & 53.75 & 69.04 & \cellcolor{worstbg}35.00 & 57.86 & 71.00 & 38.72 & \cellcolor{worstbg}78.20 \\
APOLLO & \cellcolor{worstbg}25.58 & 53.88 & 55.68 & \cellcolor{worstbg}32.25 & 46.69 & 53.67 & 69.64 & 35.60 & 58.26 & 72.00 & \cellcolor{worstbg}35.01 & 80.00 \\
\bottomrule
\end{tabular}}
\end{table}

\begin{table}[t]
\centering
\caption{Detailed Stage-2 Commonsense Reasoning results for Gated DeltaNet at 340M.}
\label{tab:appendix_stage2_cs_gated_deltanet_340m}
\tiny
\setlength{\tabcolsep}{2.1pt}
\renewcommand{\arraystretch}{0.96}
\resizebox{\linewidth}{!}{
\begin{tabular}{lcccccccccccc}
\toprule
\textbf{Optimizer} & \textbf{PPL} & \textbf{CS Avg.} & \textbf{ARC-E} & \textbf{ARC-C} & \textbf{HellaSwag} & \textbf{WinoGrande} & \textbf{PIQA} & \textbf{OBQA} & \textbf{BoolQ} & \textbf{COPA} & \textbf{LAMBADA} & \textbf{SciQ} \\
\midrule
SOAP & \cellcolor{bestbg}\textbf{23.53} & 54.77 & 56.10 & 31.57 & \cellcolor{bestbg}\textbf{47.29} & \cellcolor{bestbg}\textbf{55.33} & \cellcolor{bestbg}\textbf{68.99} & 34.20 & 61.41 & 72.00 & 39.26 & 81.60 \\
RMNP & 23.65 & 54.45 & 55.56 & 30.89 & 46.28 & 52.88 & 68.82 & 36.80 & 60.80 & 72.00 & \cellcolor{bestbg}\textbf{40.27} & 80.20 \\
MARS-AdamW & 24.17 & \cellcolor{bestbg}\textbf{54.91} & \cellcolor{bestbg}\textbf{57.95} & 30.97 & 46.49 & 54.30 & 67.41 & 35.60 & 61.62 & 74.00 & 38.68 & \cellcolor{bestbg}\textbf{82.10} \\
AdamP & 24.32 & 53.82 & 54.08 & 30.46 & 45.10 & 54.38 & 68.01 & \cellcolor{bestbg}\textbf{37.40} & 58.81 & 73.00 & 37.94 & 79.00 \\
Muon & 24.34 & 54.45 & 55.01 & \cellcolor{bestbg}\textbf{33.02} & 45.60 & 53.99 & 67.08 & 34.40 & 61.77 & 73.00 & 38.58 & 82.00 \\
\rowcolor{basebg}
AdamW & 24.47 & 53.67 & 55.26 & 31.23 & 44.63 & 51.93 & 67.25 & \cellcolor{worstbg}33.40 & 61.19 & 74.00 & 37.67 & 80.10 \\
Lion & 24.76 & 53.24 & 54.21 & 29.52 & 43.29 & 52.72 & \cellcolor{worstbg}66.97 & 36.00 & 58.99 & \cellcolor{bestbg}\textbf{75.00} & 35.92 & 79.80 \\
Adan & 24.78 & 52.83 & 54.00 & 30.97 & 45.63 & 53.83 & 67.57 & 34.60 & \cellcolor{worstbg}54.89 & \cellcolor{worstbg}68.00 & 38.00 & 80.80 \\
MARS-Lion & 25.31 & 52.96 & 53.54 & 30.03 & 43.56 & 54.46 & 67.74 & 34.20 & 60.64 & 70.00 & 37.16 & 78.30 \\
MARS-Shampoo & 25.99 & 53.37 & \cellcolor{worstbg}51.60 & 29.86 & 45.20 & 54.14 & 67.52 & 34.80 & \cellcolor{bestbg}\textbf{62.35} & 69.00 & 38.17 & 81.00 \\
Conda & 26.11 & 53.45 & 54.92 & 29.86 & 44.24 & 52.96 & 67.08 & 35.20 & 61.71 & 71.00 & 36.60 & 80.90 \\
APOLLO & \cellcolor{worstbg}30.36 & \cellcolor{worstbg}50.92 & 51.56 & \cellcolor{worstbg}29.35 & \cellcolor{worstbg}42.26 & \cellcolor{worstbg}50.91 & 67.41 & 34.00 & 57.68 & 69.00 & \cellcolor{worstbg}29.54 & \cellcolor{worstbg}77.50 \\
\bottomrule
\end{tabular}}
\end{table}

\begin{table}[t]
\centering
\caption{Detailed Stage-2 Commonsense Reasoning results for Gated DeltaNet at 1B.}
\label{tab:appendix_stage2_cs_gated_deltanet_1b}
\tiny
\setlength{\tabcolsep}{2.1pt}
\renewcommand{\arraystretch}{0.96}
\resizebox{\linewidth}{!}{
\begin{tabular}{lcccccccccccc}
\toprule
\textbf{Optimizer} & \textbf{PPL} & \textbf{CS Avg.} & \textbf{ARC-E} & \textbf{ARC-C} & \textbf{HellaSwag} & \textbf{WinoGrande} & \textbf{PIQA} & \textbf{OBQA} & \textbf{BoolQ} & \textbf{COPA} & \textbf{LAMBADA} & \textbf{SciQ} \\
\midrule
SOAP & \cellcolor{bestbg}\textbf{19.86} & 57.22 & 59.47 & 33.36 & 52.36 & 56.12 & 70.35 & 38.00 & 60.61 & 75.00 & 43.99 & 82.90 \\
MARS-AdamW & 20.04 & \cellcolor{bestbg}\textbf{58.18} & \cellcolor{bestbg}\textbf{64.44} & 35.24 & \cellcolor{bestbg}\textbf{52.80} & \cellcolor{bestbg}\textbf{58.09} & 70.67 & 37.80 & \cellcolor{bestbg}\textbf{62.57} & 72.00 & \cellcolor{bestbg}\textbf{44.05} & 84.10 \\
RMNP & 20.26 & 57.30 & 62.04 & 34.73 & 51.80 & 57.14 & 70.35 & 37.80 & 60.40 & 71.00 & 42.95 & 84.80 \\
AdamP & 20.29 & 57.07 & 60.65 & 35.24 & 50.87 & 55.49 & 69.75 & 38.00 & 61.65 & 73.00 & 43.59 & 82.50 \\
Muon & 20.32 & 57.20 & 58.54 & 33.87 & 51.40 & 57.06 & 70.89 & 38.20 & 59.48 & 74.00 & 43.70 & 84.80 \\
\rowcolor{basebg}
AdamW & 20.33 & 57.01 & 59.43 & 33.79 & 50.48 & 54.85 & 70.57 & 37.60 & 61.19 & \cellcolor{bestbg}\textbf{77.00} & 42.89 & 82.30 \\
Lion & 20.38 & 55.74 & 57.62 & 32.51 & 49.64 & 55.09 & \cellcolor{bestbg}\textbf{71.33} & 37.80 & 59.27 & 71.00 & 41.04 & 82.10 \\
Adan & 20.55 & 57.93 & 63.64 & \cellcolor{bestbg}\textbf{36.69} & 52.45 & 58.01 & 69.75 & 36.80 & \cellcolor{worstbg}57.37 & 76.00 & 43.04 & \cellcolor{bestbg}\textbf{85.60} \\
MARS-Shampoo & 20.61 & 57.58 & 58.88 & 34.47 & 51.96 & 57.93 & 70.24 & 37.60 & 61.25 & 75.00 & 43.94 & 84.50 \\
Conda & 21.07 & 57.18 & 57.83 & 33.45 & 51.24 & 55.49 & 70.19 & \cellcolor{bestbg}\textbf{39.00} & 62.08 & 74.00 & 43.37 & 85.20 \\
MARS-Lion & 21.24 & 55.50 & 57.41 & 32.34 & 48.79 & \cellcolor{worstbg}54.54 & 70.95 & 37.20 & 59.88 & 71.00 & 40.56 & 82.30 \\
APOLLO & \cellcolor{worstbg}25.61 & \cellcolor{worstbg}53.73 & \cellcolor{worstbg}55.85 & \cellcolor{worstbg}31.66 & \cellcolor{worstbg}47.68 & 54.85 & \cellcolor{worstbg}68.23 & \cellcolor{worstbg}35.80 & 61.77 & \cellcolor{worstbg}68.00 & \cellcolor{worstbg}34.31 & \cellcolor{worstbg}79.20 \\
\bottomrule
\end{tabular}}
\end{table}

\begin{table*}[htbp]
  \centering
  \caption{Dimension-A methodological taxonomy of the surveyed optimizer set.
  SF-AdamW denotes Schedule-Free AdamW.}
  \label{tab:method_taxonomy}
  \fontsize{8pt}{9pt}\selectfont
  \setlength{\tabcolsep}{1.5pt}
  \renewcommand{\arraystretch}{1.2}
  \begin{tabular}{@{}p{0.32\textwidth}p{0.11\textwidth}p{0.34\textwidth}p{0.025\textwidth}p{0.08\textwidth}@{}}
    \toprule
    Core mechanism & Pipeline site & Four-axis summary & N & Example \\
    \midrule
    \rowcolor{famT1!18}\multicolumn{5}{c}{\textbf{\textcolor{famT1!62!black}{T1: Element-wise adaptive moment and scalar control}}} \\
    \rowcolor{famT1!8}\multicolumn{5}{@{}l@{}}{\textbf{\textcolor{famT1!62!black}{T1.1 Scalar bases and direct Adam variants}}} \\
      Adam-style moments, bias correction, decoupled decay, Nesterov, rectified, belief, and cross-moment variants
      & S3 + local S5
      & Full domain, moment state, diagonal adaptive geometry, LR/WD finalize
      & 25 & AdamW \\
    \rowcolor{famT1!8}\multicolumn{5}{@{}l@{}}{\textbf{\textcolor{famT1!62!black}{T1.2 Multi-timescale, momentum correction, and variance reduction}}} \\
      Fast/slow EMA fusion, alignment-aware momentum correction, or STORM-style recursive gradient correction
      & S3; partly Axis~II
      & Full domain, temporal or VR state, diagonal geometry, LR/WD finalize
      & 10 & AdEMA\-Mix \\
    \rowcolor{famT1!8}\multicolumn{5}{@{}l@{}}{\textbf{\textcolor{famT1!62!black}{T1.3 Iterate averaging and automatic tuning}}} \\
      Schedule-free averaging, distance-based learning-rate inference, optimizer switching, initialization-time LR scaling
      & S3 + S5
      & Full domain, scalar or iterate state, base geometry, global-step finalize
      & 10 & SF-AdamW \\
    \midrule
    \rowcolor{famT2!18}\multicolumn{5}{c}{\textbf{\textcolor{famT2!62!black}{T2: Matrix-level structural methods}}} \\
    \rowcolor{famT2!8}\multicolumn{5}{@{}l@{}}{\textbf{\textcolor{famT2!62!black}{T2.1 Spectral orthogonalization}}} \\
      Orthogonalization of two-dimensional gradients or momentum
      & S1 + S2
      & Matrix domain, momentum state, polar operator, routed writeback
      & 8 & Muon \\
    \rowcolor{famT2!8}\multicolumn{5}{@{}l@{}}{\textbf{\textcolor{famT2!62!black}{T2.2 Kronecker-factored preconditioning}}} \\
      Row/column covariance, Fisher, or structured Fisher factors for matrix gradients
      & S1 + S2 + S3
      & Kronecker/eigen domain, factor state, metric geometry, damped writeback
      & 8 & SOAP \\
    \rowcolor{famT2!8}\multicolumn{5}{@{}l@{}}{\textbf{\textcolor{famT2!62!black}{T2.3 Low-rank subspace projection}}} \\
      Project gradients and states into a low-rank subspace, then reconstruct
      & S2 + S3 + S4
      & Projected subspace, projected state, adaptive geometry, projection-back
      & 3 & GaLore \\
    \midrule
    \rowcolor{famT3!18}\multicolumn{5}{c}{\textbf{\textcolor{famT3!62!black}{T3: Discretization and directional quantization}}} \\
    \rowcolor{famT3!8}\multicolumn{5}{@{}l@{}}{\textbf{\textcolor{famT3!62!black}{T3 Direction quantization}}} \\
      Sign or related maps that preserve direction and coarsen magnitude
      & S2 + simplified S3
      & Full domain, raw or momentum state, sign geometry, LR/WD radius
      & 7 & Lion \\
    \midrule
    \rowcolor{famT4!18}\multicolumn{5}{c}{\textbf{\textcolor{famT4!62!black}{T4: State compression and structural aggregation}}} \\
    \rowcolor{famT4!8}\multicolumn{5}{@{}l@{}}{\textbf{\textcolor{famT4!62!black}{T4.1 Row-column factorization}}} \\
      Approximate matrix second moments with row/column statistics
      & S3 + S4
      & Full domain, row/column state, adaptive geometry, broadcast denominator
      & 2 & CAME \\
    \rowcolor{famT4!8}\multicolumn{5}{@{}l@{}}{\textbf{\textcolor{famT4!62!black}{T4.2 Low-bit quantization and error compensation}}} \\
      Quantize optimizer state with dynamic scaling, stochastic rounding, or error control
      & S3 + S4
      & Full domain, low-bit state, base geometry, dequantized finalize
      & 2 & 8-bit Adam \\
    \rowcolor{famT4!8}\multicolumn{5}{@{}l@{}}{\textbf{\textcolor{famT4!62!black}{T4.3 Block/layer-level state sharing}}} \\
      Share moment or curvature statistics over heads, blocks, or layers
      & S1 + S3 + S4
      & Block support, shared state, block-adaptive geometry, broadcast writeback
      & 5 & Adam-mini \\
    \rowcolor{famT4!8}\multicolumn{5}{@{}l@{}}{\textbf{\textcolor{famT4!62!black}{T4.4 Fused backprop-update}}} \\
      Update parameters during backpropagation and release gradients early
      & S3 + S5
      & Full support, streaming state, base geometry, immediate writeback
      & 2 & LOMO \\
    \midrule
    \rowcolor{famT5!18}\multicolumn{5}{c}{\textbf{\textcolor{famT5!62!black}{T5: Curvature-aware and geometric regularization}}} \\
    \rowcolor{famT5!8}\multicolumn{5}{@{}l@{}}{\textbf{\textcolor{famT5!62!black}{T5.1 Adversarial perturbation (SAM family)}}} \\
      Re-evaluate gradients at a local perturbation point
      & S5 + extra gradient
      & Base domain, perturbed-gradient state, base geometry, sharpness wrapper
      & 8 & SAM \\
    \rowcolor{famT5!8}\multicolumn{5}{@{}l@{}}{\textbf{\textcolor{famT5!62!black}{T5.2 Diagonal Hessian estimation}}} \\
      Estimate Hessian/Gauss--Newton diagonals and apply clipped updates
      & S3 + S5
      & Full domain, HVP/Hessian state, clipped curvature geometry, LR/WD finalize
      & 2 & Sophia \\
    \rowcolor{famT5!8}\multicolumn{5}{@{}l@{}}{\textbf{\textcolor{famT5!62!black}{T5.3 Update post-processing and selective filtering}}} \\
      Centralize, project, normalize, mask, or sparsify computed updates
      & S2 or S5
      & Base domain, base state, base geometry, final filter or mask
      & 13 & GC \\
    \rowcolor{famT5!8}\multicolumn{5}{@{}l@{}}{\textbf{\textcolor{famT5!62!black}{T5.4 Layer-wise trust-region scaling}}} \\
      Rescale layer-wise updates by parameter/update norm ratios
      & S1 + S5
      & Layer support, base state, base geometry, trust-ratio finalize
      & 2 & LAMB \\
    \bottomrule
  \end{tabular}
\end{table*}

\end{document}